\documentclass[letterpaper]{article} % DO NOT CHANGE THIS
\usepackage{aaai25}  % DO NOT CHANGE THIS
\usepackage[utf8]{inputenc} % For handling UTF-8 encoding
\usepackage[T1]{fontenc}    % For proper font encoding
\usepackage[english]{babel} % Adjust the language as needed
\usepackage{times}  % DO NOT CHANGE THIS
\usepackage{helvet}  % DO NOT CHANGE THIS
\usepackage{courier}  % DO NOT CHANGE THIS
\usepackage[hyphens]{url}  % DO NOT CHANGE THIS
\usepackage{graphicx} % DO NOT CHANGE THIS
\usepackage{natbib}  % DO NOT CHANGE THIS AND DO NOT ADD ANY OPTIONS TO IT
\usepackage{caption} % DO NOT CHANGE THIS AND DO NOT ADD ANY OPTIONS TO IT
\usepackage{algorithm}
\usepackage{algorithmic}

\usepackage{url}            % simple URL typesetting
\usepackage{booktabs}       % professional-quality tables
\usepackage{amsfonts}       % blackboard math symbols
\usepackage{nicefrac}       % compact symbols for 1/2, etc.
\usepackage{microtype}      % microtypography
\usepackage{xcolor}         % colors
\usepackage{graphicx} 
\usepackage{multirow} 
\usepackage{caption}
\usepackage{tabularx}
\usepackage{times}
\usepackage{soul}
\usepackage{caption}
\usepackage{amsmath}
\usepackage{amsthm}
\usepackage{algorithm}
\usepackage{algorithmic}
\usepackage[switch]{lineno}
\usepackage{etoolbox}
\usepackage{subcaption}
\usepackage{newfloat}
\usepackage{listings}
\DeclareCaptionStyle{ruled}{labelfont=normalfont,labelsep=colon,strut=off} % DO NOT CHANGE THIS
\floatstyle{ruled}
\newfloat{listing}{tb}{lst}{}
\floatname{listing}{Listing}
\title{The \emph{Indoor-Training Effect}: unexpected gains from distribution shifts \\ in the transition function}

\author {
    Serena Bono\textsuperscript{\rm 1},
    Spandan Madan\textsuperscript{\rm 2},
    Ishaan Grover\textsuperscript{\rm 1}
    Mao Yasueda \textsuperscript{\rm 3},\\
    Cynthia Breazeal \textsuperscript{\rm 1},
    Hanspeter Pfister\textsuperscript{\rm 2},
    Gabriel Kreiman\textsuperscript{\rm 2}
}
\affiliations {
    \textsuperscript{\rm 1}MIT Media Lab 
    \textsuperscript{\rm 2}Harvard University
    \textsuperscript{\rm 3}Yale University 
}
\usepackage{bibentry}
\begin{document}

\maketitle

\begin{abstract}
Is it better to perform tennis training in a pristine indoor environment or a noisy outdoor one? To model this problem, here we investigate whether shifts in the transition probabilities between the training and testing environments in reinforcement learning problems can lead to better performance under certain conditions. We generate new Markov Decision Processes (MDPs) starting from a given MDP, by adding quantifiable, parametric noise into the transition function. We refer to this process as \textit{Noise Injection} and the resulting environments as \textit{$\delta$-environments}. This process allows us to create variations of the same environment with quantitative control over noise serving as a metric of distance between environments. Conventional wisdom suggests that training and testing on the same MDP should yield the best results. In stark contrast, we observe that agents can perform better when trained on the noise-free environment and tested on the noisy $\delta$-environments, compared to training and testing on the same $\delta$-environments. We confirm that this finding extends beyond noise variations: it is possible to showcase the same phenomenon in ATARI game variations including varying Ghost behaviour in PacMan, and Paddle behaviour in Pong. We demonstrate this intriguing behaviour across $60$ different variations of ATARI games, including PacMan, Pong, and Breakout. We refer to this phenomenon as the \emph{Indoor-Training Effect}. Code to reproduce our experiments and to implement \textit{Noise Injection} can be found at \url{https://bit.ly/3X6CTYk}.
\end{abstract}

% Uncomment the following to link to your code, datasets, an extended version or similar.
%
% \begin{links}
%     \link{Code}{https://aaai.org/example/code}
%     \link{Datasets}{https://aaai.org/example/datasets}
%     \link{Extended version}{https://aaai.org/example/extended-version}
% \end{links}

\begin{figure}[thb!]
\centering
        \includegraphics[width=0.33\textwidth]{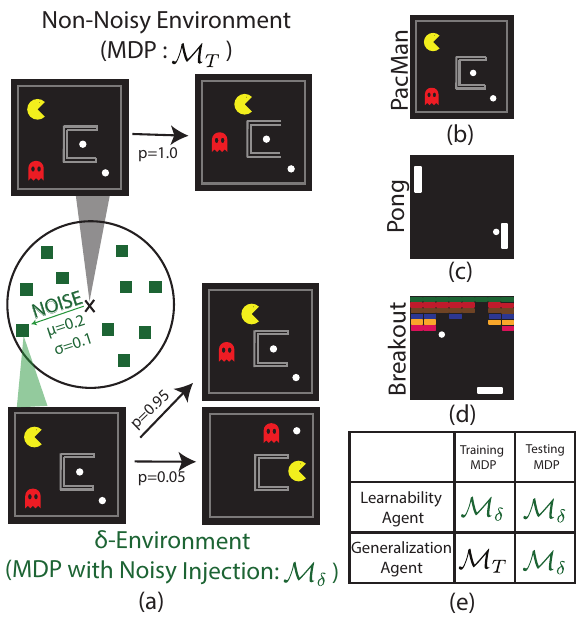}
        \caption{\textit{ATARI games modified with Noise Injection.} (a) In the original Target Environment($\mathcal{M}_T$), when the agent (PacMan) moves right, PacMan moves right with probability $1.0$. Noise Injection allows us to create multiple worlds in the vicinity of this environment by adding controlled Gaussian noise ($\delta$) to the original Transition Function ($T$). When the agent takes the action \textit{right} in these $\delta-$environments, with a low probability the game may transition to a state which would not be possible in non-noisy PacMan. For brevity, we refer to these transitions as non-standard transitions which are $0$ probability in the original Target, but are now possible. Experiments with noise injection are presented on three ATARI games---(b) PacMan, (c) Pong, and (d) Breakout. (e) We compare two agents with these environments---a Learnability agent trained and tested on the same target environment ($\mathcal{M}_\delta)$, and a Generalization agent trained on a different MDP ($\mathcal{M}_T$) and tested on $\mathcal{M}_\delta$.} 
        \label{fig:fig_schematic}
\end{figure}

\begin{figure*}[thb!]
\centering
        \includegraphics[width=0.8\textwidth]{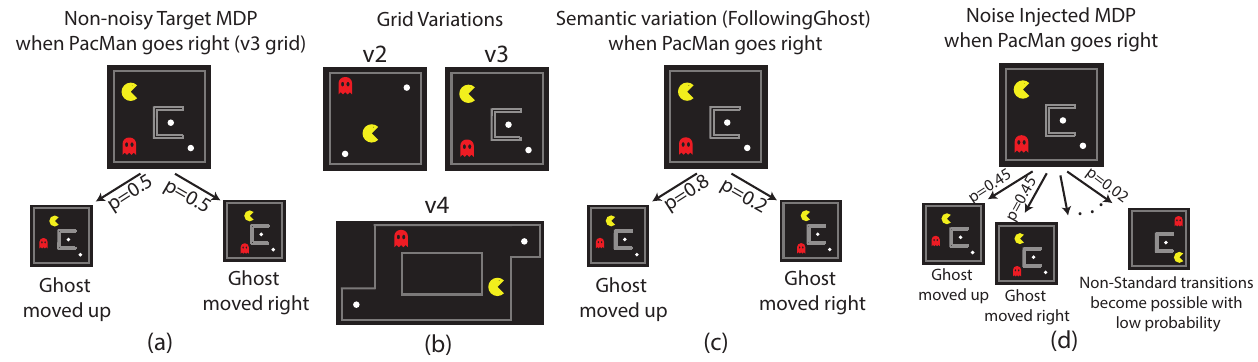} 
        \caption{\emph{Schematic illustration of variations for Pacman.}(a) 
        Game dynamics when the agent picks the action \textit{right} in a standard, non-noisy MDP for the v3 grid. The ghosts' actions follow a uniform probability distribution over possible moves and move \textit{up} or \textit{right} with an equal probability of $0.5$. This is referred to as a RandomGhost. (b) Grid variations for Pacman---v2, v3, and v4. These grids vary in size, positions of walls, and positions of food pellets. v2, v3, and v4 are designed to be increasingly harder. (c) Semantic variations whereby there is a meaningful change in the distribution of game elements. 
        Here, a FollowingGhost is depicted which has a higher probability of taking a move that brings it closer to the Pacman ($0.8$). 
        (d) Noise injected MDP generated by adding Gaussian noise to the standard transition function. Alongside states reachable by the ghost taking a legal move, non-standard transitions now become possible which result in the game reaching states otherwise unreachable. 
        }
        \label{fig:atari_variations}
\end{figure*}

\section{Introduction}

Consider the process of learning how to play tennis. You might think that the best way to prepare for an outdoor match is to train under the same outdoor conditions you will face during the match. However, training in a calm, noise-free indoor environment instead can help focus on mastering the fundamentals of tennis without the added challenge of sources of noise like wind. We refer to this phenomenon as the \emph{Indoor-Training Effect}. Here we model this problem using reinforcement learning (RL) agents. Surprisingly, we found that under certain conditions, training in a noise-free environment can lead to better performance when tested in a noisy environment---just like tennis. This phenomenon challenges our intuitions about the standard way to train  RL agents where conventional wisdom would suggest that the best approach to perform well on a target problem is to train an RL agent on the same test environment.

Environments in RL are usually described using Markov Decision Process (MDP). An MDP is defined by a State Space $\mathcal{S}$, an Action Space $\mathcal{A}$, a Transition Function $\mathcal{T}$, %that specifies the transition probabilities between states given an action
and a Reward Function $\mathcal{R}$. In practice, these parameters are assumed to be known or approximated with reasonable precision~\cite{bauerle2022distributionally,goyal2023robust}. A significant challenge in RL is generalizing to environments that differ from the training environment~\cite{cobbe2019quantifying,kang2019generalization, devin2018deep}. To address this, the RL community has focused on training agents capable of learning policies that perform well in novel, unseen environments at deployment time~\cite{DBLP:journals/corr/abs-2111-09794,make4010013,DBLP:journals/corr/abs-1910-07113,DBLP:journals/corr/abs-2006-14911,biedenkapp-ecai20}. The complexity of this task has called for ingenious ways of aligning the policy learned by the agent in training environments with the testing optimal policy. Notable approaches include using human feedback \cite{rummery1994line}, using language \cite{tellex2011understanding, walter2013learning, squire2015grounding}, and using vision \cite{guss2019minerl, osinski2020simulation, gopalan2017planning}.

To study this, we explored zero-shot policy transfer where a policy trained in one environment is tested on a different environment. We extended past works which focused on uncertainty in the transition probabilities~\cite{nilim2005robust,moos2022robust,goyal2023robust}, and propose a novel framework for studying zero-shot policy transfer in environments with controlled, quantifiable distribution shifts in the transition probabilities. 

Our framework introduces these shifts by computing the transition function of an MDP, and adding small Gaussian noise to its entries.
%, as shown in Fig.~\ref{fig:fig_schematic}(a). 
Starting with an environment ($\mathcal{M}_T$), noise is sampled and added to it to obtain a new MDP ($\mathcal{M}_\delta$). We refer to this approach as \textit{Noise Injection} and the resulting new MDPs as \textit{$\delta$-environments} as in Fig.~\ref{fig:fig_schematic}. Noise injection introduces several non-standard transitions, which had zero probability in the original MDP. Multiple such environments can be created by sampling noise and the noise serves as a metric of distance between environments. This approach allows us to create multiple worlds starting from the same MDP, with quantitative control over the variations in the transition probabilities. An increase in the standard deviation of the Gaussian noise results in increasingly perturbed MDPs. We report experiments with Noise Injection on multiple domains across three ATARI games---PacMan, Pong, and Breakout.
%as shown in Fig.~\ref{fig:fig_schematic} (b), (c), and (d), respectively. 

\begin{table*}
\centering
\footnotesize
{\renewcommand{\arraystretch}{1}
\begin{tabular}{p{2cm}|p{2.4cm}|c|c|p{0.6cm}}
    \hline
    \hline
    \multirow{2}{*}{\textbf{ATARI Game}} & \multirow{2}{*}{\textbf{Grid Variations}} & \multirow{2}{*}{\textbf{Noise Injected Variations}} & \multirow{2}{*}{\textbf{Semantic Variation}} & \multirow{2}{*}{\textbf{Total}}\\
    & & & \\
     \hline
     \multirow{4}{*}{\textbf{PacMan}} & & $\delta = 0$ (No Noise) &RandomGhost & \\
      & v2, v3, v4 & $\delta \sim \mathcal{N}(0,0.1)$ & FollowingGhost ($p=0.3,0.6$) &  33\\  
      & & $\delta \sim \mathcal{N}(0,0.5)$ & TeleportingGhost ($p=0.5,0.2$) & \\  
      \hline
      \multirow{4}{*}{\textbf{Pong}} & & $\delta = 0$ (No Noise)  & RandomPaddle & \\
      &  p1, p2 &  $\delta \sim \mathcal{N}(0,0.1) $ & FollowingPaddle ($p=0.3,0.6$) & 18\\
      & & $\delta \sim \mathcal{N}(0,0.5)$ & & \\
      \hline
     \multirow{4}{*}{\textbf{Breakout}} &  & $\delta = 0$ (No Noise)  & & \\
      & b1, b2, b3 & $\delta \sim \mathcal{N}(0,0.1)$ & - & 9\\ 
      & & $\delta \sim \mathcal{N}(0,0.5)$& & \\
      \hline
\end{tabular}
\caption{\textbf{Overview of experimental protocol.} 
Our experiments include multiple variations of three ATARI games---PacMan, Pong, and Breakout. For each game, we have multiple grid variations. 
When introducing variations in these grids with noise injection, we report results for two levels of added noise---a low-noise setting: $\delta \sim \mathcal{N}(0,0.1)$, and a high-noise setting: $\delta \sim \mathcal{N}(0,0.5)$. 
Furthermore, for each grid we introduce further variations by modifying the distribution of the stochastic game element (ghost in PacMan, and the computer paddle in Pong). In all, we report results on $60$ MDPs across these games.
}\label{table:protocol}
}
\end{table*}
To study policy transfer we define two agents: 
a \textbf{\textit{Learnability Agent}} ($\mathcal{L}_\delta$) which is \textbf{trained and tested on the same \textit{$\delta$-environment} ($\mathcal{M}_\delta$)}, and a \textbf{\textit{Generalization Agent}} ($\mathcal{G}_T$) which is \textbf{trained on the original noise-free environment ($\mathcal{M}_T$) but tested on the \textit{$\delta$-environment} ($\mathcal{M}_\delta$)}. Conventional wisdom suggests that the Learnability Agent should perform better as it is trained and tested on the same environment. However, our study across $60$ MDPs built on ATARI games reveals a surprising finding---there are several cases where the Generalization Agent outperformed the Learnability Agent. 
We confirmed that this finding extends beyond our setup of noise injection and $\delta$-environments and also holds true for game variations including varying the Ghost behaviour in PacMan, and Paddle behaviour in Pong. We refer to these as semantic variations in MDPs.

In conclusion, to better understand this phenomenon we analyzed the exploration patterns of the Learnability and Generalization Agents, and the corresponding policies learned by them. Our analyses revealed that $\mathcal{L}_\delta$ agents outperformed $\mathcal{G}_T$ agents, as expected from the literature, when $\mathcal{G}_T$ agents fail to explore the same State-Action pairs as the $\mathcal{L}_\delta$ agents. In contrast, when there were no large differences in their exploration patterns, the performance of  $\mathcal{G}_T$ aligned or exceeded that of  $\mathcal{L}_\delta$ agent.

\subsection*{Preliminaries: Reinforcement Learning}

Similar to~\cite{Cederborg2015PolicySW}, our work considers Reinforcement Learning (RL) as a group of algorithms designed to solve problems formulated as Markov Decision Processes (MDPs). A Markov Decision Process is characterized by the tuple $(\mathcal{S}, \mathcal{A}, \mathcal{T}, \mathcal{R}, \lambda)$, representing the collection of potential world states ($\mathcal{S}$), space of actions ($\mathcal{A}$), the transition function ($\mathcal{T} : \mathcal{S} \times \mathcal{A} \rightarrow \mathcal{P}(\mathcal{S})$), the reward function ($\mathcal{R} : \mathcal{S} \times \mathcal{A} \rightarrow \mathcal{R}$), and a discount factor ($0 < \gamma \leq 1$). The objective is to identify policies ($\pi : \mathcal{S} \times \mathcal{A} \rightarrow \mathcal{R}$) that maximize cumulative rewards.

Q-learning \cite{Watkins1992} and SARSA \cite{Kaelbling1996ReinforcementLA} are two algorithms to learn such policies.  
Both Q-Learning and SARSA algorithms update the Q-values of state-action pairs, but they differ in their approaches. Q-Learning focuses on the maximum expected future rewards, and updates Q-values using the formula:
\begin{equation}
    Q(s, a) \leftarrow Q(s, a) + \alpha \left[ r + \gamma \max_{a'} Q(s', a') - Q(s, a) \right]
\end{equation}
where $\alpha$ is the learning rate, $\gamma$ is the discount factor, and $s, s', a, a', r$ represent the current state, next state, current action, next action, and immediate reward, respectively.

On the other hand, SARSA updates Q-values based on the actual policy's actions with the formula:
\begin{equation}
    Q(s, a) \leftarrow Q(s, a) + \alpha \left[ r + \gamma Q(s', a') - Q(s, a) \right]
\end{equation}
Here the update incorporates both immediate rewards and the Q-value of the actual next action taken.

Agents need to balance two critical aspects: exploration and exploitation. Exploration involves trying potentially less optimal actions to understand the environment better. Conversely, exploitation means choosing actions known to yield high rewards. We report results with the Boltzmann and the $\epsilon$-greedy exploration strategies. 
Boltzmann exploration determines the probability of selecting an action as follows:
\begin{equation}
  Pr_q(a) = \frac{e^{Q(s,a)/\tau}}{\sum_{a'}e^{Q(s,a')/\tau}}  
\end{equation}

The constant $\tau$ is referred to as the temperature. On the other hand, the $\epsilon$-greedy strategy is simpler and more direct---the agent selects a random action with probability $\epsilon$, and the action with the highest Q-value with probability $1-\epsilon$.

\section{Related Works}
Generalization benchmarks in RL involve training and testing across different subsets of tasks, levels, or environments. Recent years have seen several generalization benchmarks, which include variations in the state space~\cite{DBLP:journals/corr/abs-2109-06780}, dynamics \cite{DBLP:journals/corr/abs-1904-12901}, observation \cite{DBLP:journals/corr/abs-2009-12293}, reward function \cite{DBLP:journals/corr/abs-1904-03177}, and new game levels~\cite{justesen2018illuminating}, among others. There has also been recent work investigating generalization in Deep Reinforcement Learning~\cite{zhu2023transfer,packer2018assessing,cobbe2019quantifying,lyle2022learning}. Combined, these tasks require explicit modeling of variations to effectively assess generalization, highlighting the need for robust evaluation protocols.

Contextual Markov Decision Processes (CMDP) provide a formal structure for this, where environments are sampled from a class of contexts, with agents trained on a subset and tested on a disjoint subset. These contexts are generated through two primary methods: Procedural Content Generation (PCG), which relies on a seed value for environment generation, and Controllable Environments (CE), which allow for manipulation of individual components. The integration of a suitable evaluation protocol with these contexts helps define the relationship between training and testing sets, which can range from interpolation to full extrapolation. Some examples of benchmarks using these frameworks include the OpenAI Procgen benchmark \cite{cobbe2020leveraging} and the Distracting Control Suite \cite{stone2021distracting} for PCG 
%\GK{What is PCG? Define}, and CausalWorld 
\cite{DBLP:journals/corr/abs-2010-04296} and RWRL \cite{DBLP:journals/corr/abs-1904-12901} for controllable environments. A major drawback in these benchmarks is the lack of a clearly defined metric for measuring how the distance between different contexts affects agent performance.

To solve this issue we draw inspiration from work studying generalization under controlled, quantifiable distribution shifts in computer vision. These studies include shifts in 3D rotation~\cite{mondal2022generalization,madan2023adversarial}, category-viewpoint combinations~\cite{madan2022and}, incongruent scene context~\cite{bomatter2021pigs}, novel light and viewpoint combinations~\cite{sakai2022three}, object materials~\cite{madan2022makes} and texturess~\cite{geirhos2018imagenet, michaelis2019benchmarking}, and non-canonical viewpoints~\cite{barbu2019objectnet}, among others.

\section*{Generating MDPs for investigating generalization}\label{sec:generating_mdps}
We created $60$ different MDPs across three ATARI games (PacMan, Pong, and Breakout) by varying grid layouts, distributions defining the stochasticity of different game elements, and modifying transition probabilities using Noise Injection (Fig.~\ref{fig:atari_variations} and Table.~\ref{table:protocol}). Here we outline these variations.

\subsubsection*{Domains}
We implemented all three ATARI games from scratch, building on the Berkeley PacMan Projects \cite{berkeleypacproj}. PacMan was modelled as an MDP characterized by the tuple $(\mathcal{S}, \mathcal{A}, \mathcal{T}, \mathcal{R}, \lambda)$. 
%Below we define these:

\noindent \textbf{State ($s$) and State Space ($\mathcal{S}$):} We represented a grid of size $M \times N$ as a matrix of the same shape with the entries corresponding to the game element occupying the position in the grid---p (PacMan), g (Ghost), f (Food), w (Wall), or e (Empty). %States were computed by unrolling this matrix into a vector of length $M*N$. 
The state space $\mathcal{S}$ refers to the set of all possible states.

\noindent\textbf{Action Space ($\mathcal{A}(s)$):} Set of legal actions PacMan could take in state $s$. PacMan can move Left, Right, Up, or Down but not enter walls. Thus, when the PacMan is at the top left position the set of legal actions was only \{Right, Down\}. 

\noindent \textbf{Transition Matrix ($\mathcal{T}(s_i, a, s_j)$):} Probability of moving to state $s_j$ if the agent took action $a$ at state $s_i$ (Fig.~\ref{fig:atari_variations}a).

\noindent \textbf{Reward Function ($\mathcal{R}(s)$):} PacMan received +20 for eating a food pellet, -1 for every time step, -200 when it was killed, and +500 for finishing the game. ~\cite{Cederborg2015PolicySW}.

\noindent \textbf{Game Stochasticity:} The motion of PacMan is deterministic---a left action (if legal) will ensure that PacMan moves left. However, ghosts move stochastically according to a prefixed distribution. For instance, a RandomGhost moves in all directions with equal probability (accounting for walls). Thus, the game is nondeterministic.

MDPs for Pong and Breakout are defined analogously. For additional details, please refer to Supplementary Section~\textbf{Domains}.

\subsection*{Noise Injection Variation: Generating new controlled environments}\label{sec:TFcomputation}
We generate controlled variations of an original MDP by explicitly computing its Transition Function and then adding sampled noise to it. %Below we outline this process:

\noindent \textbf{Explicit enumeration of all states:}
States are defined by the position of the game elements. The probability of transitioning from one state to another is computed by multiplying the probability that each game element is able to reach the final configuration independently. Therefore, we visualize the game as a tree, each state is a node, and the edges represent the transition probabilities of the game elements independently reaching their final configuration. By rolling out all possible moves by each game elements at each step, we enumerate all possible reachable states.
% Should this be mentioned as a procedural limitation because this becomes prohibitive for more complex games?

\noindent \textbf{Explicit computation of Transition Function:} 
Once we have all possible states, we can calculate the transition function, denoted as $\mathcal{T}(s_i,a,s_j)$. This function is determined by calculating the probability of each game character moving from one state to another independently.

\noindent \textbf{Creating $\delta-$environments:} We introduce variations in the game environment by modifying the transition function to $\mathcal{T}_{\delta} = \mathcal{T} + \delta$. Here, $\delta$ is a variable that follows a normal distribution, randomly chosen before each game to add unpredictability (Fig.~\ref{fig:atari_variations}c). The modified transition function, $\mathcal{T}_\delta$, is then adjusted to make sure the total probability of moving from any state $s_i$ using action $a$ to any other state $s_j$ sums to 1.

\begin{equation}
            \mathcal{T}_\delta(s_{j},a,s_{i}) = \frac{|\mathcal{S}| p_{i,j} + \delta_{i,j}}{|\mathcal{S}| + \sum_{j}\delta_{i,j}}
\label{eq:norm}
\end{equation}

\noindent $|\mathcal{S}|$ denotes the number of states, and guarantees the probability of legal successors does not approach $0$ as the state space grows. We investigated two settings---(i) \textit{Low-Noise} with $\delta \sim \mathcal{N}(0,0.1)$, where some non-standard transitions previously impossible without noise are now possible with a low probability. (ii) \emph{High-Noise} with $\delta \sim \mathcal{N}(0,0.5)$, where non-standard transitions are possible with higher probability. We further analyze minimum $\mathcal{N}(0,0)$, and maximum $\mathcal{N}(0,1)$ perturbation settings in the Supplement Sec.~\textbf{Perturbation Bounds}.
\section{Experimental Details}\label{section:experimental-details}
We compared the mean reward curve of Learnability and Generalization agents. An agent $\mathcal{G}_T$ is said to generalize well with respect to $\mathcal{M}_\delta$, if its mean reward is as good as the corresponding Learnability agent $\mathcal{L}_\delta$.

Agents are trained with both tabular Q-Learning~\cite{Watkins1992} and SARSA Q-learning \cite{Kaelbling1996ReinforcementLA}, using Boltzmann or $\epsilon$-greedy exploration strategies. In particular, we trained agents for $1,000$ episodes and averaged results over $500$ trained agents. After every $10$ training episodes, agents were evaluated using $10$ testing episodes. We report the mean reward curves at convergence. Hyperparameters were inherited from past work \cite{Cederborg2015PolicySW} and are available in the Supplement in Sec.~\textbf{Training Parameters}. We extended the analysis to DQN~\cite{cobbe2019quantifying} and reported the results in the Supplement in Sec.~\textbf{DQN}. The experiments were conducted on a system with an Intel(R) Xeon(R) CPU E5-2683 v4 @ 2.10GHz.
\section{Results}
We report findings from the Generalization and Learnability agents trained with the multiple variations of PacMan, Pong, and Breakout as described in Sec.~\textbf{Generating MDPs for investigating generalization} and Table \ref{table:protocol}.

\begin{figure}[hbt!]
\centering
        \includegraphics[width=0.37\textwidth]{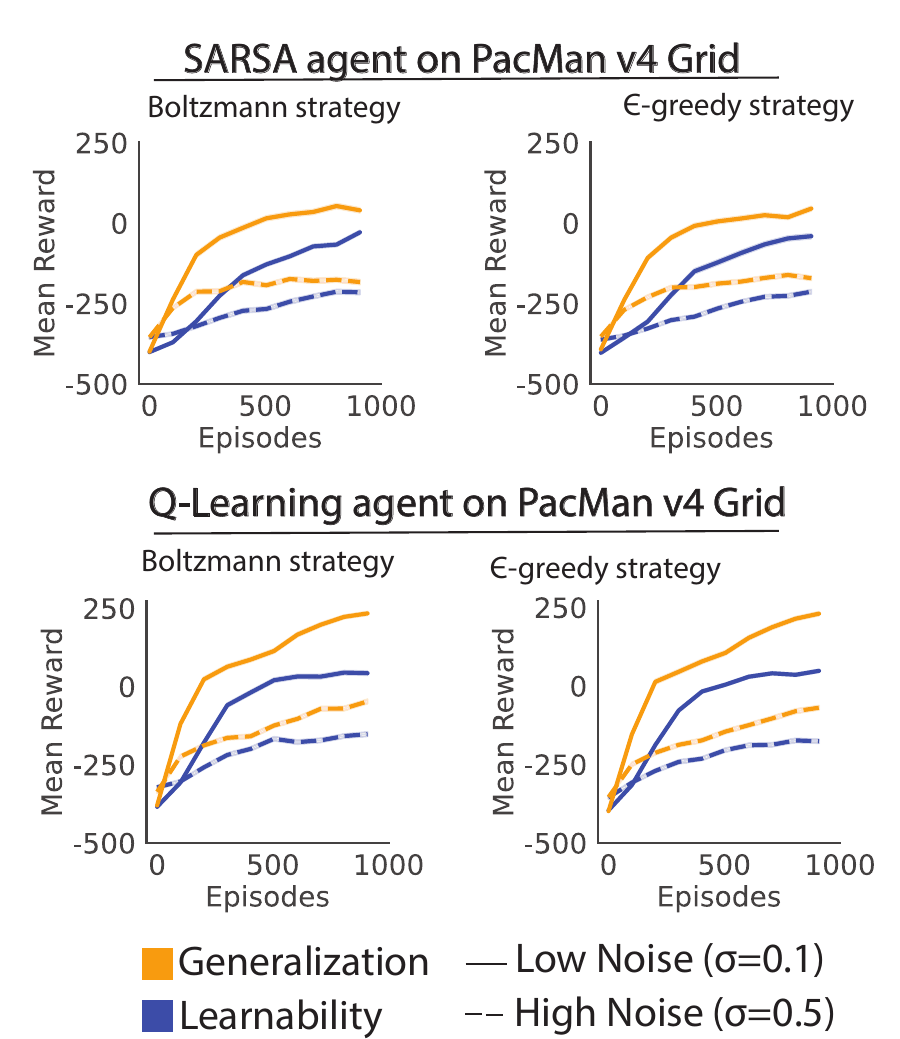}
        \caption{ 
        \textit{Generalization agents can outperform Learnability agents.} Results for PacMan v4 grid reporting mean reward as a function of episode number. (a) SARSA agent trained with a Boltzmann exploration strategy for Target MDPs generated with both high (solid line) and low (line with `x' markers) level noise injection. The Generalization Agent (red) beats the Learnability Agent (green) (two-sided t-test, p$<$0.001). (b) The same result holds for a SARSA agent trained with the $\epsilon-$greedy exploration strategy. This finding also holds for Q-Learning agents trained with (c) Boltzmann and (d) $\epsilon-$greedy exploration strategies. 
        %Noise added to the transition function is sampled $\delta \sim \mathcal{N}(0,0.1)$ in Low-Noise and $\delta \sim \mathcal{N}(0,0.5)$ in High-Noise settings. 
        Standard deviation across the 500 agents is reported as the error bar in all figures. However, the standard deviation is too small for these error bars to be visible. 
        }
        \label{fig:fig_sarsa_q_learning}
\end{figure}

\begin{figure}[htb!]
\centering
        \includegraphics[width=0.47\textwidth]{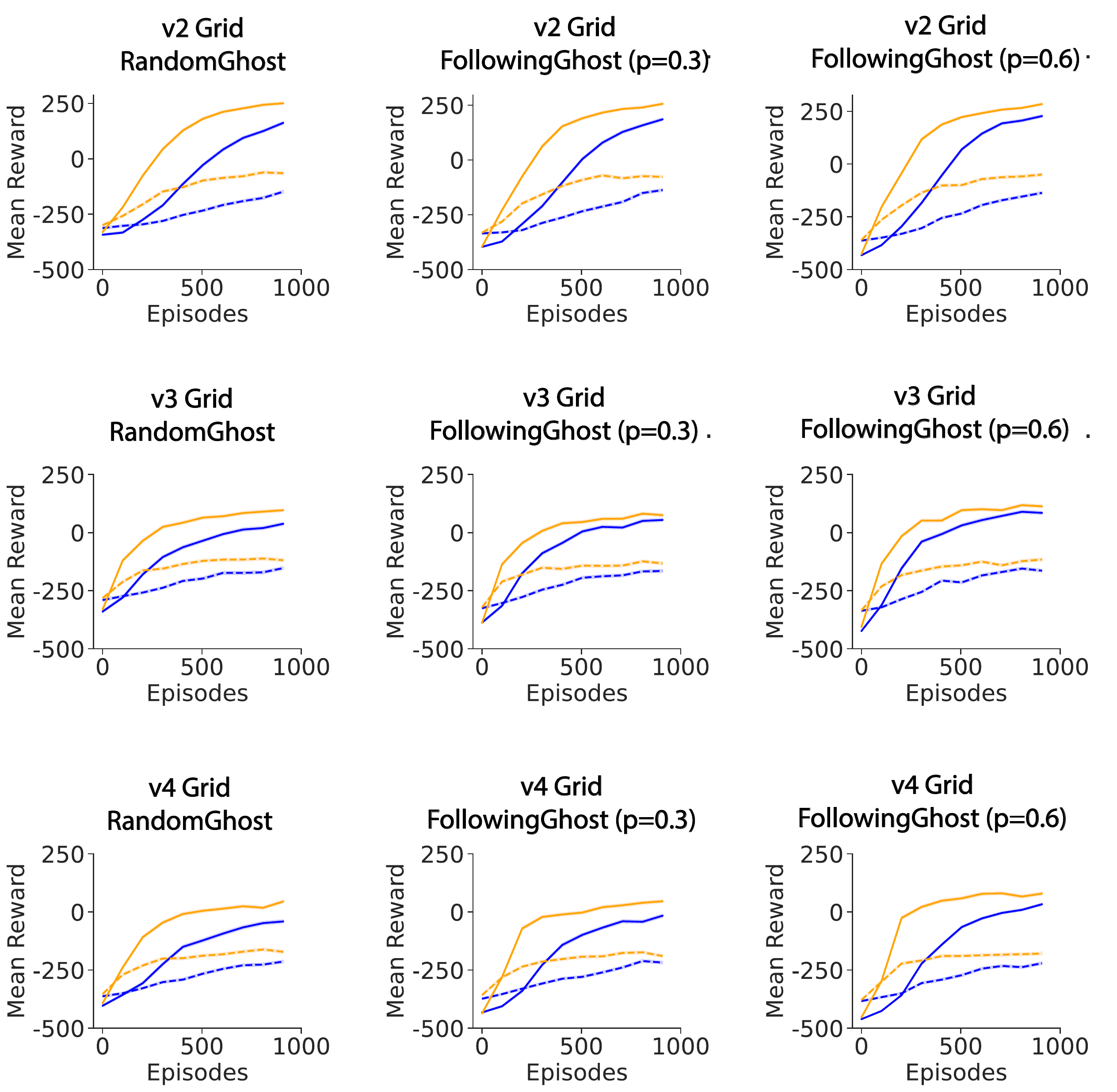} % Reduce the figure size so that it is slightly narrower than the column.
        \caption{
        \textit{Generalization can outperform Learnability across multiple variations of PacMan.}
        Format and conventions as in Fig.~\ref{fig:fig_sarsa_q_learning}.
        (a) Agents trained on the PacMan v2 grid with the Ghost dynamics set to the RandomGhost setting.
        %---the ghost picks a move from all possible legal moves with an equal probability. 
        (b) Agents trained on v2 with a DirectionalGhost with $p=0.3$.
        %---ghost has probability of $0.3$ to move in a pre-specified direction (here \textit{right}), and a probability of $0.7$ of picking one of the remaining moves (each remaining move equally likely). 
        (c) DirectionalGhost with $p=0.6$. (d),(e),(f) Variations with the v3 grid with RandomGhost, DirectionGhost ($p=0.3$) and DirectionalGhost ($p=0.6$), respectively. All experiments are shown for SARSA agents trained with the $\epsilon-$greedy exploration strategy. Generalization agents consistently beat Learnability Agents (two-sided t-test, p$<$0.001).
        Corresponding results for agents trained with SARSA + Boltzmann exploration strategy, and for Q-Learning with both $\epsilon-$greedy and Boltzmann exploration strategies are shown in Figures ~\ref{fig:atari_variations-pacman-sarsa-boltzmann}\text{-}~\ref{fig:atari_variations-pacman-qlearning-egreedy}.        
        }
        \label{fig:fig_pacman_variations}
\end{figure}

\begin{figure}[hbt!]
\centering
        \includegraphics[width=0.4\textwidth]{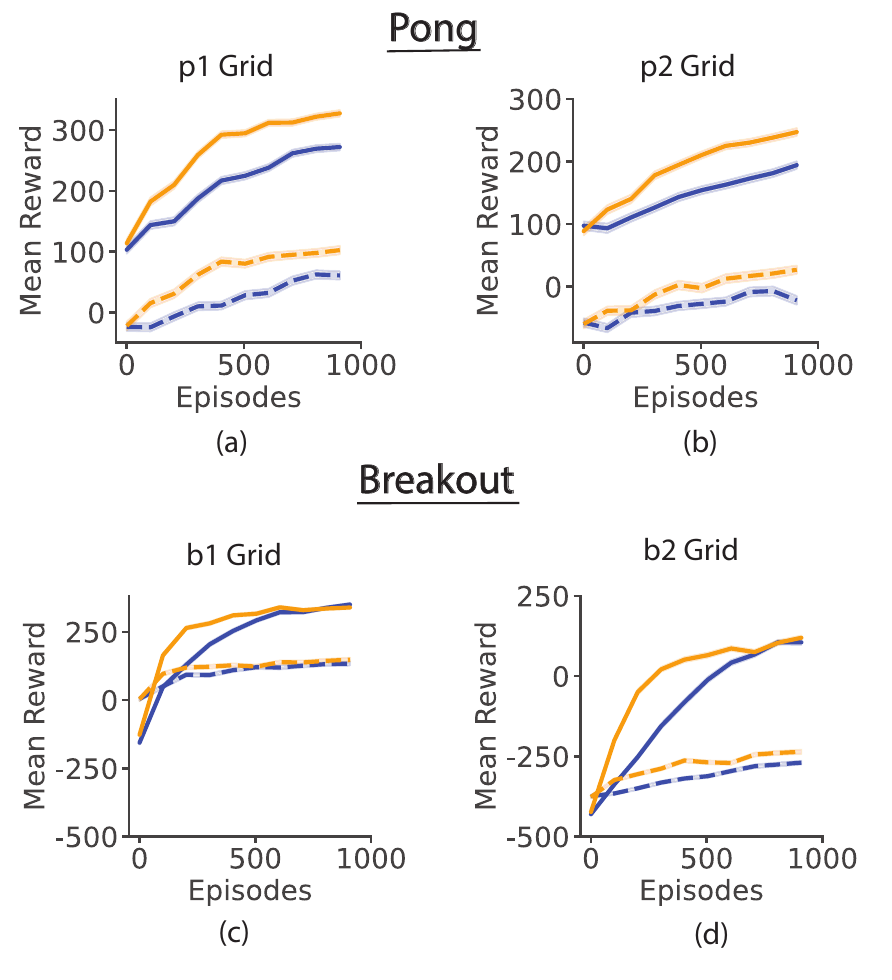}
        \caption{\textit{Generalization agents outperform Learnability agents on Pong and Breakout as well.}         
        Format and conventions as in Fig.~\ref{fig:fig_sarsa_q_learning}.
Performance of SARSA agents trained with an $\epsilon-$greedy exploration strategy on (a) Pong p1 grid, (b) Pong p2 grid, (c) Breakout b1 grid, and (d) Breakout b2 grid. The Generalization Agent consistently beats the Learnability Agent (two-sided t-test, p$<$0.001).}
        \label{fig:fig_pong_breakout}
\end{figure}

\begin{figure}[hbt!]
\centering
        \includegraphics[width=0.38\textwidth]{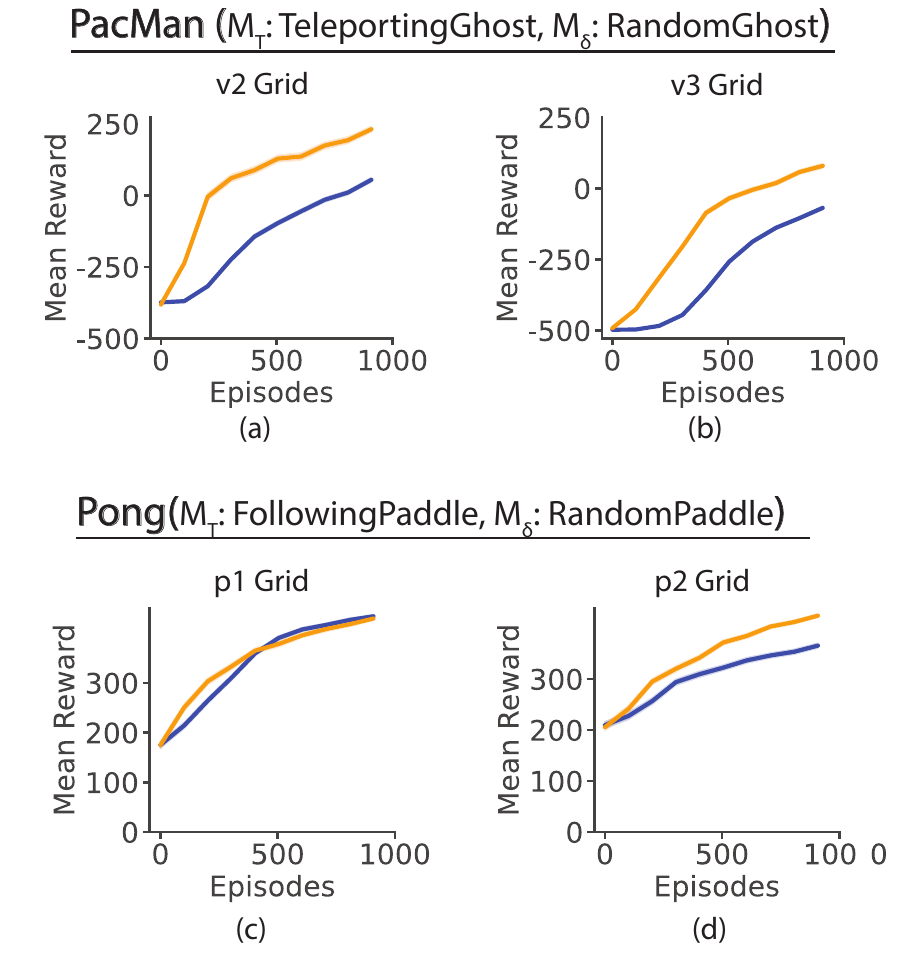}
        \caption{\textit{Generalization agents outperform Learnability agents on semantic variations of PacMan and Pong as well.} Format and conventions as in Fig.~\ref{fig:fig_sarsa_q_learning}. (a) Given the target PacMan MDP with the v2 grid and TeleportingGhost, the Generalization trained on the RandomGhost outperformed the Learnability agent that was trained and tested on the same Target MDP (TeleportingGhost) (two-sided t-test, p$<$0.001). (b) This finding extends to TeleportingGhost and RandomGhost MDPs with the PacMan v3 Grid as well. (c) For the Pong p1 grid, Generalization agents trained on an MDP with DirectionalPaddle performed better on the RandomPaddle MDP during testing, as compared to the Learnability Agent trained and tested on the RandomPaddle MDP. (d) The same finding extends to the p2 grid as well.
        }
        \label{fig:fig_semantically_interpretable}
\end{figure}

\begin{figure}[hbt!]
\centering        \includegraphics[width=0.40\textwidth]{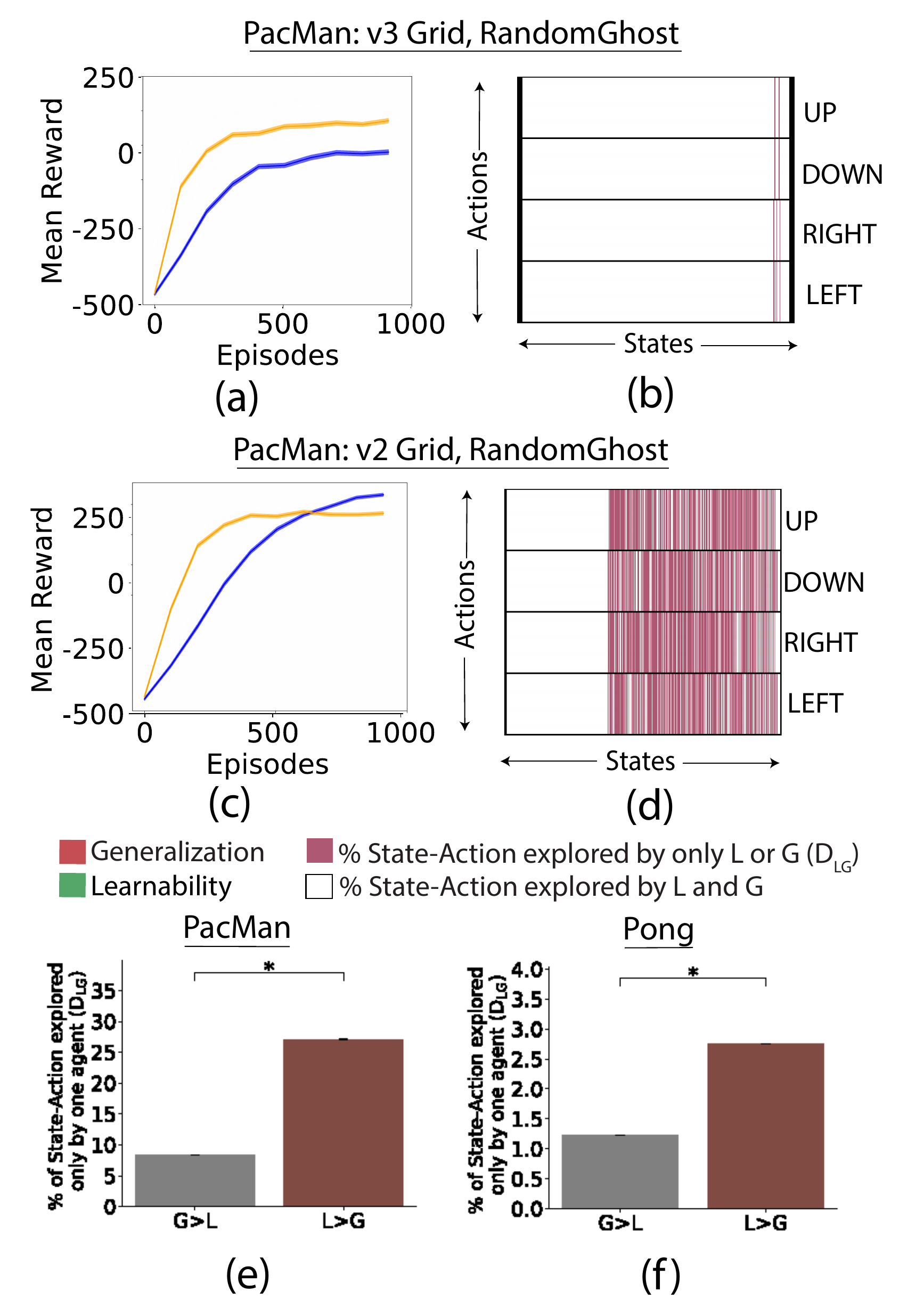} 
        \caption{\textit{The exploration patterns predict the reward gap between $\mathcal{L}_\delta$ and $\mathcal{G}_T$}. (a) Reward for agents trained on Pacman v3, where $\mathcal{G}_T$ outperforms $\mathcal{L}_\delta$ (format as in Fig.~\ref{fig:fig_sarsa_q_learning}). (b) \textit{Exploration grid} visualizing the difference in State-Action ($\mathcal{S}$-$\mathcal{A}$) pairs explored by these agents ($D_{LG}$). The grid shows Sates on the x-axis and Actions on the y-axis. The black lines separate the Actions for clarity. Each cell corresponds to one $\mathcal{S}$-$\mathcal{A}$ pair.
        In this case, a negligible fraction of $\mathcal{S}$-$\mathcal{A}$ pairs were visited only by one agent (pink).
        (c) Rewards for agents trained on PacMan v3. Here, $\mathcal{G}_T$ performs worse than $\mathcal{L}_\delta$ at the end of training. (d) A large fraction of pairs were only visited by either one or the other agent but not both (contrast with part (b)). (e) $D_{LG}$ averaged over PacMan grids where $\mathcal{G}_T$ outperformed $\mathcal{L}_\delta$ (gray) and vice-versa (brown) %\GK{are those the colors?}. 
        (f) $D_{LG}$ averaged over Pong grids. The "*" is for statistical significance (two-sided t-test, p$<$0.001). %\GK{Explain what the error bars are. Explain what the asterisk is. }
        %\GK{y-axis in e, f not clear. Do you mean \% that the agent explored? What about '\% S-A explored by agent'? Also, the text makes it sound like these are proportions not percentages. For example, you say in the text that they add up to 1, not 100.}
        %\GK{I am not sure how convincing it is that L > G in (c). Are there cases where the green line is mostly above red or where the gap is larger?} 
        }
        \label{fig:fig_explaining_behaviour}
\end{figure}

\subsection*{Generalization agents can outperform Learnability agents in several instances of the \emph{Indoor-Training Effect}}

The mean reward increased with training, as expected, (Fig.~\ref{fig:fig_sarsa_q_learning}a), both for the Generalization agent (red) and for the Learnability agent (green). Also, as intuitively expected, 
%agents trained on the noise-free worlds and tested on the $\delta-$environments achieved higher performance under conditions of low noise compared to high noise (Fig.~\ref{fig:fig_sarsa_q_learning}, compare solid lines versus lines with `x' markers). 
both agents performed better under low-noise conditions (solid lines) compared to high-noise conditions (lines with '\text{--}' markers).
Less intuitive was the relationship between Generalization and Learnability agents. 
Intriguingly, the Generalization agent consistently outperformed the Learnability agent (two-sided t-test, p$<$0.001). This gap continued until convergence at $1,000$ episodes, was observed both across low and high noise levels (solid lines versus '\text{--}' lines), when using a Boltzmann strategy (Fig.~\ref{fig:fig_sarsa_q_learning}a, c) or an $\epsilon-$greedy strategy (Fig.~\ref{fig:fig_sarsa_q_learning}b, d), and when using  SARSA agents (Fig.~\ref{fig:fig_sarsa_q_learning}a, b) or 
Q-Learning agents (Fig.~\ref{fig:fig_sarsa_q_learning}c, d). Another metric commonly used to assess performance is the \emph{Area Under the Curve} (AUC). We compute the ratio between the learnability and generalization agents' AUC and normalize it using regret. (see Supp. Sec.~\textbf{Regret Normalization}).

To assess whether this observation was dependent on the target MDP, we replicated these findings on multiple PacMan grids and noise variations (Fig.~\ref{fig:fig_pacman_variations}a-f).
In (Fig.~\ref{fig:fig_pacman_variations}), the Generalization agents beat the Learnability agents, for both low and high levels of noise (see Supp. Sec.~\textbf{Additional Graphs Non-Semantic Variations}: Figs. ~\ref{fig:atari_variations-pacman-sarsa-boltzmann}\text{-}~\ref{fig:atari_variations-pacman-qlearning-egreedy} for Boltzmann strategy and Q-learning results). 

We also extended these findings to two additional ATARI games, Pong Fig.~\ref{fig:atari_variations-pong} and Breakout Fig.~\ref{fig:atari_variations-breakout}, to assess their applicability across different games (Fig.~\ref{fig:fig_pong_breakout}). Consistent with the results described for Pacman, the Generalization agent was on par with or better than the Learnability agent in Pong Fig.~\ref{fig:fig_pong_breakout}a,b) and Breakout Fig.~\ref{fig:fig_pong_breakout}c,d) (two-sided t-test, p$<$0.001; see Figs.~\ref{fig:atari_variations-pong-sarsa-boltzmann}\textit{-}~\ref{fig:atari_variations-breakout-qlearning-egreedy} for results with Q-Learning, Sarsa and different sampling strategies).

In sum, there exist several MDPs where it is better to train on a different MDP than the target. 
%To the best of our knowledge, 
These results provide novel intriguing evidence suggesting that training on a different MDP can enable more efficient policy learning than training on the target environment. 

\subsection*{Instances of the \emph{Indoor-Training Effect} in semantic variations of ATARI games}

The results presented so far focused on altered MDPs generated by noise injection. Next, we evaluated \emph{semantic} variations, where the changes are more meaningful and interpretable. Specifically, we modified the transition probabilities of Pacman so that ghosts could teleport to new locations, and Pong so that the paddle could follow the ball. We refer to these alternate semantic environments as $\mathcal{M}_{T'}$ (semantic noise), in contrast to $\mathcal{M}_\delta$ used for noise injection.

%We further confirmed that analogous trends also hold true for semantic variations of two ATARI Games (see Sec.~\ref{subsec:semantic_variations} \GK{section reference missing}). 

For PacMan, Learnability agents were trained and tested using TeleportingGhosts ($\mathcal{M}_{T'}$), while the Generalization agents were trained with PacMan with RandomGhosts ($\mathcal{M}_{T}$) and then tested on TeleportingGhosts ($\mathcal{M}_{T'}$) (Fig.~\ref{fig:fig_semantically_interpretable}a, b, Supp. Sec.~\textbf{Additional Graphs Non-Semantic Variations}). 
Even under these semantic noise conditions, Generalization agents outperformed Learnability agents  (two-sided t-test, p$<$0.001; see
Figs. \ref{fig:atari_variations-semantic-pacman-sarsa-boltzmann}-\ref{fig:atari_variations-semantic-pacman-qlearning-egreedy} for results with Q-Learning, Sarsa and different sampling strategies).
%These results provide compelling evidence suggesting these findings extend beyond the noise-variations.
In the case of Pong, we report analogous results with $\mathcal{M}_{T'}$ set to FollowingPaddle, and $\mathcal{M}_{T}$ set to RandomPaddle. Generalization agents also outperformed Learnability agents (by a smaller margin) in both the p1 and p2 grids (Fig.~\ref{fig:fig_semantically_interpretable}c,d). 
%These results are for SARSA agents trained with $\epsilon-$greedy exploration strategy. 
Analogous results for Pong with Q-Learning and other exploration strategies are reported in Figs.~\ref{fig:atari_variations-semantic-breakout-sarsa-boltzmann_0.3}\text{-}~\ref{fig:atari_variations-semantic-pong-qlearning-egreedy}.

\subsection*{The exploration patterns of state-action pairs can predict differences between Generalization and Learning agents} 

%So far, we have presented comprehensive analyses with several examples of our main finding---Generalization Agents beating Learnability Agents. Here, we present a potential explanation for this counter-intuitive phenomenon. 

To better understand how Generalization agents could outperform Learnability agents, we investigated the exploration patterns for $\mathcal{L}_\delta$ and
$\mathcal{G}_T$. We enumerated all State ($\mathcal{S}$)-Action ($\mathcal{A}$) Pairs, and divided them into three groups---(i) Percentage of $\mathcal{S}$-$\mathcal{A}$ pairs explored by both agents ($P_{LG}$), (ii) Percentage of pairs explored only by the Learnability agent ($P_L$), and (iii) Percentage of pairs explored only by the Generalization agent ($P_G)$. Thus, $P_{LG} + P_L + P_G = 100$. We defined $D_{LG} = P_L + P_G$, the divergence in the exploration patterns between these two agents. 

In Fig.~\ref{fig:fig_explaining_behaviour} we visualize $D_{LG}$ for grids where $\mathcal{G}_T$ outperformed $\mathcal{L}_\delta$  agents and compare it to cases where it did not. Fig.~\ref{fig:fig_explaining_behaviour}a shows an agent trained with Q-Learning and Boltzmann exploration strategy for the Pacman v3 grid with RandomGhost stochasticity, where $\mathcal{G}_T$ beat the $\mathcal{L}_\delta$ agent. The corresponding panel Fig.~\ref{fig:fig_explaining_behaviour}b depicts $D_{LG}$ ---each entry of this grid represents an $\mathcal{S}$-$\mathcal{A}$ pair. We refer to this plot as the \textit{exploration grid} for these agents. 
The exploration grid shows that most $\mathcal{S}$-$\mathcal{A}$ pairs were explored by both agents, with almost no pairs explored only by one type of agent and therefore no significant differences in their exploration patterns. In contrast, Fig.~\ref{fig:fig_explaining_behaviour}c, d report $\mathcal{S}$-$\mathcal{A}$ pairs for PacMan v2, here the $\mathcal{G}_T$ agent performs worse than the $\mathcal{L}_\delta$ agent. The exploration grid reveals that there is a high fraction of $\mathcal{S}$-$\mathcal{A}$ pairs explored either by one \emph{or} the other agent but not both. 
%These figures depict the relationship between $D_{LG}$ and the Reward Gap between the two agents which we denote as $R_{LG} = R_L - R_G$.

%\GK{I changed most of the signs here because it was the other way around.}
%\GK{I am still confused about this. G>L is gray. L>G is brown. }\SB{YES! Sorry for the confusion, I changed it!}
We grouped all the cases where $\mathcal{L}_\delta > \mathcal{G}_T$ (
Fig.~\ref{fig:fig_explaining_behaviour}e, brown) and all the cases where 
$\mathcal{L}_\delta < \mathcal{G}_T$ (
Fig.~\ref{fig:fig_explaining_behaviour}e, gray) and computed $D_{LG}$. On average, $D_{LG}$ was significantly higher in MDPs where $\mathcal{L}_\delta > \mathcal{G}_T$  
%\GK{\textbf{Isn't this the other way around???}},than when it did not 
(two-sided t-test, $p<0.05$). The same result holds true for Pong MDPs as reported in Fig.~\ref{fig:fig_explaining_behaviour}f. Exploration grids and additional results for variations of PacMan (\ref{fig:atari_variations-exploration-pacman-qlearning-boltzmann}-\ref{fig:atari_variations-exploration-semantic-pacman-sarsa-egreedy}), Pong (\ref{fig:atari_variations-exploration-pong-qlearning-boltzmann}-\ref{fig:atari_variations-exploration-semantic-pong-sarsa-egreedy}), and Breakout (\ref{fig:atari_variations-exploration-breakout-qlearning-boltzmann}-\ref{fig:atari_variations-exploration-breakout-sarsa-egreedy}) can be found in the Supplement in Sec.~\textbf{Additional Graphs State-Action Pairs}.
Instead of grouping MDPs, we also conducted a correlation analysis. 
We defined the Reward Gap: $R_{LG} = R_G - R_L$. The Spearman correlation coefficient between $D_{LG}$ and $R_{LG}$ was $0.43$ 
 ($p<0.005$) for PacMan and $0.26$ ($p<0.005$) for Pong. Combined, these analyses show that the \emph{Indoor-Training Effect} is associated with similar exploration patterns in the training and testing environments.

 \section*{Discussion}

In this work, our objective is to understand the paradoxical \emph{Indoor-Training Effect} - where agents perform better when trained in a noise-free environment and tested in noisy $\delta$-environments, compared to being trained and tested in the same $\delta$-environments. Similarly to how training in a quiet, noise-free indoor environment helps athletes focus on mastering the fundamentals of tennis, we explore whether training in certain environments is more conducive to learning than training on the same testing environment.

To investigate this, we propose a new methodology to generate modified MDPs from a given MDP, along with a metric to quantify the distance between different environments. We demonstrate the Indoor-Training Effect across various algorithms and exploration strategies (Fig.~\ref{fig:fig_sarsa_q_learning}), grid layouts and game stochasticity (Fig.~\ref{fig:fig_pacman_variations}), and multiple ATARI games (Fig.~\ref{fig:fig_pong_breakout}). We also showed that this phenomenon extends beyond Noise Injected environments, and can also occur when semantic changes are introduced in the game elements (Fig.~\ref{fig:fig_semantically_interpretable}).

To gain deeper insights into these environments, we examine the exploration patterns of agents under different transition probabilities. Similarly to a tennis player who has never encountered a smash serve during their training and develops an optimal playing style that does not anticipate or respond to such powerful shots, the suboptimal performance of the agents could be caused by a divergence in exploration patterns. We show that the performance gap between agents is indeed correlated with their exploration patterns under different transition probabilities (Fig.~\ref{fig:fig_explaining_behaviour}). 

The Indoor-Training Effect is particularly relevant to robotics, where robots often operate in complex, dynamic environments. The Indoor-Training Effect opens new avenues of research, whereby robotic systems could be trained in simplified, controlled settings to master essential skills without the interference of noise. This finding could also enhance their ability to adapt and perform in real-world conditions where unpredictability and noise are prevalent. Such training strategies could lead to more robust, adaptable robots capable of navigating and executing tasks effectively in diverse and challenging environments. 

%Another interesting avenue for future research is that of modifying training environments to induce emergent behaviours conducive to learning. By interacting with the environment the agent could be biased towards exploring states that serve as good testbeds for learning skills that would eventually ease generalization. 

We note that these findings are reminiscent of results with biological agents. For example, recent experiments with the \emph{C. Elegans} worm have shown that biological agents perform best when cross-trained on different environments as compared to being tested on the environments they were trained on~\cite{chenguangworms}. 

Despite the evidence provided in this study, we would like to highlight two main limitations. Firstly, our experiments were conducted solely in the context of ATARI games. We hope that future research can extend and examine the findings in real-world environments. Secondly, it will be interesting to assess whether the conclusions drawn from classical Reinforcement Learning methods extend to deep RL approaches.

These findings raise fundamental questions about our understanding of RL algorithms. Typically, RL practitioners have strived to train agents in environments that closely resemble their deployment conditions. This approach assumes that matching the training and testing environments is critical for optimal performance. However, the Indoor-Training Effect challenges this assumption by showing that agents trained in noise-free, controlled environments can sometimes outperform those trained in more chaotic, realistic settings when faced with noisy, unpredictable scenarios during testing. 

\bibliography{aaai25} % Ensure 'aaai25' matches your .bib filename

\begin{thebibliography}{44}
\providecommand{\natexlab}[1]{#1}

\bibitem[{Ahmed et~al.(2020)Ahmed, Tr{\"{a}}uble, Goyal, Neitz, W{\"{u}}thrich, Bengio, Sch{\"{o}}lkopf, and Bauer}]{DBLP:journals/corr/abs-2010-04296}
Ahmed, O.; Tr{\"{a}}uble, F.; Goyal, A.; Neitz, A.; W{\"{u}}thrich, M.; Bengio, Y.; Sch{\"{o}}lkopf, B.; and Bauer, S. 2020.
\newblock CausalWorld: {A} Robotic Manipulation Benchmark for Causal Structure and Transfer Learning.
\newblock \emph{CoRR}, abs/2010.04296.

\bibitem[{Bapst et~al.(2019)Bapst, Sanchez{-}Gonzalez, Doersch, Stachenfeld, Kohli, Battaglia, and Hamrick}]{DBLP:journals/corr/abs-1904-03177}
Bapst, V.; Sanchez{-}Gonzalez, A.; Doersch, C.; Stachenfeld, K.~L.; Kohli, P.; Battaglia, P.~W.; and Hamrick, J.~B. 2019.
\newblock Structured agents for physical construction.
\newblock \emph{CoRR}, abs/1904.03177.

\bibitem[{Barbu et~al.(2019)Barbu, Mayo, Alverio, Luo, Wang, Gutfreund, Tenenbaum, and Katz}]{barbu2019objectnet}
Barbu, A.; Mayo, D.; Alverio, J.; Luo, W.; Wang, C.; Gutfreund, D.; Tenenbaum, J.; and Katz, B. 2019.
\newblock Objectnet: A large-scale bias-controlled dataset for pushing the limits of object recognition models.
\newblock \emph{Advances in neural information processing systems}, 32.

\bibitem[{B{\"a}uerle and Glauner(2022)}]{bauerle2022distributionally}
B{\"a}uerle, N.; and Glauner, A. 2022.
\newblock Distributionally robust Markov decision processes and their connection to risk measures.
\newblock \emph{Mathematics of Operations Research}, 47(3): 1757--1780.

\bibitem[{Biedenkapp et~al.(2020)Biedenkapp, Bozkurt, Eimer, Hutter, and Lindauer}]{biedenkapp-ecai20}
Biedenkapp, A.; Bozkurt, H.~F.; Eimer, T.; Hutter, F.; and Lindauer, M. 2020.
\newblock Dynamic Algorithm Configuration: Foundation of a New Meta-Algorithmic Framework.
\newblock In \emph{Proceedings of the Twenty-fourth European Conference on Artificial Intelligence ({ECAI}'20)}.

\bibitem[{Bomatter et~al.(2021)Bomatter, Zhang, Karev, Madan, Tseng, and Kreiman}]{bomatter2021pigs}
Bomatter, P.; Zhang, M.; Karev, D.; Madan, S.; Tseng, C.; and Kreiman, G. 2021.
\newblock When Pigs Fly: Contextual Reasoning in Synthetic and Natural Scenes.
\newblock arXiv:2104.02215.

\bibitem[{Cederborg et~al.(2015)Cederborg, Grover, Isbell, and Thomaz}]{Cederborg2015PolicySW}
Cederborg, T.; Grover, I.; Isbell, C.~L.; and Thomaz, A.~L. 2015.
\newblock Policy Shaping with Human Teachers.
\newblock In \emph{International Joint Conference on Artificial Intelligence}.

\bibitem[{Cobbe et~al.(2020)Cobbe, Hesse, Hilton, and Schulman}]{cobbe2020leveraging}
Cobbe, K.; Hesse, C.; Hilton, J.; and Schulman, J. 2020.
\newblock Leveraging Procedural Generation to Benchmark Reinforcement Learning.
\newblock arXiv:1912.01588.

\bibitem[{Cobbe et~al.(2019)Cobbe, Klimov, Hesse, Kim, and Schulman}]{cobbe2019quantifying}
Cobbe, K.; Klimov, O.; Hesse, C.; Kim, T.; and Schulman, J. 2019.
\newblock Quantifying generalization in reinforcement learning.
\newblock In \emph{International conference on machine learning}, 1282--1289. PMLR.

\bibitem[{DeNero, Klein, and Abbeel(2014)}]{berkeleypacproj}
DeNero, J.; Klein, D.; and Abbeel, P. 2014.
\newblock CS188: Berkeley Pacman Projects.
\newblock {http://ai.berkeley.edu/home.html} (Spring 2014).

\bibitem[{Devin et~al.(2018)Devin, Abbeel, Darrell, and Levine}]{devin2018deep}
Devin, C.; Abbeel, P.; Darrell, T.; and Levine, S. 2018.
\newblock Deep object-centric representations for generalizable robot learning.
\newblock In \emph{2018 IEEE International Conference on Robotics and Automation (ICRA)}, 7111--7118. IEEE.

\bibitem[{Dulac{-}Arnold, Mankowitz, and Hester(2019)}]{DBLP:journals/corr/abs-1904-12901}
Dulac{-}Arnold, G.; Mankowitz, D.~J.; and Hester, T. 2019.
\newblock Challenges of Real-World Reinforcement Learning.
\newblock \emph{CoRR}, abs/1904.12901.

\bibitem[{Filos et~al.(2020)Filos, Tigas, McAllister, Rhinehart, Levine, and Gal}]{DBLP:journals/corr/abs-2006-14911}
Filos, A.; Tigas, P.; McAllister, R.; Rhinehart, N.; Levine, S.; and Gal, Y. 2020.
\newblock Can Autonomous Vehicles Identify, Recover From, and Adapt to Distribution Shifts?
\newblock \emph{CoRR}, abs/2006.14911.

\bibitem[{Geirhos et~al.(2018)Geirhos, Rubisch, Michaelis, Bethge, Wichmann, and Brendel}]{geirhos2018imagenet}
Geirhos, R.; Rubisch, P.; Michaelis, C.; Bethge, M.; Wichmann, F.~A.; and Brendel, W. 2018.
\newblock ImageNet-trained CNNs are biased towards texture; increasing shape bias improves accuracy and robustness.
\newblock \emph{arXiv preprint arXiv:1811.12231}.

\bibitem[{Gopalan et~al.(2017)Gopalan, Littman, MacGlashan, Squire, Tellex, Winder, Wong et~al.}]{gopalan2017planning}
Gopalan, N.; Littman, M.; MacGlashan, J.; Squire, S.; Tellex, S.; Winder, J.; Wong, L.; et~al. 2017.
\newblock Planning with abstract Markov decision processes.
\newblock In \emph{Proceedings of the International Conference on Automated Planning and Scheduling}, volume~27, 480--488.

\bibitem[{Goyal and Grand-Clement(2023)}]{goyal2023robust}
Goyal, V.; and Grand-Clement, J. 2023.
\newblock Robust Markov decision processes: Beyond rectangularity.
\newblock \emph{Mathematics of Operations Research}, 48(1): 203--226.

\bibitem[{Guss et~al.(2019)Guss, Codel, Hofmann, Houghton, Kuno, Milani, Mohanty, Liebana, Salakhutdinov, Topin et~al.}]{guss2019minerl}
Guss, W.~H.; Codel, C.; Hofmann, K.; Houghton, B.; Kuno, N.; Milani, S.; Mohanty, S.; Liebana, D.~P.; Salakhutdinov, R.; Topin, N.; et~al. 2019.
\newblock The MineRL 2019 competition on sample efficient reinforcement learning using human priors.
\newblock \emph{arXiv preprint arXiv:1904.10079}.

\bibitem[{Hafner(2021)}]{DBLP:journals/corr/abs-2109-06780}
Hafner, D. 2021.
\newblock Benchmarking the Spectrum of Agent Capabilities.
\newblock \emph{CoRR}, abs/2109.06780.

\bibitem[{Justesen et~al.(2018)Justesen, Torrado, Bontrager, Khalifa, Togelius, and Risi}]{justesen2018illuminating}
Justesen, N.; Torrado, R.~R.; Bontrager, P.; Khalifa, A.; Togelius, J.; and Risi, S. 2018.
\newblock Illuminating generalization in deep reinforcement learning through procedural level generation.
\newblock \emph{arXiv preprint arXiv:1806.10729}.

\bibitem[{Kaelbling, Littman, and Moore(1996)}]{Kaelbling1996ReinforcementLA}
Kaelbling, L.~P.; Littman, M.~L.; and Moore, A.~W. 1996.
\newblock Reinforcement Learning: A Survey.
\newblock \emph{J. Artif. Intell. Res.}, 4: 237--285.

\bibitem[{Kang et~al.(2019)Kang, Belkhale, Kahn, Abbeel, and Levine}]{kang2019generalization}
Kang, K.; Belkhale, S.; Kahn, G.; Abbeel, P.; and Levine, S. 2019.
\newblock Generalization through simulation: Integrating simulated and real data into deep reinforcement learning for vision-based autonomous flight.
\newblock In \emph{2019 international conference on robotics and automation (ICRA)}, 6008--6014. IEEE.

\bibitem[{Kirk et~al.(2021)Kirk, Zhang, Grefenstette, and Rockt{\"{a}}schel}]{DBLP:journals/corr/abs-2111-09794}
Kirk, R.; Zhang, A.; Grefenstette, E.; and Rockt{\"{a}}schel, T. 2021.
\newblock A Survey of Generalisation in Deep Reinforcement Learning.
\newblock \emph{CoRR}, abs/2111.09794.

\bibitem[{Li, Kreiman, and Ramanathan(2024)}]{chenguangworms}
Li, C.; Kreiman, G.; and Ramanathan, S. 2024.
\newblock Discovering neural policies to drive behavior by integrating deep reinforcement learning agents with biological neural networks.
\newblock \emph{Nature Machine Intelligence}, In Press.

\bibitem[{Lyle et~al.(2022)Lyle, Rowland, Dabney, Kwiatkowska, and Gal}]{lyle2022learning}
Lyle, C.; Rowland, M.; Dabney, W.; Kwiatkowska, M.; and Gal, Y. 2022.
\newblock Learning dynamics and generalization in deep reinforcement learning.
\newblock In \emph{International Conference on Machine Learning}, 14560--14581. PMLR.

\bibitem[{Madan et~al.(2022{\natexlab{a}})Madan, Henry, Dozier, Ho, Bhandari, Sasaki, Durand, Pfister, and Boix}]{madan2022and}
Madan, S.; Henry, T.; Dozier, J.; Ho, H.; Bhandari, N.; Sasaki, T.; Durand, F.; Pfister, H.; and Boix, X. 2022{\natexlab{a}}.
\newblock When and how convolutional neural networks generalize to out-of-distribution category--viewpoint combinations.
\newblock \emph{Nature Machine Intelligence}, 4(2): 146--153.

\bibitem[{Madan et~al.(2023)Madan, Sasaki, Pfister, Li, and Boix}]{madan2023adversarial}
Madan, S.; Sasaki, T.; Pfister, H.; Li, T.-M.; and Boix, X. 2023.
\newblock Adversarial examples within the training distribution: A widespread challenge.
\newblock arXiv:2106.16198.

\bibitem[{Madan et~al.(2022{\natexlab{b}})Madan, You, Zhang, Pfister, and Kreiman}]{madan2022makes}
Madan, S.; You, L.; Zhang, M.; Pfister, H.; and Kreiman, G. 2022{\natexlab{b}}.
\newblock What makes domain generalization hard?
\newblock arXiv:2206.07802.

\bibitem[{Michaelis et~al.(2019)Michaelis, Mitzkus, Geirhos, Rusak, Bringmann, Ecker, Bethge, and Brendel}]{michaelis2019benchmarking}
Michaelis, C.; Mitzkus, B.; Geirhos, R.; Rusak, E.; Bringmann, O.; Ecker, A.~S.; Bethge, M.; and Brendel, W. 2019.
\newblock Benchmarking robustness in object detection: Autonomous driving when winter is coming.
\newblock \emph{arXiv preprint arXiv:1907.07484}.

\bibitem[{Mondal, Dulberg, and Cohen(2022)}]{mondal2022generalization}
Mondal, S.~S.; Dulberg, Z.; and Cohen, J. 2022.
\newblock Generalization to Out-of-Distribution transformations.

\bibitem[{Moos et~al.(2022{\natexlab{a}})Moos, Hansel, Abdulsamad, Stark, Clever, and Peters}]{make4010013}
Moos, J.; Hansel, K.; Abdulsamad, H.; Stark, S.; Clever, D.; and Peters, J. 2022{\natexlab{a}}.
\newblock Robust Reinforcement Learning: A Review of Foundations and Recent Advances.
\newblock \emph{Machine Learning and Knowledge Extraction}, 4(1): 276--315.

\bibitem[{Moos et~al.(2022{\natexlab{b}})Moos, Hansel, Abdulsamad, Stark, Clever, and Peters}]{moos2022robust}
Moos, J.; Hansel, K.; Abdulsamad, H.; Stark, S.; Clever, D.; and Peters, J. 2022{\natexlab{b}}.
\newblock Robust reinforcement learning: A review of foundations and recent advances.
\newblock \emph{Machine Learning and Knowledge Extraction}, 4(1): 276--315.

\bibitem[{Nilim and El~Ghaoui(2005)}]{nilim2005robust}
Nilim, A.; and El~Ghaoui, L. 2005.
\newblock Robust control of Markov decision processes with uncertain transition matrices.
\newblock \emph{Operations Research}, 53(5): 780--798.

\bibitem[{OpenAI et~al.(2019)OpenAI, Akkaya, Andrychowicz, Chociej, Litwin, McGrew, Petron, Paino, Plappert, Powell, Ribas, Schneider, Tezak, Tworek, Welinder, Weng, Yuan, Zaremba, and Zhang}]{DBLP:journals/corr/abs-1910-07113}
OpenAI; Akkaya, I.; Andrychowicz, M.; Chociej, M.; Litwin, M.; McGrew, B.; Petron, A.; Paino, A.; Plappert, M.; Powell, G.; Ribas, R.; Schneider, J.; Tezak, N.; Tworek, J.; Welinder, P.; Weng, L.; Yuan, Q.; Zaremba, W.; and Zhang, L. 2019.
\newblock Solving Rubik's Cube with a Robot Hand.
\newblock \emph{CoRR}, abs/1910.07113.

\bibitem[{Osi{\'n}ski et~al.(2020)Osi{\'n}ski, Jakubowski, Zi{\k{e}}cina, Mi{\l}o{\'s}, Galias, Homoceanu, and Michalewski}]{osinski2020simulation}
Osi{\'n}ski, B.; Jakubowski, A.; Zi{\k{e}}cina, P.; Mi{\l}o{\'s}, P.; Galias, C.; Homoceanu, S.; and Michalewski, H. 2020.
\newblock Simulation-based reinforcement learning for real-world autonomous driving.
\newblock In \emph{2020 IEEE International Conference on Robotics and Automation (ICRA)}, 6411--6418. IEEE.

\bibitem[{Packer et~al.(2018)Packer, Gao, Kos, Kr{\"a}henb{\"u}hl, Koltun, and Song}]{packer2018assessing}
Packer, C.; Gao, K.; Kos, J.; Kr{\"a}henb{\"u}hl, P.; Koltun, V.; and Song, D. 2018.
\newblock Assessing generalization in deep reinforcement learning.
\newblock \emph{arXiv preprint arXiv:1810.12282}.

\bibitem[{Rummery and Niranjan(1994)}]{rummery1994line}
Rummery, G.~A.; and Niranjan, M. 1994.
\newblock \emph{On-line Q-learning using connectionist systems}, volume~37.
\newblock University of Cambridge, Department of Engineering Cambridge, UK.

\bibitem[{Sakai et~al.(2022)Sakai, Sunagawa, Madan, Suzuki, Katoh, Kobashi, Pfister, Sinha, Boix, and Sasaki}]{sakai2022three}
Sakai, A.; Sunagawa, T.; Madan, S.; Suzuki, K.; Katoh, T.; Kobashi, H.; Pfister, H.; Sinha, P.; Boix, X.; and Sasaki, T. 2022.
\newblock Three approaches to facilitate invariant neurons and generalization to out-of-distribution orientations and illuminations.
\newblock \emph{Neural Networks}, 155: 119--143.

\bibitem[{Squire et~al.(2015)Squire, Tellex, Arumugam, and Yang}]{squire2015grounding}
Squire, S.; Tellex, S.; Arumugam, D.; and Yang, L. 2015.
\newblock Grounding English commands to reward functions.
\newblock In \emph{Robotics: Science and Systems}.

\bibitem[{Stone et~al.(2021)Stone, Ramirez, Konolige, and Jonschkowski}]{stone2021distracting}
Stone, A.; Ramirez, O.; Konolige, K.; and Jonschkowski, R. 2021.
\newblock The Distracting Control Suite -- A Challenging Benchmark for Reinforcement Learning from Pixels.
\newblock arXiv:2101.02722.

\bibitem[{Tellex et~al.(2011)Tellex, Kollar, Dickerson, Walter, Banerjee, Teller, and Roy}]{tellex2011understanding}
Tellex, S.; Kollar, T.; Dickerson, S.; Walter, M.; Banerjee, A.; Teller, S.; and Roy, N. 2011.
\newblock Understanding natural language commands for robotic navigation and mobile manipulation.
\newblock In \emph{Proceedings of the AAAI Conference on Artificial Intelligence}, volume~25, 1507--1514.

\bibitem[{Walter et~al.(2013)Walter, Hemachandra, Homberg, Tellex, and Teller}]{walter2013learning}
Walter, M.~R.; Hemachandra, S.~M.; Homberg, B.~S.; Tellex, S.; and Teller, S. 2013.
\newblock Learning semantic maps from natural language descriptions.
\newblock Robotics: Science and Systems.

\bibitem[{Watkins and Dayan(1992)}]{Watkins1992}
Watkins, C. J. C.~H.; and Dayan, P. 1992.
\newblock Q-learning.
\newblock \emph{Machine Learning}, 8(3): 279--292.

\bibitem[{Zhu et~al.(2020)Zhu, Wong, Mandlekar, and Mart{\'{\i}}n{-}Mart{\'{\i}}n}]{DBLP:journals/corr/abs-2009-12293}
Zhu, Y.; Wong, J.; Mandlekar, A.; and Mart{\'{\i}}n{-}Mart{\'{\i}}n, R. 2020.
\newblock robosuite: {A} Modular Simulation Framework and Benchmark for Robot Learning.
\newblock \emph{CoRR}, abs/2009.12293.

\bibitem[{Zhu et~al.(2023)Zhu, Lin, Jain, and Zhou}]{zhu2023transfer}
Zhu, Z.; Lin, K.; Jain, A.~K.; and Zhou, J. 2023.
\newblock Transfer learning in deep reinforcement learning: A survey.
\newblock \emph{IEEE Transactions on Pattern Analysis and Machine Intelligence}.

\end{thebibliography}
\bibstyle{aaai}

\clearpage
\renewcommand{\thefigure}{Sup\arabic{figure}}
\renewcommand{\thetable}{Sup\arabic{table}}
\renewcommand{\theequation}{Sup\arabic{equation}}

\setcounter{figure}{0}
\setcounter{table}{0}
\setcounter{equation}{0}

\section*{Reproducibility Checklist}

\begin{itemize}
    \item \textbf{Includes a conceptual outline and/or pseudocode description of AI methods introduced:} Yes (preliminaries: Reinforcement Learning)
    
    \item \textbf{Clearly delineates statements that are opinions, hypothesis, and speculation from objective facts and results:} Yes (Introduction, Results)
    
    \item \textbf{Provides well-marked pedagogical references for less-familiar readers to gain background necessary to replicate the paper:} Yes (Preliminaries, Related Works)
    
    \item \textbf{Does this paper make theoretical contributions?} No
\end{itemize}

\begin{itemize}
    \item \textbf{Does this paper rely on one or more datasets?} No
\end{itemize}

\begin{itemize}
    \item \textbf{Does this paper include computational experiments?} Yes (Experimental Details)
\end{itemize}

\textbf{If yes, please complete the list below.}

\begin{itemize}
    \item \textbf{Any code required for pre-processing data is included in the appendix:} Yes (Code Availability)
    
    \item \textbf{All source code required for conducting and analyzing the experiments is included in a code appendix:} Yes (Code Availability)
    
    \item \textbf{All source code required for conducting and analyzing the experiments will be made publicly available upon publication of the paper with a license that allows free usage for research purposes:} Yes (Code Availability)
    
    \item \textbf{All source code implementing new methods have comments detailing the implementation, with references to the paper where each step comes from:} Yes (Code Availability)
    
    \item \textbf{If an algorithm depends on randomness, then the method used for setting seeds is described in a way sufficient to allow replication of results:} Yes (Experimental Details)
    
    \item \textbf{This paper specifies the computing infrastructure used for running experiments (hardware and software), including GPU/CPU models; amount of memory; operating system; names and versions of relevant software libraries and frameworks:} Yes (Experimental Details)
    
    \item \textbf{This paper formally describes evaluation metrics used and explains the motivation for choosing these metrics:} Yes (Experimental Details)
    
    \item \textbf{This paper states the number of algorithm runs used to compute each reported result:} Yes (Experimental Details)
    
    \item \textbf{Analysis of experiments goes beyond single-dimensional summaries of performance (e.g., average; median) to include measures of variation, confidence, or other distributional information:} Yes (Results)
    
    \item \textbf{The significance of any improvement or decrease in performance is judged using appropriate statistical tests (e.g., Wilcoxon signed-rank):} Yes (Results)
    
    \item \textbf{This paper lists all final (hyper-)parameters used for each model/algorithm in the paper’s experiments:} Yes (Experimental Details, Supplement)
    
    \item \textbf{This paper states the number and range of values tried per (hyper-) parameter during development of the paper, along with the criterion used for selecting the final parameter setting:} Yes (Experimental Details, Supplement)
\end{itemize}

% !TEX program = pdflatex
% \maketitle
\renewcommand{\thefigure}{Sup\arabic{figure}}
\renewcommand{\thetable}{Sup\arabic{table}}
\renewcommand{\theequation}{Sup\arabic{equation}}

\setcounter{figure}{0}
\setcounter{table}{0}
\setcounter{equation}{0}
\clearpage
\section*{Appendix}
\addcontentsline{toc}{section}{Appendix}
\section{Domains}\label{supp-sec:domain}
We present details for the ATARI PacMan, Pong and Breakout domains. 
\subsection{PacMan}
PacMan is set in a two-dimensional grid that contains food, walls, ghosts, and the PacMan character. The game concludes with a +500 reward when all food pellets are consumed, while encountering a ghost results in a -500 penalty and game over. Each consumed food pellet awards +10 points, and PacMan incurs a -1 penalty for every time step. The available actions for PacMan are moving Up, Down, Right, or Left. The game's state includes the location of PacMan, the position and direction of any ghosts, and the distribution of food pellets. In this iteration of the game, ghosts move according to some distributions. 
\subsection{Pong}
\begin{figure*}[!hbt]
\centering
        \includegraphics[width=0.5\textwidth]{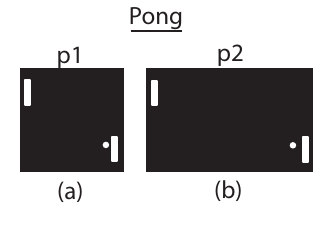} 
        \caption{\emph{Grid variations for Pong.}}
        \label{fig:atari_variations-pong}
\end{figure*}

In this one-player version of Pong, the player competes against a computer-controlled paddle. The game is set on a two-dimensional grid, with the player controlling one paddle and the computer controlling the other. The game concludes with a +500 reward when the ball reaches the grid boundaries on the computer controlled paddle side, while if the grid boundary is reached on the agent's side, a -500 penalty is applied and game over. The agent incurs a -1 penalty for every time step. The available actions for the paddles are moving Right and Left or to Stop. The game's state includes the location of the ball and the position and direction of any paddle. In this iteration of the game, the computer controlled paddle moves according to some distribution. Visualizations of the grids are presented in \ref{fig:atari_variations-pong}.

\subsection{Breakout}
\begin{figure*}[t]
\centering
        \includegraphics[width=0.5\textwidth]{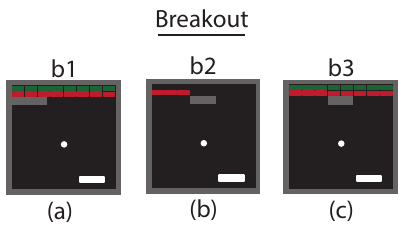} 
        \caption{\emph{Grid variations for Breakout.}}
        \label{fig:atari_variations-breakout}
\end{figure*}

In this version of Breakout, the agent competes against a wall of bricks using a horizontally-moving paddle and a ball. The game is set on a two-dimensional grid, with the agent controlling the paddle located at the bottom of the screen. The objective is to break bricks by hitting them with the ball, which bounces back after each hit. The game concludes with a +500 reward when all bricks are destroyed, but if the ball passes the paddle and reaches the bottom grid boundary, a -500 penalty is applied, resulting in game over. Each hit brick awards +10 points, and the agent incurs a -1 penalty for every time step. The available actions for the agent's paddle are moving Right or Left, or choosing to Stop. The game's state includes the position of the ball, the location of the paddle, and the configuration and status of the bricks. Visualizations of the grids are presented in \ref{fig:atari_variations-breakout}.

\section{Training Parameters}\label{supp-sec:params}
In our experiments, the parameters for Q-Learning and SARSA are inherited by \cite{Cederborg2015PolicySW}. In particular, $\mathcal{T}=1.5$, $\alpha=0.05$, and $\lambda=0.9$.
\section{Additional Graphs Non-Semantic Variations}\label{supp-sec:semantic-variation}
In this section we present supplementary results showing the Generalization Agent and Learnability Agent behavior for Semantic variations of grids throughout Pacman and Pong.
\subsection{PacMan}
\begin{figure*}[t]
\centering
        \includegraphics[width=0.9\textwidth]{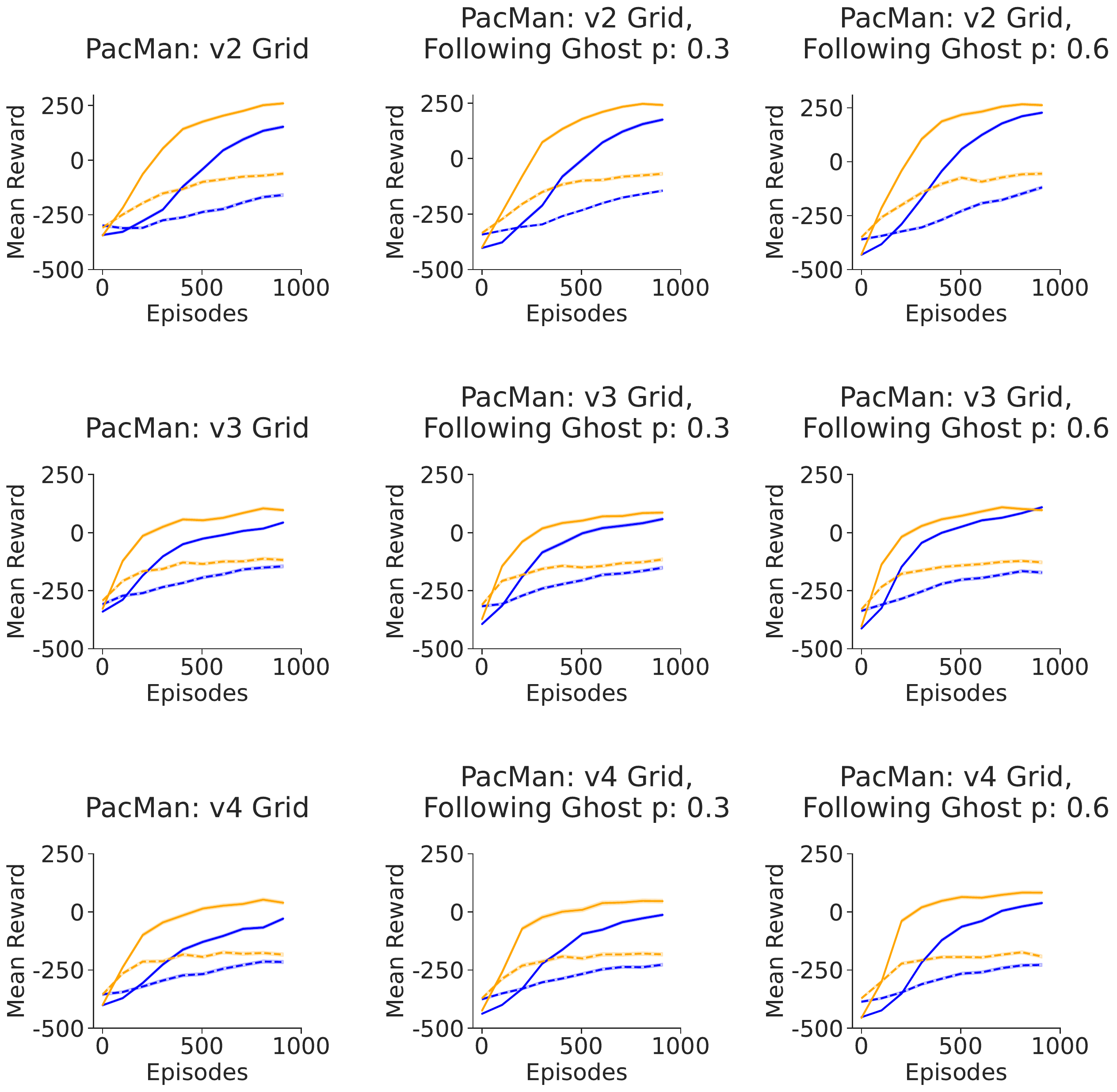} 
        \caption{\emph{SARSA Agent with Boltzmann exploration strategy}: Results for PacMan v2, v3, v4 grids reporting mean reward as a function of episode number. The agent is trained on the non-noisy version of the environment and tested on different level of noise ($\delta \sim \mathcal{N}(0,0.1)$ in Low-Noise and $\delta \sim \mathcal{N}(0,0.5)$ in High-Noise settings).
        }
        \label{fig:atari_variations-pacman-sarsa-boltzmann}
\end{figure*}

\begin{figure*}[t]
\centering
        \includegraphics[width=0.9\textwidth]{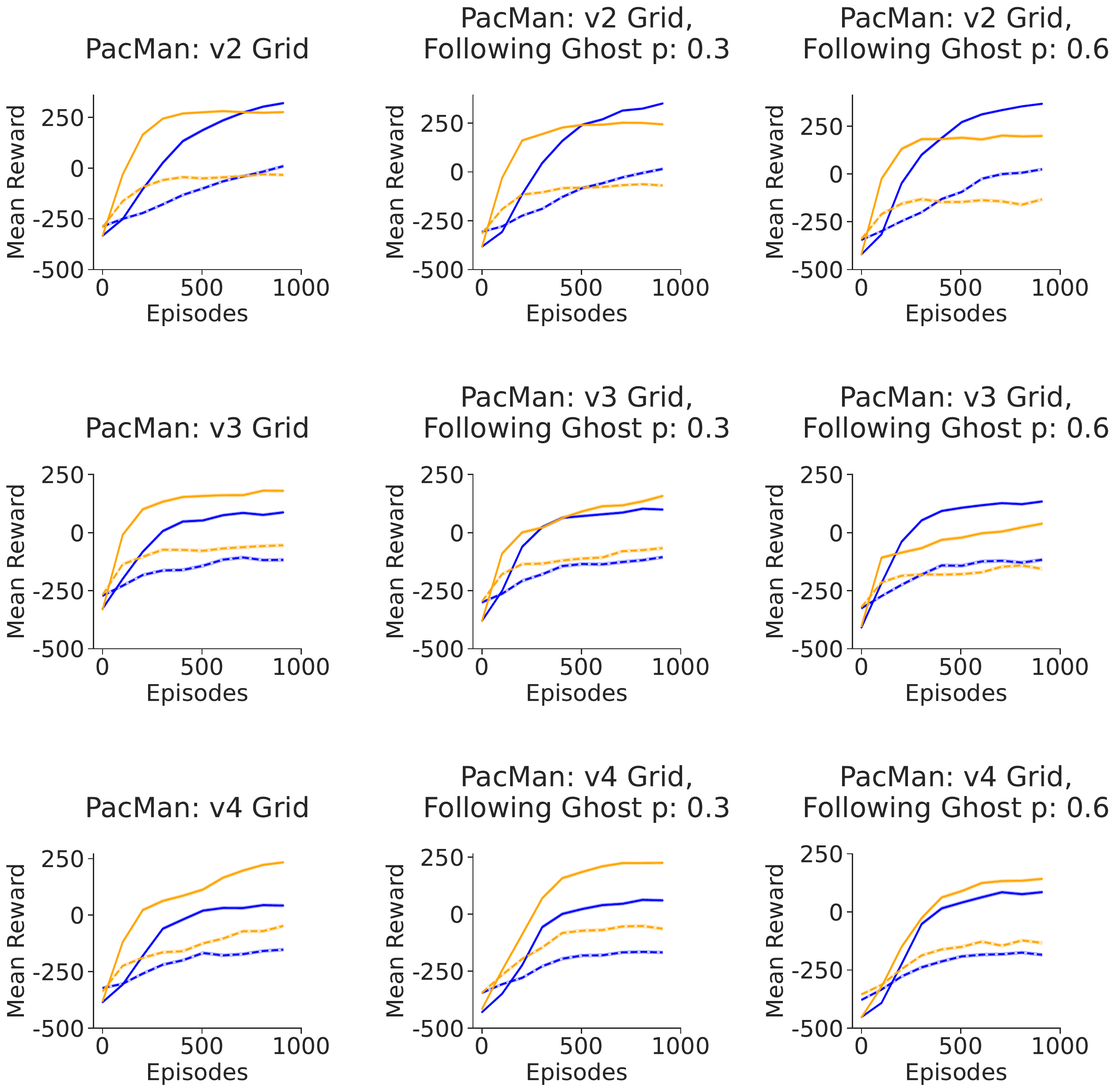} 
        \caption{\emph{Q-learning Agent with Boltzmann exploration strategy}: Results for PacMan v2, v3, v4 grids reporting mean reward as a function of episode number. The agent is trained on the non-noisy version of the environment and tested on different level of noise ($\delta \sim \mathcal{N}(0,0.1)$ in Low-Noise and $\delta \sim \mathcal{N}(0,0.5)$ in High-Noise settings).}
        \label{fig:atari_variations-pacman-qlearning-boltzmann}
\end{figure*}

\begin{figure*}[t]
\centering
        \includegraphics[width=0.9\textwidth]{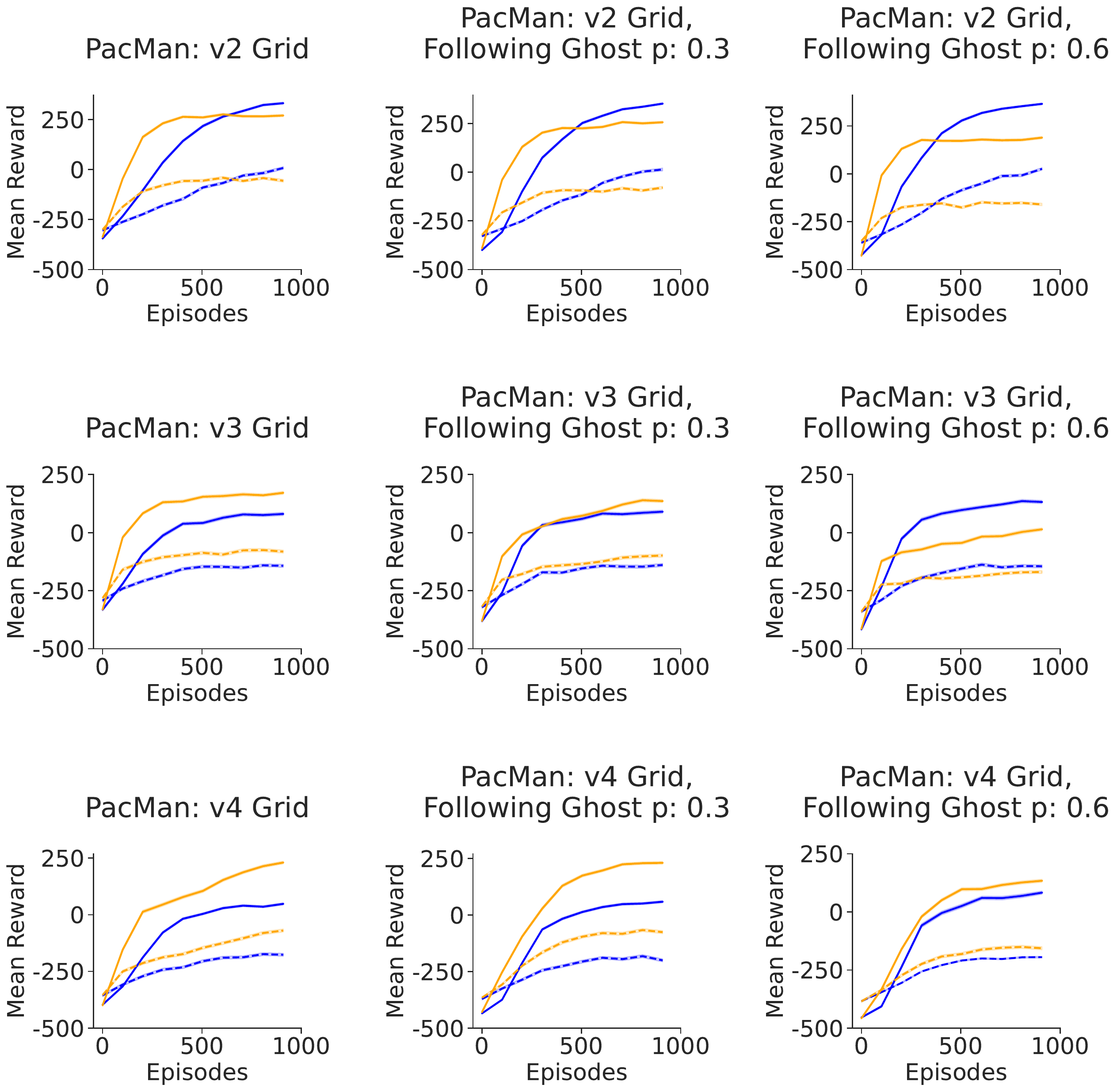} 
        \caption{\emph{Q-learning Agent with $\epsilon\text{-}$greedy exploration strategy}: Results for PacMan v2, v3, v4 grids reporting mean reward as a function of episode number. The agent is trained on the non-noisy version of the environment and tested on different level of noise ($\delta \sim \mathcal{N}(0,0.1)$ in Low-Noise and $\delta \sim \mathcal{N}(0,0.5)$ in High-Noise settings).
        }
        \label{fig:atari_variations-pacman-qlearning-egreedy}
\end{figure*}

Additional results showing the Generalization Agent and Learnability Agent behaviour in Pacman for grids \emph{v2}, \emph{v3}, \emph{v4}, are presented in the Supplementary figures. In particular, results for SARSA Agent with Boltzmann exploration strategy are presented in \ref{fig:atari_variations-pacman-sarsa-boltzmann}. \ref{fig:atari_variations-pacman-qlearning-boltzmann}, \ref{fig:atari_variations-pacman-qlearning-egreedy} show Q-learning Agent with Boltzmann and $\epsilon$-greedy exploration strategies respectively.
\subsection{Pong}
\begin{figure*}[t]
        \centering
        \includegraphics[width=0.9\textwidth]{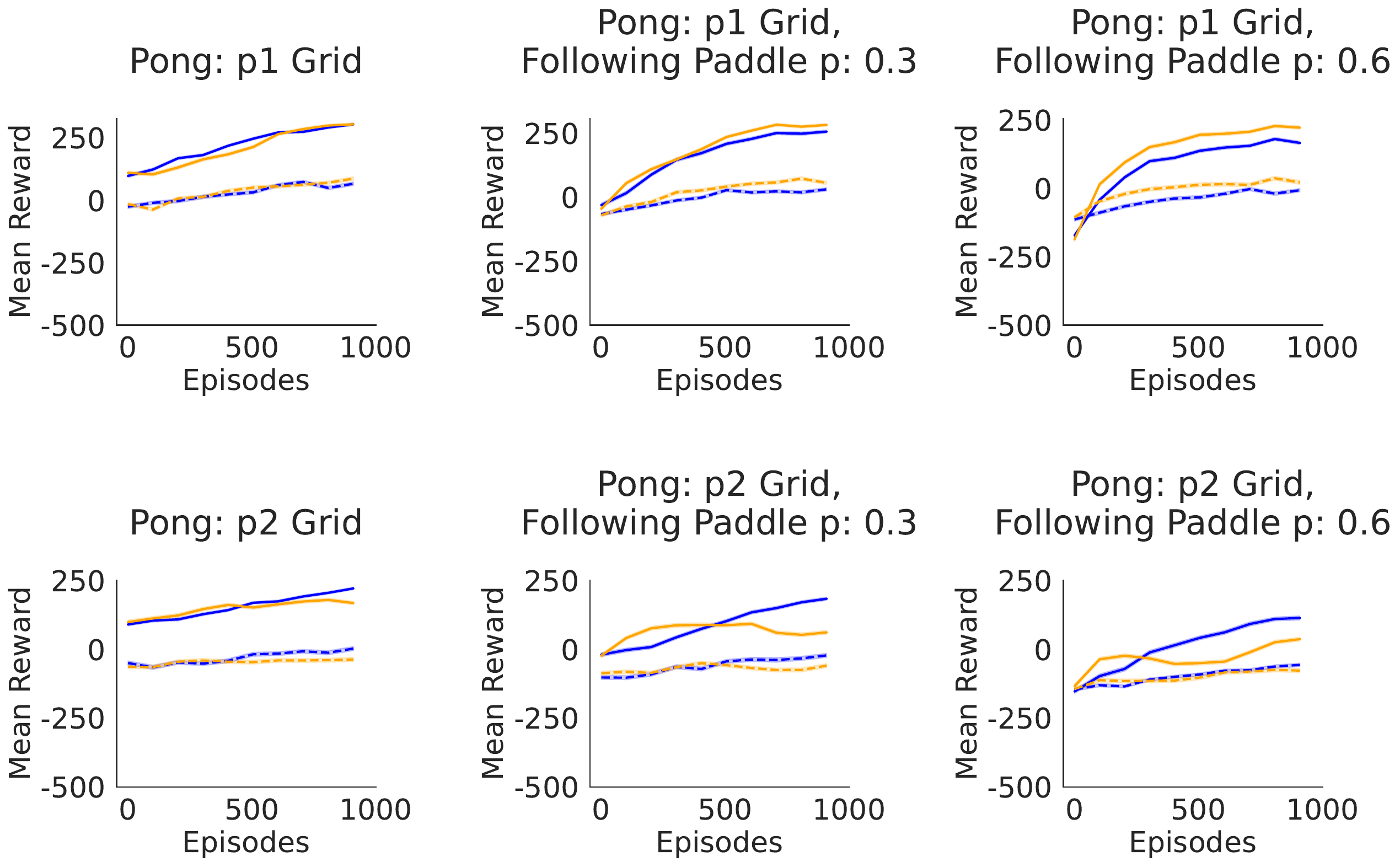} 
        \caption{\emph{Q-learning Agent with Boltzmann exploration strategy}: Results for Pong p1, p2 grids reporting mean reward as a function of episode number. The agent is trained on the non-noisy version of the environment and tested on different level of noise ($\delta \sim \mathcal{N}(0,0.1)$ in Low-Noise and $\delta \sim \mathcal{N}(0,0.5)$ in High-Noise settings).
        }
        \label{fig:atari_variations-pong-qlearning-boltzmann}
\end{figure*}

\begin{figure*}[t]
\centering
        \includegraphics[width=0.9\textwidth]{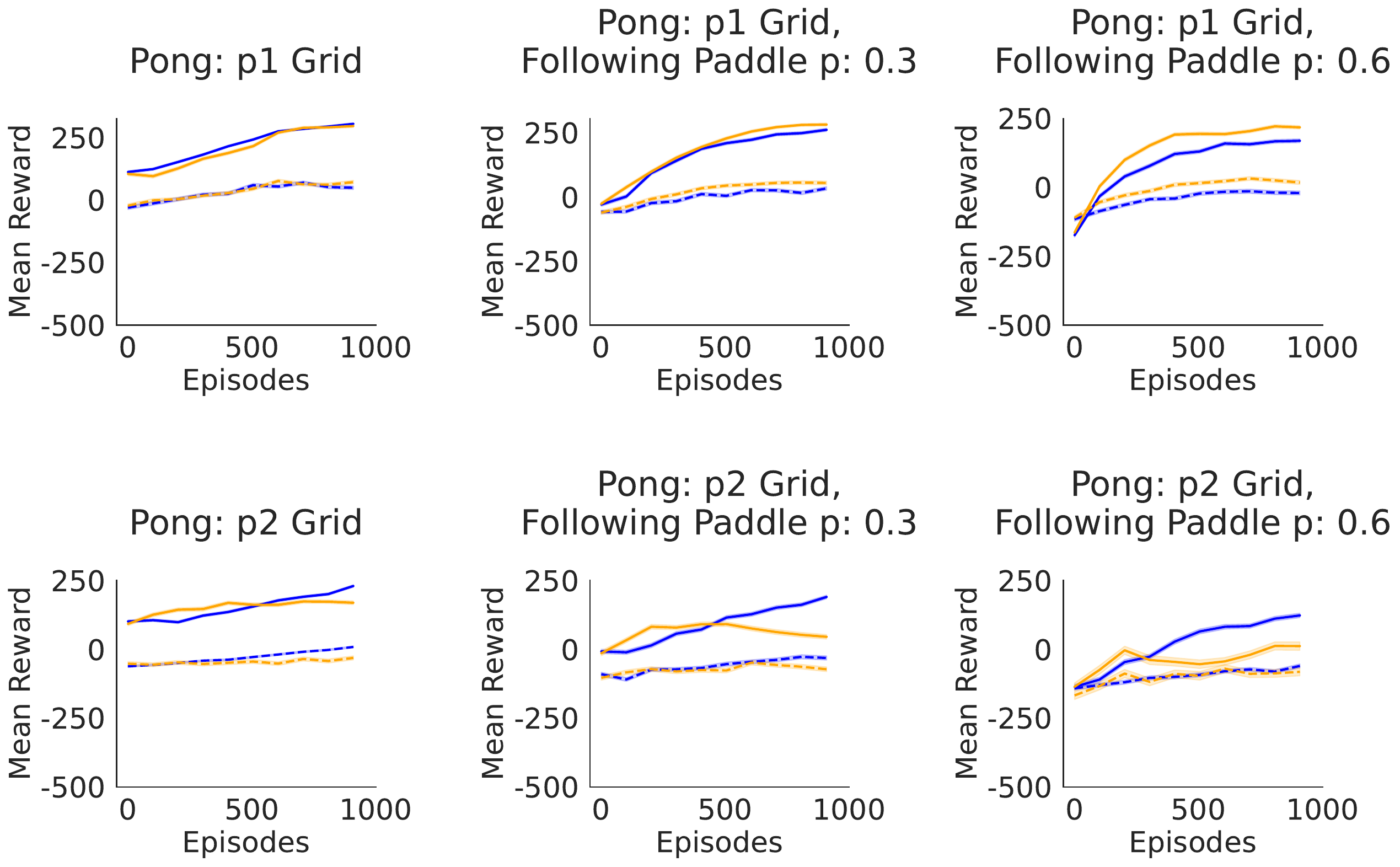} 
        \caption{\emph{Q-learning Agent with $\epsilon\text{-}$greedy exploration strategy}: Results for Pong p1, p2 grids reporting mean reward as a function of episode number. The agent is trained on the non-noisy version of the environment and tested on different level of noise ($\delta \sim \mathcal{N}(0,0.1)$ in Low-Noise and $\delta \sim \mathcal{N}(0,0.5)$ in High-Noise settings).
        }
        \label{fig:atari_variations-pong-qlearning-egreedy}
\end{figure*}

Similarly, for Pong grids \emph{p1}, \emph{p2} results are presented in the Supplementary figures \ref{fig:atari_variations-pong-sarsa-boltzmann} for SARSA Agent and  \ref{fig:atari_variations-pong-qlearning-boltzmann}, \ref{fig:atari_variations-pong-qlearning-egreedy} for Q-learning Agent.
\subsection{Breakout}
\begin{figure*}[t]
\centering
        \includegraphics[width=0.9\textwidth]{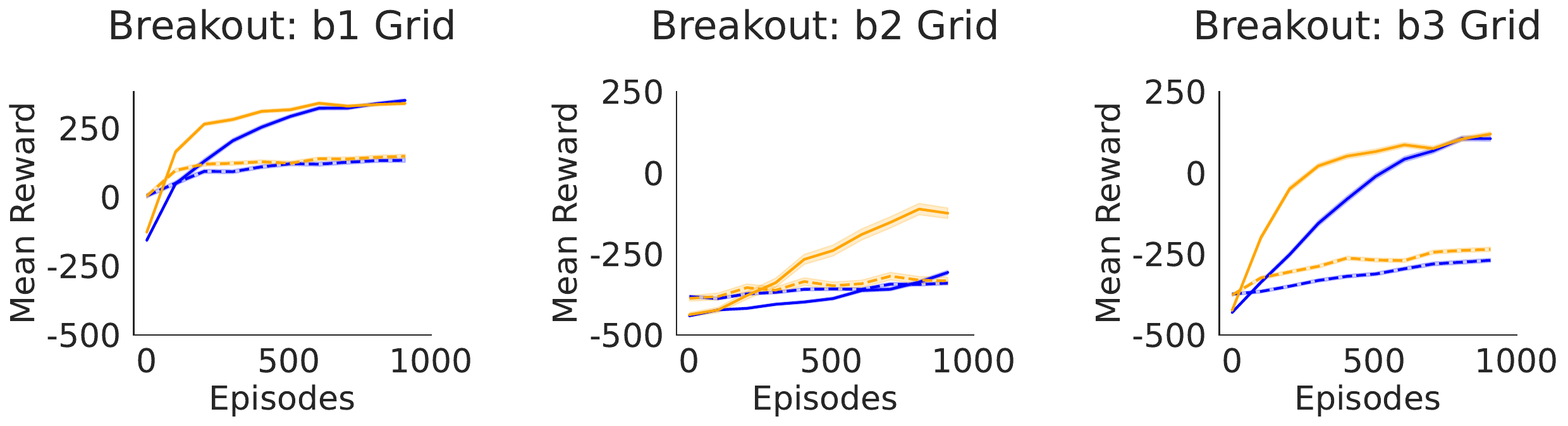} 
        \caption{\emph{SARSA Agent with $\epsilon\text{-}$greedy exploration strategy}: Results for Breakout b1, b2, b3 grids reporting mean reward as a function of episode number. The agent is trained on the non-noisy version of the environment and tested on different level of noise ($\delta \sim \mathcal{N}(0,0.1)$ in Low-Noise and $\delta \sim \mathcal{N}(0,0.5)$ in High-Noise settings).
        }
        \label{fig:atari_variations-breakout-sarsa-egreedy}
\end{figure*}

\begin{figure*}[t]
\centering
        \includegraphics[width=0.9\textwidth]{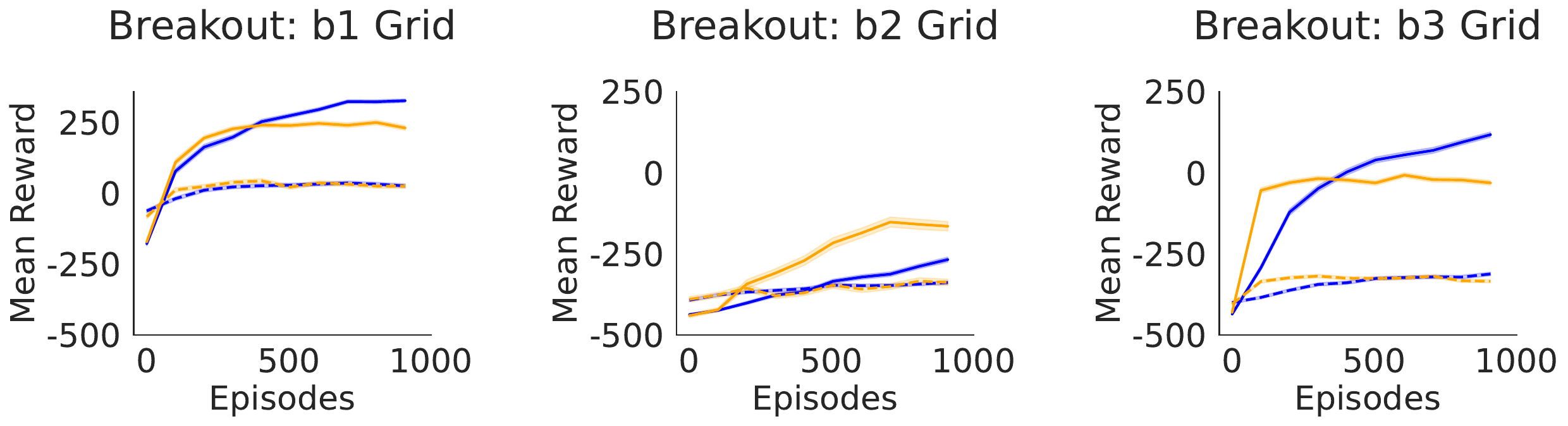} 
        \caption{\emph{Q-learning Agent with Boltzmann exploration strategy}: Results for Breakout b1, b2, b3 grids reporting mean reward as a function of episode number. The agent is trained on the non-noisy version of the environment and tested on different level of noise ($\delta \sim \mathcal{N}(0,0.1)$ in Low-Noise and $\delta \sim \mathcal{N}(0,0.5)$ in High-Noise settings).
        }
        \label{fig:atari_variations-breakout-qlearning-boltzmann}
\end{figure*}

\begin{figure*}[t]
\centering
        \includegraphics[width=0.9\textwidth]{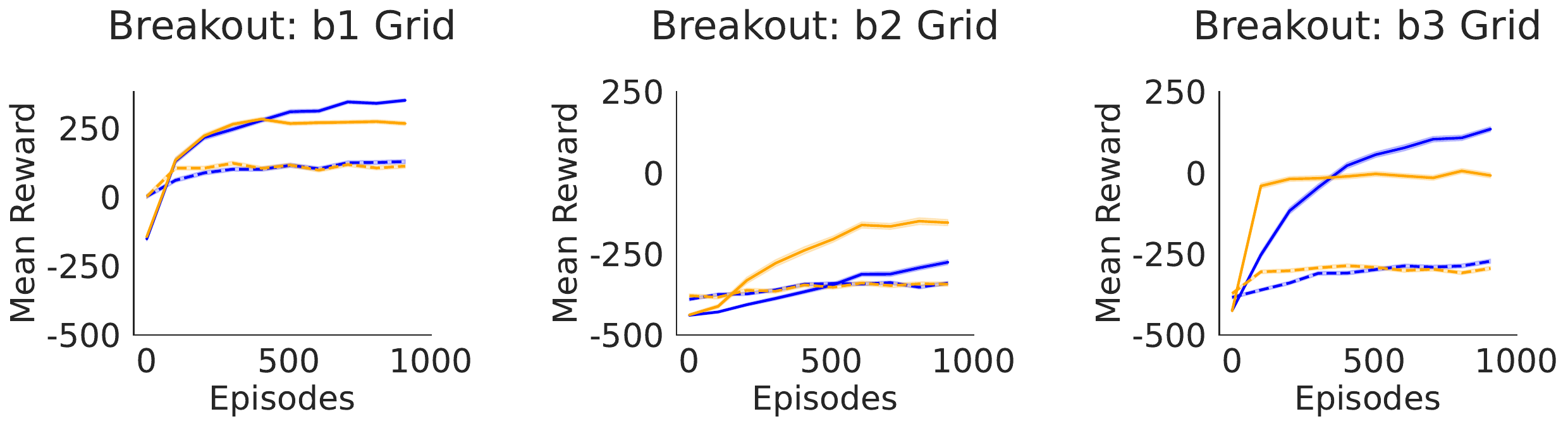} 
        \caption{\emph{Q-learning Agent with $\epsilon\text{-}$greedy exploration strategy}: Results for Breakout b1, b2, b3 grids reporting mean reward as a function of episode number. The agent is trained on the non-noisy version of the environment and tested on different level of noise ($\delta \sim \mathcal{N}(0,0.1)$ in Low-Noise and $\delta \sim \mathcal{N}(0,0.5)$ in High-Noise settings).
        }
        \label{fig:atari_variations-breakout-qlearning-egreedy}
\end{figure*}

Analogously, for Breakout grids \emph{b1}, \emph{b2}, \emph{b3} results are presented in the Supplementary figures \ref{fig:atari_variations-breakout-sarsa-boltzmann}, \ref{fig:atari_variations-breakout-sarsa-egreedy} for SARSA Agent and  \ref{fig:atari_variations-breakout-qlearning-boltzmann}, \ref{fig:atari_variations-breakout-qlearning-egreedy} for Q-learning Agent.
%%%% Semantic Variatioxns
\section{Additional Graphs Semantic variations}
In this section we present supplementary results showing the Generalization Agent and Learnability Agent behavior for Semantic variations of grids throughout Pacman and Pong.
\subsection{PacMan}
\begin{figure*}[t]
\centering
        \includegraphics[width=0.9\textwidth]{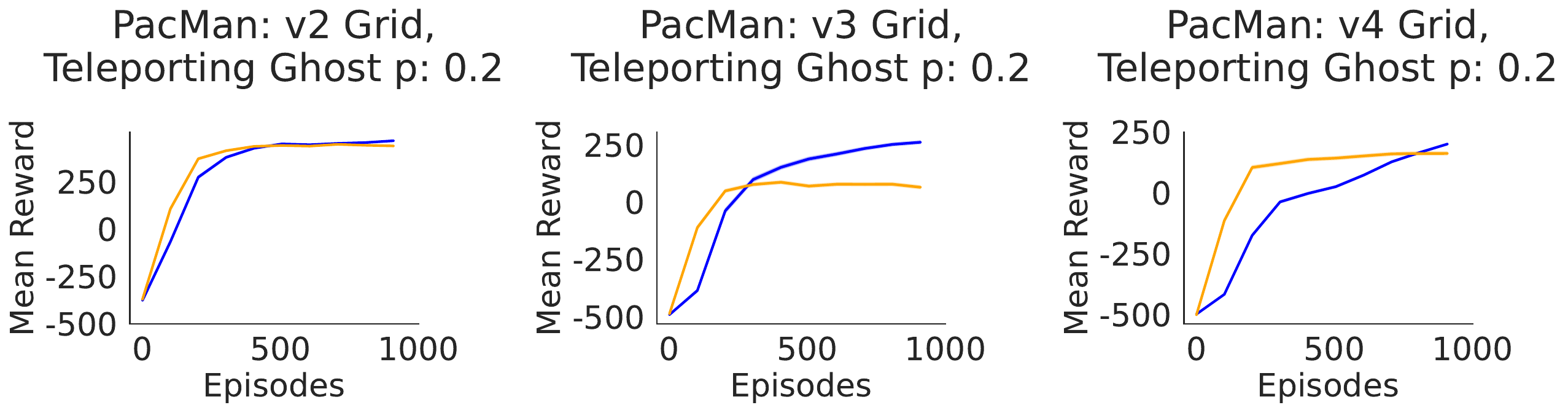} 
        \caption{\emph{Q-learning Agent with Boltzmann exploration strategy}: Results for PacMan v2, v3, v4 grids reporting mean reward as a function of episode number. The agent is trained on the Random Ghost environment and tested on the Teleporting Ghost variation ($p=0.2$, $p=0.5$)}
        \label{fig:atari_variations-semantic-pacman-qlearning-boltzmann}
\end{figure*}

\begin{figure*}[t]
\centering
        \includegraphics[width=0.9\textwidth]{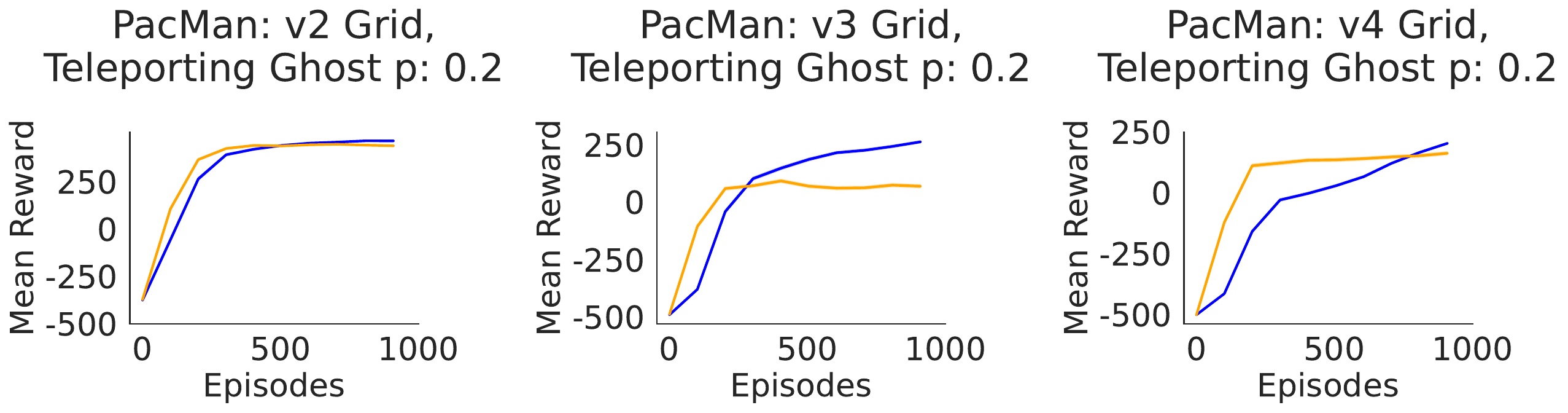} 
        \caption{\emph{Q-learning Agent with $\epsilon\text{-}$greedy exploration strategy}: Results for PacMan v2, v3, v4 grids reporting mean reward as a function of episode number. The agent is trained on the Random Ghost environment and tested on the Teleporting Ghost variation ($p=0.2$, $p=0.5$)}
        \label{fig:atari_variations-semantic-pacman-qlearning-egreedy}
\end{figure*}

The behavior of the Generalization and Learnabilty Agents under semantic variations of PacMan on grids \emph{v2}, \emph{v3}, \emph{v4} are presented in Supplementary figures \ref{fig:atari_variations-semantic-pacman-sarsa-boltzmann} for SARSA Agent and \ref{fig:atari_variations-semantic-pacman-qlearning-boltzmann} and \ref{fig:atari_variations-semantic-pacman-qlearning-egreedy} for Q-learning Agent. 
\subsection{Pong}
\begin{figure*}[t]
\centering
        \includegraphics[width=0.65\textwidth]{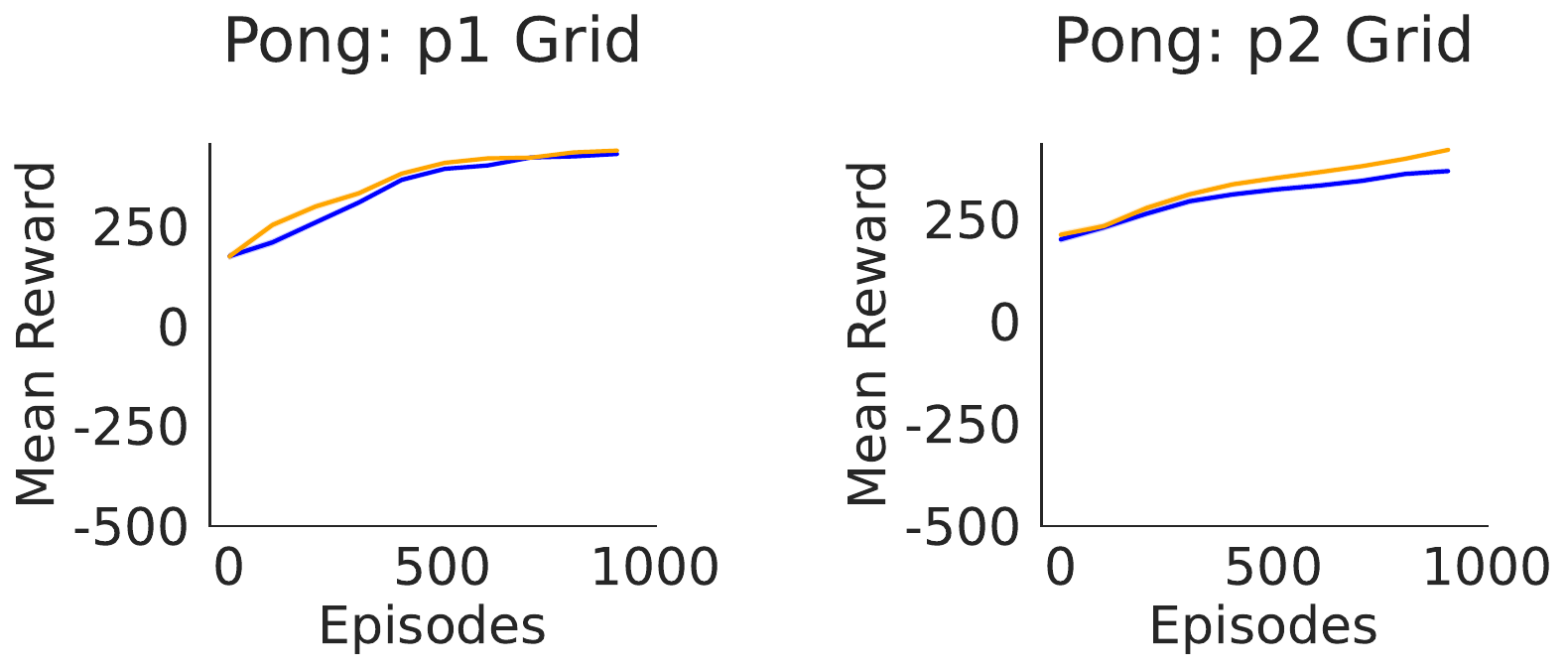} 
        \caption{\emph{SARSA Agent with Boltzmann exploration strategy}: Results for Pong p1, p2 grids reporting mean reward as a function of episode number. The agent is trained on the Directional Ghost ($p=0.3$) environment and tested on the Random Ghost variation.}
        \label{fig:atari_variations-semantic-breakout-sarsa-boltzmann_0.3}
\end{figure*}

\begin{figure*}[t]
\centering
        \includegraphics[width=0.65\textwidth]{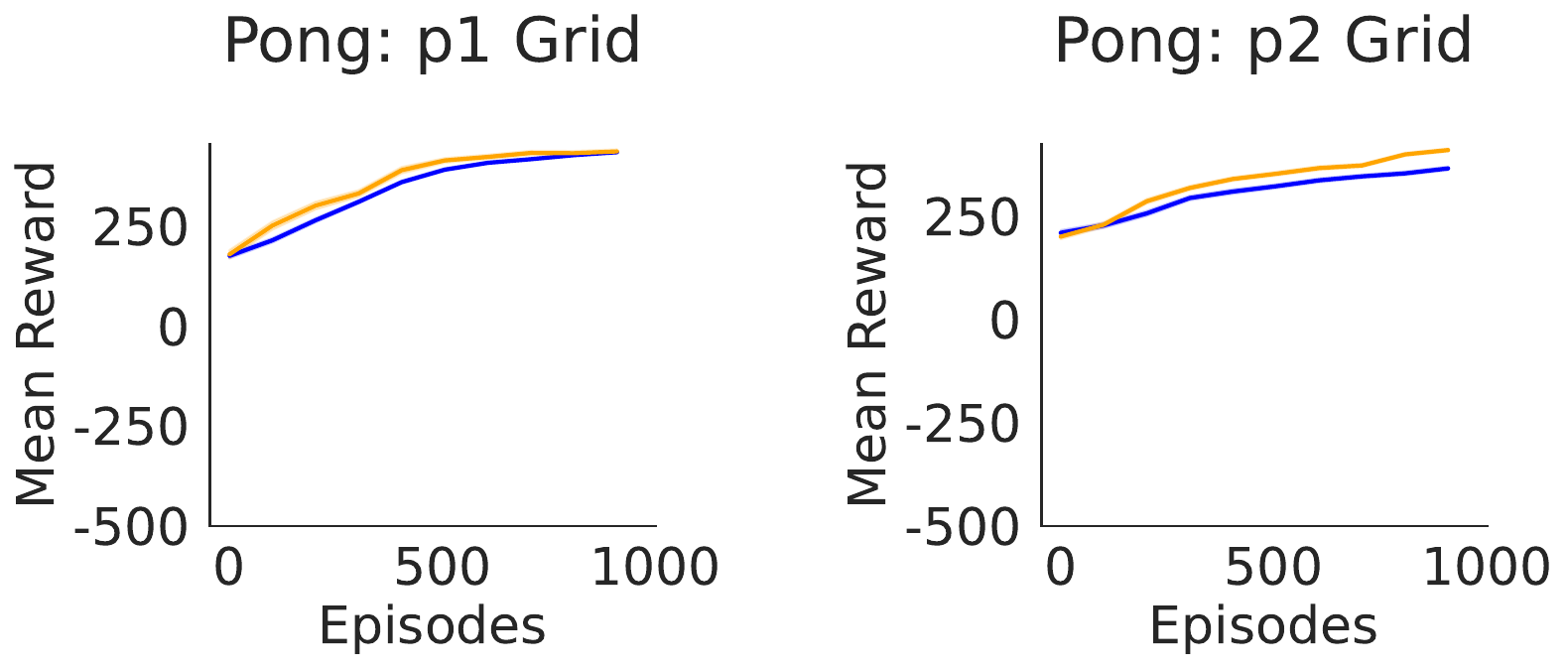} 
        \caption{\emph{SARSA Agent with $\epsilon\text{-}$greedy  exploration strategy}: Results for Pong p1, p2 grids reporting mean reward as a function of episode number. The agent is trained on the Directional Ghost ($p=0.3$) environment and tested on the Random Ghost variation.}
        \label{fig:atari_variations-semantic-breakout-sarsa-egreedy_0.3}
\end{figure*}

\begin{figure*}[t]
\centering
        \includegraphics[width=0.65\textwidth]{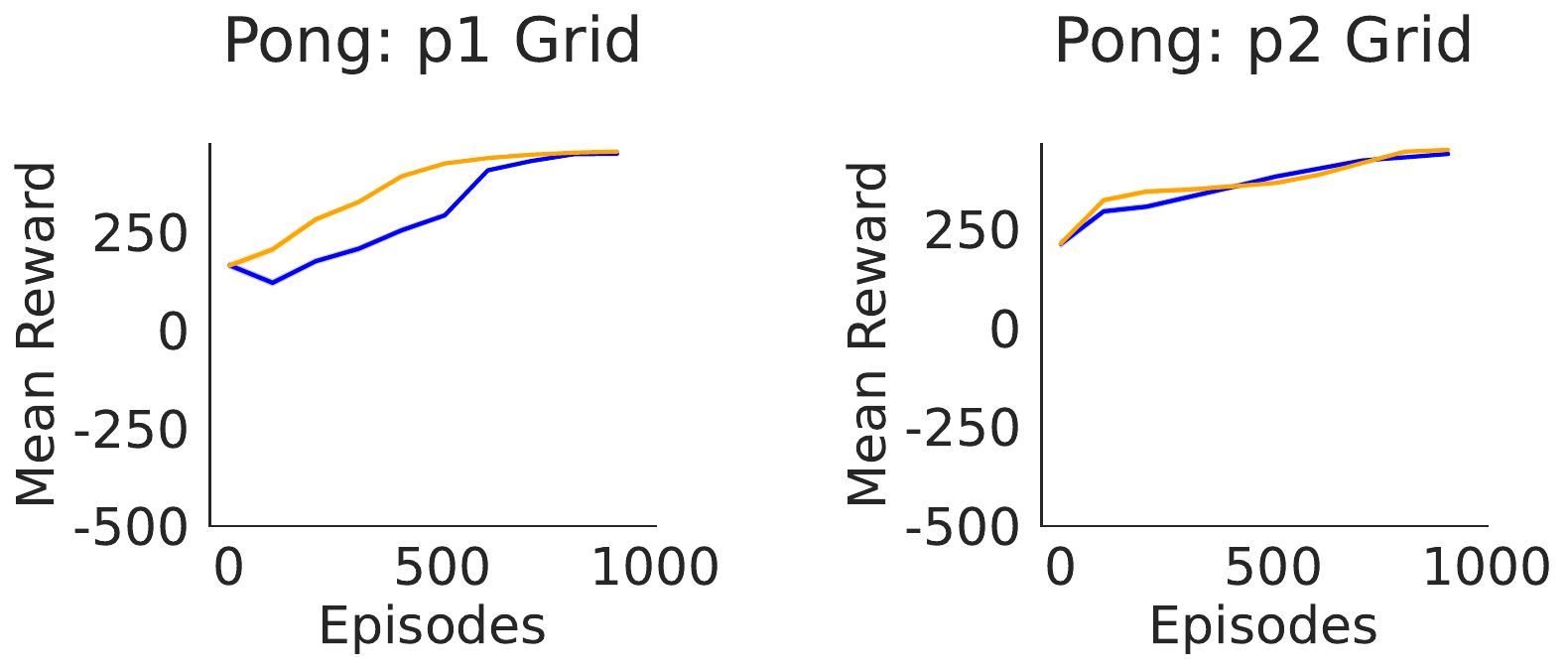} 
        \caption{\emph{Q-learning Agent with Boltzmann exploration strategy}: Results for Pong p1, p2 grids reporting mean reward as a function of episode number. The agent is trained on the Directional Ghost ($p=0.3$) environment and tested on the Random Ghost variation.}
        \label{fig:atari_variations-semantic-breakout-qlearning-boltzmann_0.3}
\end{figure*}

\begin{figure*}[t]
\centering
        \includegraphics[width=0.65\textwidth]{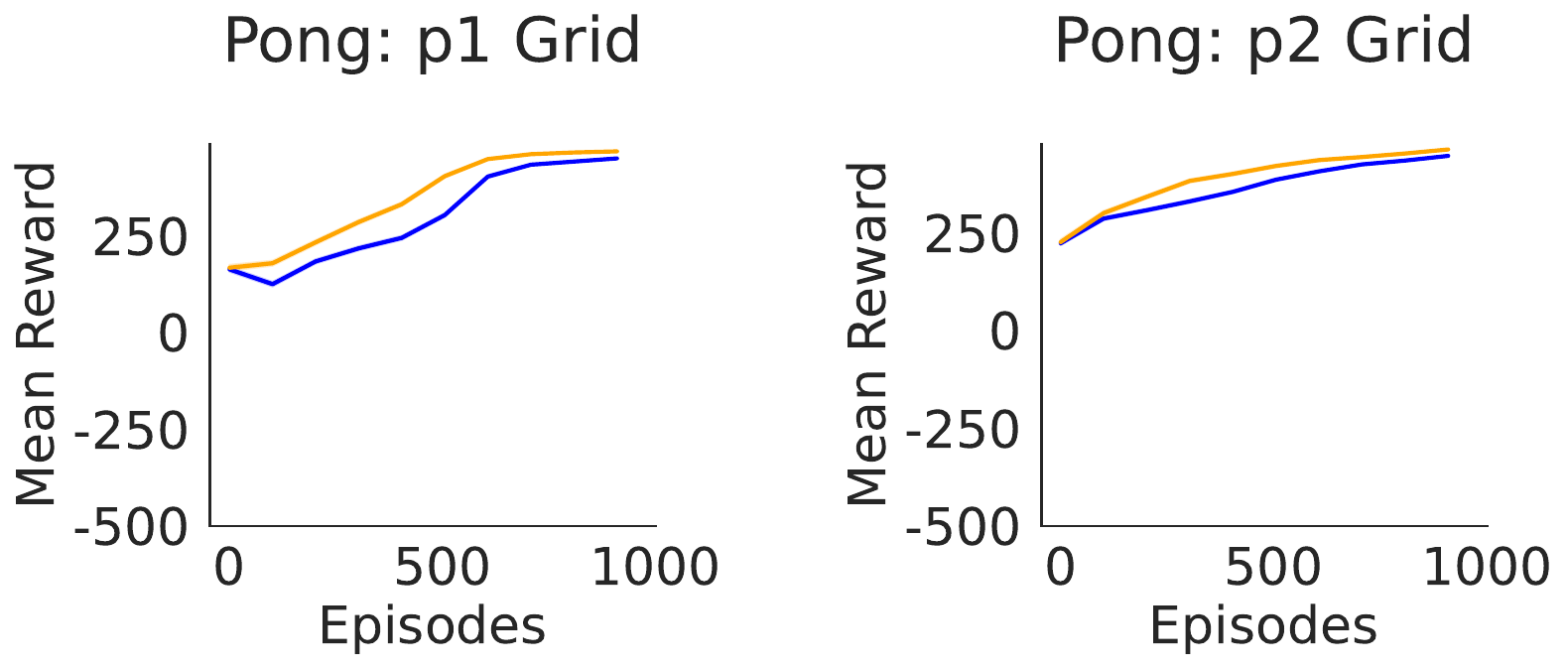} 
        \caption{\emph{Q-learning Agent with $\epsilon\text{-}$greedy exploration strategy}: Results for Pong p1, p2 grids reporting mean reward as a function of episode number. The agent is trained on the Directional Ghost ($p=0.3$) environment and tested on the Random Ghost variation.}
        \label{fig:atari_variations-semantic-pong-qlearning-egreedy_0.3}
\end{figure*}

\begin{figure*}[t]
\centering
        \includegraphics[width=0.65\textwidth]{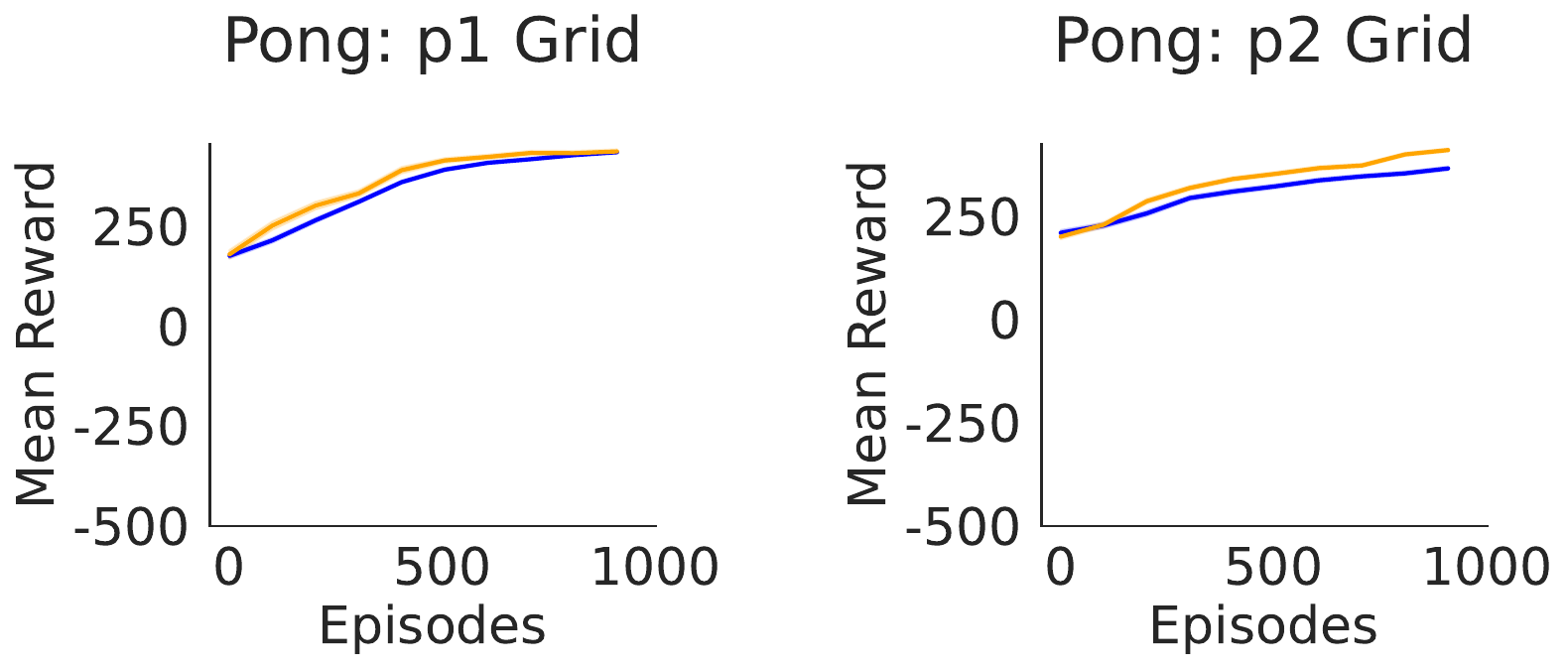} 
        \caption{\emph{SARSA Agent with $\epsilon\text{-}$greedy exploration strategy}: Results for Pong p1, p2 grids reporting mean reward as a function of episode number. The agent is trained on the Directional Ghost ($p=0.3$) environment and tested on the Random Ghost variation.}
        \label{fig:atari_variations-semantic-pong-sarsa-egreedy}
\end{figure*}

\begin{figure*}[t]
\centering
        \includegraphics[width=0.65\textwidth]{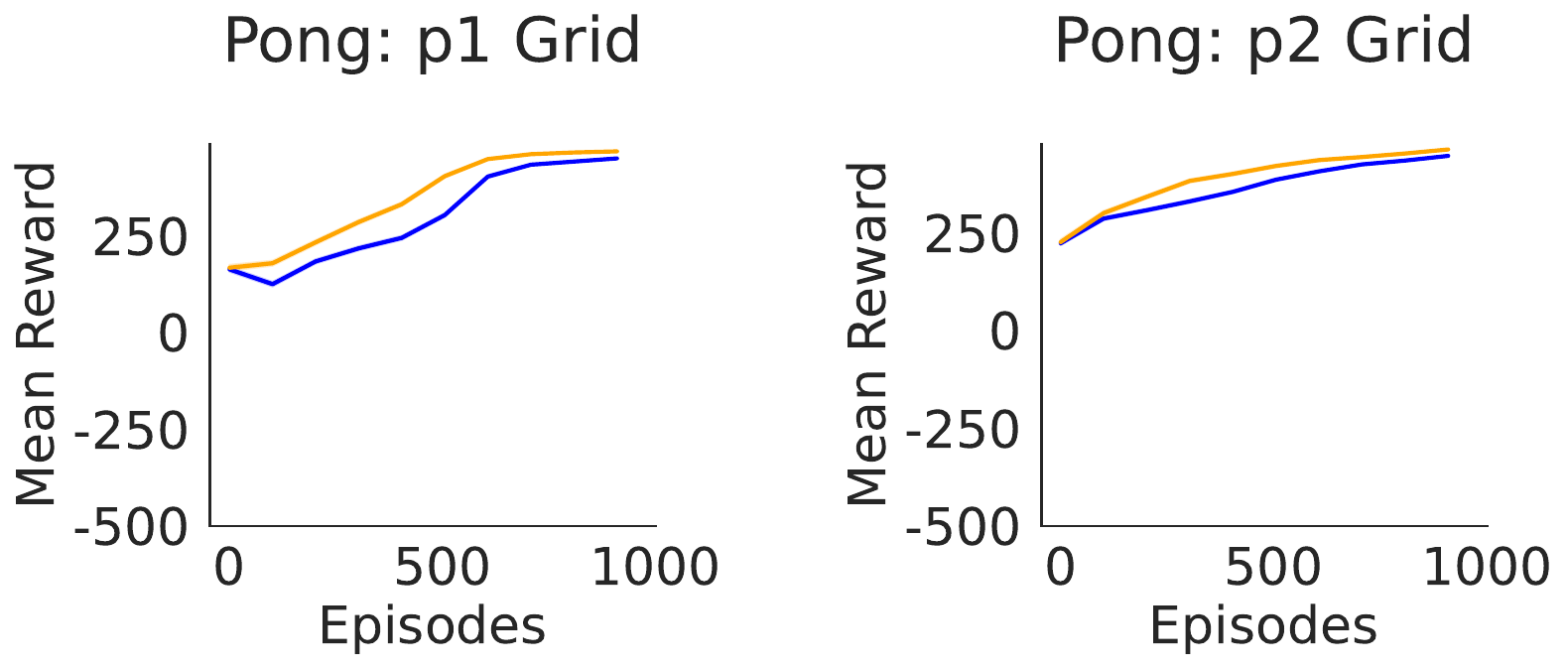} 
        \caption{\emph{SARSA Agent with $\epsilon\text{-}$greedy exploration strategy}: Results for Pong p1, p2 grids reporting mean reward as a function of episode number. The agent is trained on the Directional Ghost ($p=0.3$) environment and tested on the Random Ghost variation.}
        \label{fig:atari_variations-semantic-pong-qlearning-egreedy}
\end{figure*}

\begin{figure*}[t]
\centering
        \includegraphics[width=0.65\textwidth]{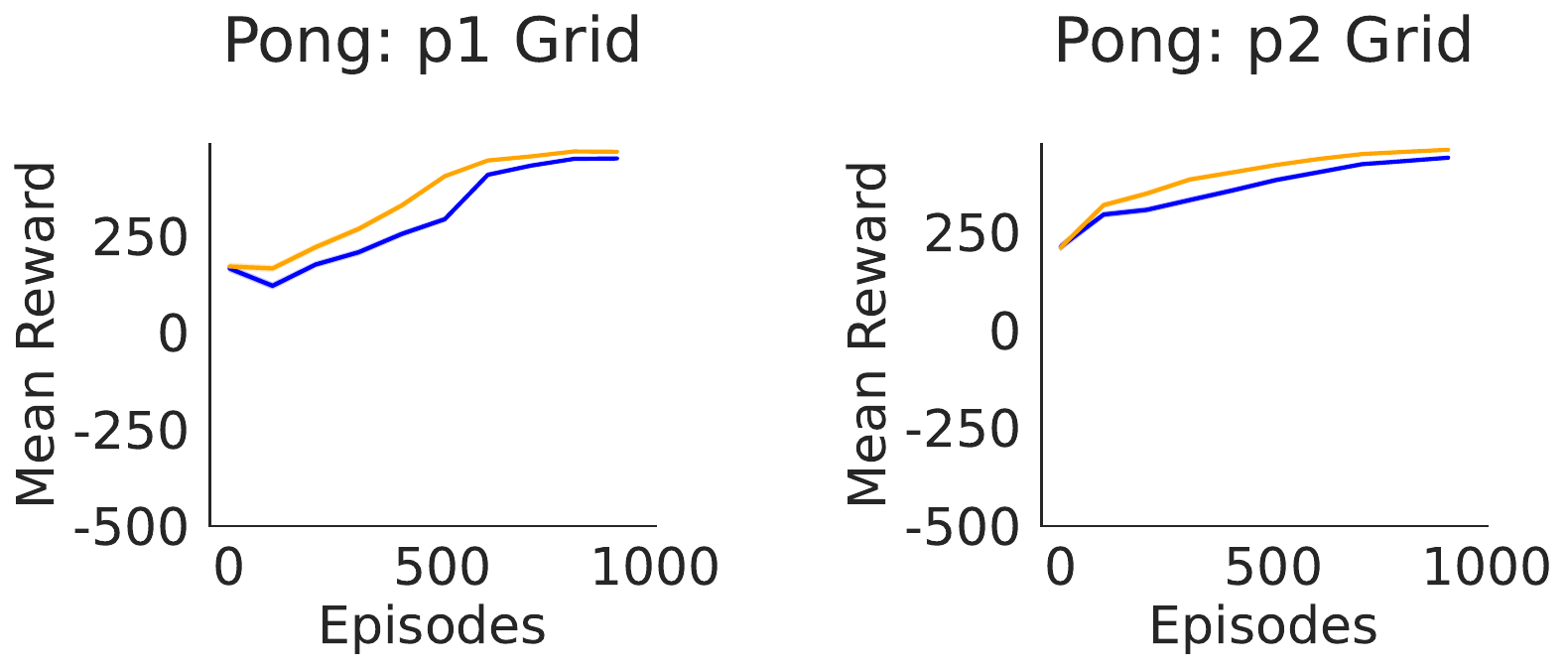} 
        \caption{\emph{SARSA Agent with $\epsilon\text{-}$greedy exploration strategy}: Results for Pong p1, p2 grids reporting mean reward as a function of episode number. The agent is trained on the Directional Ghost ($p=0.3$) environment and tested on the Random Ghost variation.}
        \label{fig:atari_variations-semantic-pong-qlearning-boltzmann}
\end{figure*}

Similarly, for Pong grids p1, p2 results are presented in the Supplementary figures. In particular, semantic variations featuring Directional Ghost $p=0.3$ are presented in \ref{fig:atari_variations-semantic-breakout-sarsa-boltzmann_0.3}, \ref{fig:atari_variations-semantic-breakout-sarsa-egreedy_0.3} for SARSA Agent and \ref{fig:atari_variations-semantic-breakout-qlearning-boltzmann_0.3}, \ref{fig:atari_variations-semantic-pong-qlearning-egreedy_0.3} for Q-learning Agent. While semantic variations featuring Directional Ghost $p=0.6$ are shown in \ref{fig:atari_variations-semantic-pong-sarsa-boltzmann}, \ref{fig:atari_variations-semantic-pong-sarsa-egreedy} for SARSA Agent and \ref{fig:atari_variations-semantic-pong-qlearning-boltzmann}, \ref{fig:atari_variations-semantic-pong-qlearning-egreedy} for Q-learning.
\section{Additional Graphs State-Action Pairs}\label{supp-sec:exploration-patterns}
This section shows the supplementary results for the \textit{exploration grid} visualizing the difference in State-Action (S-A) pairs explored by these agents ($D_{LG}$) throughout the analyzed domains.
\subsection{PacMan}
\begin{figure*}[t]
  %\centering
  \begin{subfigure}{0.3\textwidth}
    \includegraphics[width=\linewidth]{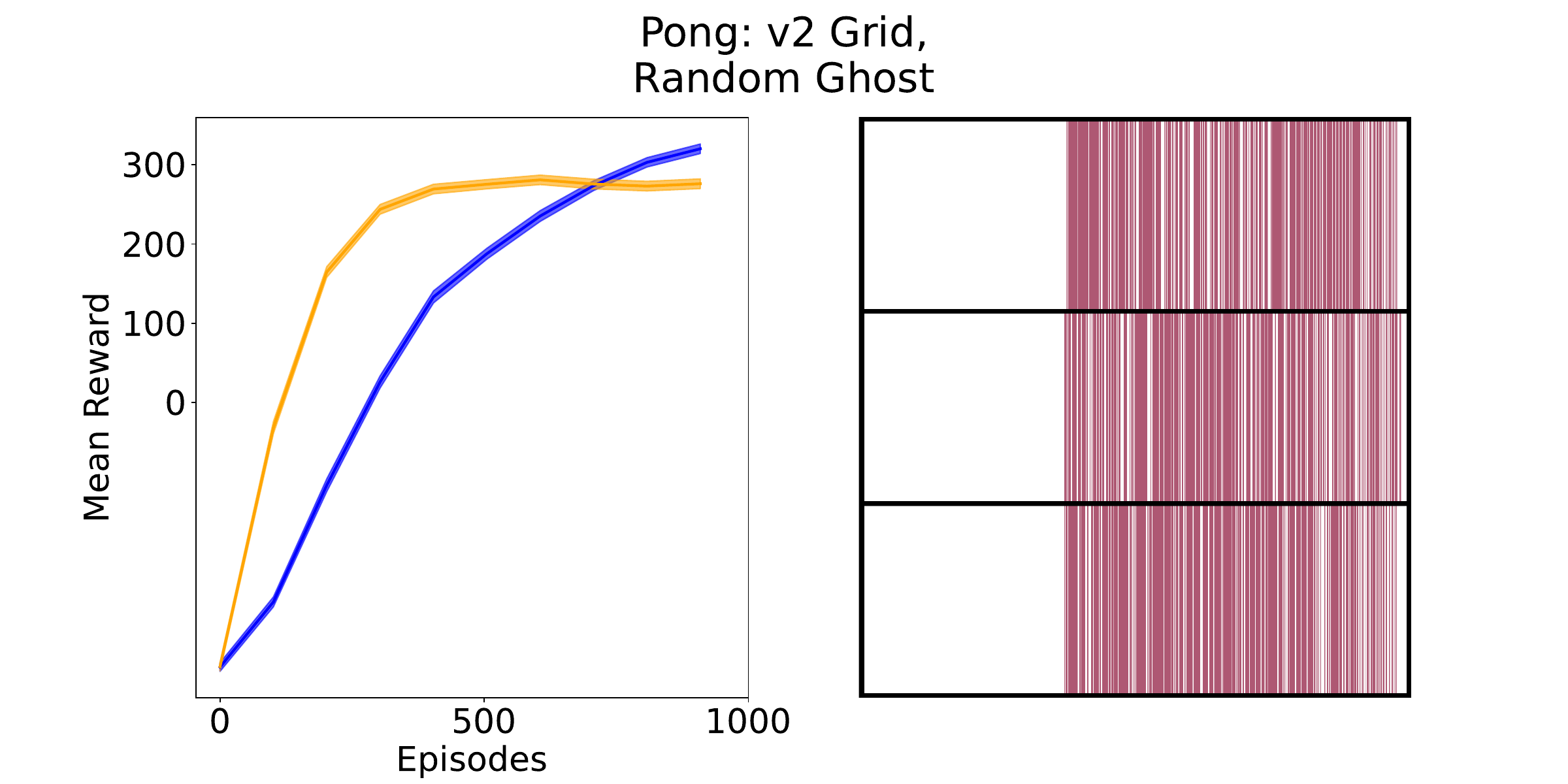}
  \end{subfigure}
  \hfill
  \begin{subfigure}{0.3\textwidth}
    \includegraphics[width=\linewidth]{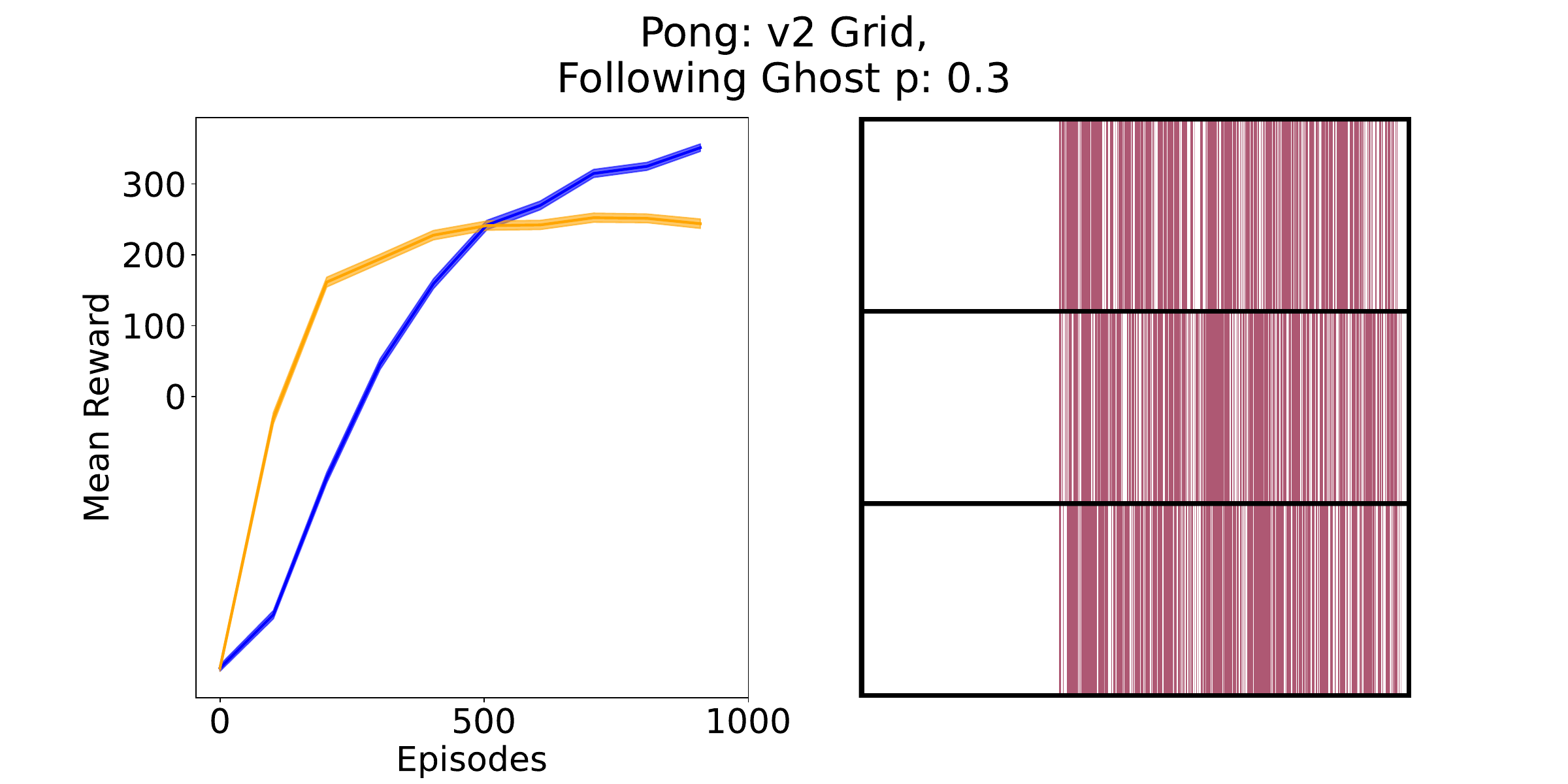}
  \end{subfigure}
  \hfill
  \begin{subfigure}{0.3\textwidth}
    \includegraphics[width=\linewidth]{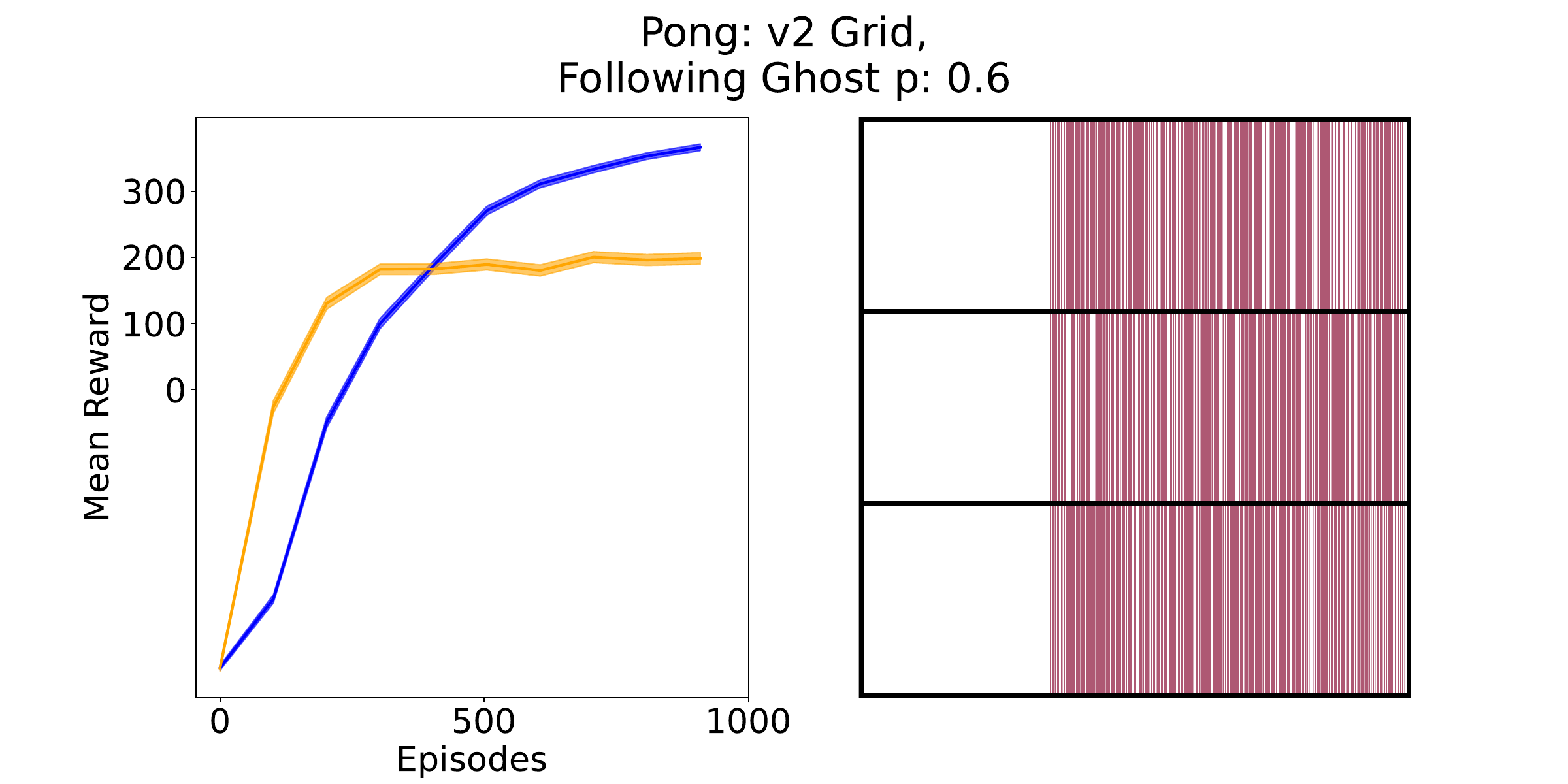}
  \end{subfigure}
  
  \begin{subfigure}{0.3\textwidth}
    \includegraphics[width=\linewidth]{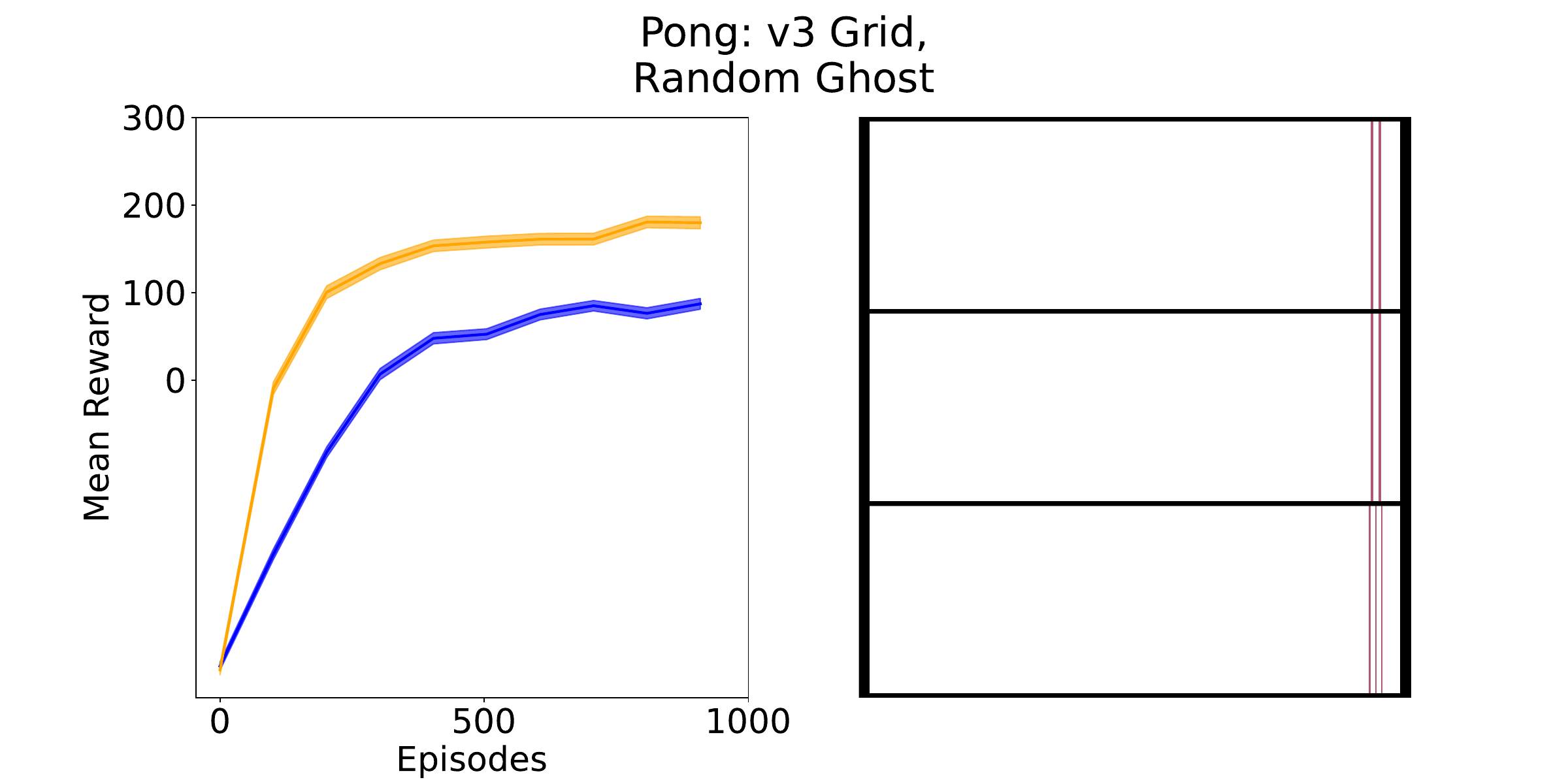}
  \end{subfigure}
  \hfill
  \begin{subfigure}{0.3\textwidth}
    \includegraphics[width=\linewidth]{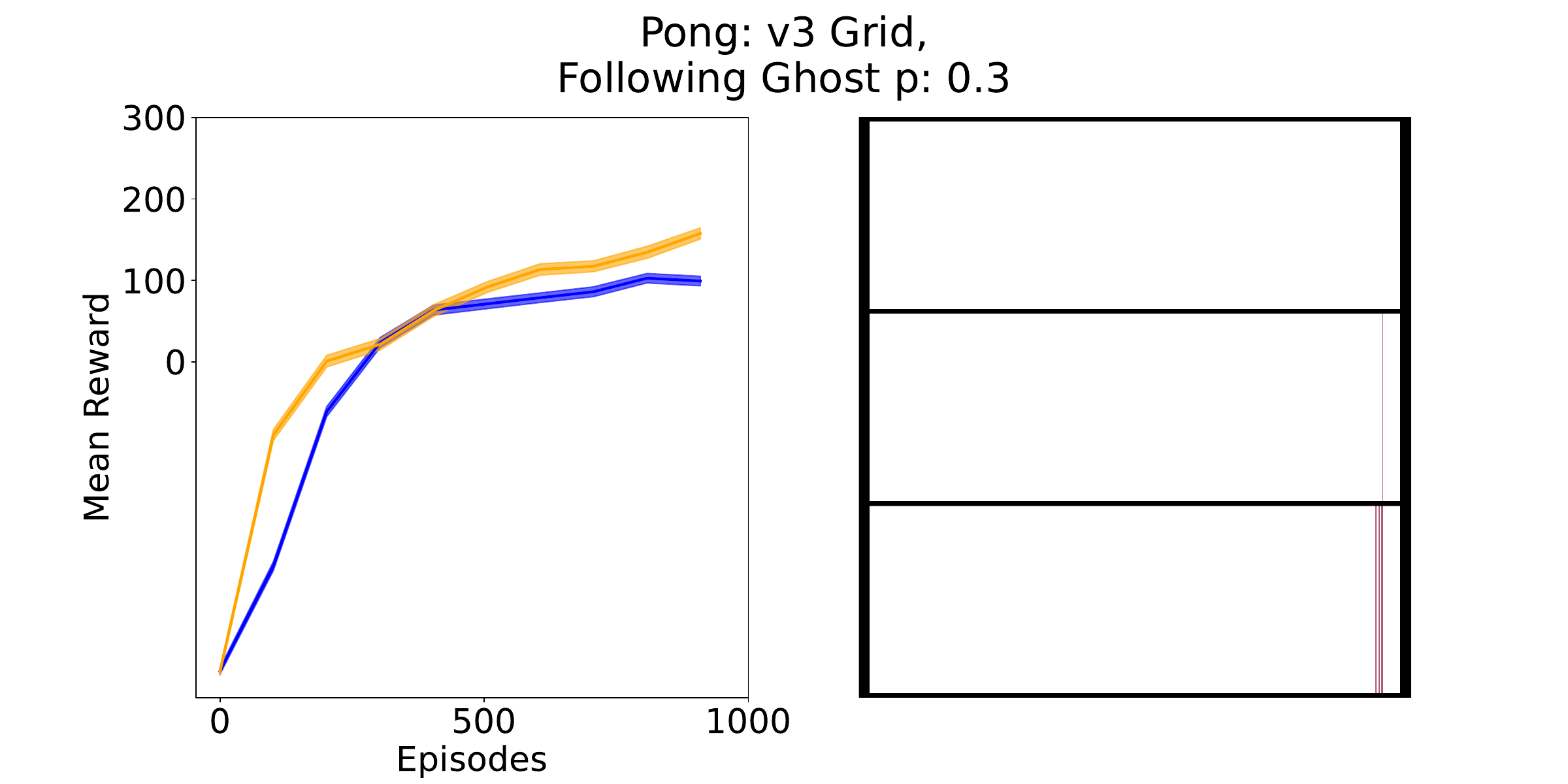}
  \end{subfigure}
  \hfill
  \begin{subfigure}{0.3\textwidth}
    \includegraphics[width=\linewidth]{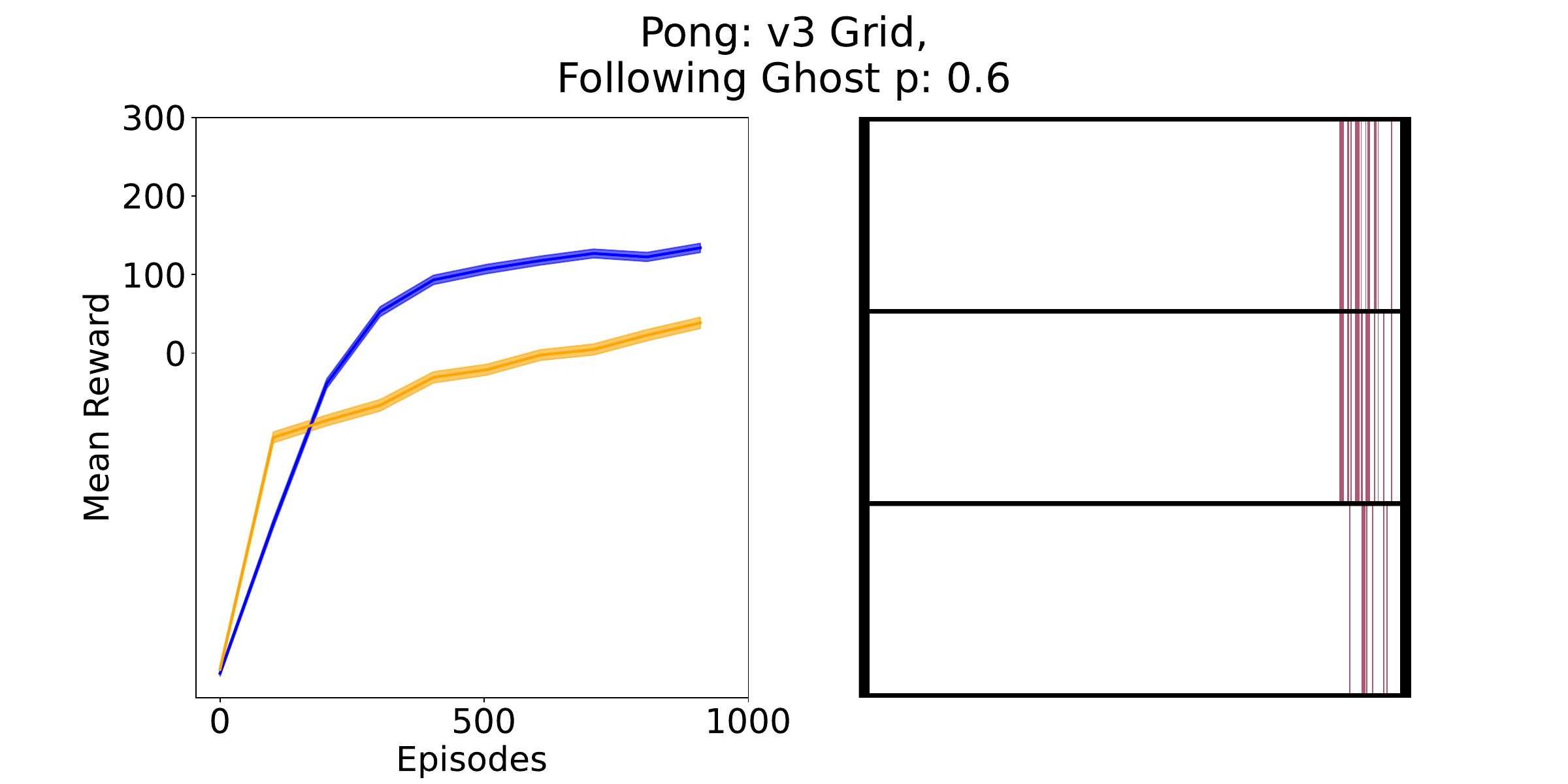}
  \end{subfigure}\\
  \hfill
  \begin{subfigure}{0.30\textwidth}
    \includegraphics[width=\linewidth]{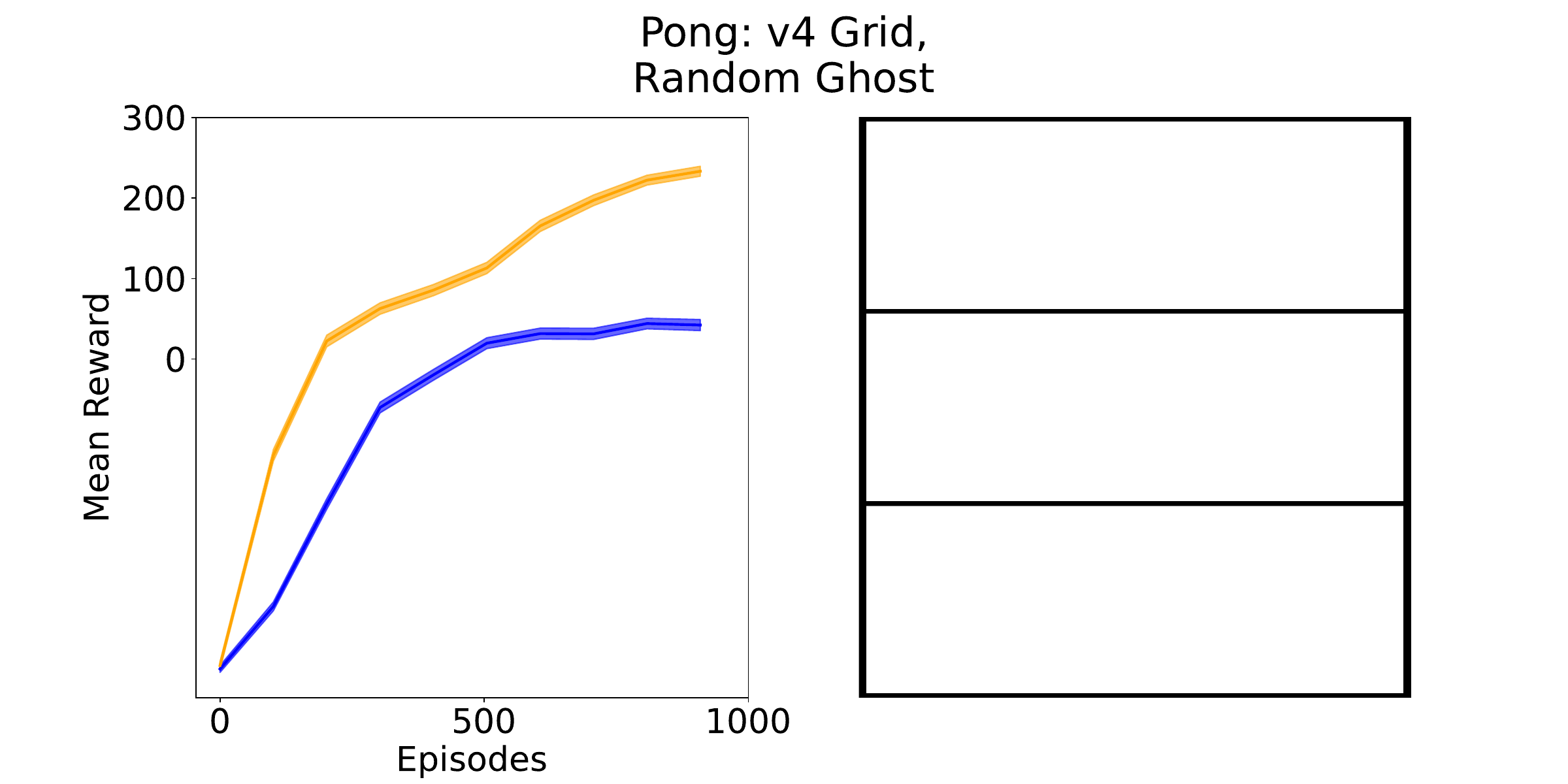}
  \end{subfigure}
  \hfill
  \begin{subfigure}{0.30\textwidth}
    \includegraphics[width=\linewidth]{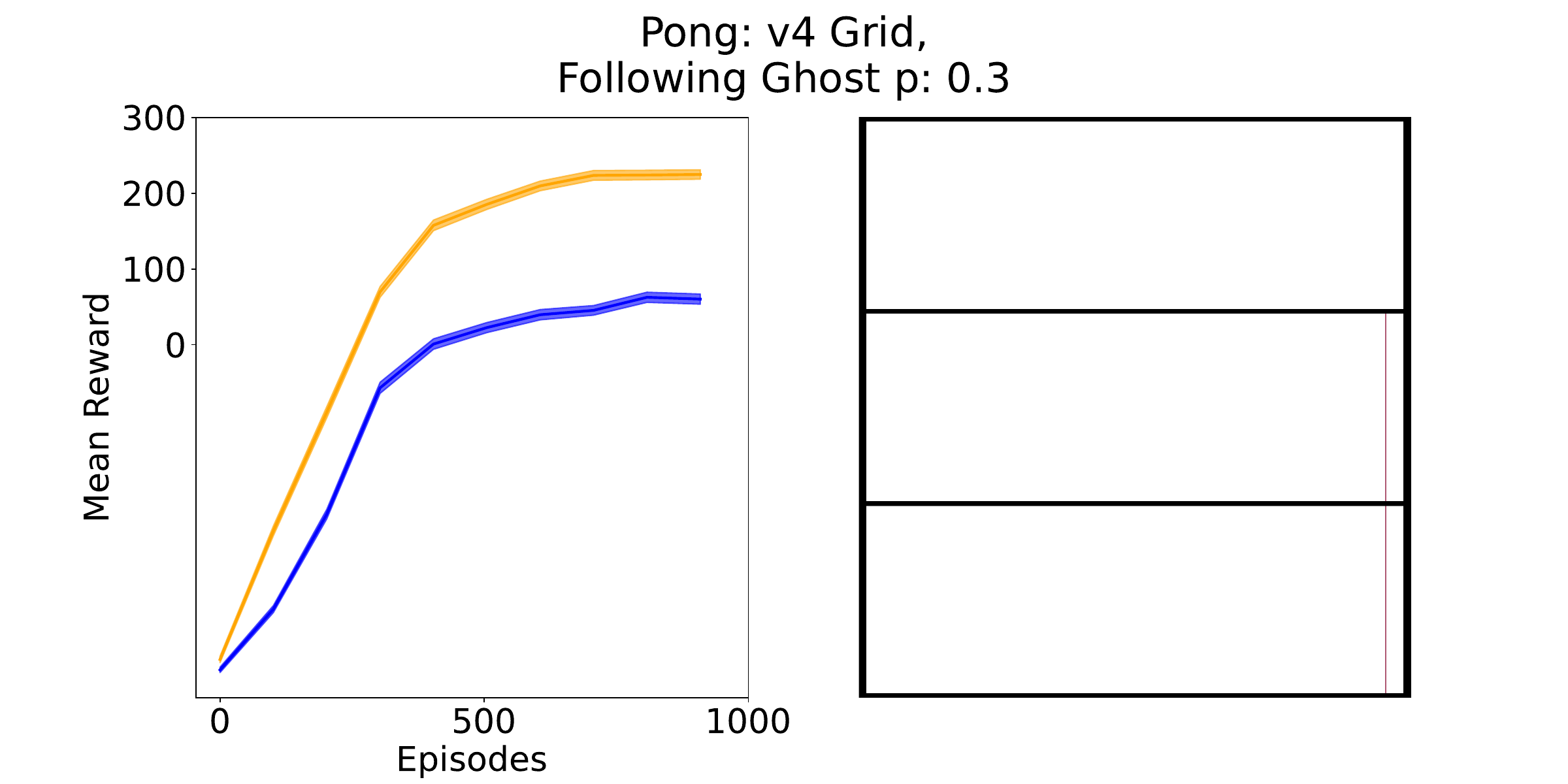}
  \end{subfigure}
  \hfill
  \begin{subfigure}{0.30\textwidth}
    \includegraphics[width=\linewidth]{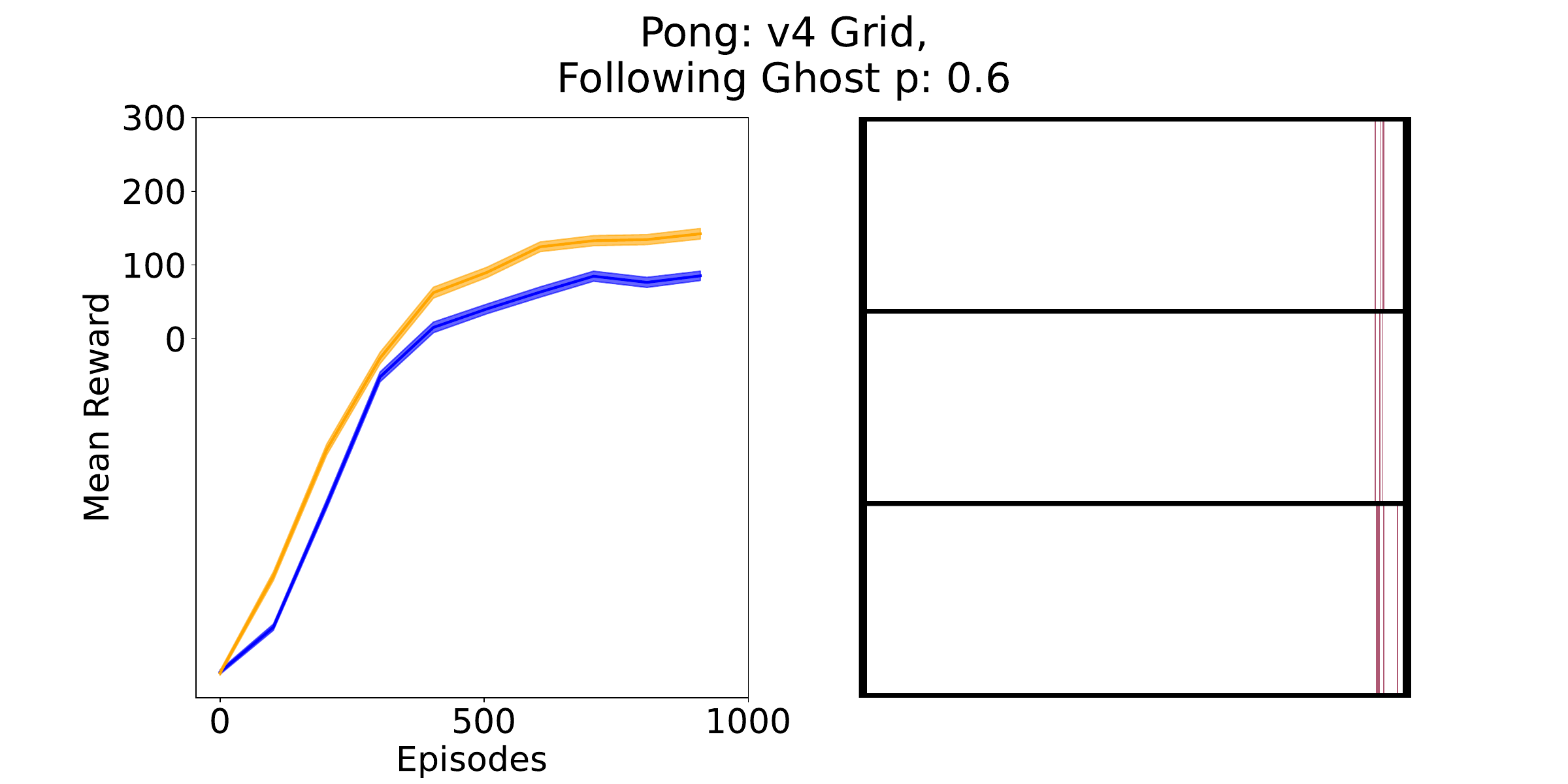}
  \end{subfigure}\\

  \caption{\emph{Q-learning Agent with Boltzmann exploration strategy}: The \textit{exploration grid} visualizing the difference in State-Action (S-A) pairs explored by these agents ($D_{LG}$). Results for PacMan v2, v3, v4 grids, the agent is trained on non-noisy variations of different environments (reported in the headings) and tested in the Low-Noise regime. Rows in the right figure represents agent's actions Left, Right, Up, or Down.}
  \label{fig:atari_variations-exploration-pacman-qlearning-boltzmann}
\end{figure*}

\begin{figure*}[t]
  %\centering
  \begin{subfigure}{0.3\textwidth}
    \includegraphics[width=\linewidth]{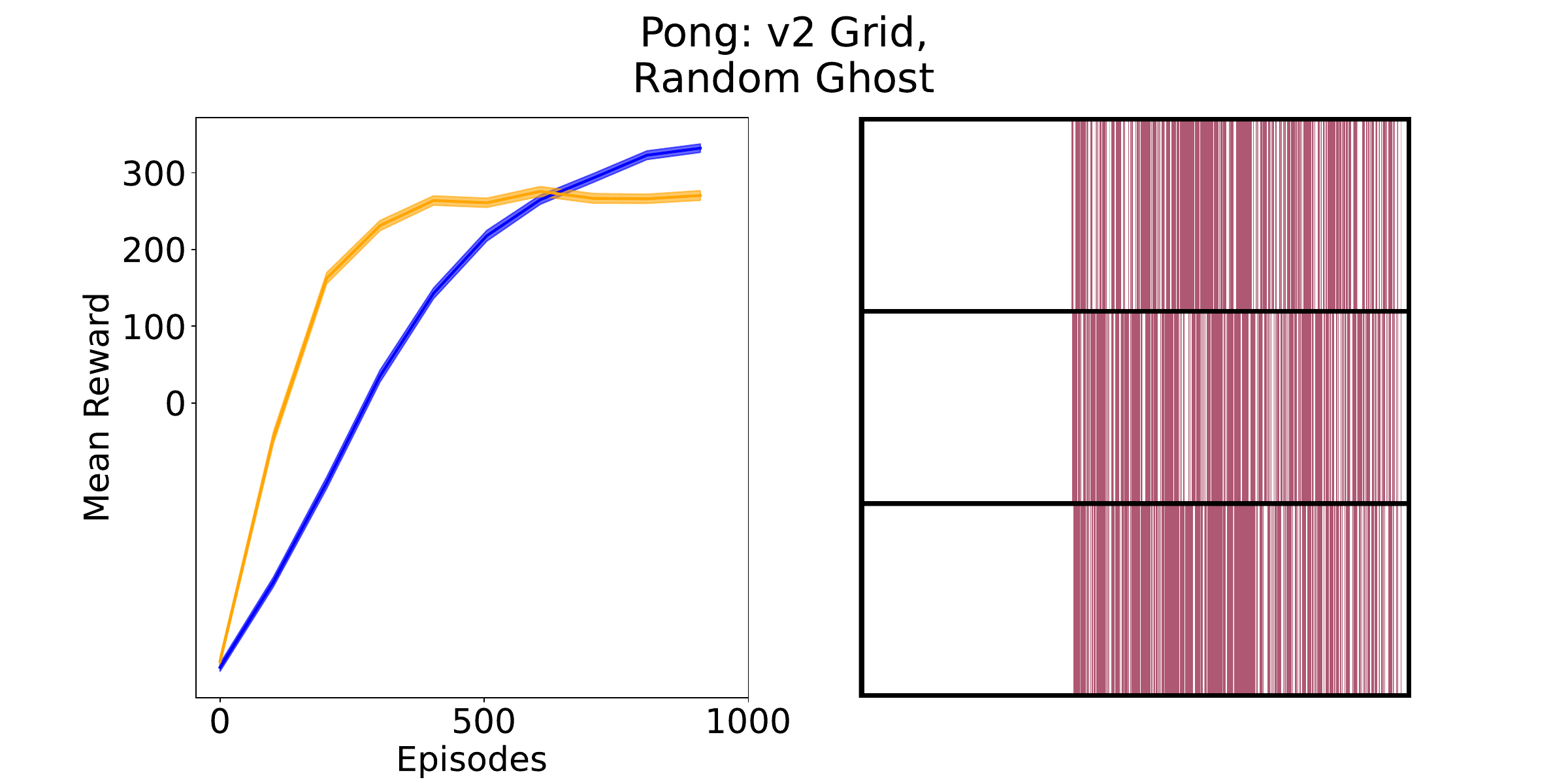}
  \end{subfigure}
  \hfill
  \begin{subfigure}{0.3\textwidth}
    \includegraphics[width=\linewidth]{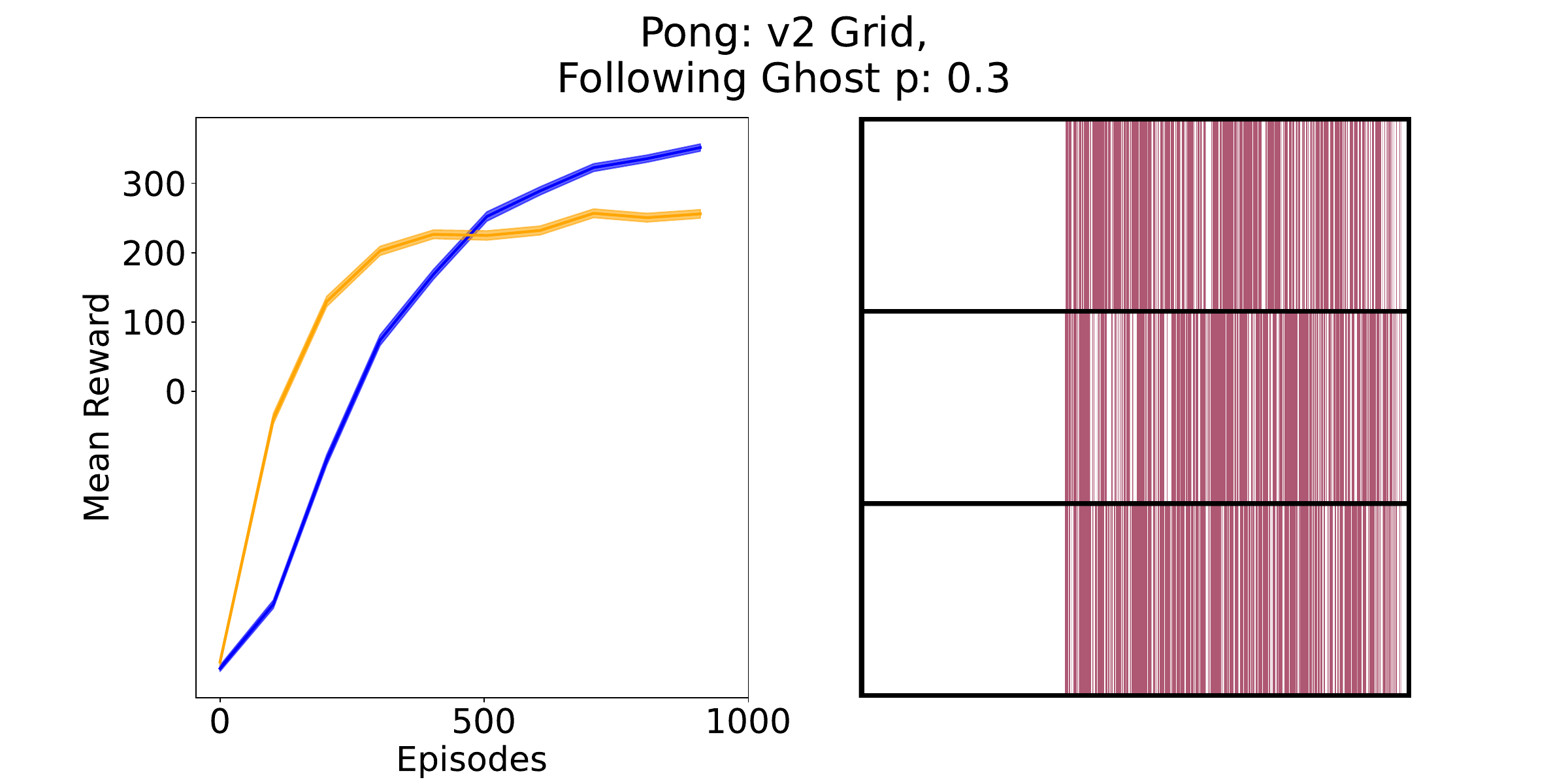}
  \end{subfigure}
  \hfill
  \begin{subfigure}{0.3\textwidth}
    \includegraphics[width=\linewidth]{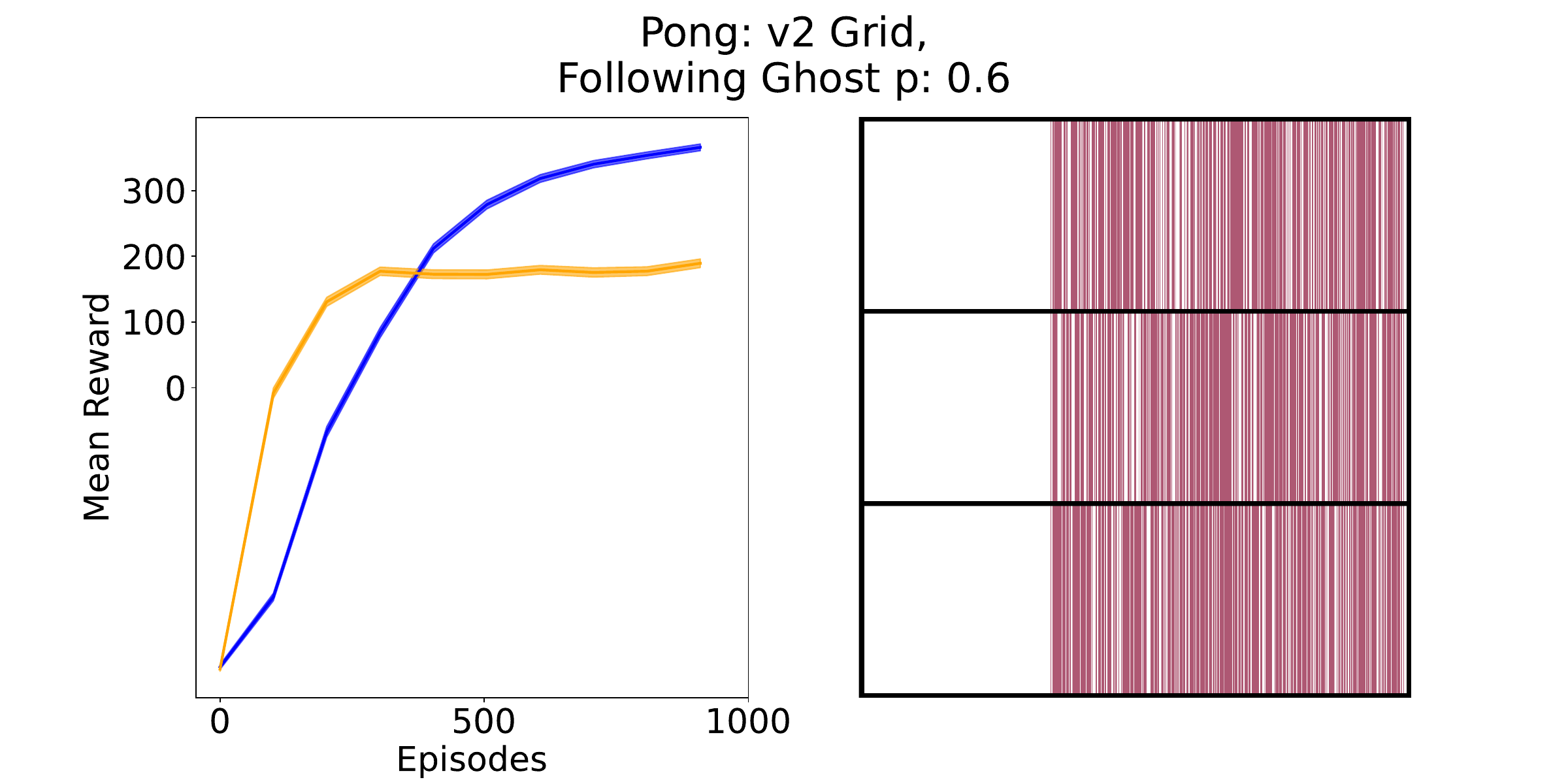}
  \end{subfigure}
  
  \begin{subfigure}{0.3\textwidth}
    \includegraphics[width=\linewidth]{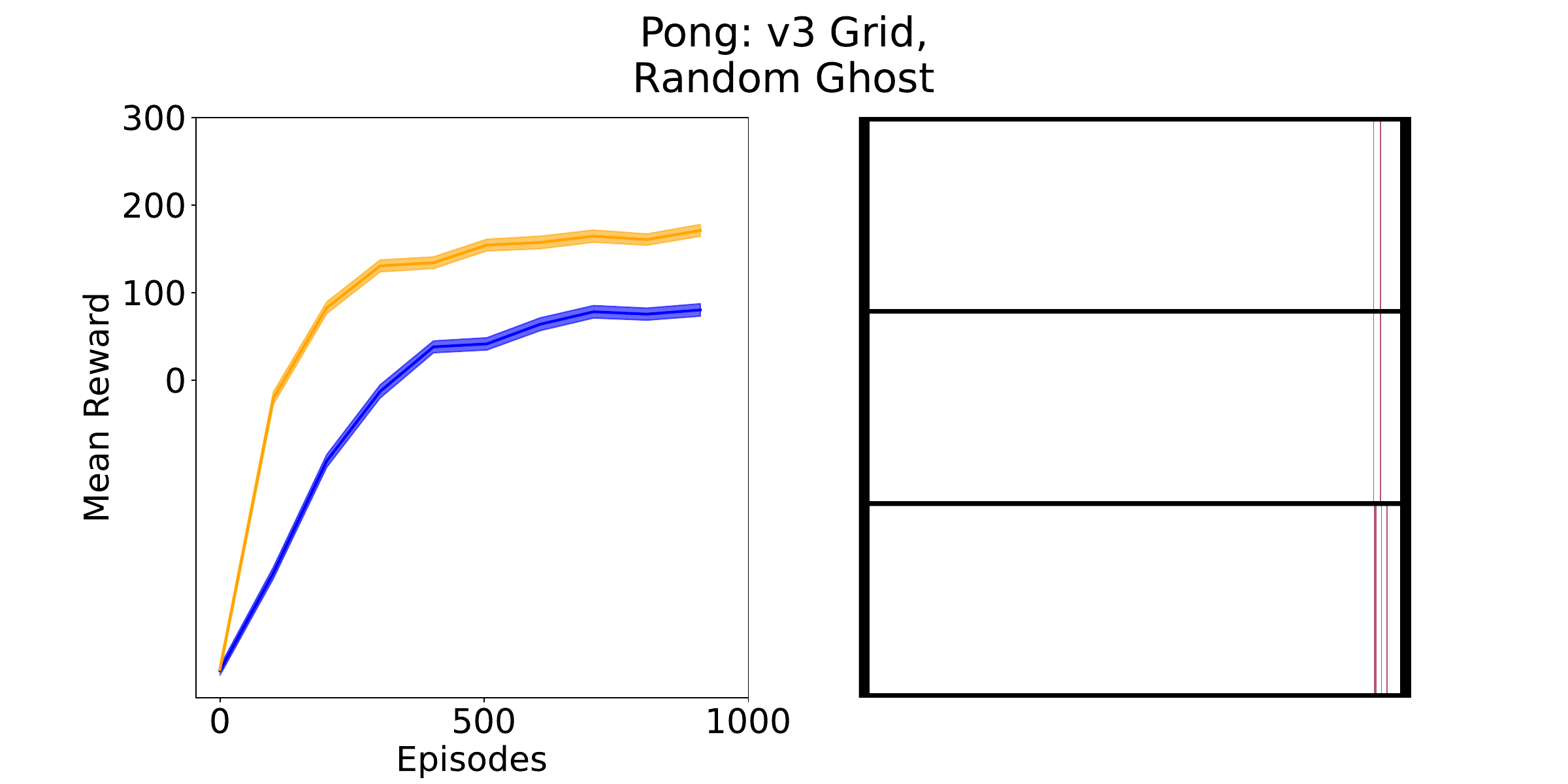}
  \end{subfigure}
  \hfill
  \begin{subfigure}{0.3\textwidth}
    \includegraphics[width=\linewidth]{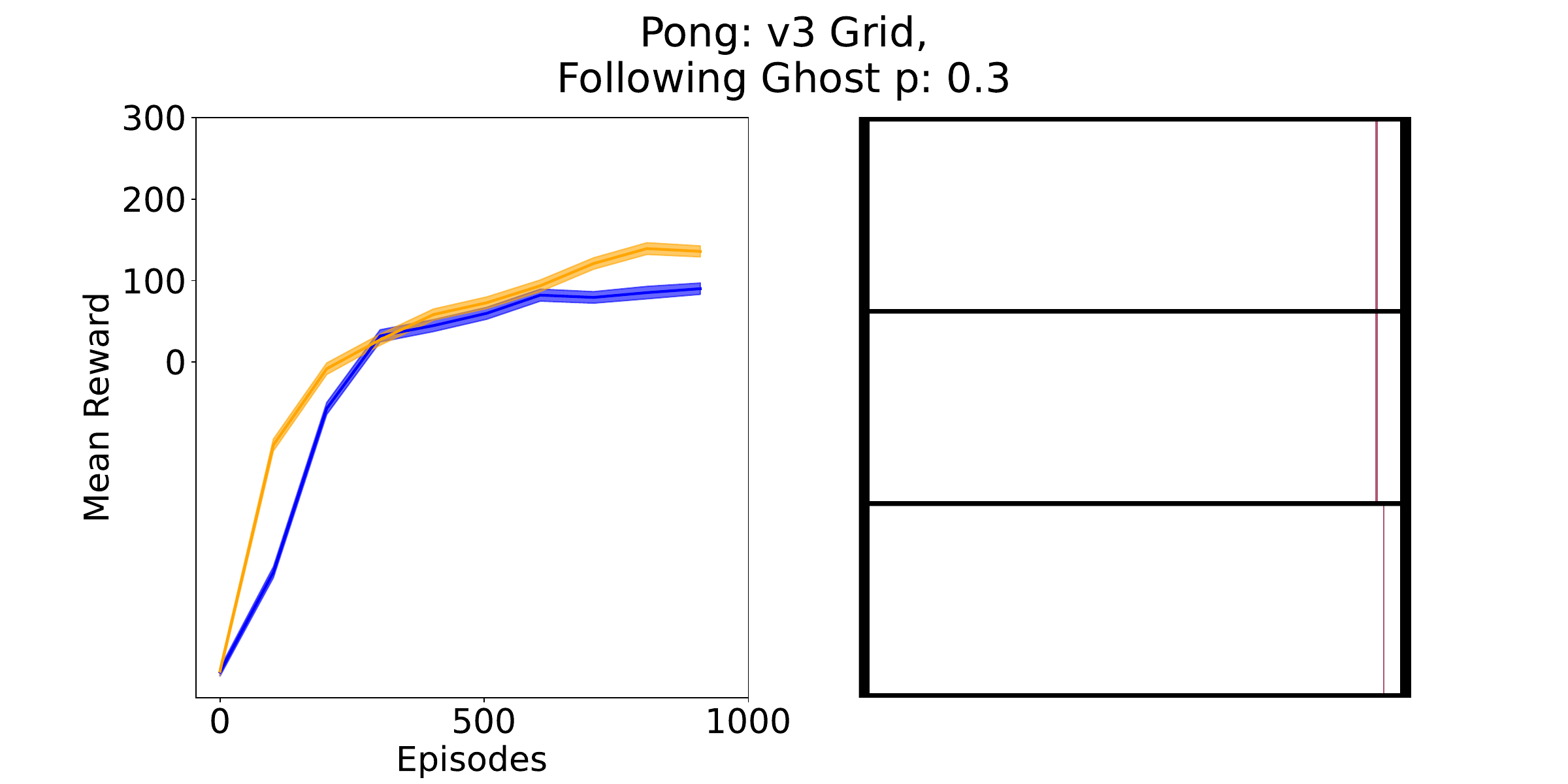}
  \end{subfigure}
  \hfill
  \begin{subfigure}{0.3\textwidth}
    \includegraphics[width=\linewidth]{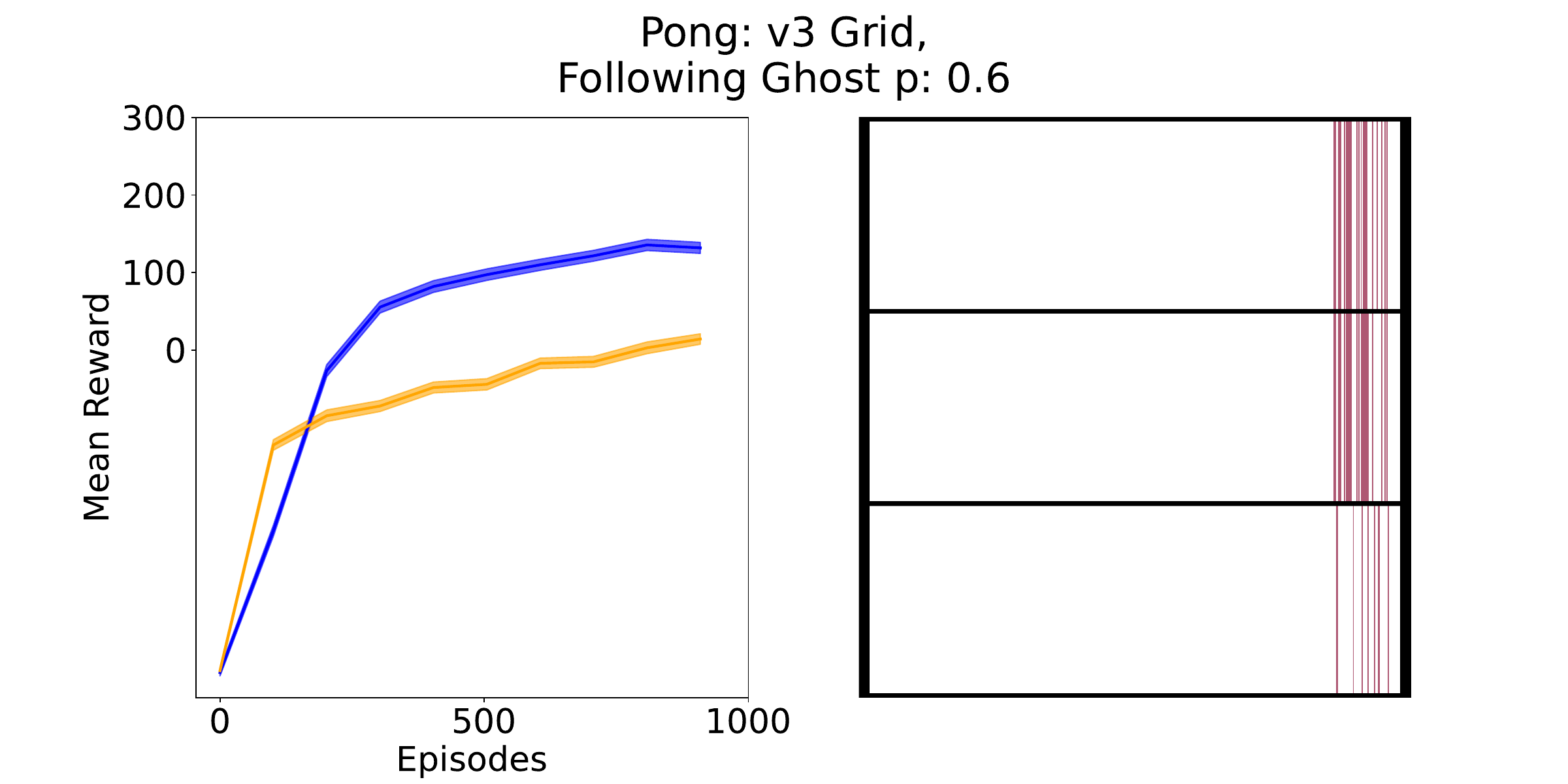}
  \end{subfigure}\\
  \hfill
  \begin{subfigure}{0.30\textwidth}
    \includegraphics[width=\linewidth]{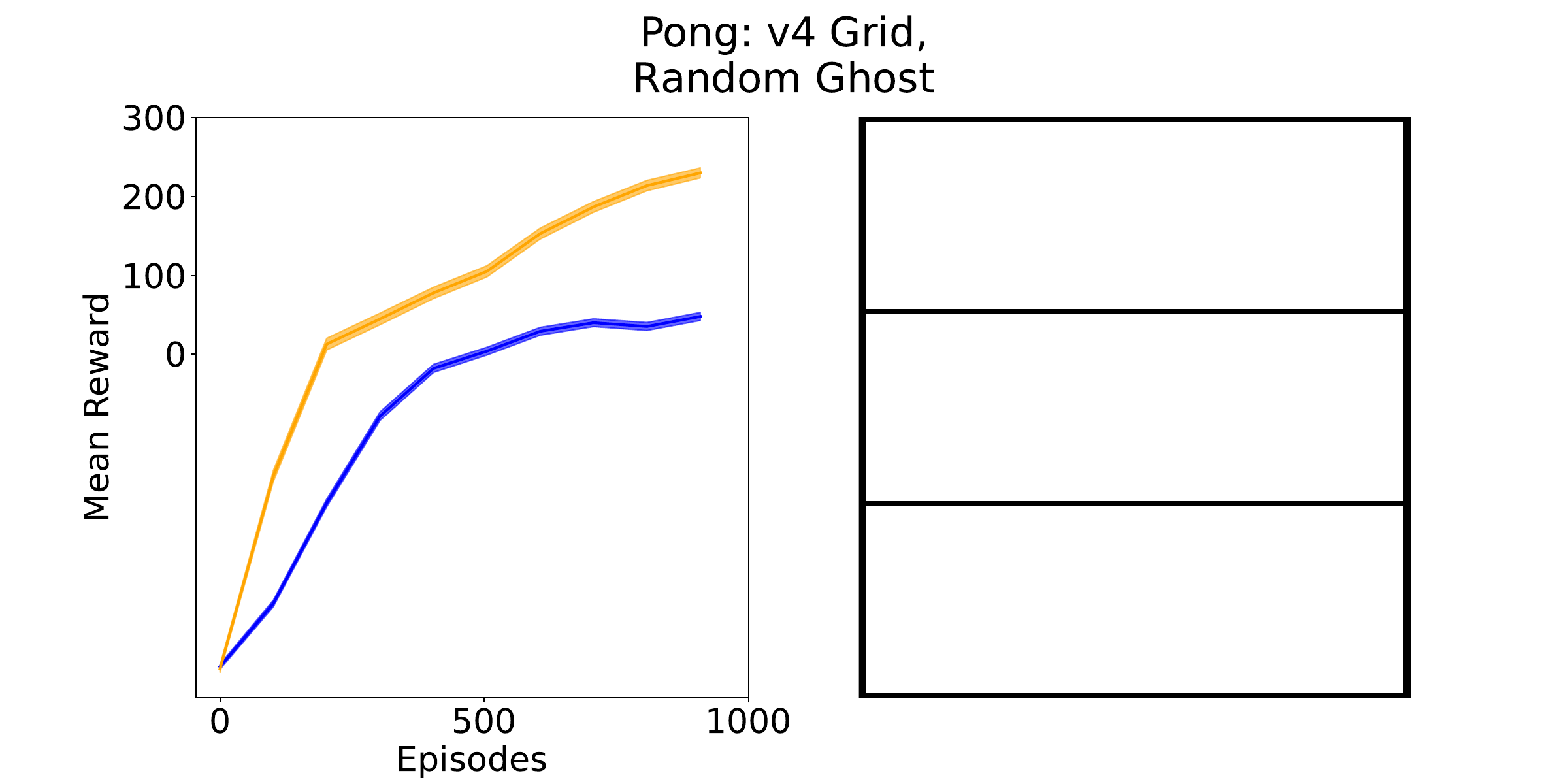}
  \end{subfigure}
  \hfill
  \begin{subfigure}{0.30\textwidth}
    \includegraphics[width=\linewidth]{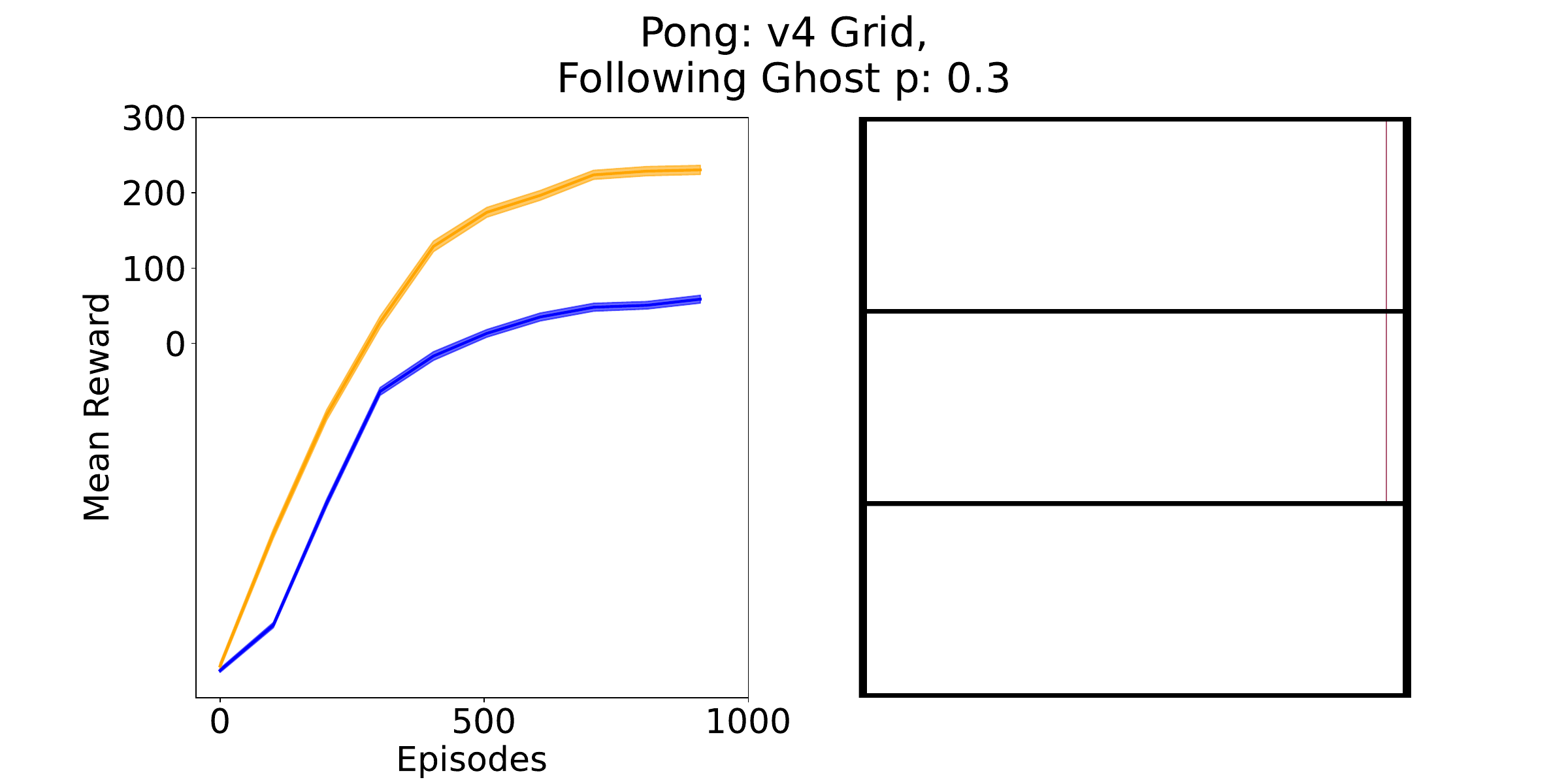}
  \end{subfigure}
  \hfill
  \begin{subfigure}{0.30\textwidth}
    \includegraphics[width=\linewidth]{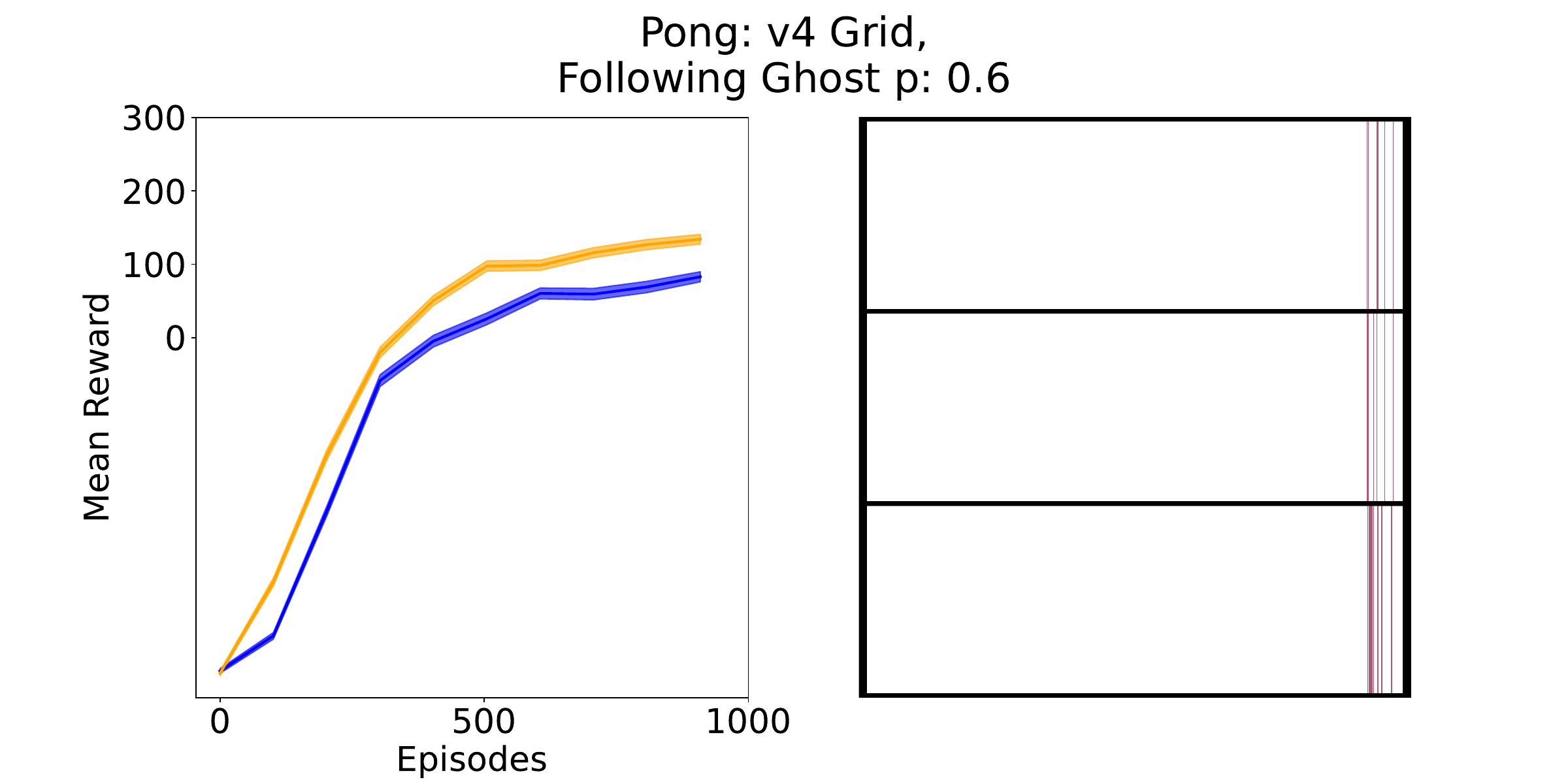}
  \end{subfigure}\\

  \caption{\emph{Q-learning Agent with $\epsilon\text{-}$greedy exploration strategy}: The \textit{exploration grid} visualizing the difference in State-Action (S-A) pairs explored by these agents ($D_{LG}$). Results for PacMan v2, v3, v4 grids, the agent is trained on non-noisy variations of different environments (reported in the headings) and tested in the Low-Noise regime. Rows in the right figure represents agent's actions Left, Right, Up, or Down.}
  \label{fig:atari_variations-exploration-pacman-qlearning-egreedy}
\end{figure*}

\begin{figure*}[t]
  %\centering
  \begin{subfigure}{0.3\textwidth}
    \includegraphics[width=\linewidth]{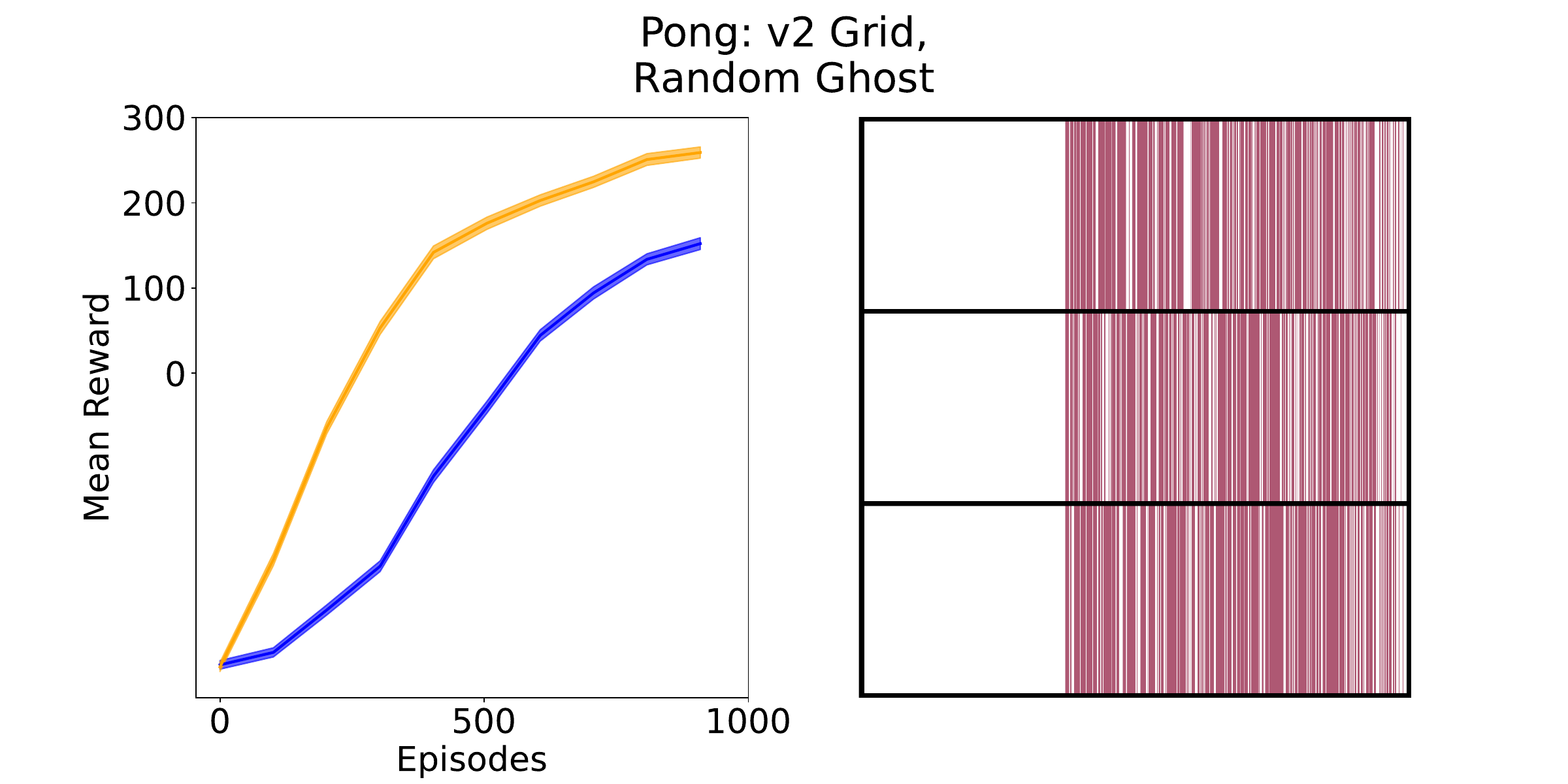}
  \end{subfigure}
  \hfill
  \begin{subfigure}{0.3\textwidth}
    \includegraphics[width=\linewidth]{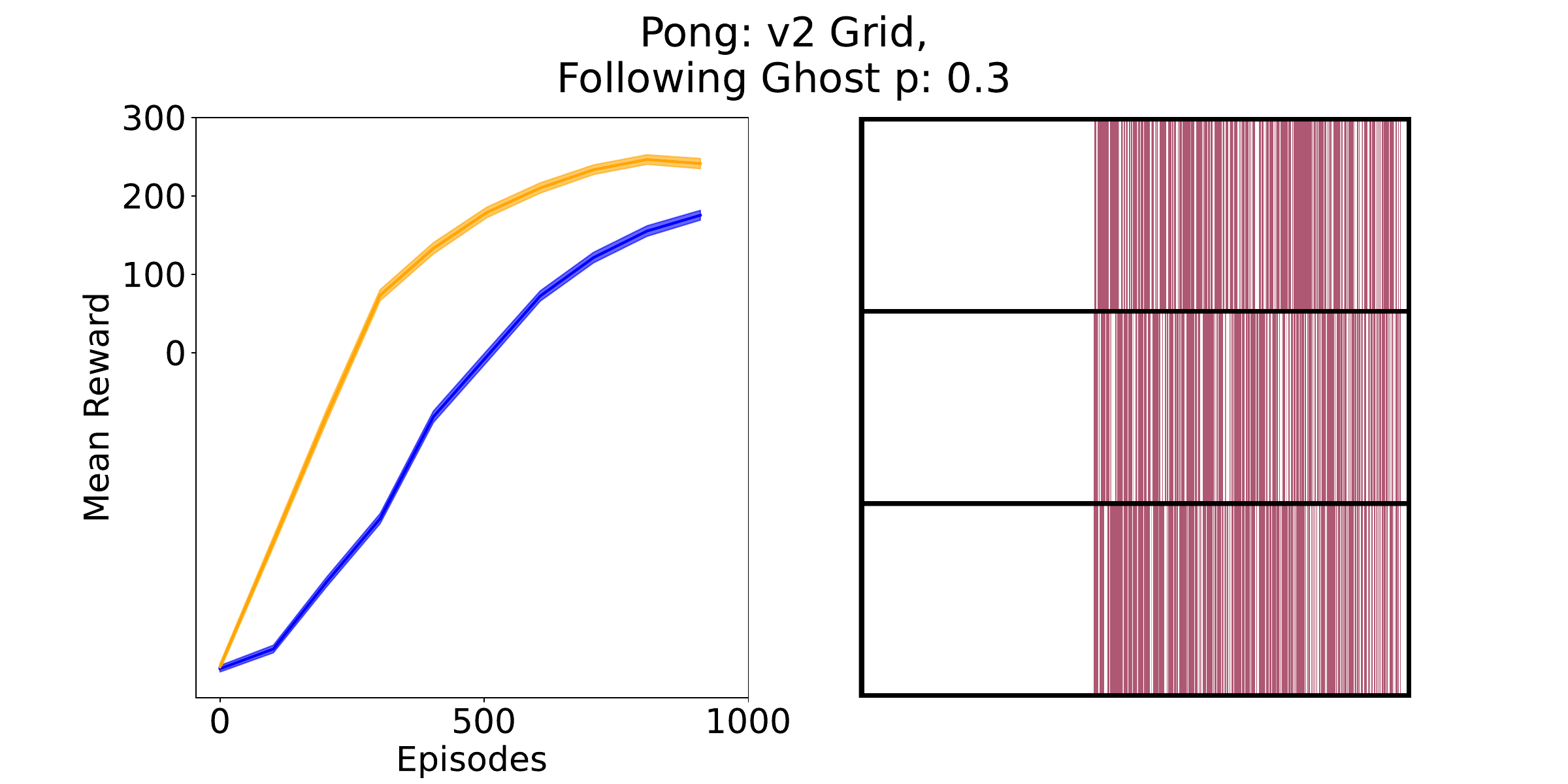}
  \end{subfigure}
  \hfill
  \begin{subfigure}{0.3\textwidth}
    \includegraphics[width=\linewidth]{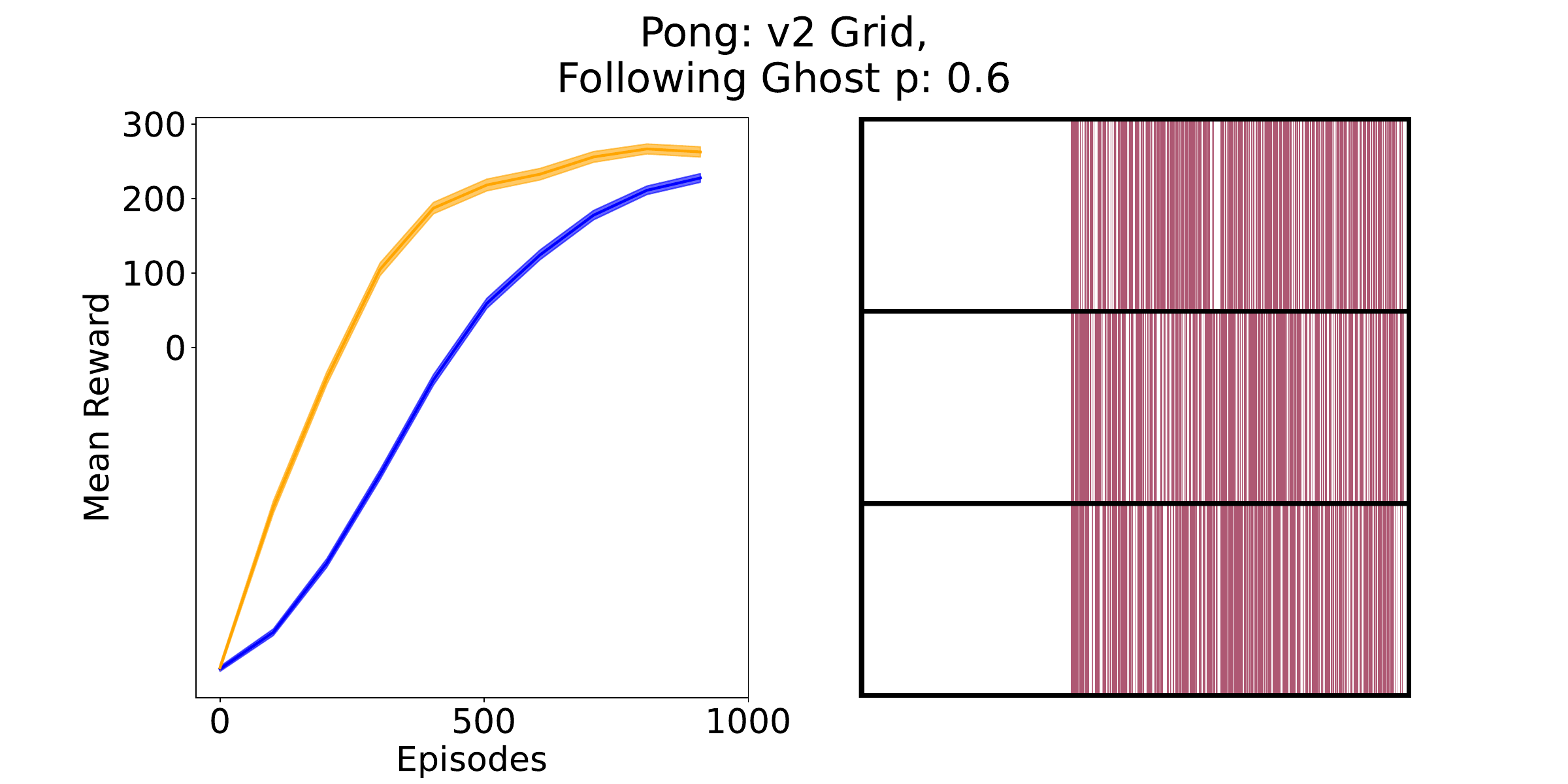}
  \end{subfigure}
  
  \begin{subfigure}{0.3\textwidth}
    \includegraphics[width=\linewidth]{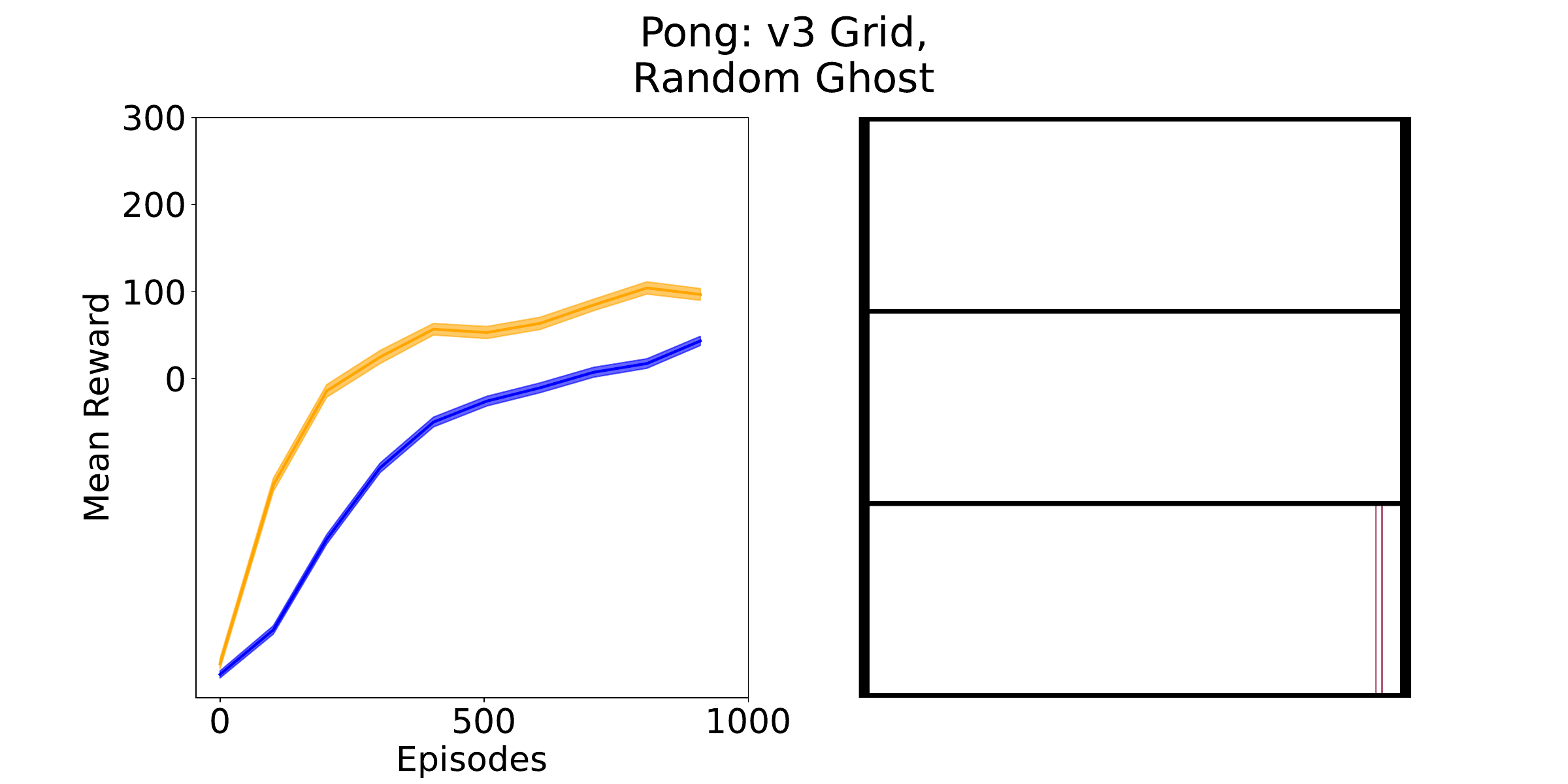}
  \end{subfigure}
  \hfill
  \begin{subfigure}{0.3\textwidth}
    \includegraphics[width=\linewidth]{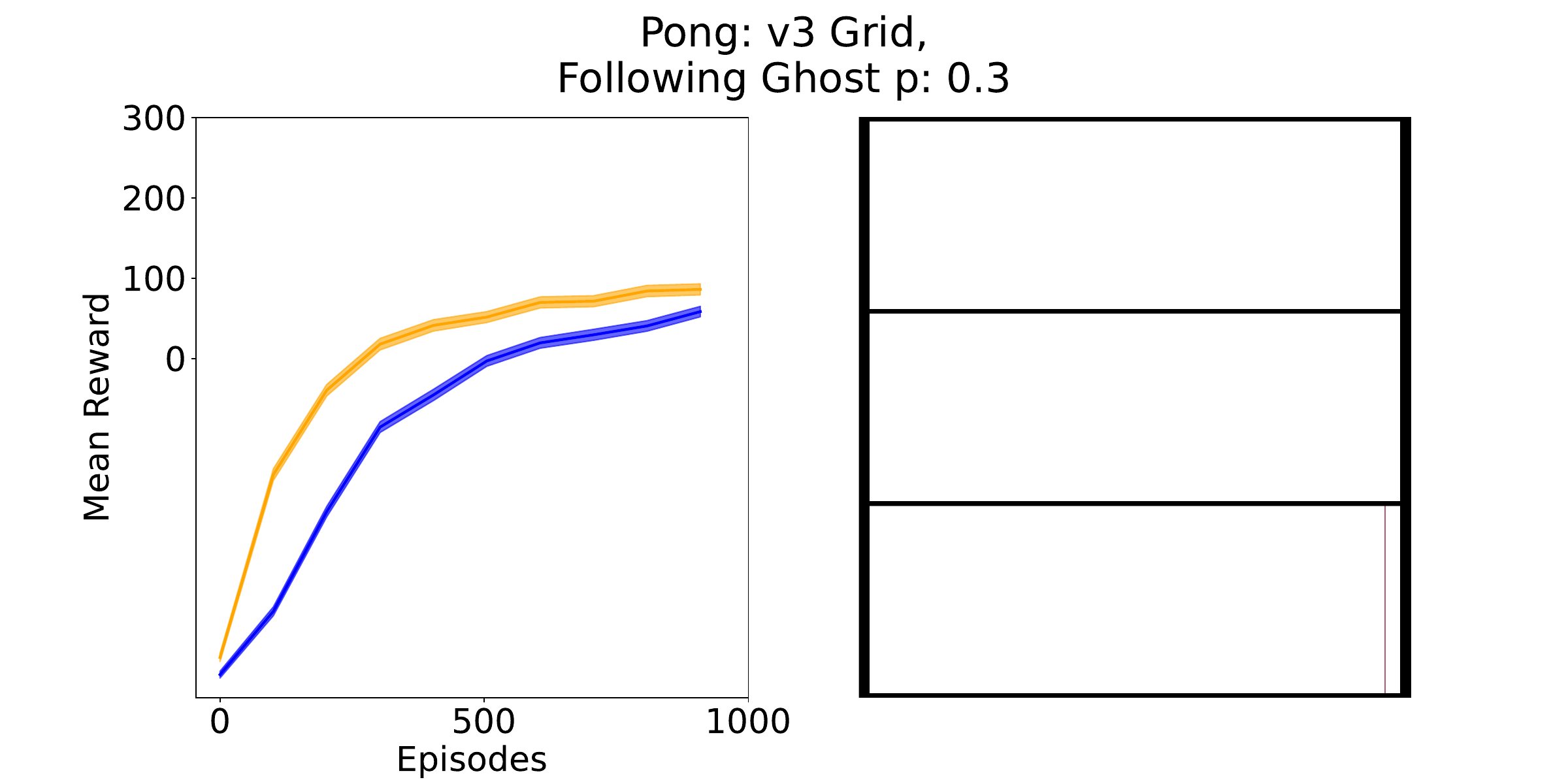}
  \end{subfigure}
  \hfill
  \begin{subfigure}{0.3\textwidth}
    \includegraphics[width=\linewidth]{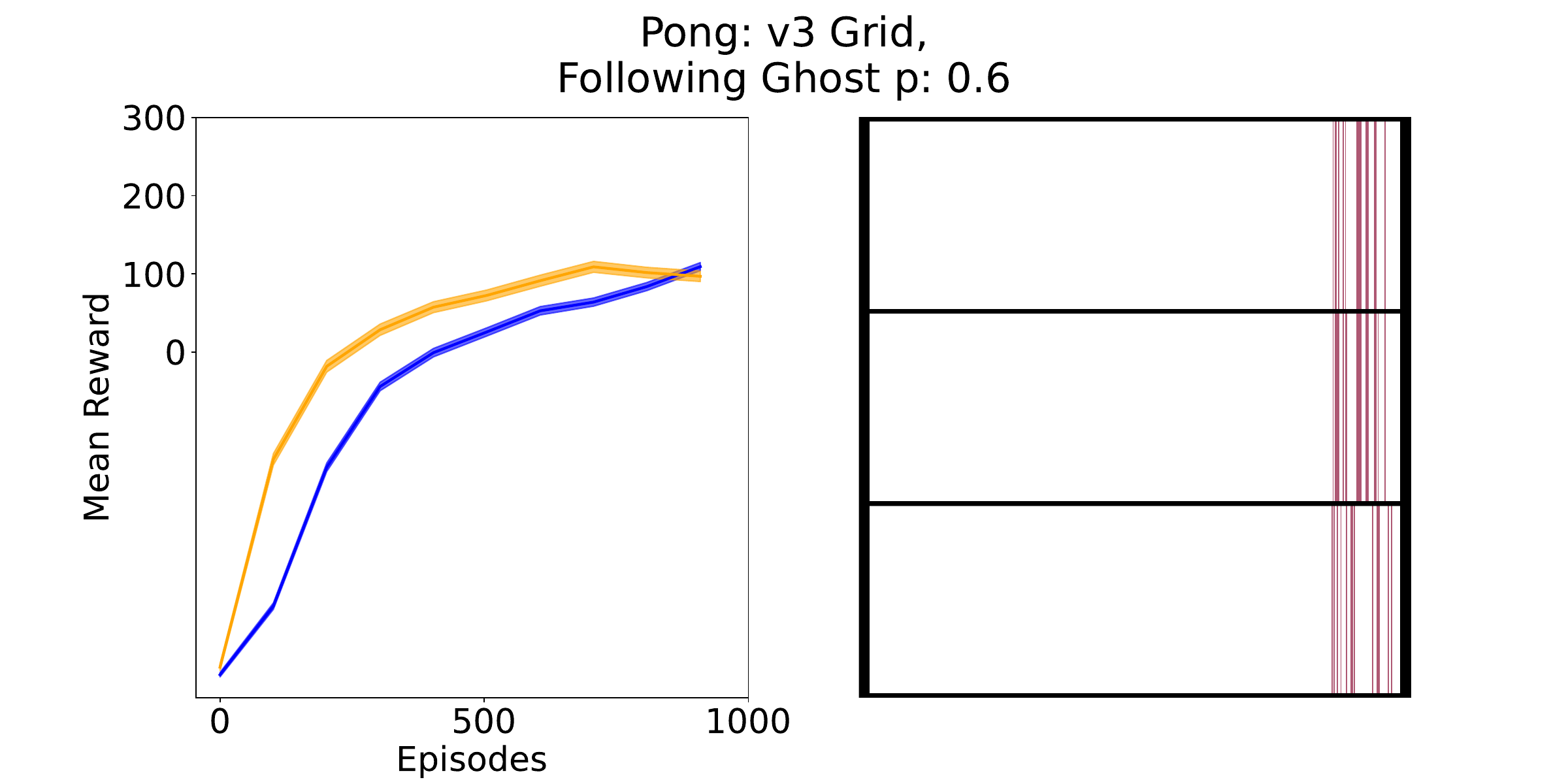}
  \end{subfigure}\\
  \hfill
  \begin{subfigure}{0.30\textwidth}
    \includegraphics[width=\linewidth]{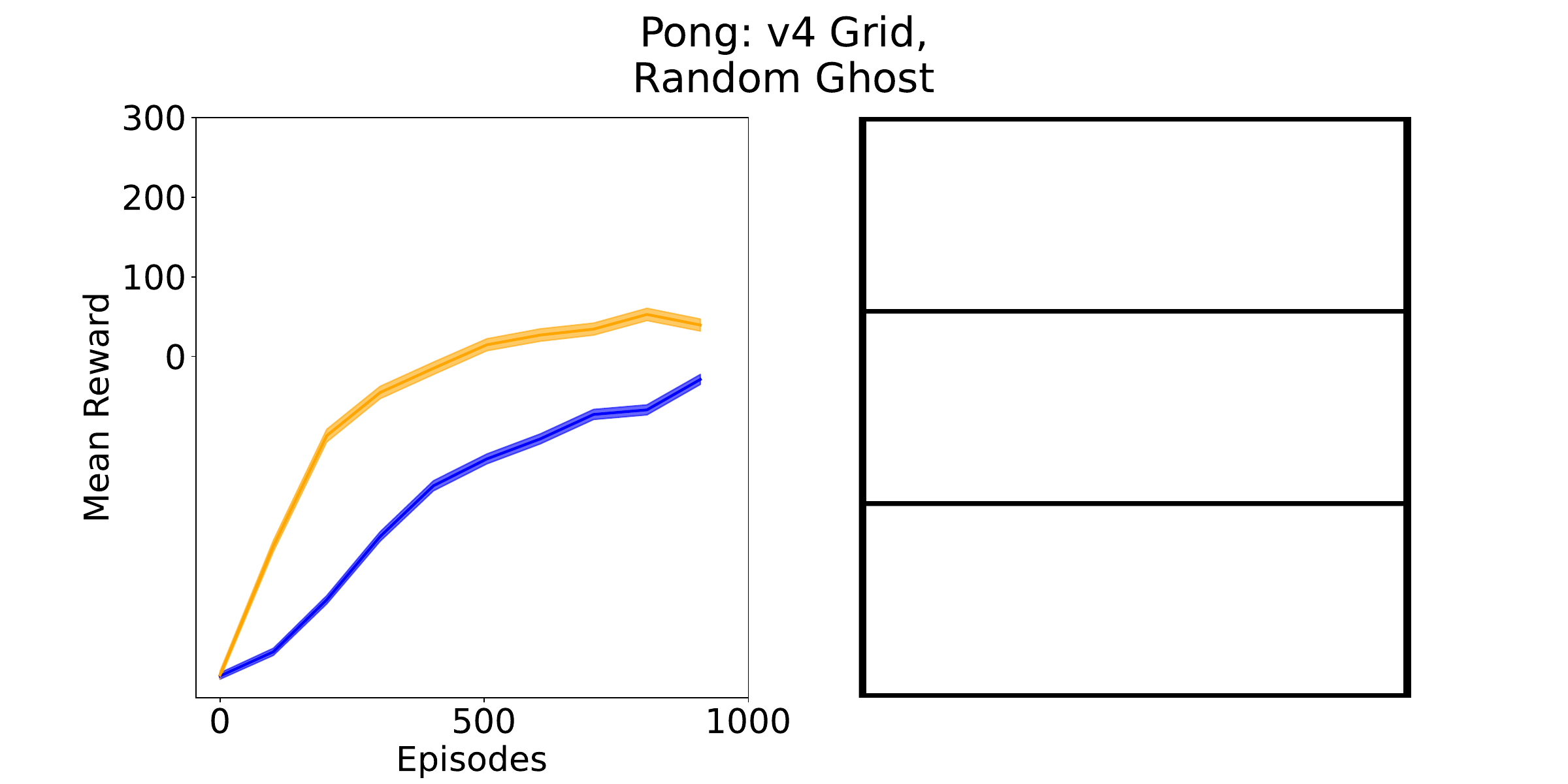}
  \end{subfigure}
  \hfill
  \begin{subfigure}{0.30\textwidth}
    \includegraphics[width=\linewidth]{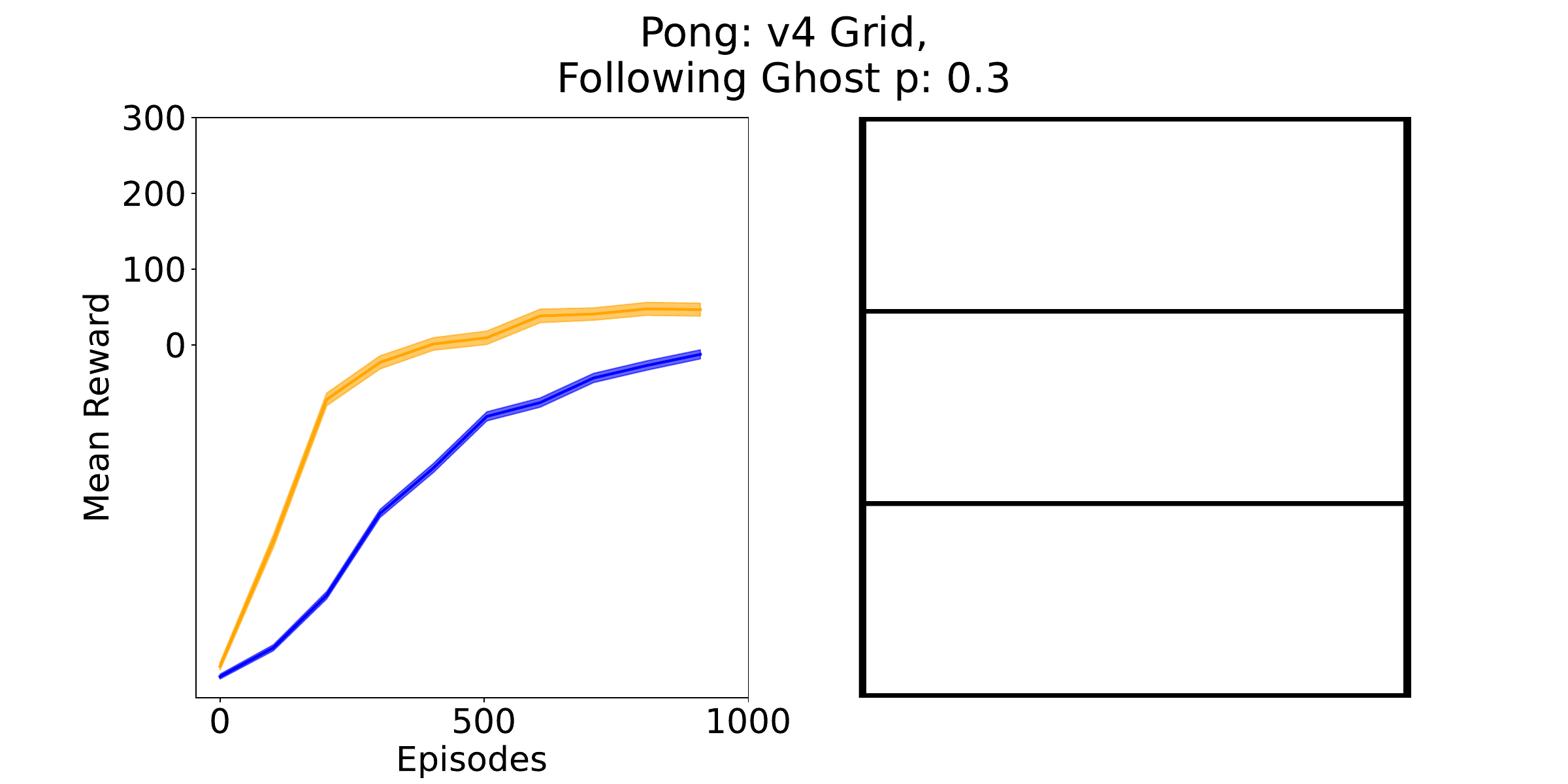}
  \end{subfigure}
  \hfill
  \begin{subfigure}{0.30\textwidth}
    \includegraphics[width=\linewidth]{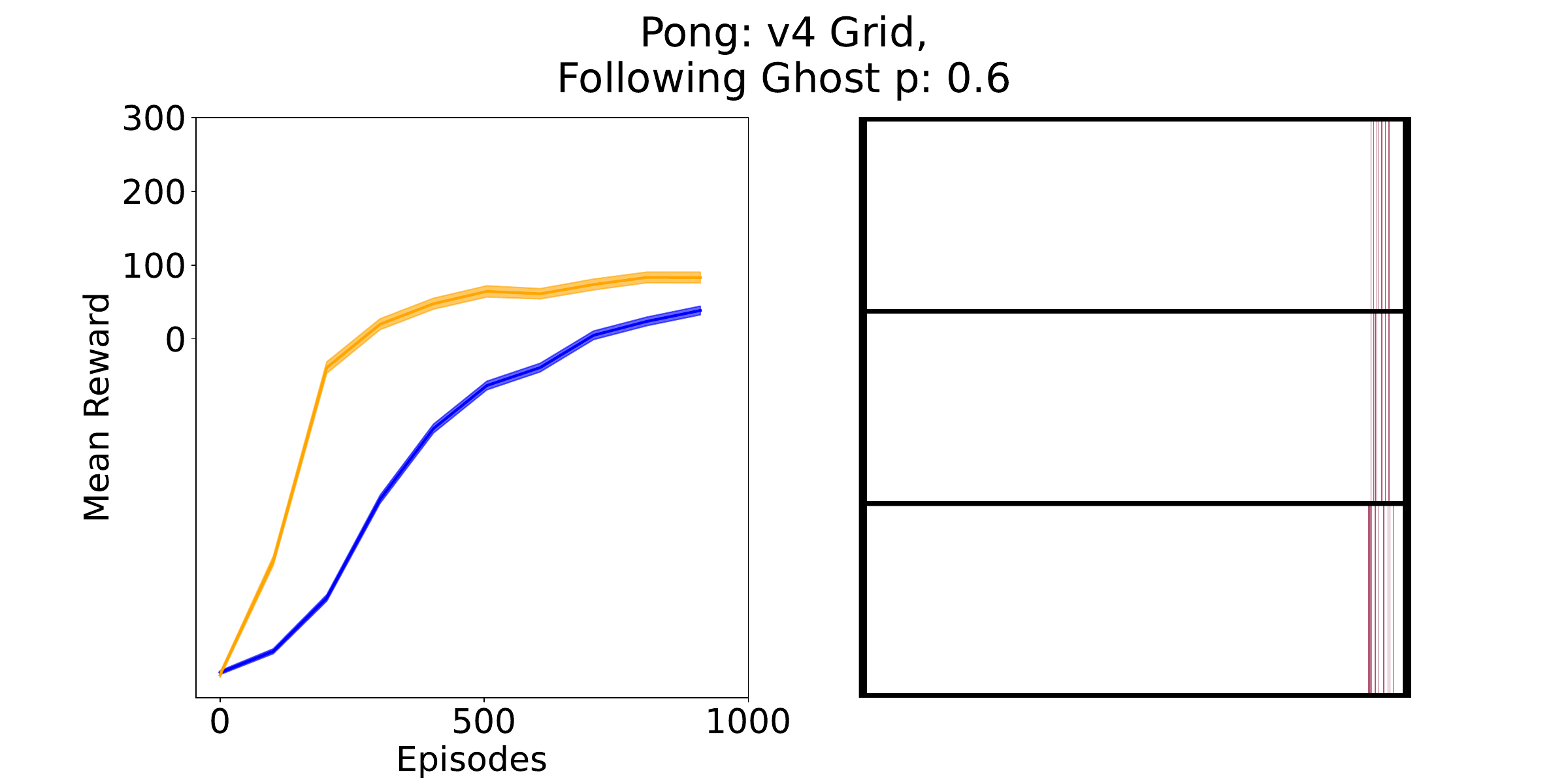}
  \end{subfigure}\\

  \caption{\emph{SARSA Agent with Boltzmann exploration strategy}: The \textit{exploration grid} visualizing the difference in State-Action (S-A) pairs explored by these agents ($D_{LG}$). Results for PacMan v2, v3, v4 grids, the agent is trained on non-noisy variations of different environments (reported in the headings) and tested in the Low-Noise regime. Rows in the right figure represents agent's actions Left, Right, Up, or Down.}
  \label{fig:atari_variations-exploration-pacman-sarsa-boltzmann}
\end{figure*}

\begin{figure*}[t]
  %\centering
  \begin{subfigure}{0.3\textwidth}
    \includegraphics[width=\linewidth]{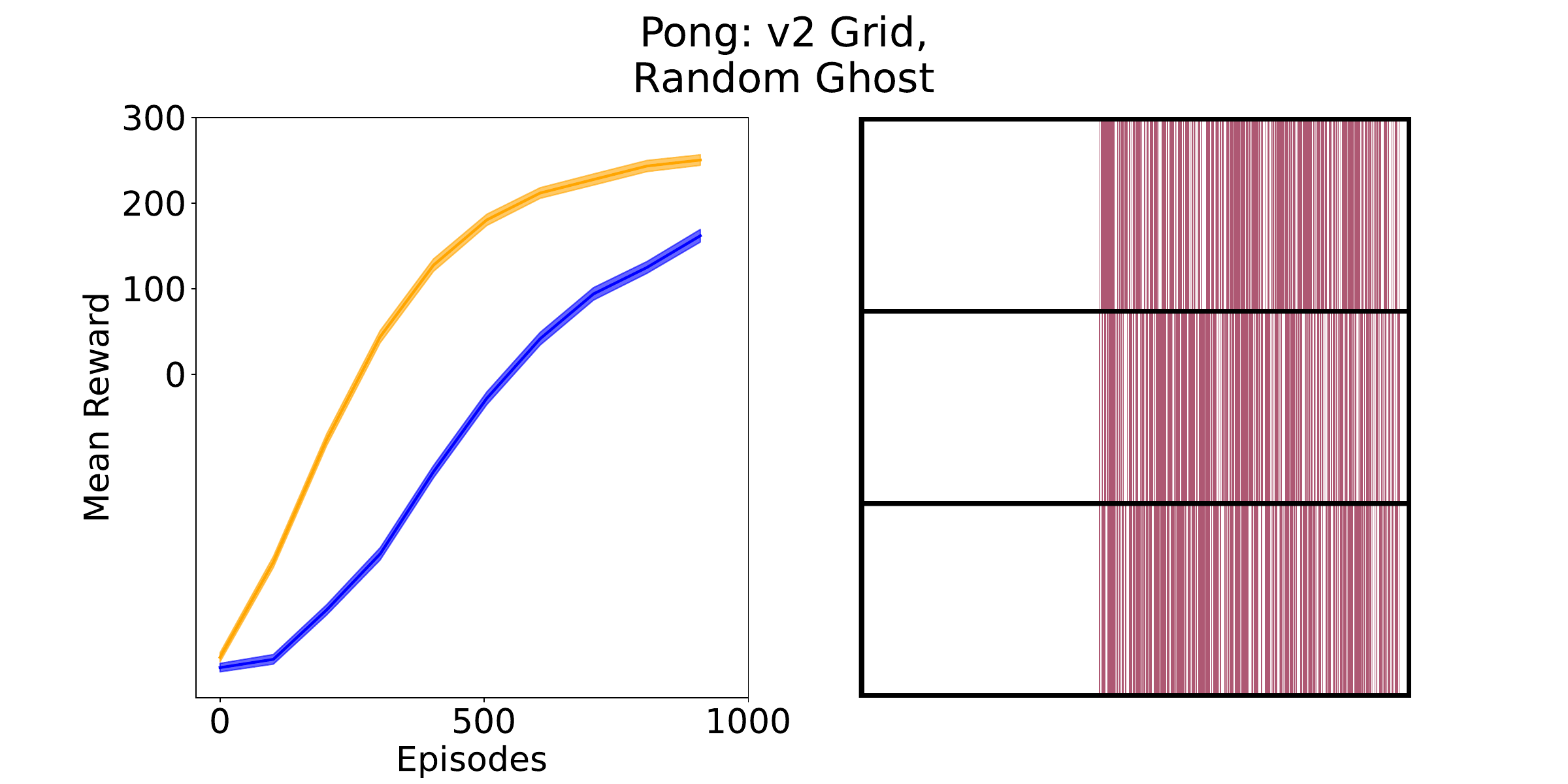}
  \end{subfigure}
  \hfill
  \begin{subfigure}{0.3\textwidth}
    \includegraphics[width=\linewidth]{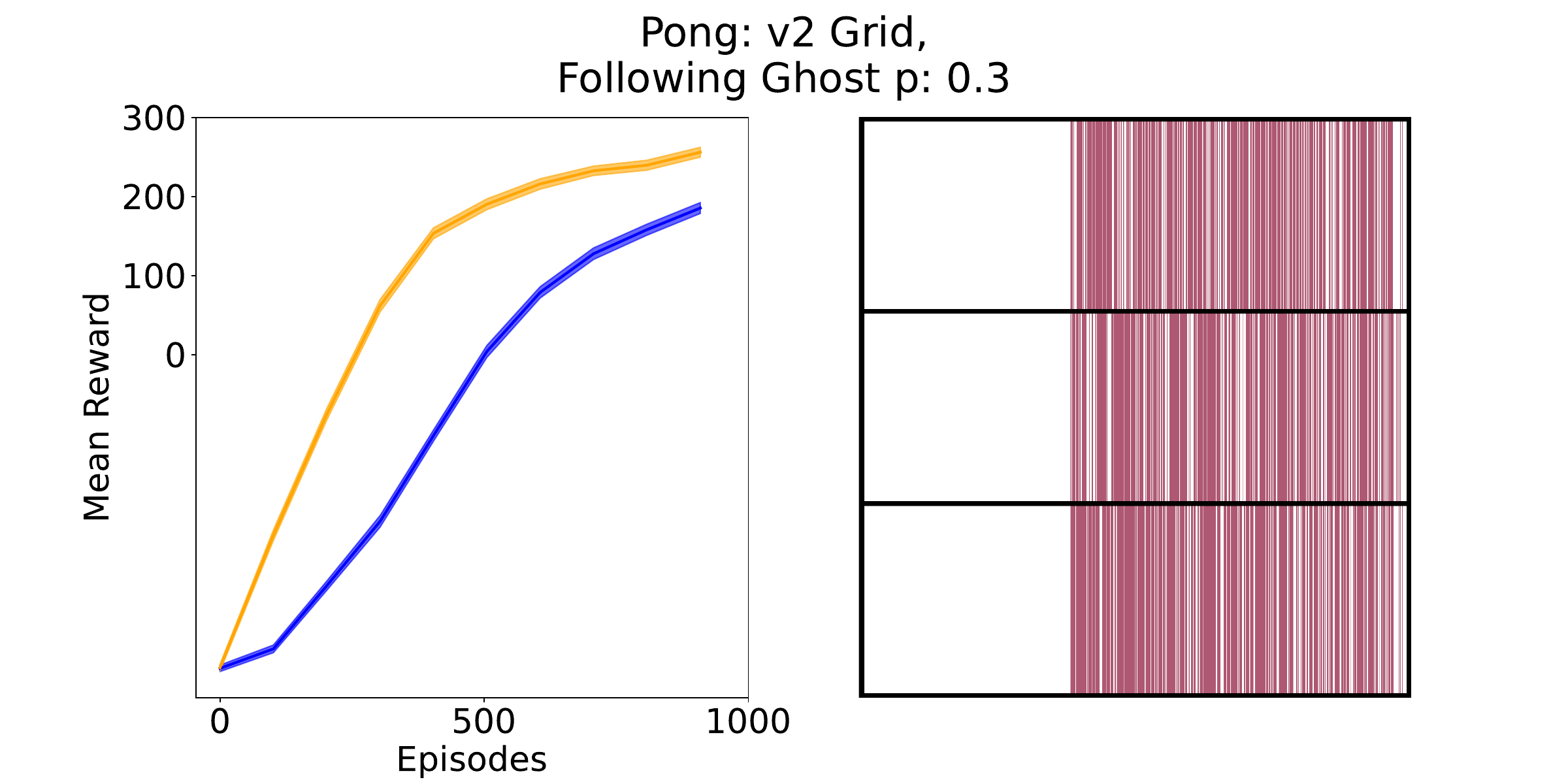}
  \end{subfigure}
  \hfill
  \begin{subfigure}{0.3\textwidth}
    \includegraphics[width=\linewidth]{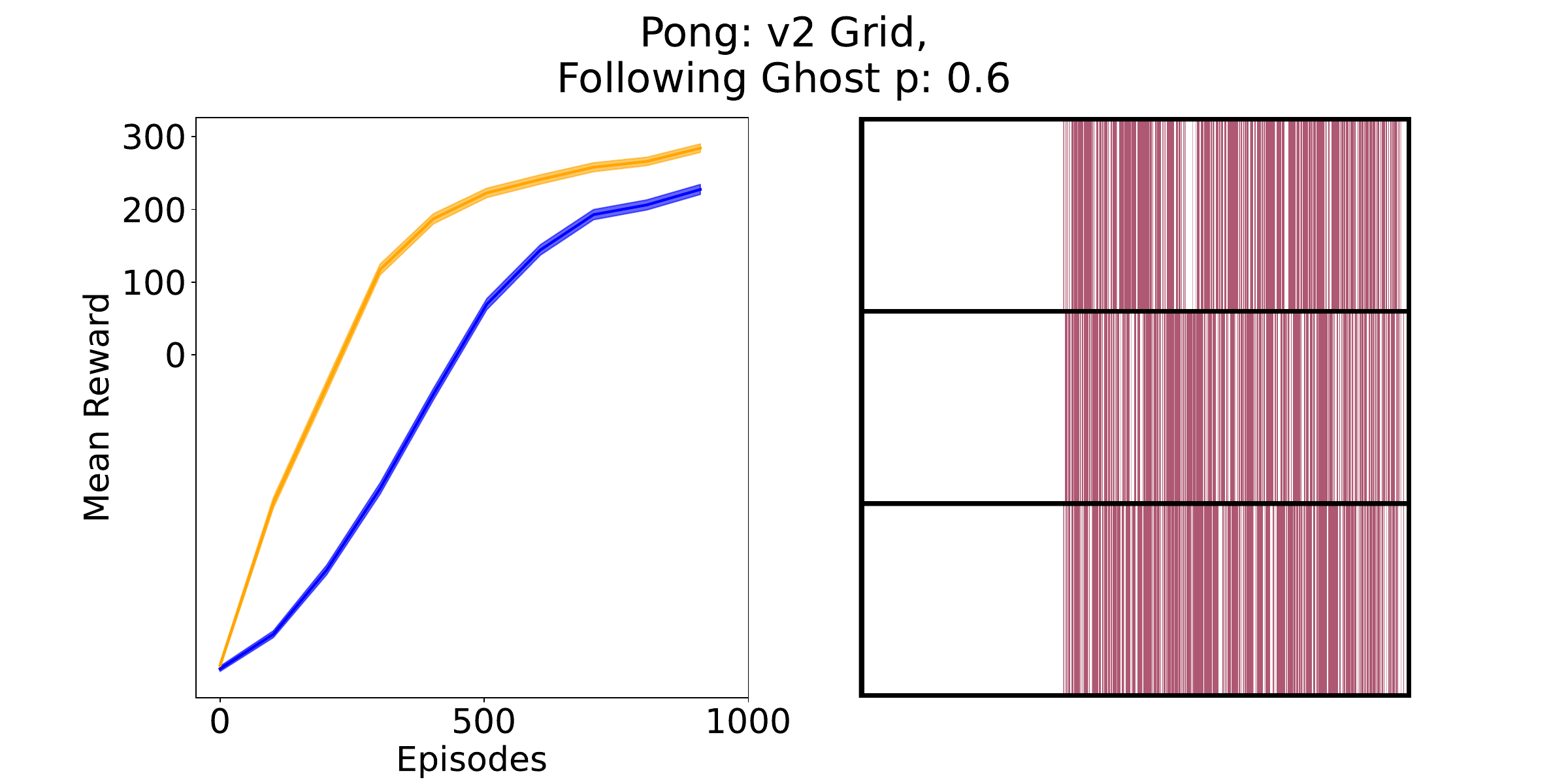}
  \end{subfigure}
  
  \begin{subfigure}{0.3\textwidth}
    \includegraphics[width=\linewidth]{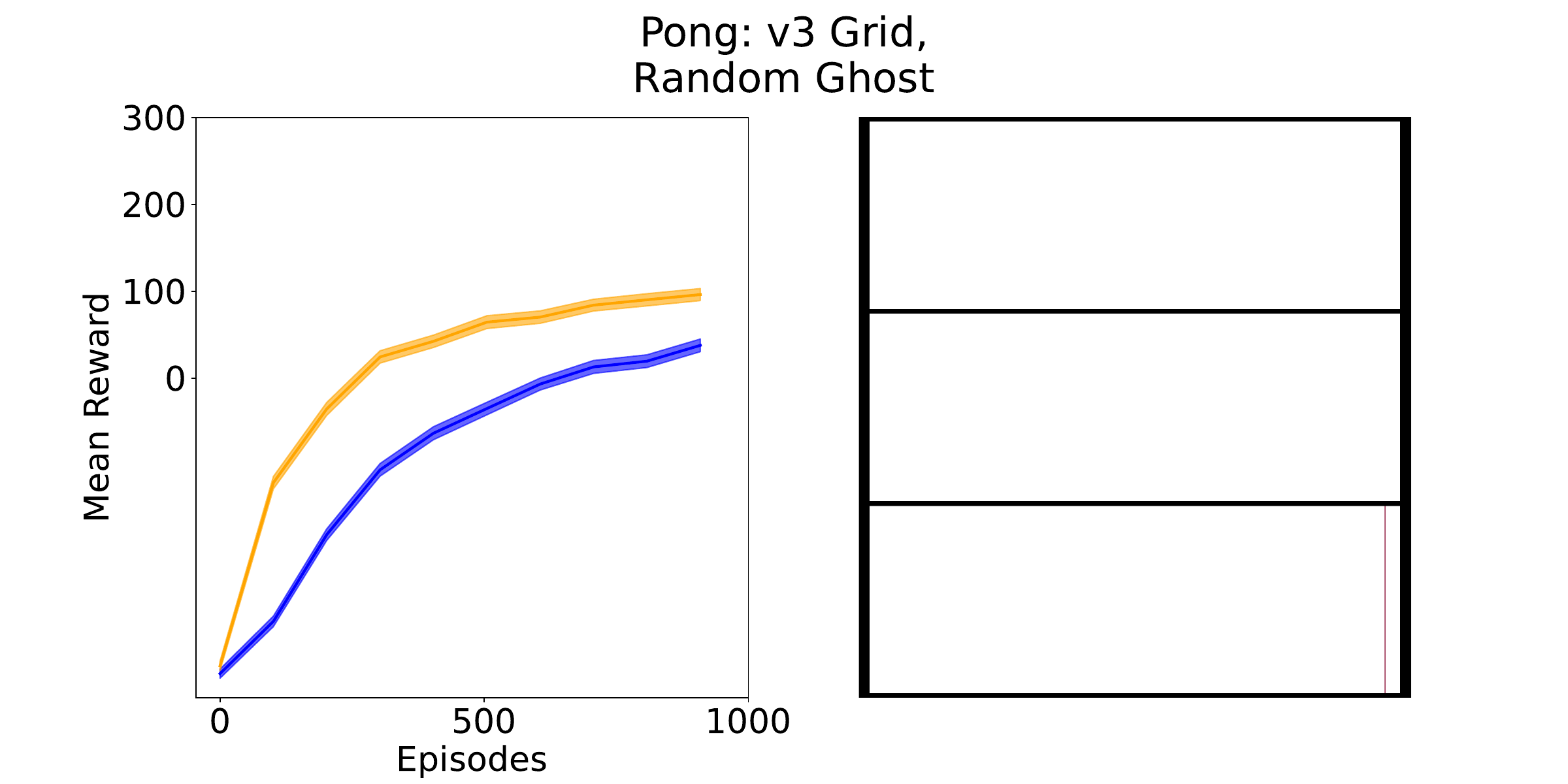}
  \end{subfigure}
  \hfill
  \begin{subfigure}{0.3\textwidth}
    \includegraphics[width=\linewidth]{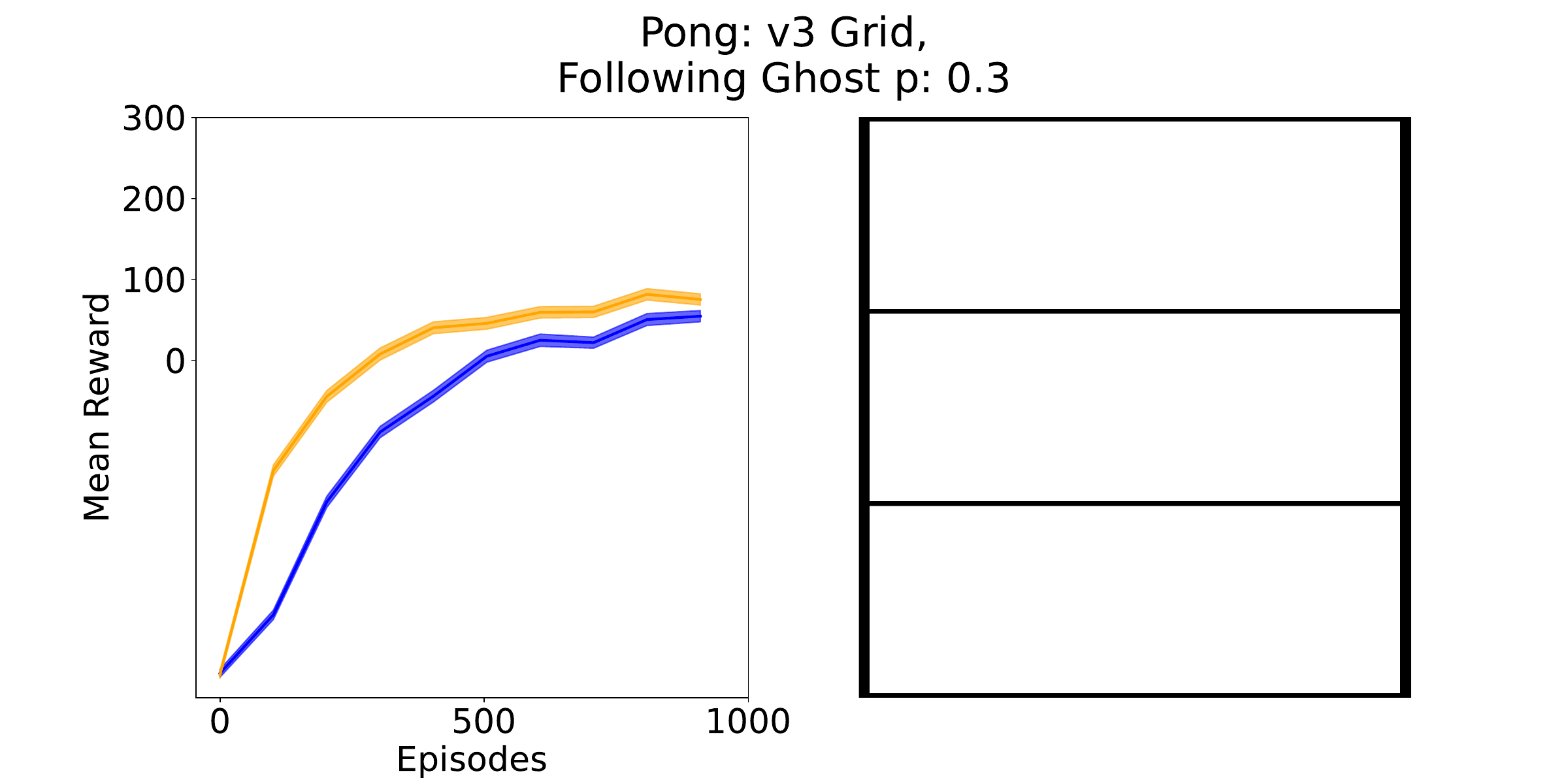}
  \end{subfigure}
  \hfill
  \begin{subfigure}{0.3\textwidth}
    \includegraphics[width=\linewidth]{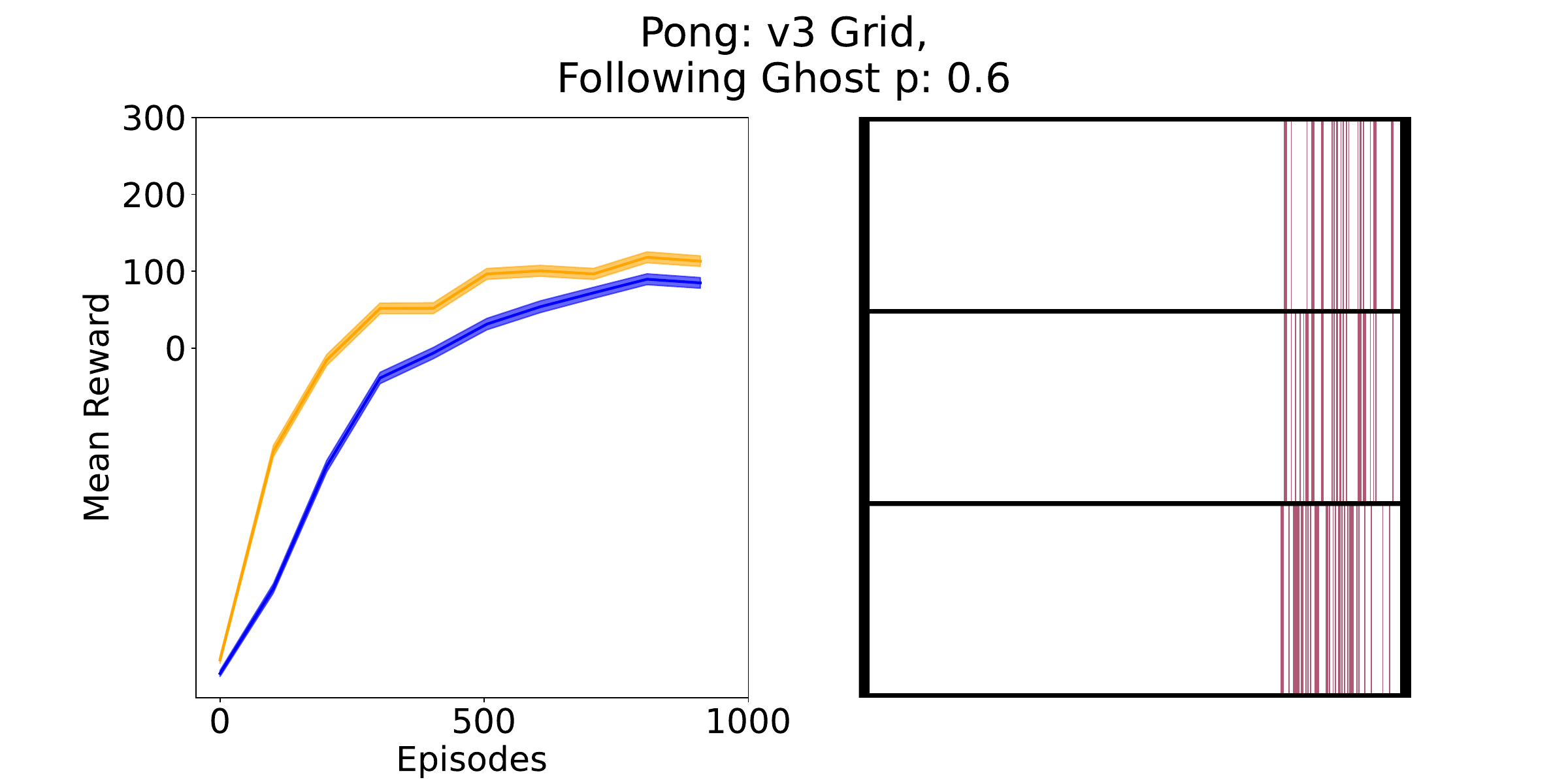}
  \end{subfigure}\\
  \hfill
  \begin{subfigure}{0.30\textwidth}
    \includegraphics[width=\linewidth]{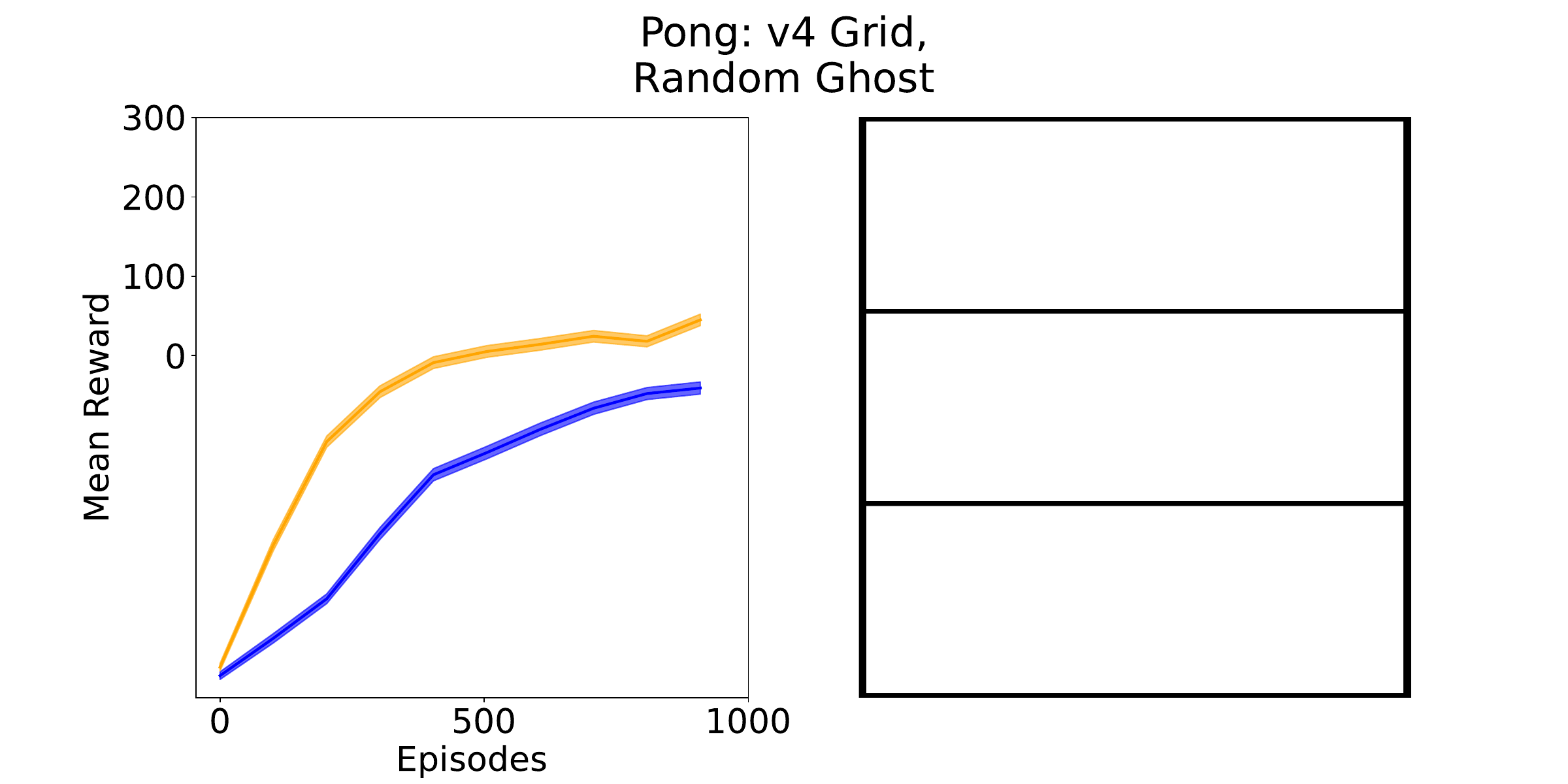}
  \end{subfigure}
  \hfill
  \begin{subfigure}{0.30\textwidth}
    \includegraphics[width=\linewidth]{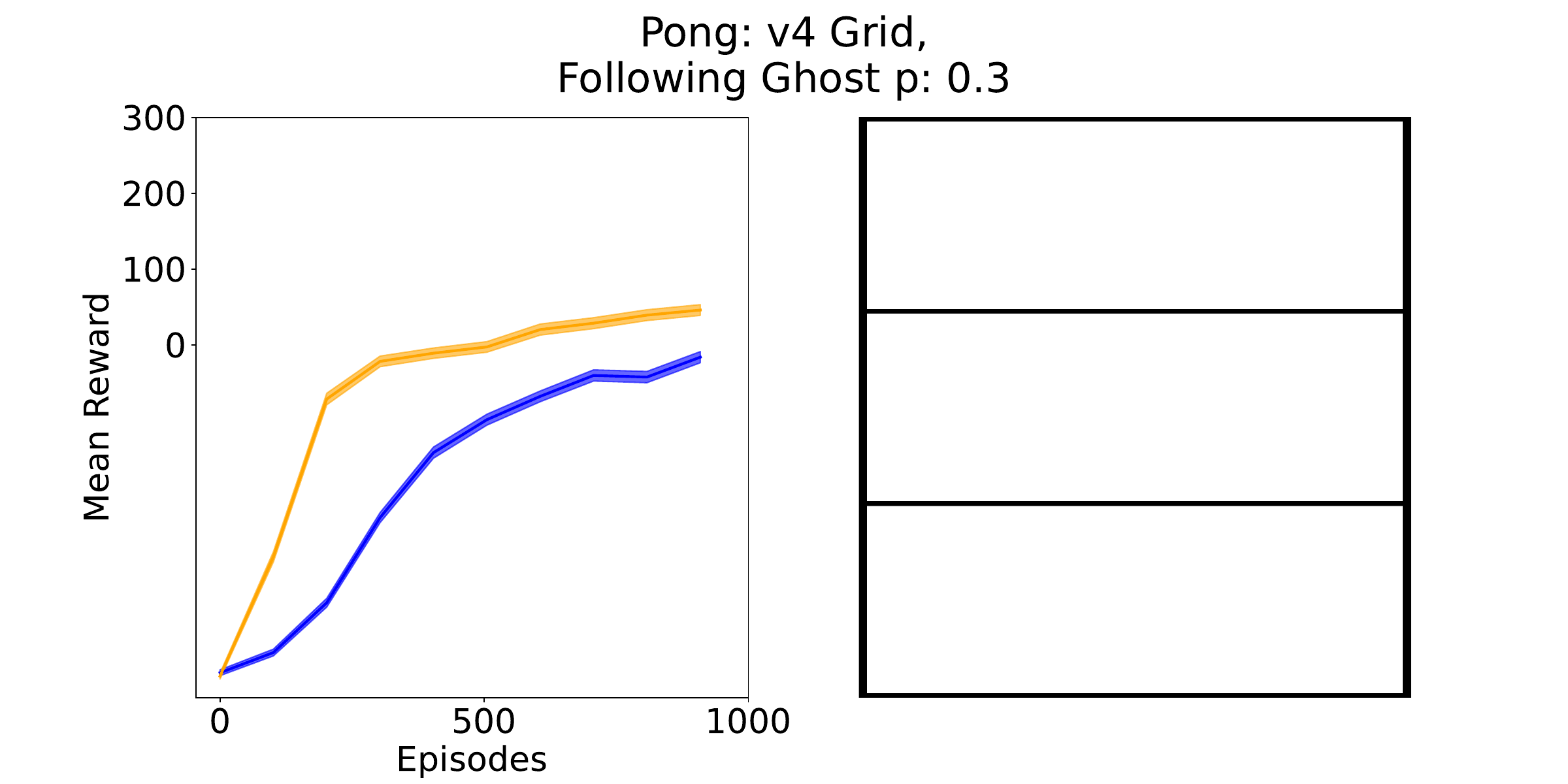}
  \end{subfigure}
  \hfill
  \begin{subfigure}{0.30\textwidth}
    \includegraphics[width=\linewidth]{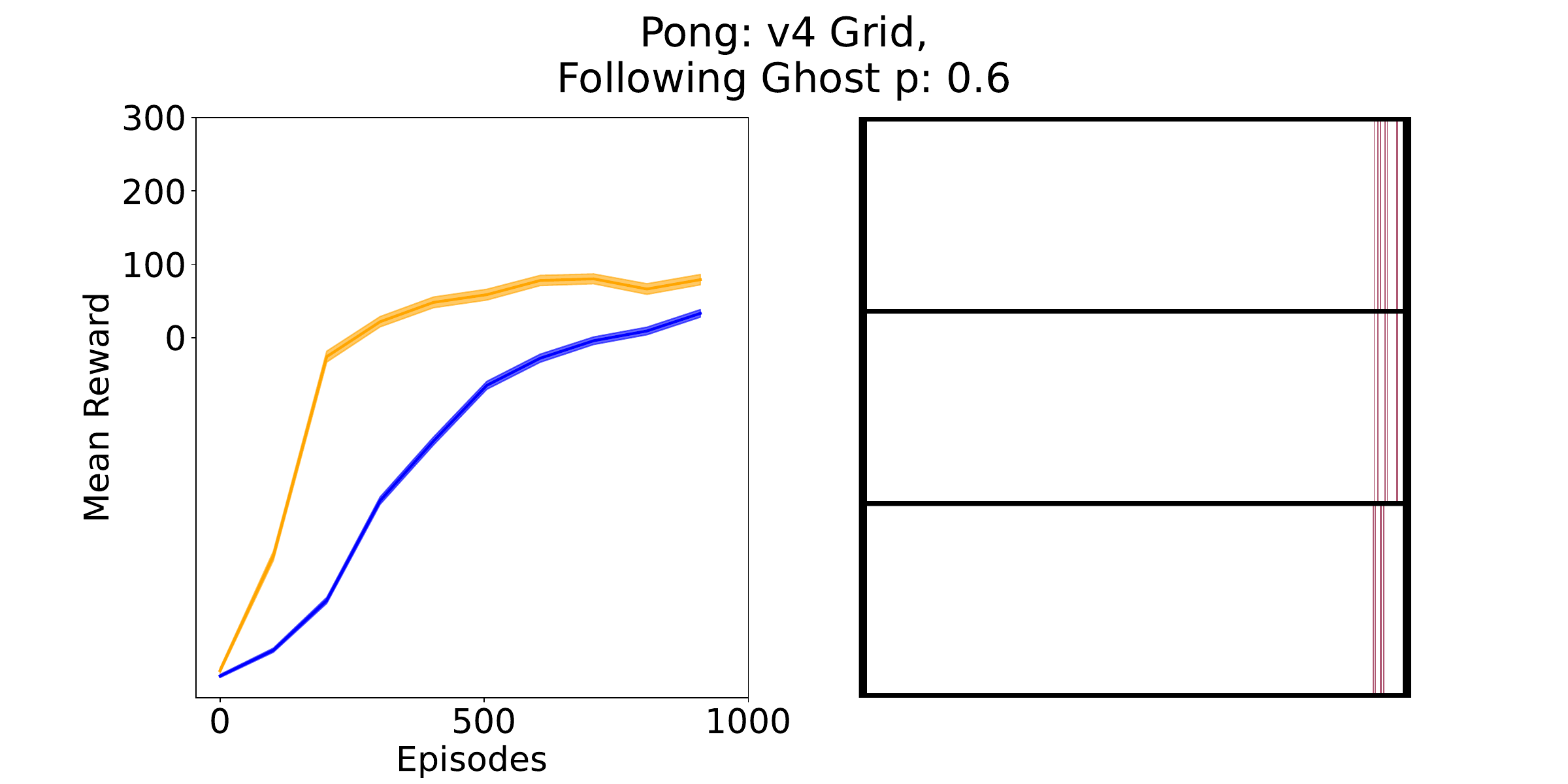}
    
  \end{subfigure}\\

  \caption{\emph{SARSA Agent with $\epsilon\text{-}$greedy exploration strategy}: The \textit{exploration grid} visualizing the difference in State-Action (S-A) pairs explored by these agents ($D_{LG}$). Results for PacMan v2, v3, v4 grids, the agent is trained on non-noisy variations of different environments (reported in the headings) and tested in the Low-Noise regime. Rows in the right figure represents agent's actions Left, Right, Up, or Down.}
  \label{fig:atari_variations-exploration-pacman-sarsa-egreedy}
\end{figure*}

\begin{figure*}[t]
  %\centering
  \begin{subfigure}{0.32\textwidth}
    \includegraphics[width=\linewidth]{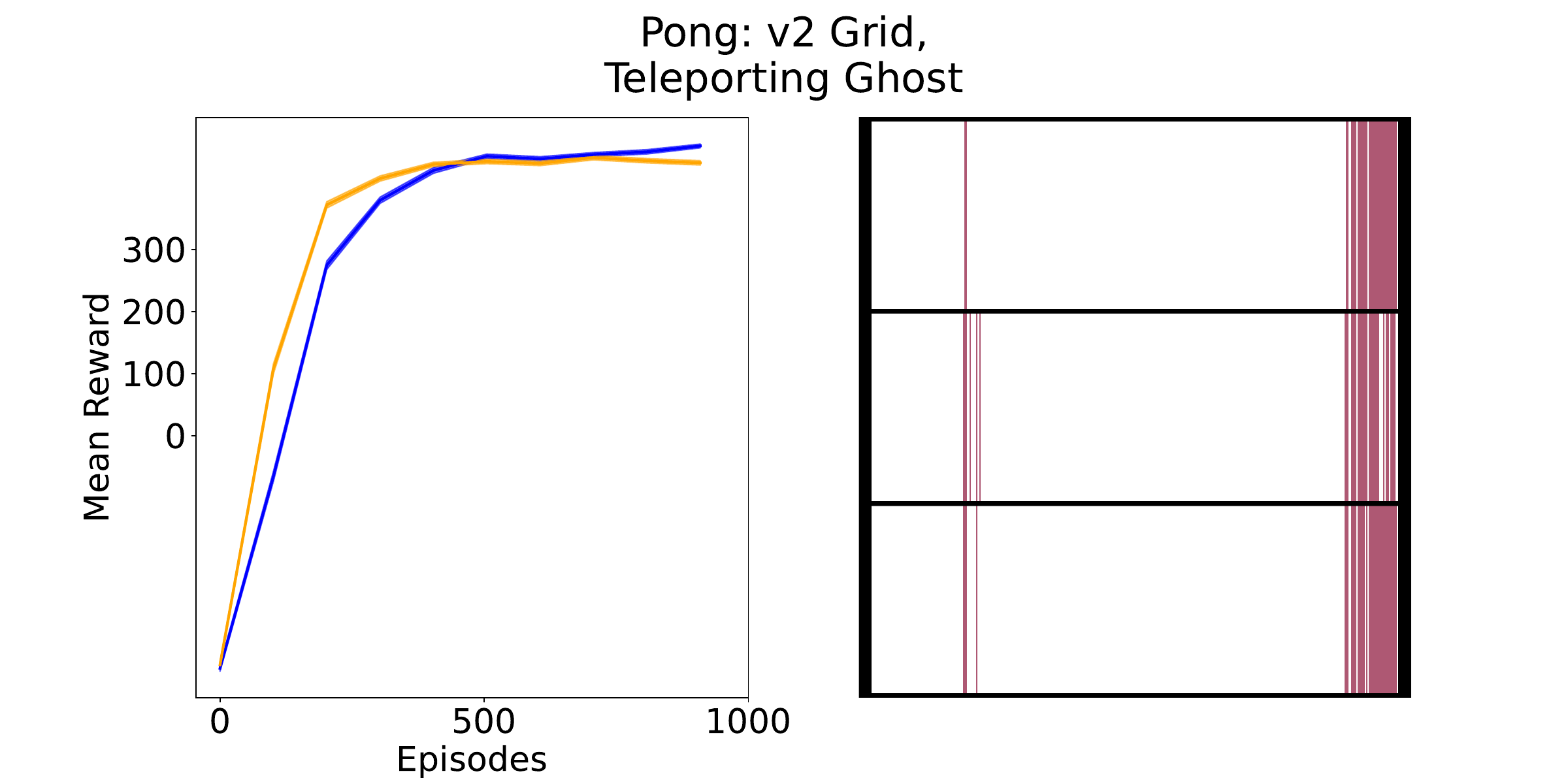}
  \end{subfigure}
  \hfill
  \begin{subfigure}{0.32\textwidth}
    \includegraphics[width=\linewidth]{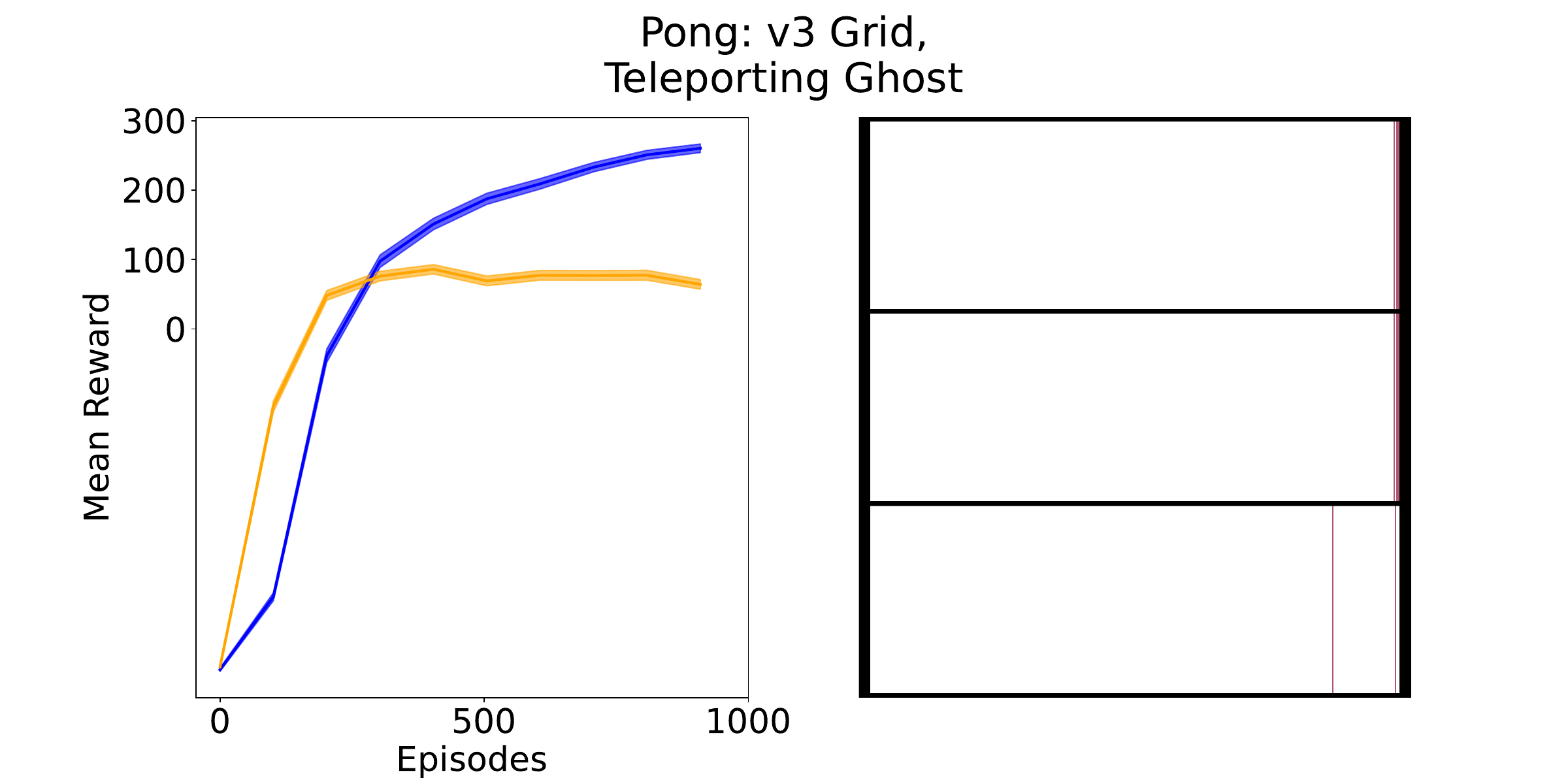}
  \end{subfigure}
  \hfill
  \begin{subfigure}{0.32\textwidth}
    \includegraphics[width=\linewidth]{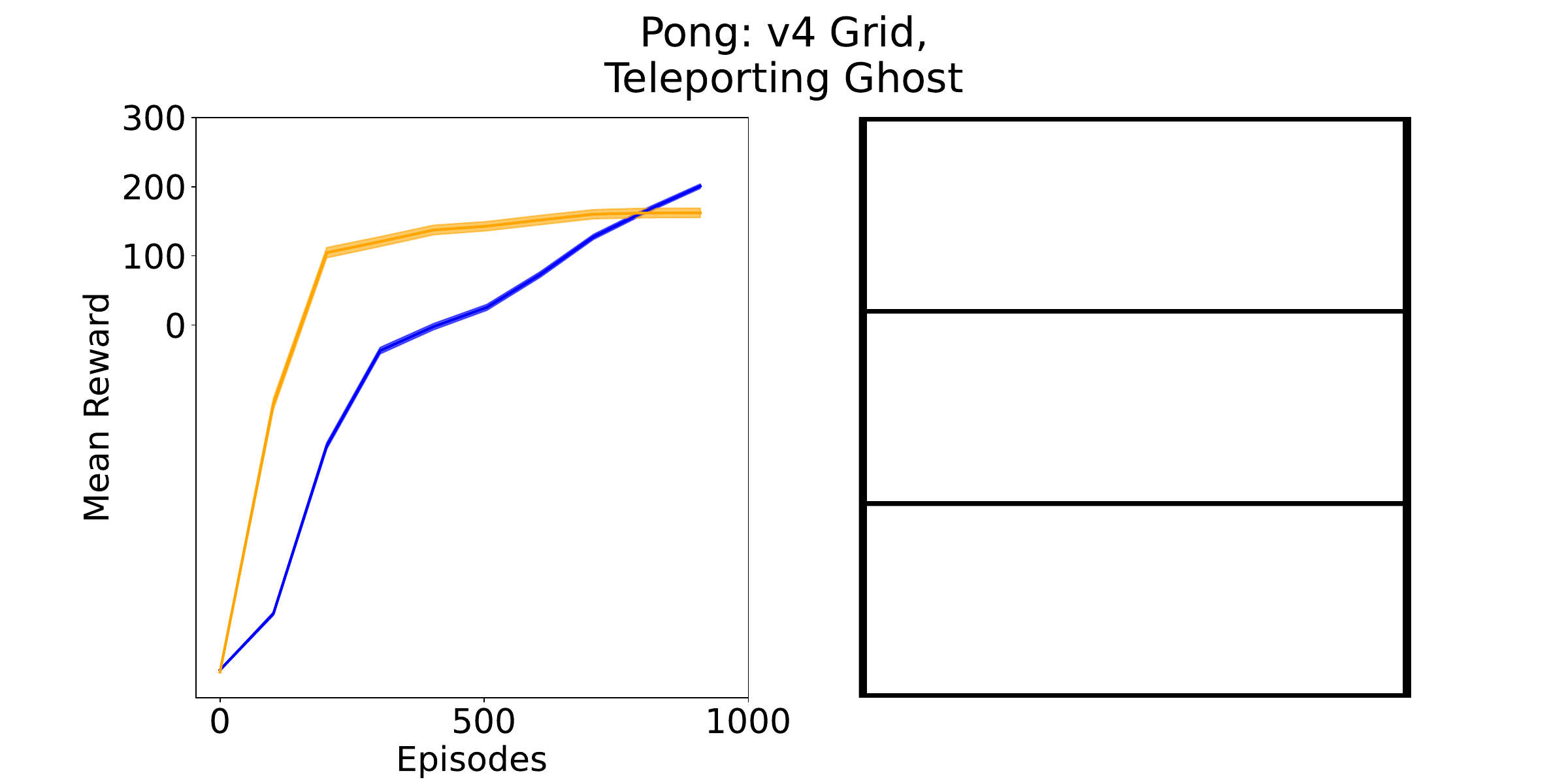}
  \end{subfigure}

  \begin{subfigure}{0.32\textwidth}
    \includegraphics[width=\linewidth]{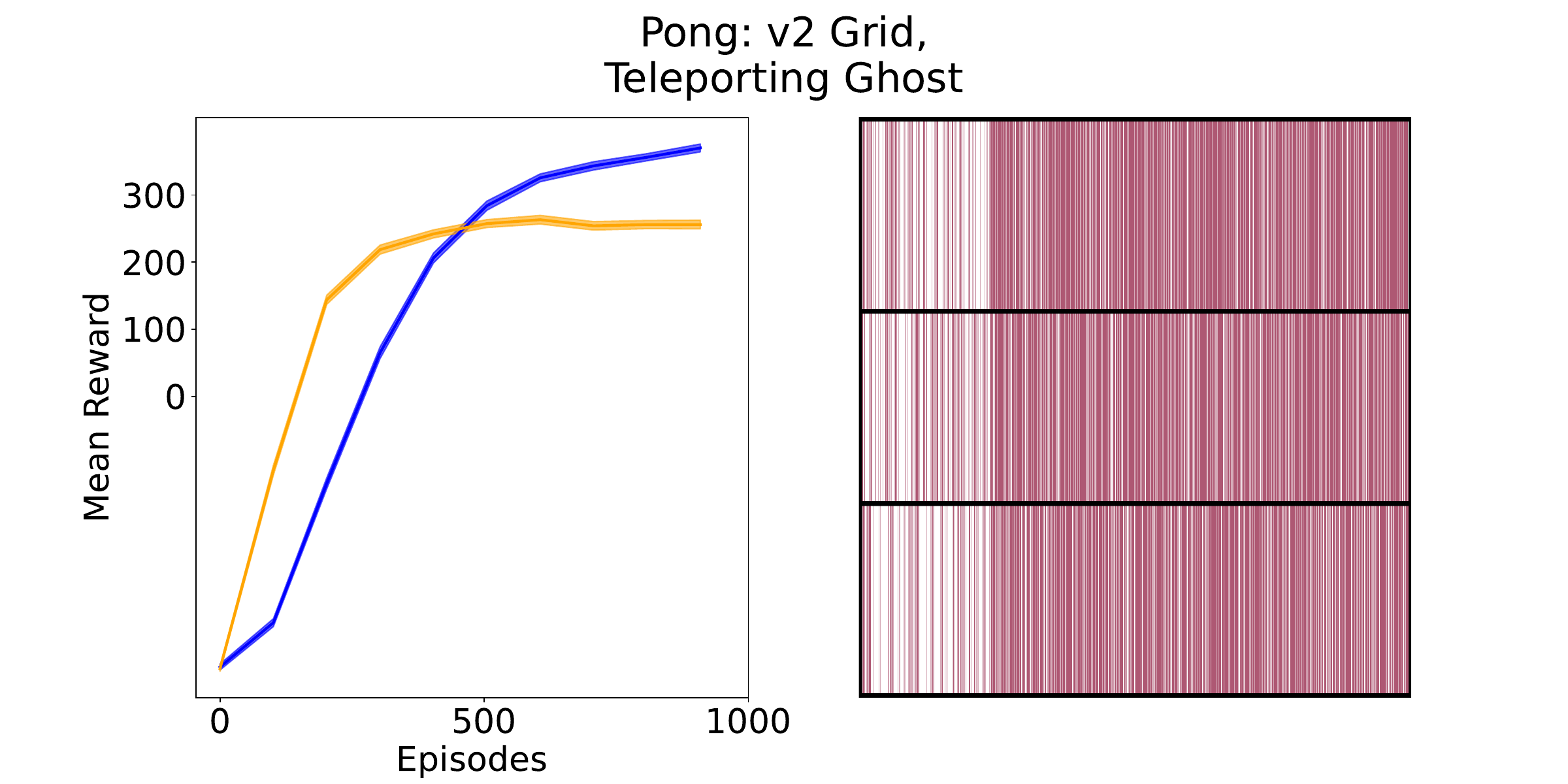}
  \end{subfigure}
  \hfill
  \begin{subfigure}{0.32\textwidth}
    \includegraphics[width=\linewidth]{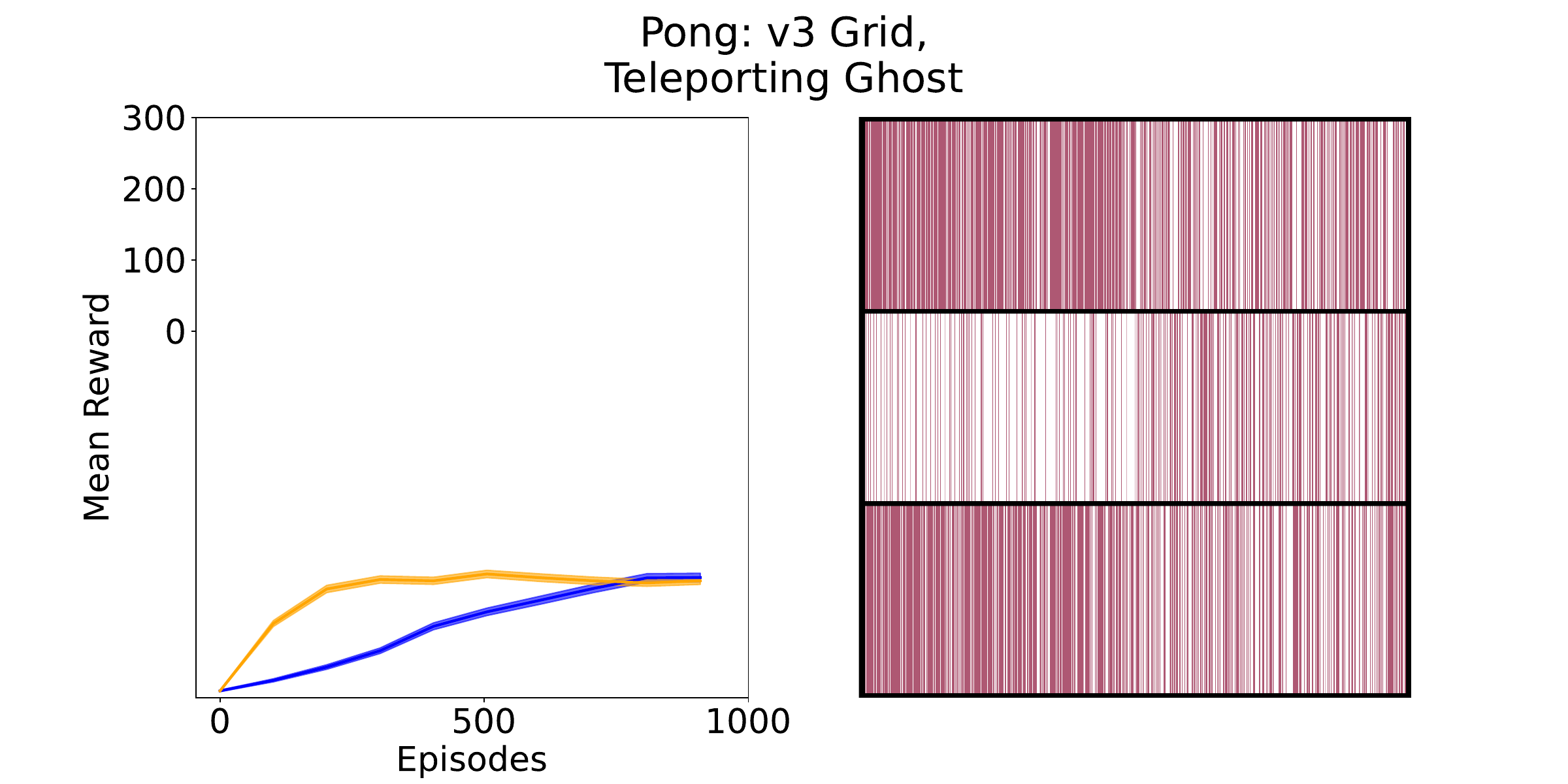}
  \end{subfigure}
  \hfill
  \begin{subfigure}{0.32\textwidth}
    \includegraphics[width=\linewidth]{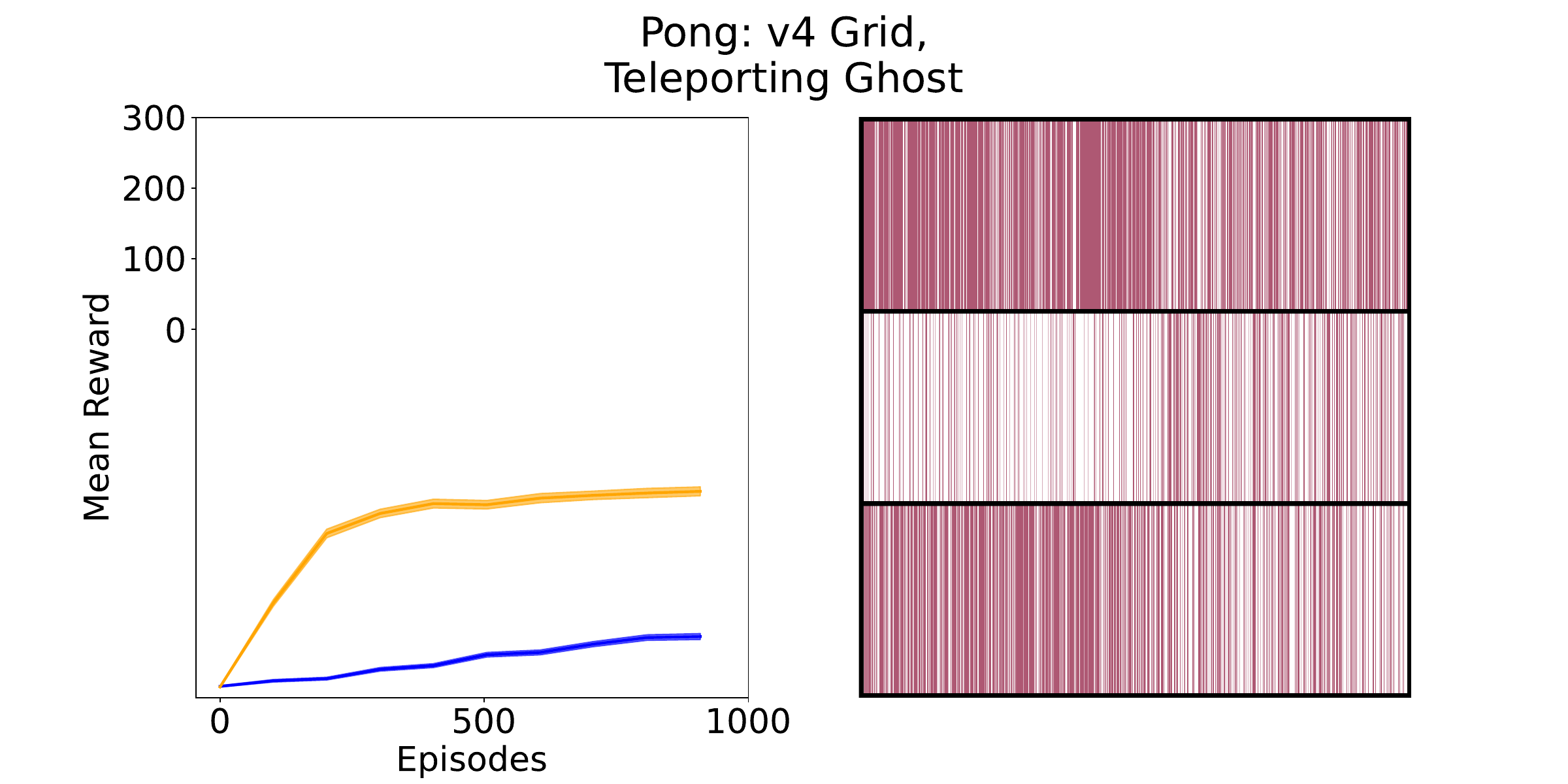}
  \end{subfigure}

  \caption{\emph{Q-learning Agent with Boltzmann exploration strategy}: The \textit{exploration grid} visualizing the difference in State-Action (S-A) pairs explored by these agents ($D_{LG}$). Results for PacMan v2, v3, v4 grids, the agent is trained on Teleporting Ghost variation ($p=0.2$, $p=0.5$) and tested in different environments (reported in the headings). Rows in the right figure represents agent's actions Left, Right, Up, or Down.}
  \label{fig:atari_variations-exploration-semantic-pacman-qlearning-boltzmann}
\end{figure*}

\begin{figure*}[t]
  %\centering
  \begin{subfigure}{0.32\textwidth}
    \includegraphics[width=\linewidth]{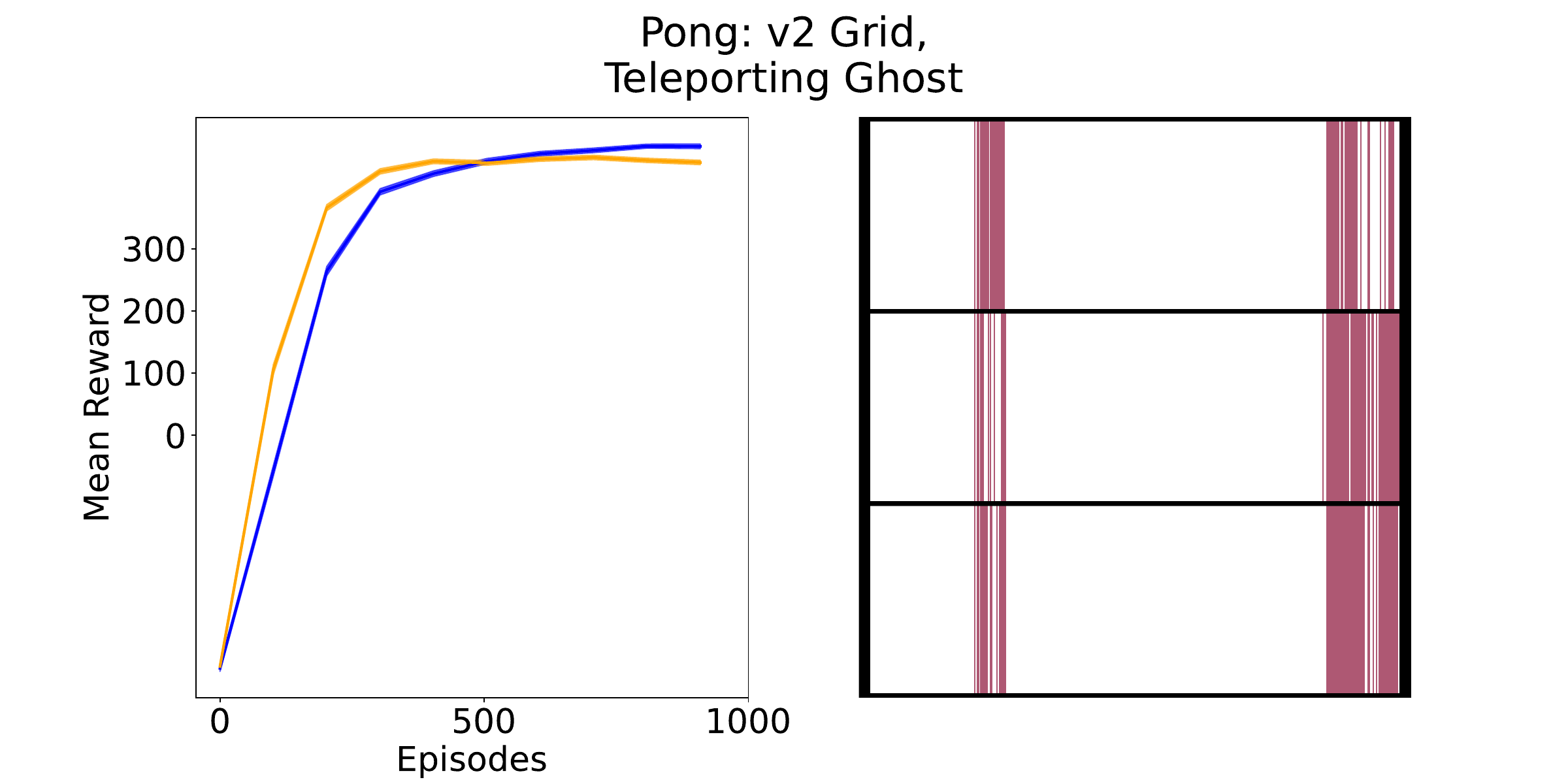}
  \end{subfigure}
  \hfill
  \begin{subfigure}{0.32\textwidth}
    \includegraphics[width=\linewidth]{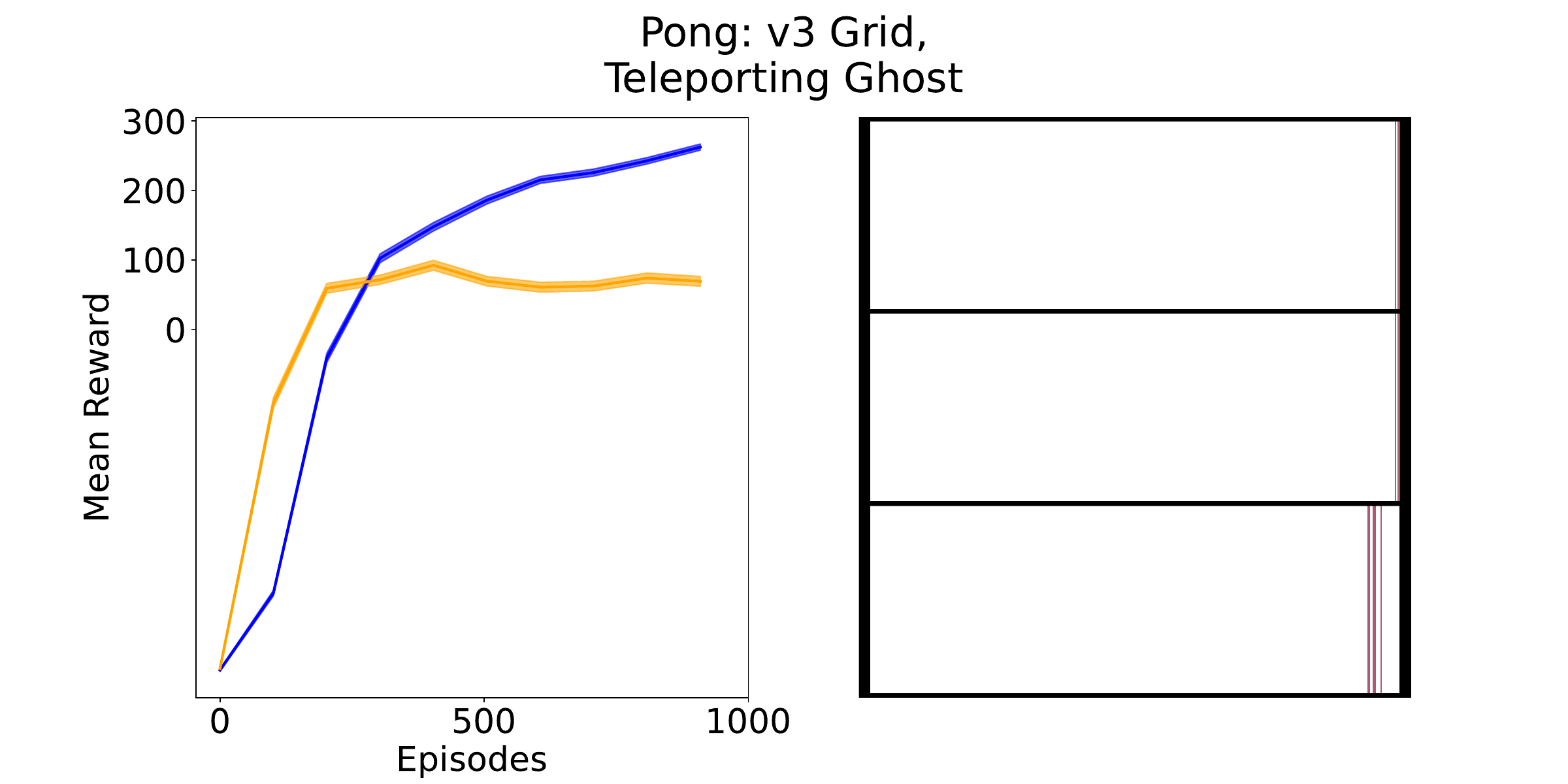}
  \end{subfigure}
  \hfill
  \begin{subfigure}{0.32\textwidth}
    \includegraphics[width=\linewidth]{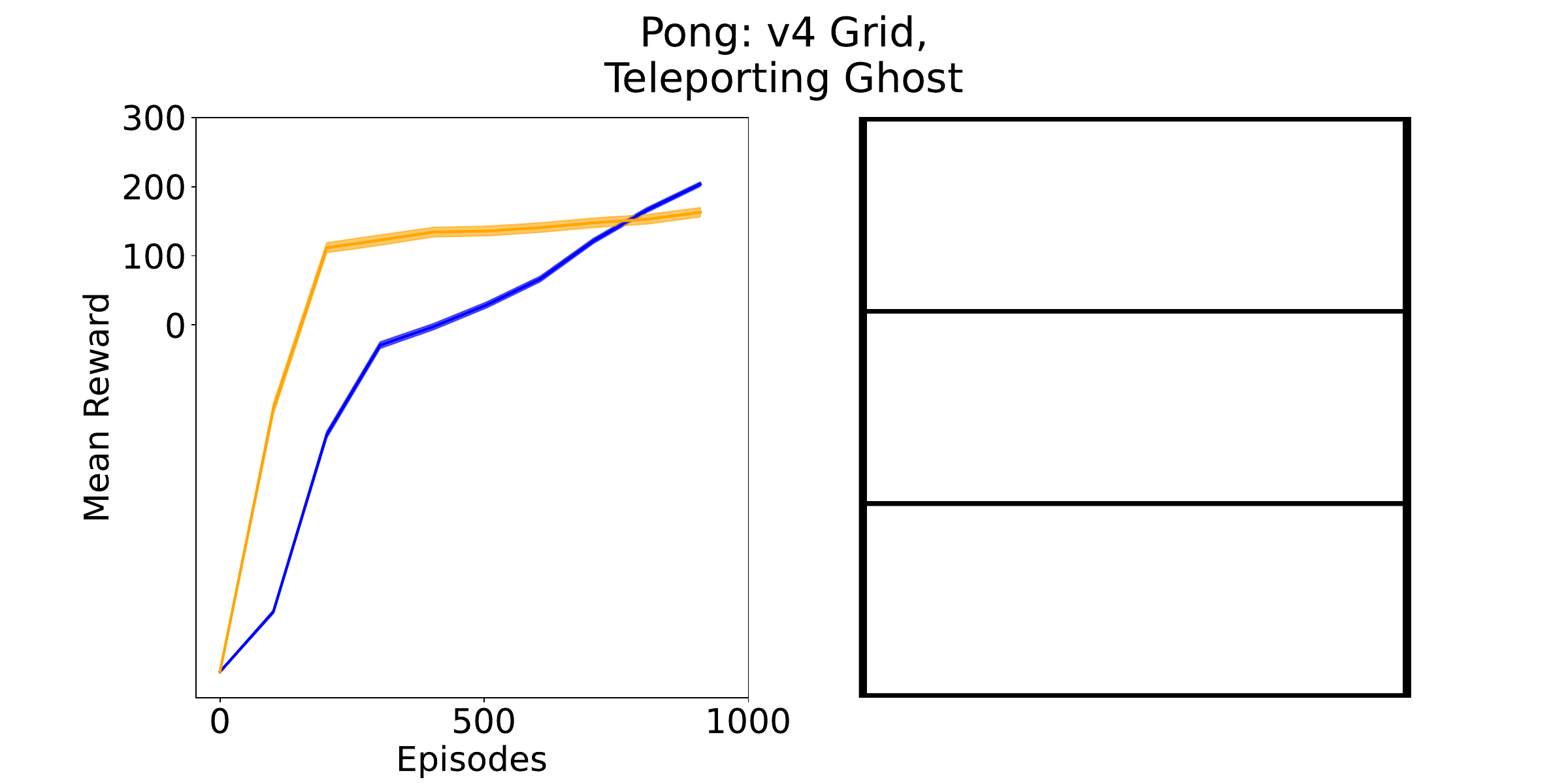}
  \end{subfigure}

  \begin{subfigure}{0.32\textwidth}
    \includegraphics[width=\linewidth]{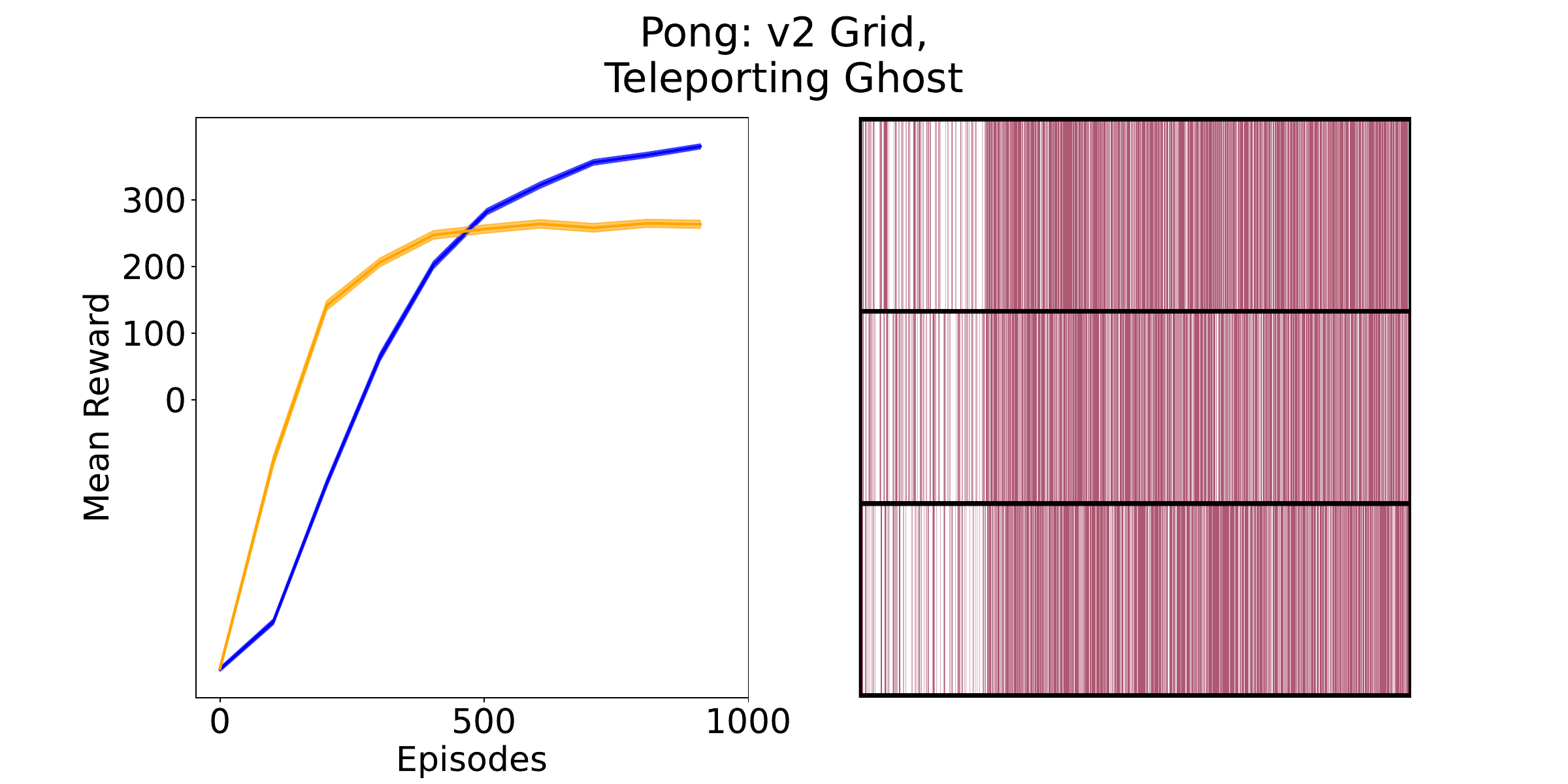}
  \end{subfigure}
  \hfill
  \begin{subfigure}{0.32\textwidth}
    \includegraphics[width=\linewidth]{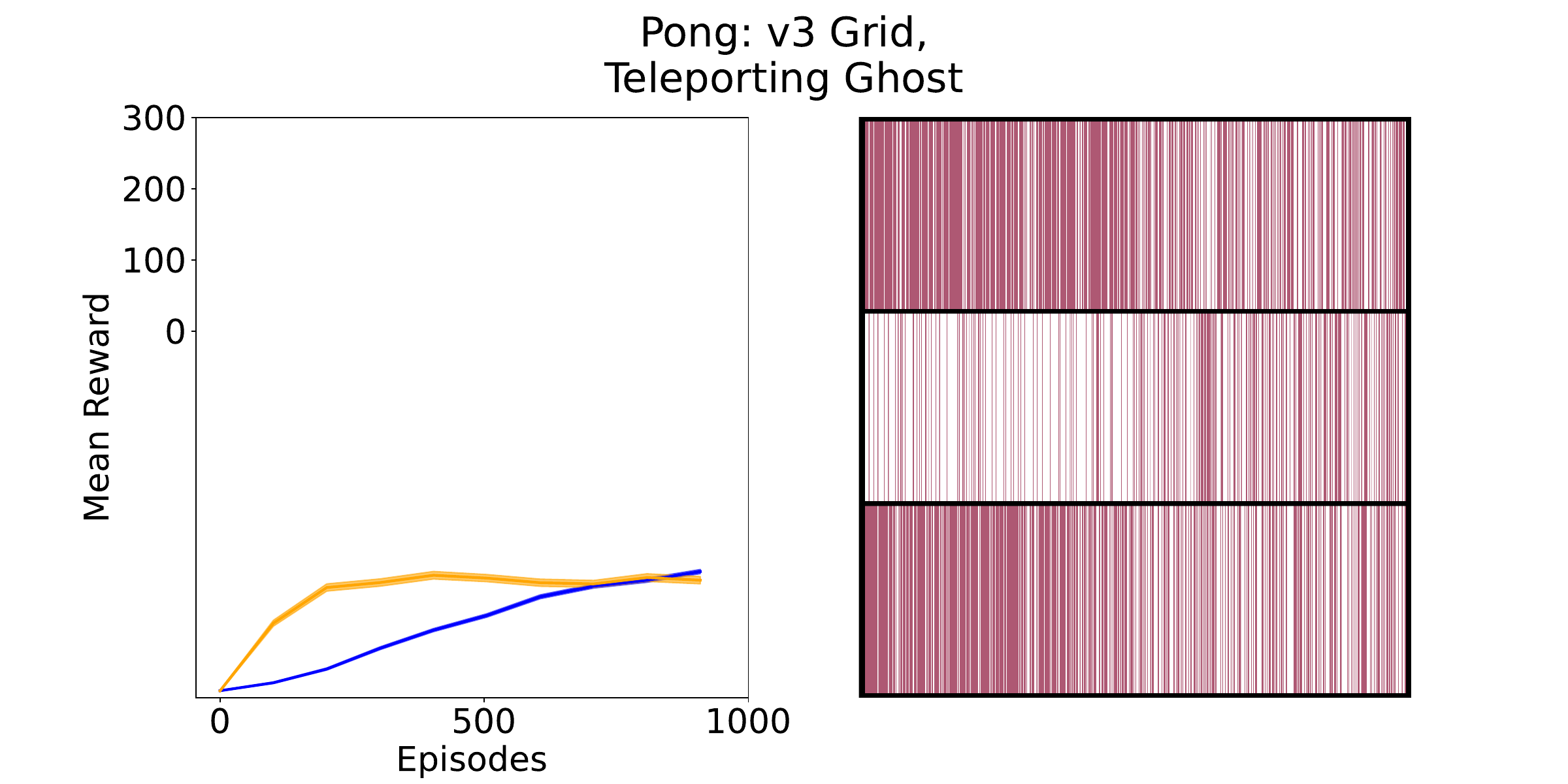}
  \end{subfigure}
  \hfill
  \begin{subfigure}{0.32\textwidth}
    \includegraphics[width=\linewidth]{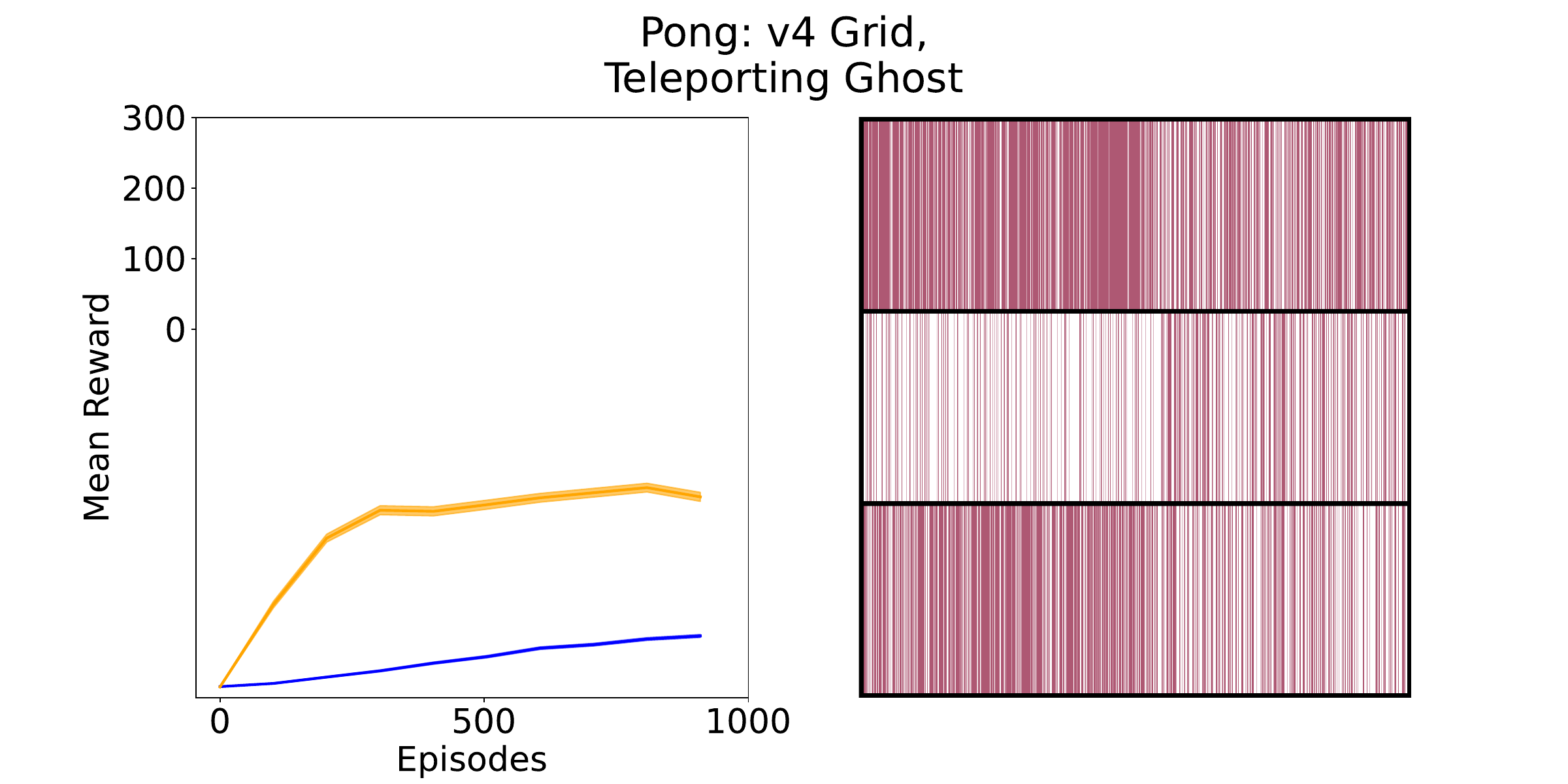}
  \end{subfigure}

  \caption{\emph{Q-learning Agent with $\epsilon\text{-}$greedy exploration strategy}: The \textit{exploration grid} visualizing the difference in State-Action (S-A) pairs explored by these agents ($D_{LG}$). Results for PacMan v2, v3, v4 grids, the agent is trained on Teleporting Ghost variation ($p=0.2$, $p=0.5$) and tested in different environments (reported in the headings). Rows in the right figure represents agent's actions Left, Right, Up, or Down.}
  \label{fig:atari_variations-exploration-semantic-pacman-qlearning-egreedy}
\end{figure*}

\begin{figure*}[t]
  %\centering
  \begin{subfigure}{0.32\textwidth}
    \includegraphics[width=\linewidth]{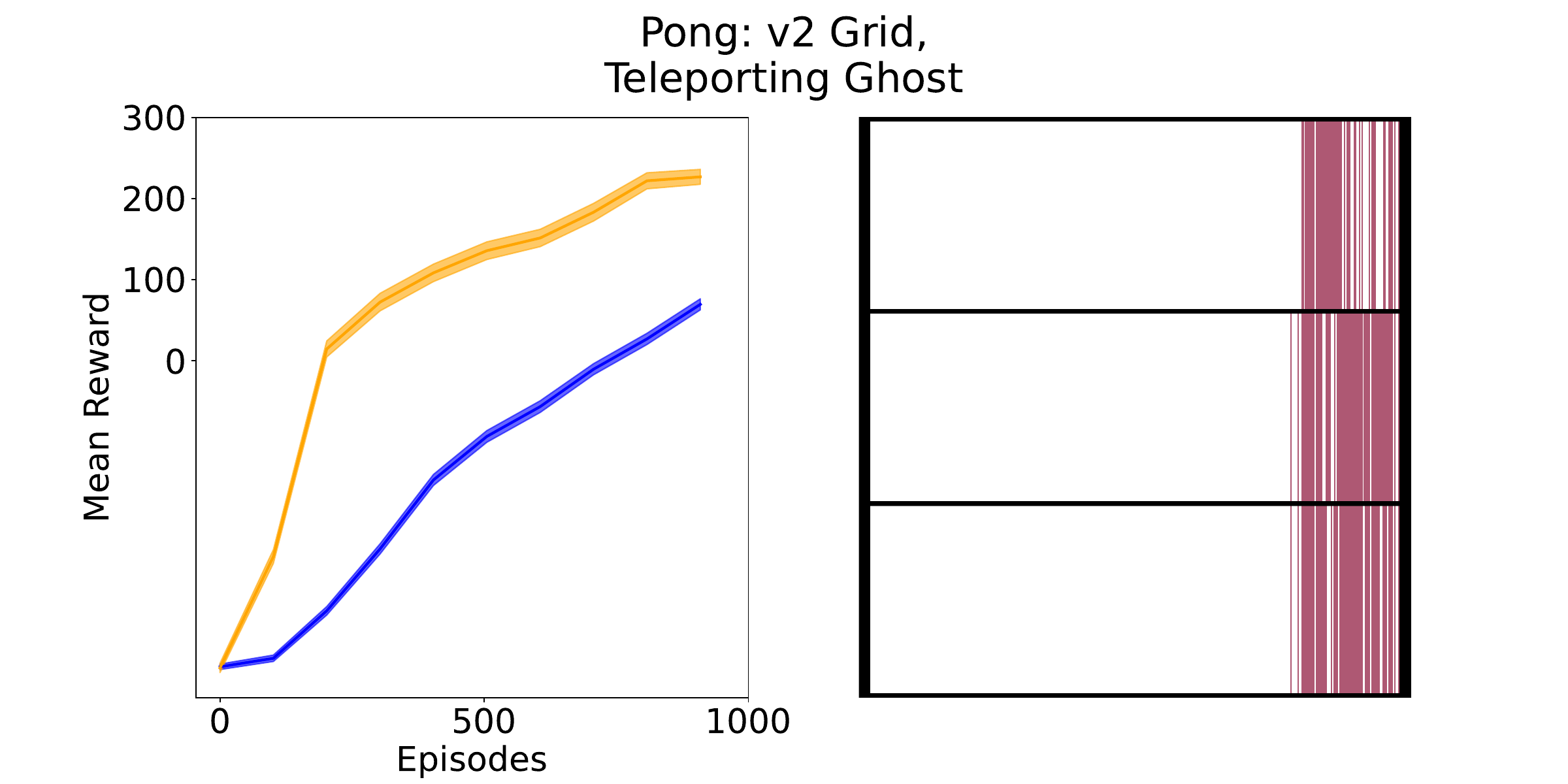}
  \end{subfigure}
  \hfill
  \begin{subfigure}{0.32\textwidth}
    \includegraphics[width=\linewidth]{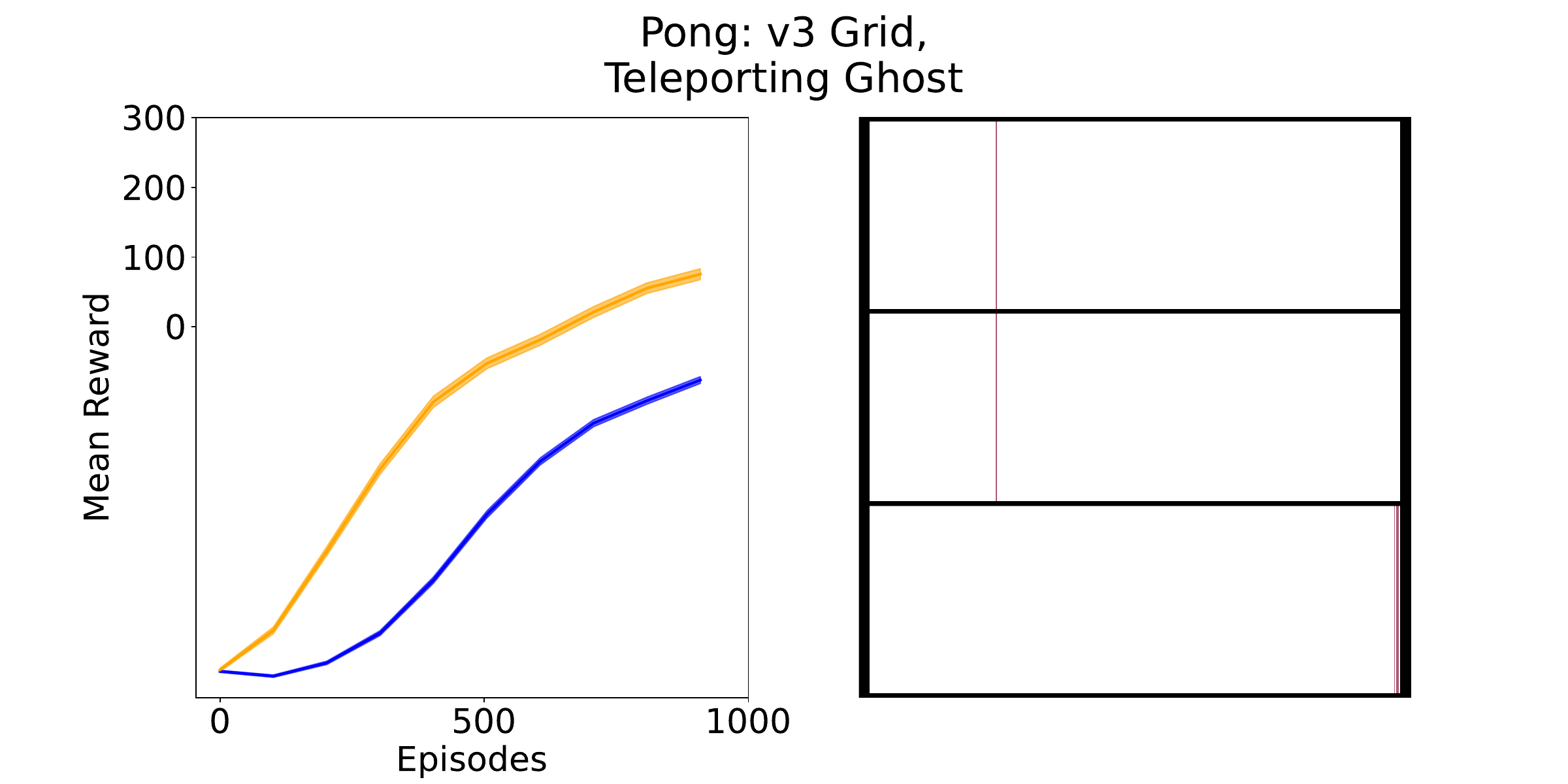}
  \end{subfigure}
  \hfill
  \begin{subfigure}{0.32\textwidth}
    \includegraphics[width=\linewidth]{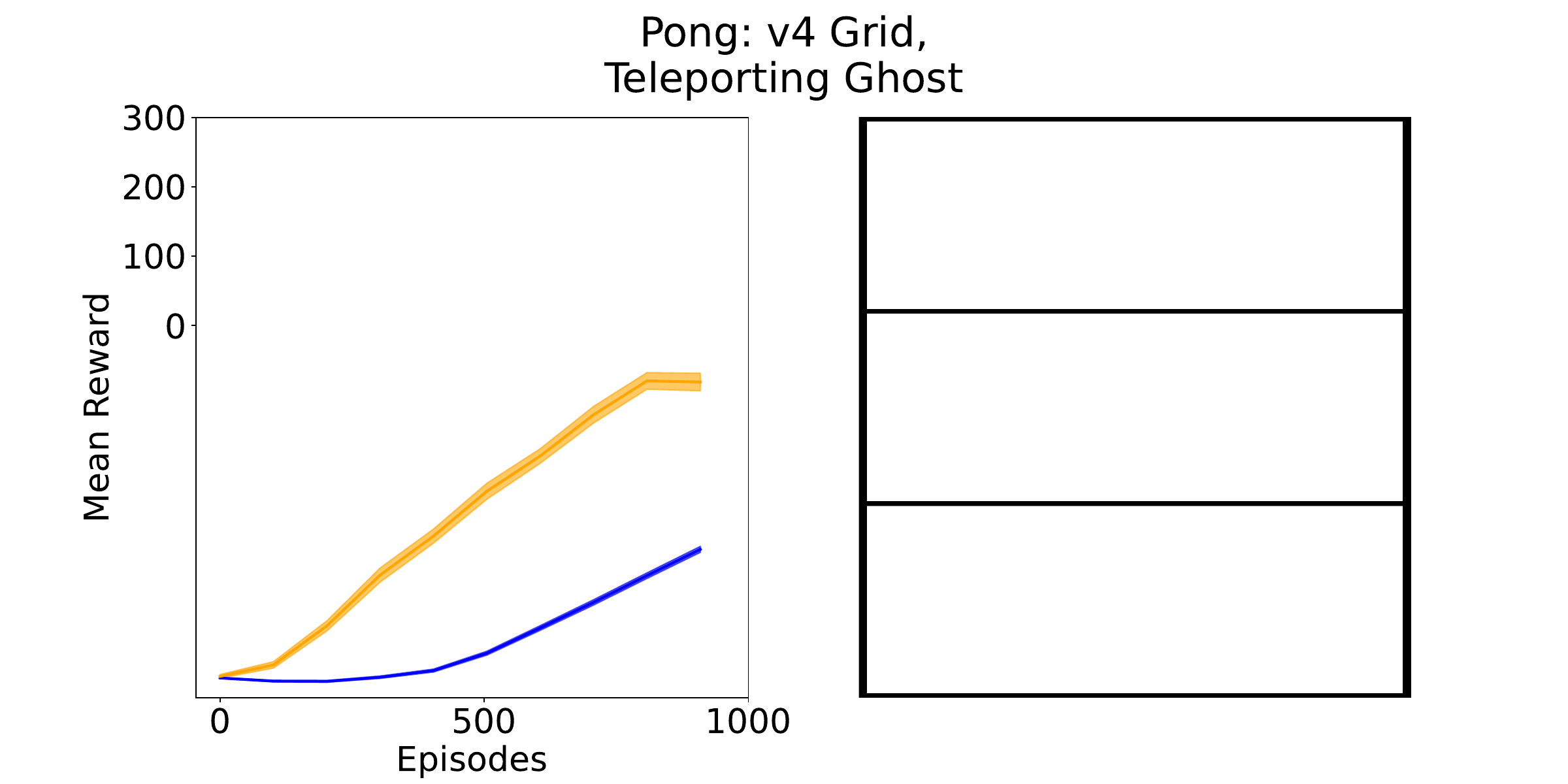}
  \end{subfigure}

  \begin{subfigure}{0.32\textwidth}
    \includegraphics[width=\linewidth]{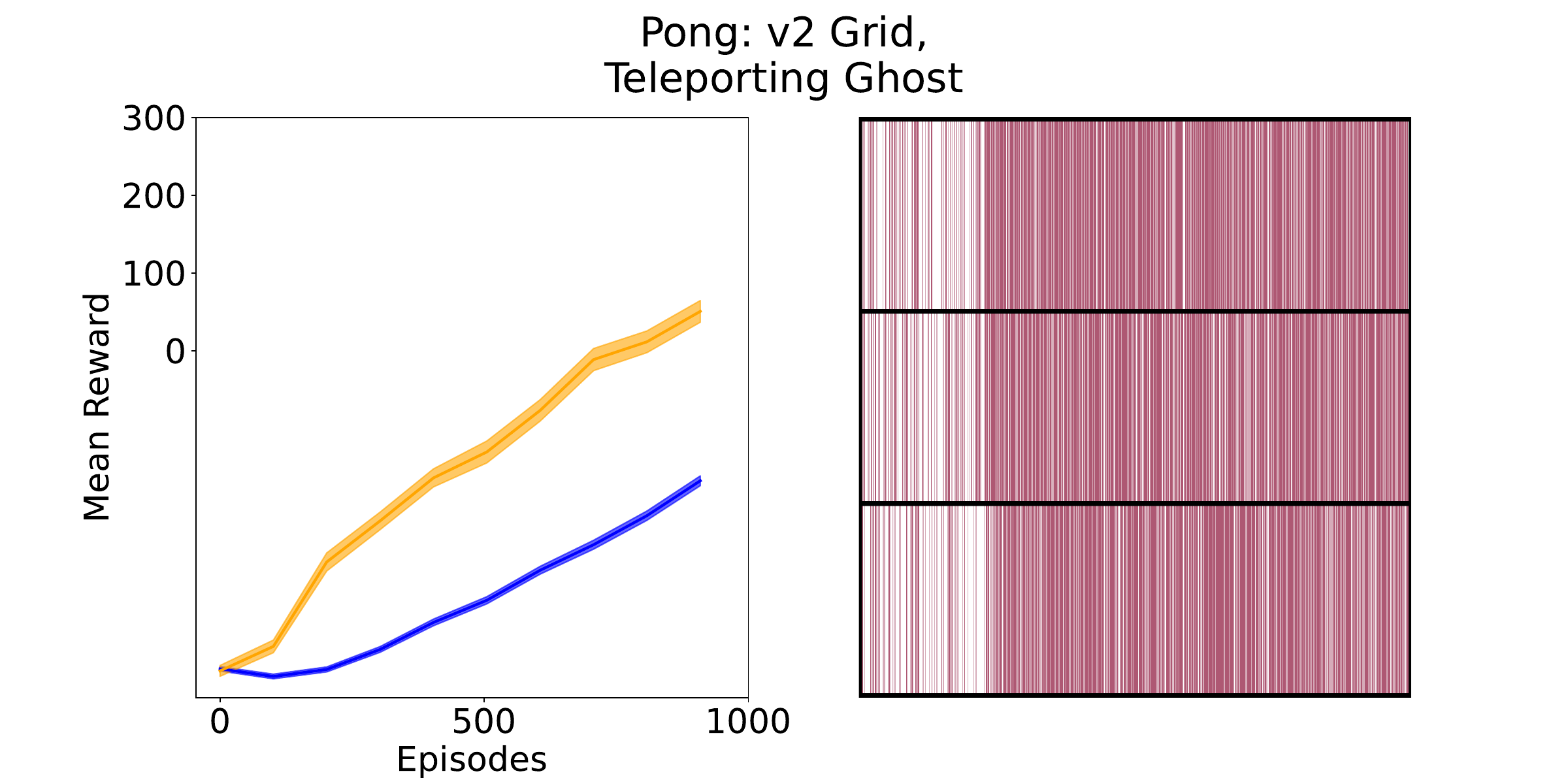}
  \end{subfigure}
  \hfill
  \begin{subfigure}{0.32\textwidth}
    \includegraphics[width=\linewidth]{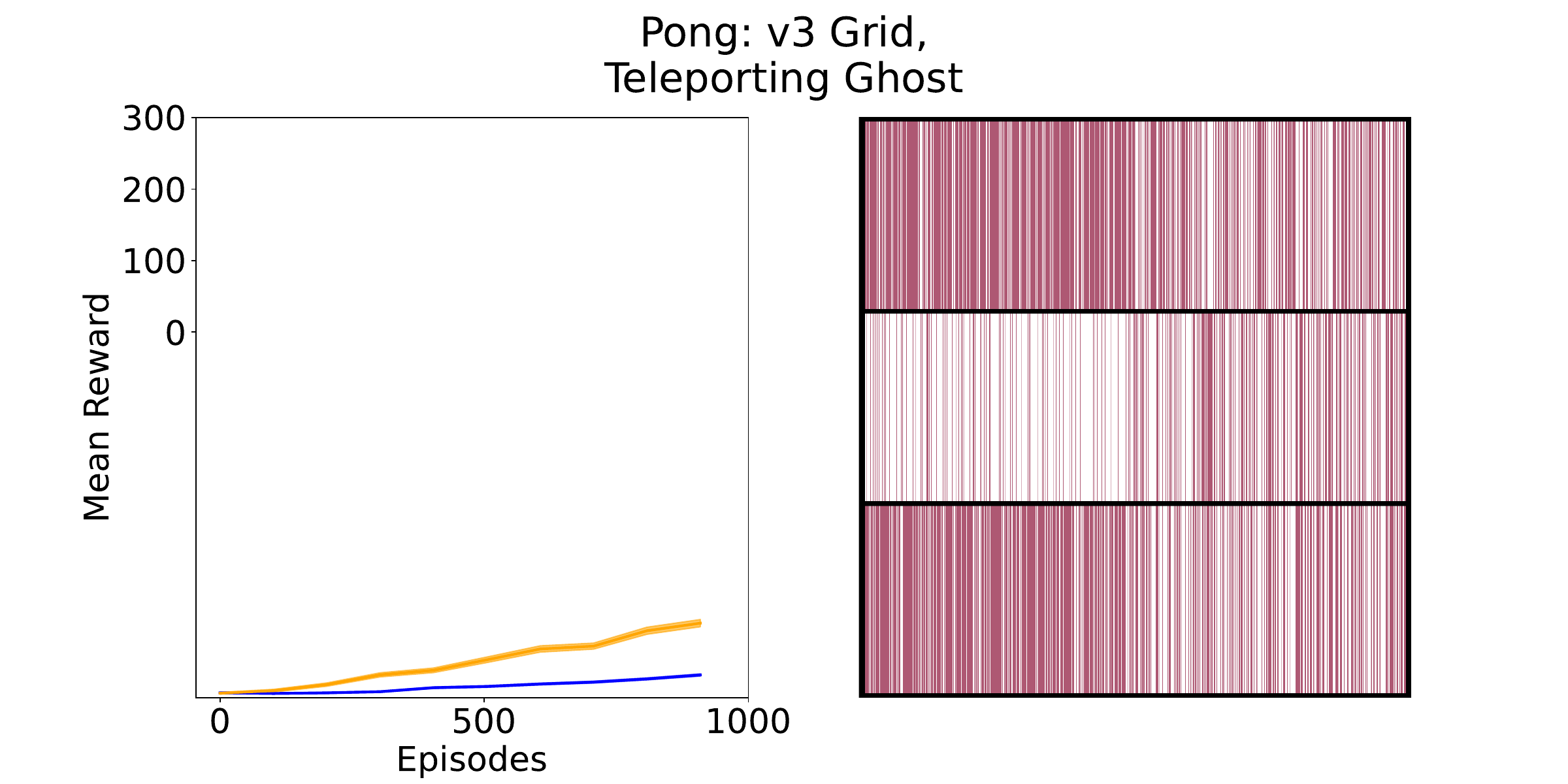}
  \end{subfigure}
  \hfill
  \begin{subfigure}{0.32\textwidth}
    \includegraphics[width=\linewidth]{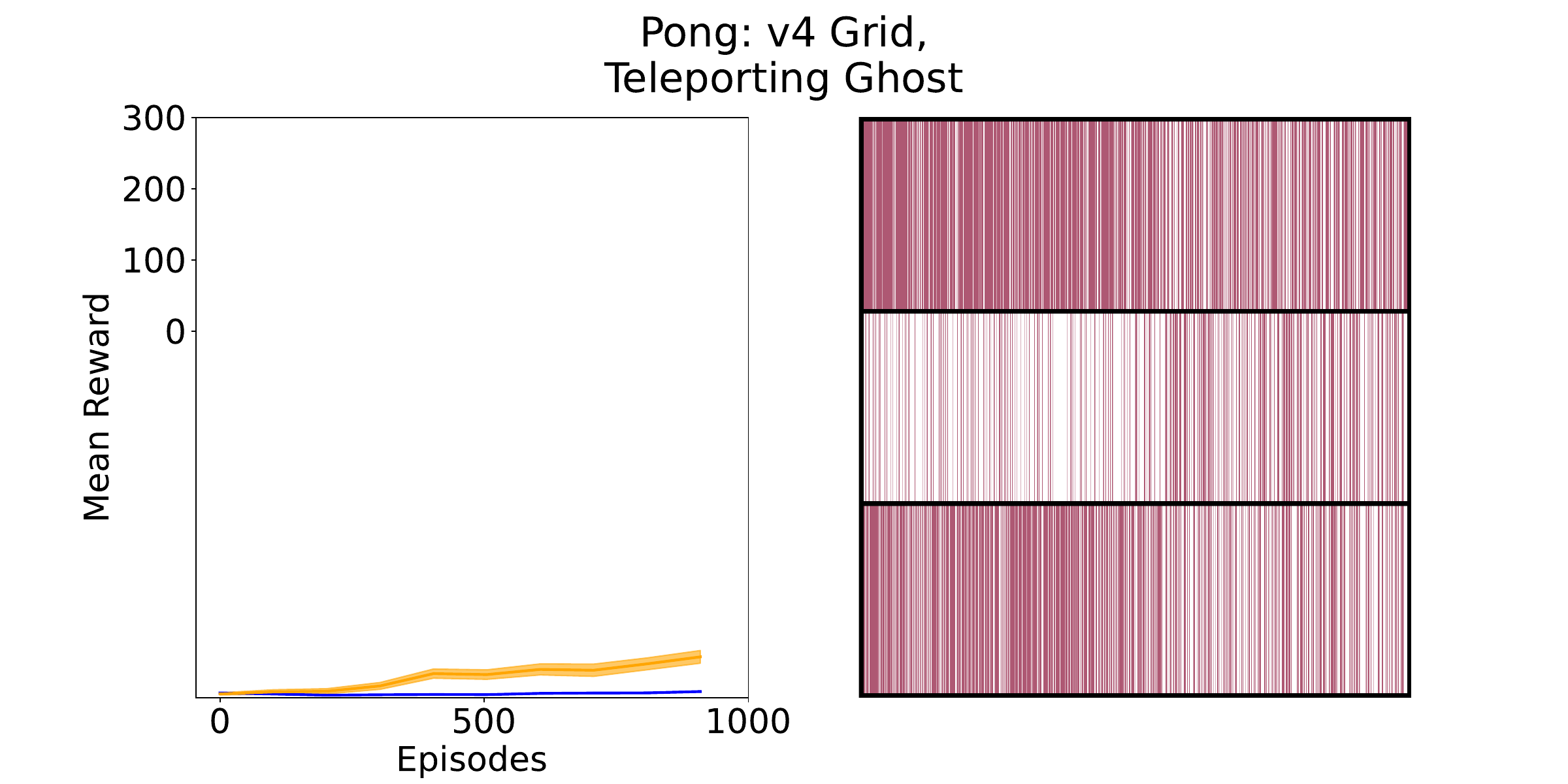}
  \end{subfigure}

  \caption{\emph{SARSA Agent with Boltzmann exploration strategy}: The \textit{exploration grid} visualizing the difference in State-Action (S-A) pairs explored by these agents ($D_{LG}$). Results for PacMan v2, v3, v4 grids, the agent is trained on Teleporting Ghost variation ($p=0.2$, $p=0.5$) and tested in different environments (reported in the headings). Rows in the right figure represents agent's actions Left, Right, Up, or Down.}
  \label{fig:atari_variations-exploration-semantic-pacman-sarsa-boltzmann}
\end{figure*}

\begin{figure*}[t]
  %\centering
  \begin{subfigure}{0.32\textwidth}
    \includegraphics[width=\linewidth]{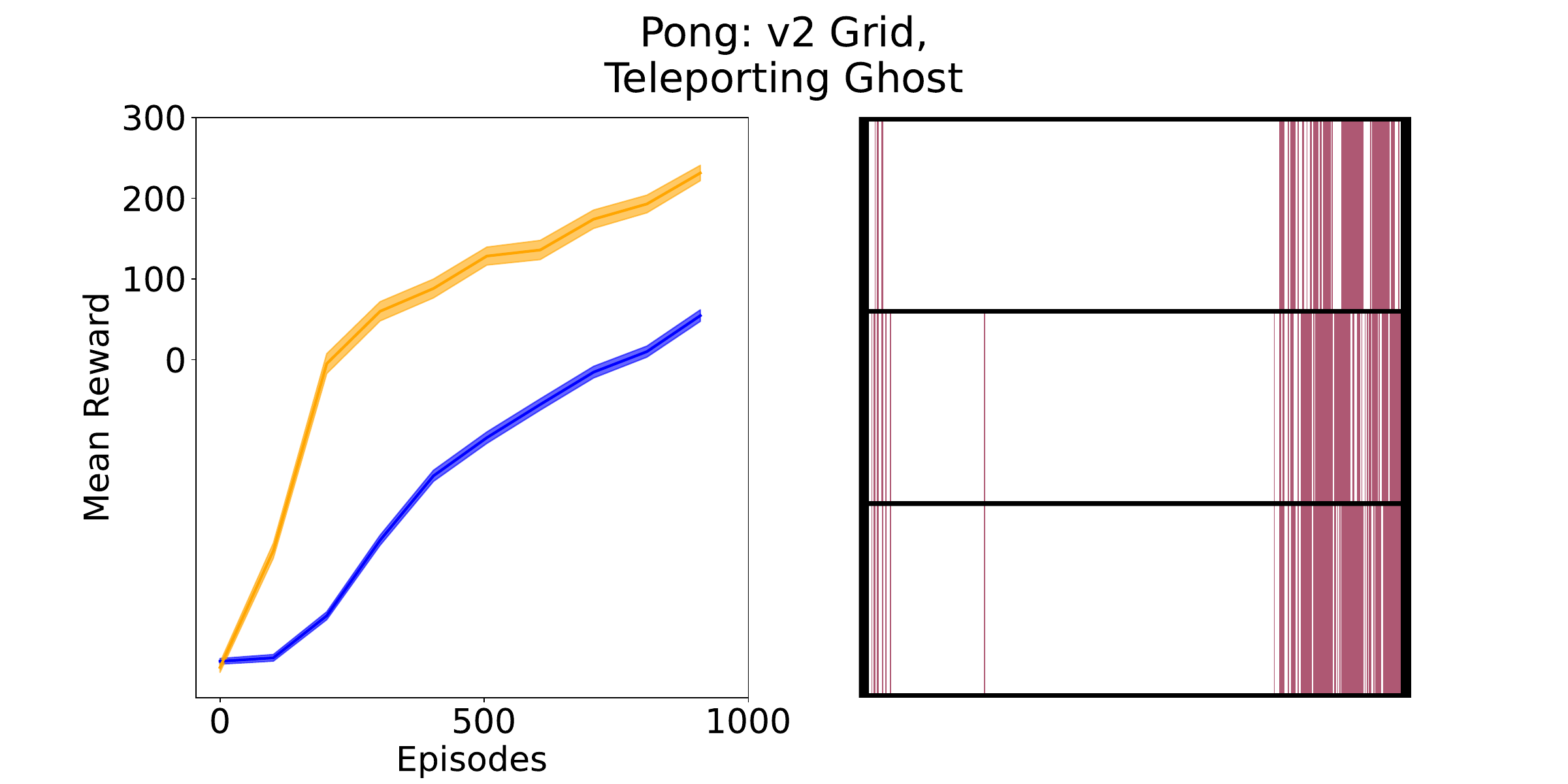}
  \end{subfigure}
  \hfill
  \begin{subfigure}{0.32\textwidth}
    \includegraphics[width=\linewidth]{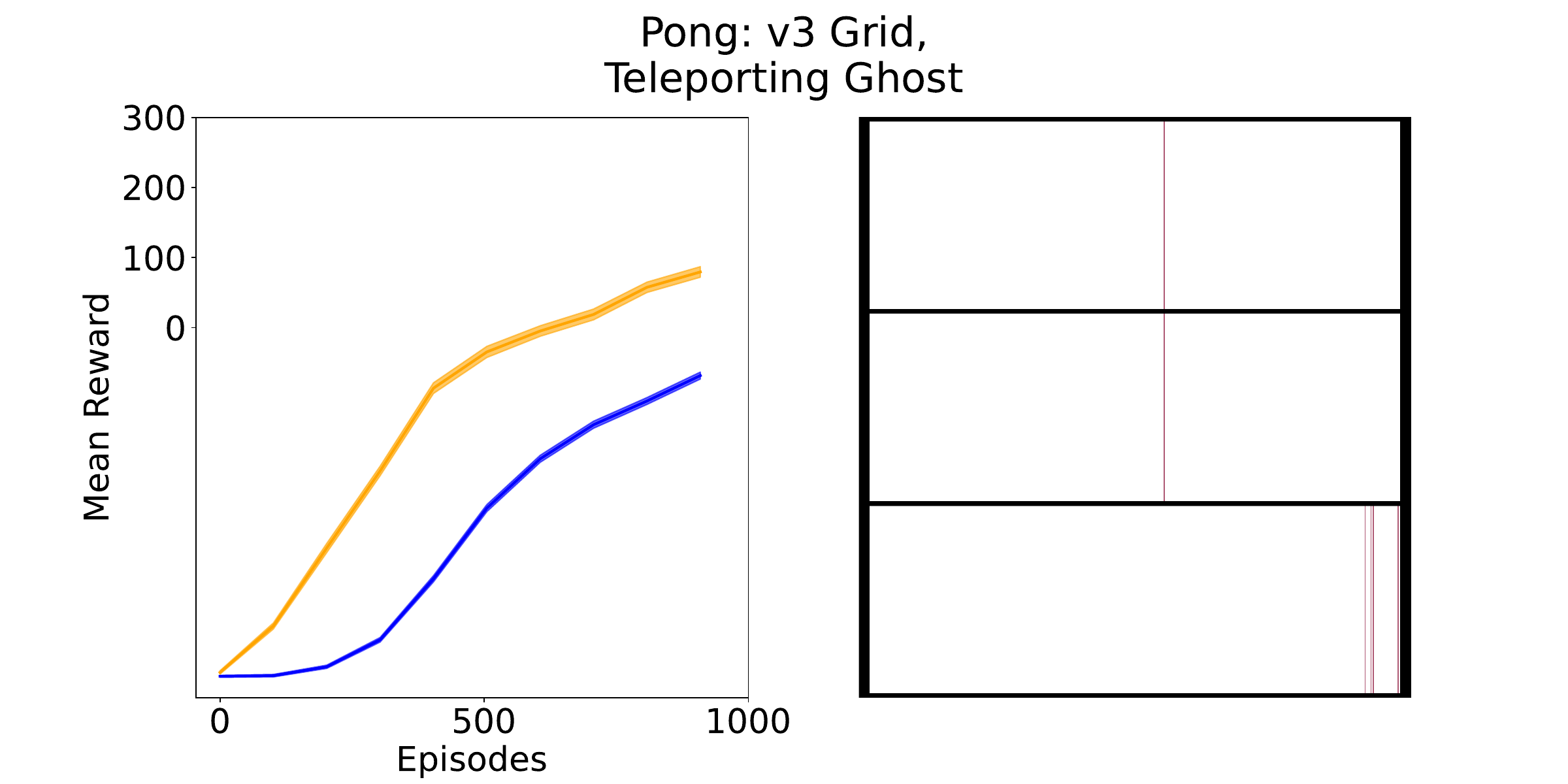}
  \end{subfigure}
  \hfill
  \begin{subfigure}{0.32\textwidth}
    \includegraphics[width=\linewidth]{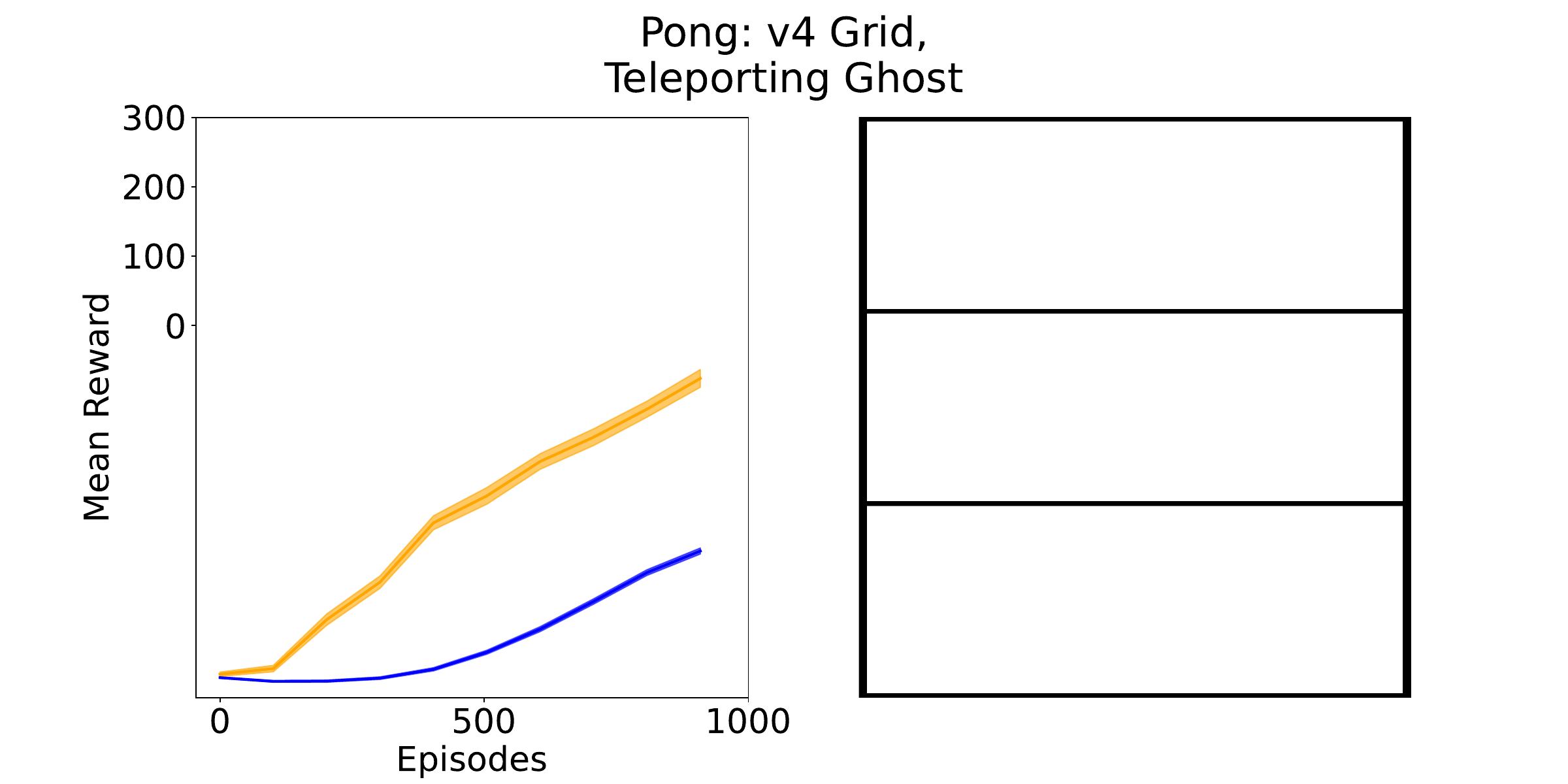}
  \end{subfigure}

  \begin{subfigure}{0.32\textwidth}
    \includegraphics[width=\linewidth]{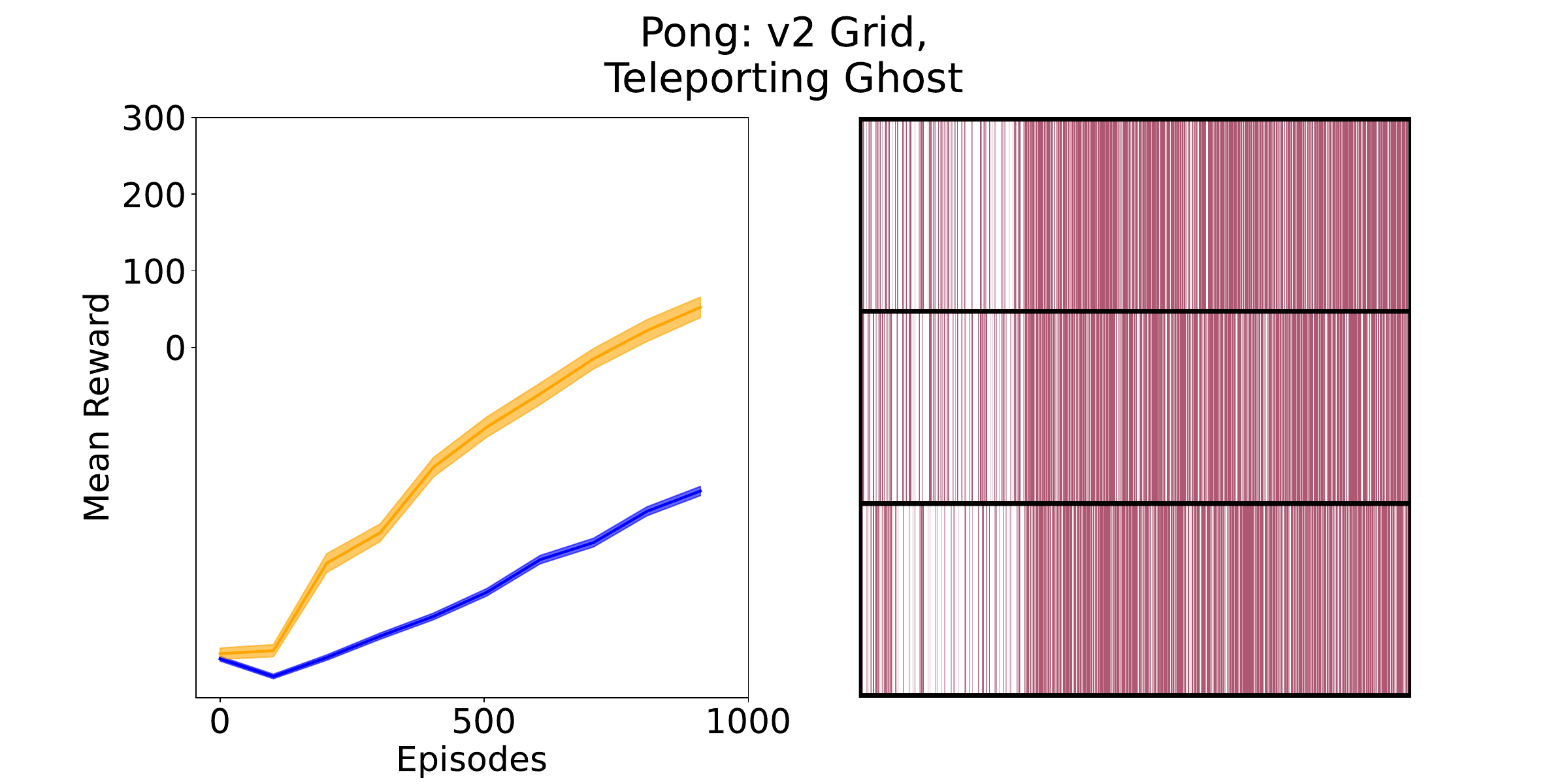}
  \end{subfigure}
  \hfill
  \begin{subfigure}{0.32\textwidth}
    \includegraphics[width=\linewidth]{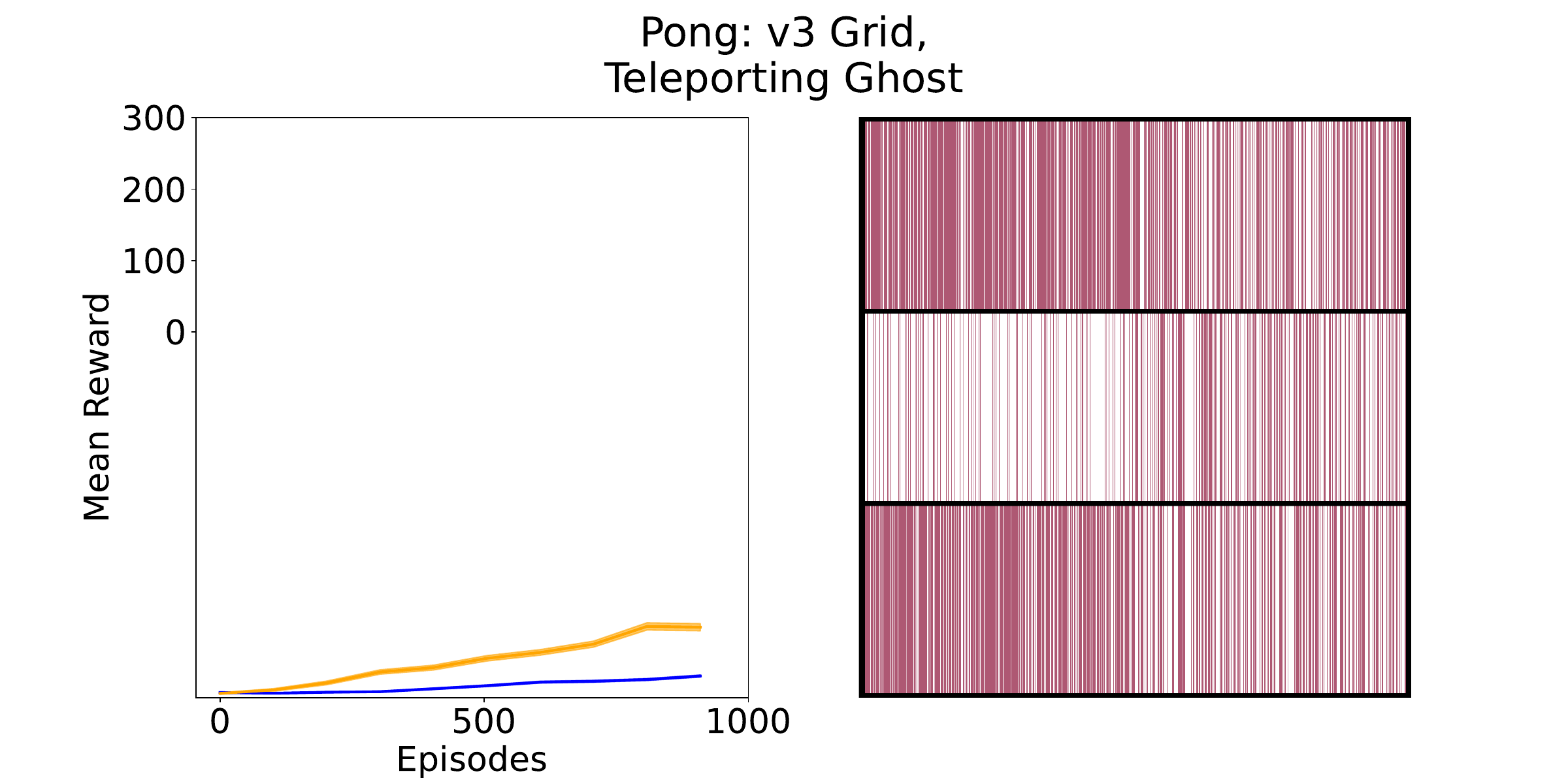}
  \end{subfigure}
  \hfill
  \begin{subfigure}{0.32\textwidth}
    \includegraphics[width=\linewidth]{images/occupancy_grid_pacman/BBv4RandomGhostindex1probmean0std0RandomGhostTeleportingNearWallsindex1probmean0std0.pdf}
  \end{subfigure}

  \caption{\emph{SARSA Agent with $\epsilon\text{-}$greedy exploration strategy}: The \textit{exploration grid} visualizing the difference in State-Action (S-A) pairs explored by these agents ($D_{LG}$). Results for PacMan v2, v3, v4 grids, the agent is trained on Teleporting Ghost variation ($p=0.2$, $p=0.5$) and tested in different environments (reported in the headings). Rows in the right figure represents agent's actions Left, Right, Up, or Down.}
  \label{fig:atari_variations-exploration-semantic-pacman-sarsa-egreedy}
\end{figure*}

Results of the \textit{exploration grid} for PacMan \emph{v2}, \emph{v3}, \emph{v4} are shown in Supplementary figures. In particular, for non-semantic grid variations, \ref{fig:atari_variations-exploration-pacman-qlearning-boltzmann} and \ref{fig:atari_variations-exploration-pacman-qlearning-egreedy} report grid exploration graphs for Q-learning Agent and \ref{fig:atari_variations-exploration-pacman-sarsa-boltzmann} and \ref{fig:atari_variations-exploration-pacman-sarsa-egreedy} for SARSA Agent. Additionally, for semantic games variations, \ref{fig:atari_variations-exploration-semantic-pacman-qlearning-boltzmann} and \ref{fig:atari_variations-exploration-semantic-pacman-qlearning-egreedy} report grid exploration graphs for Q-learning Agent and \ref{fig:atari_variations-exploration-semantic-pacman-sarsa-boltzmann} and \ref{fig:atari_variations-exploration-semantic-pacman-sarsa-egreedy} for SARSA Agent.
\subsection{Pong}
\begin{figure*}[t]
  \centering
  \begin{subfigure}{0.3\textwidth}
    \includegraphics[width=\linewidth]{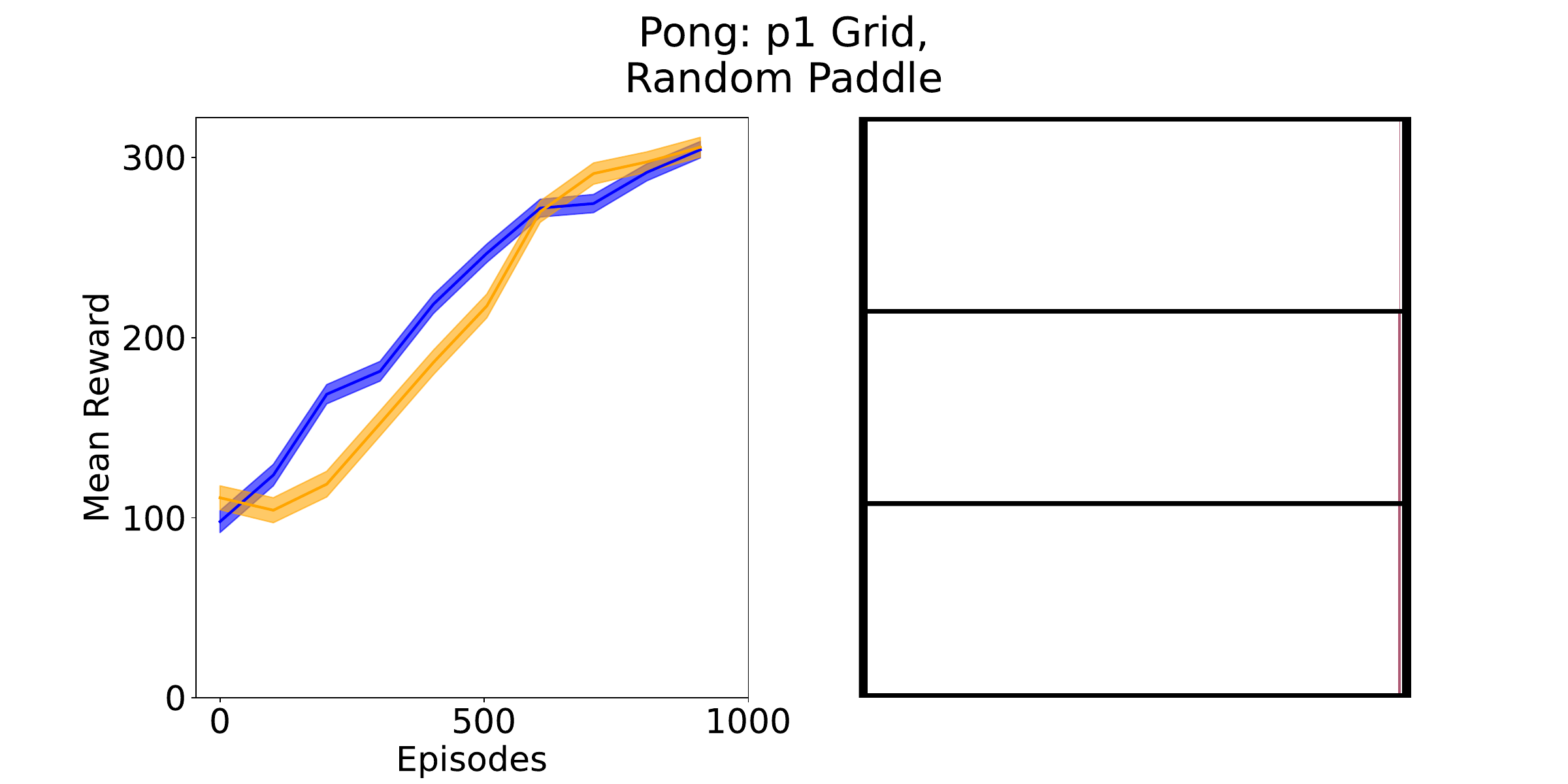}
  \end{subfigure}
  \hfill
  \begin{subfigure}{0.3\textwidth}
    \includegraphics[width=\linewidth]{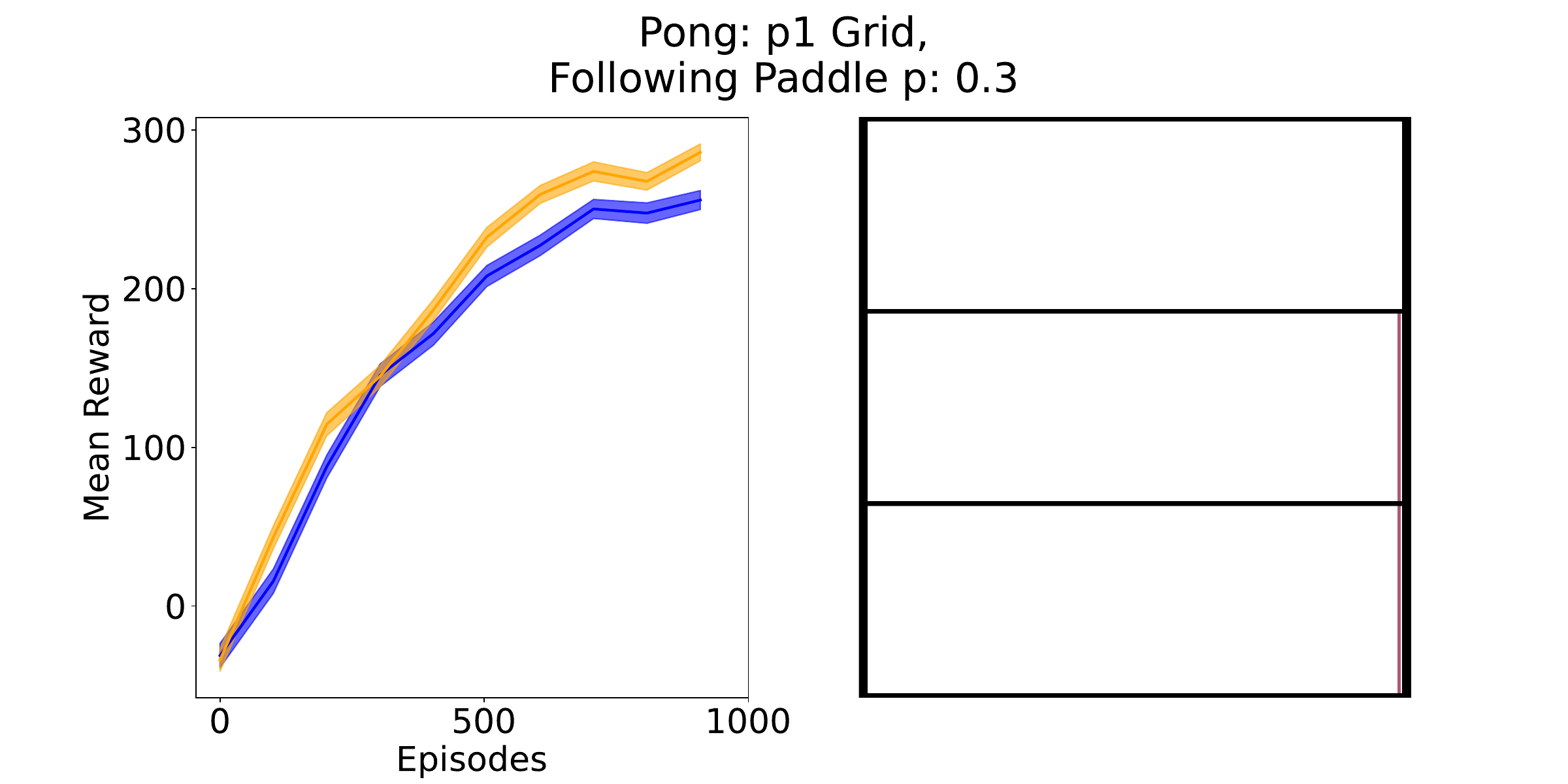}
  \end{subfigure}
  \hfill
  \begin{subfigure}{0.3\textwidth}
    \includegraphics[width=\linewidth]{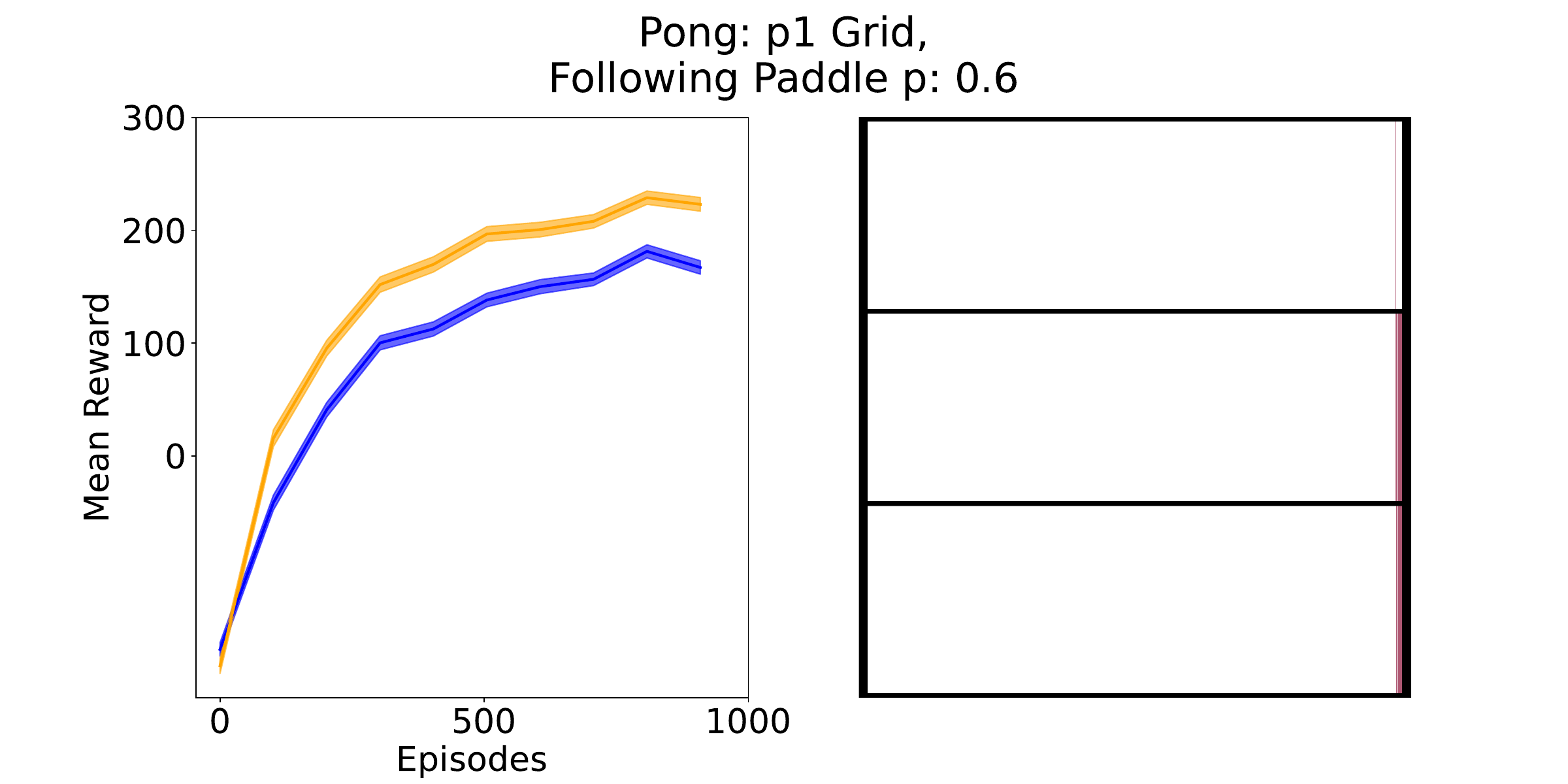}
  \end{subfigure}\\
  
  \begin{subfigure}{0.3\textwidth}
    \includegraphics[width=\linewidth]{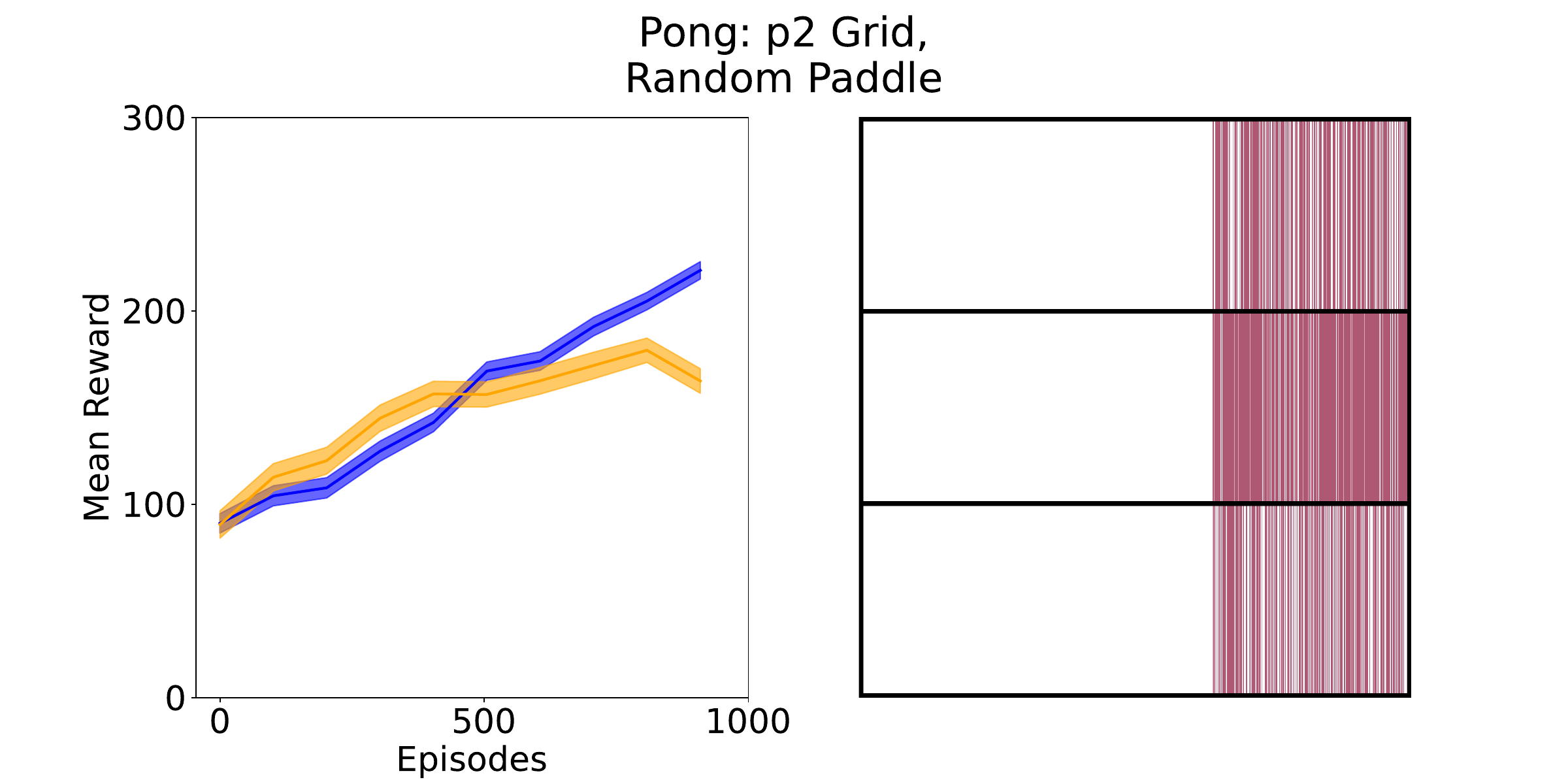}
  \end{subfigure}
  \hfill
  \begin{subfigure}{0.3\textwidth}
    \includegraphics[width=\linewidth]{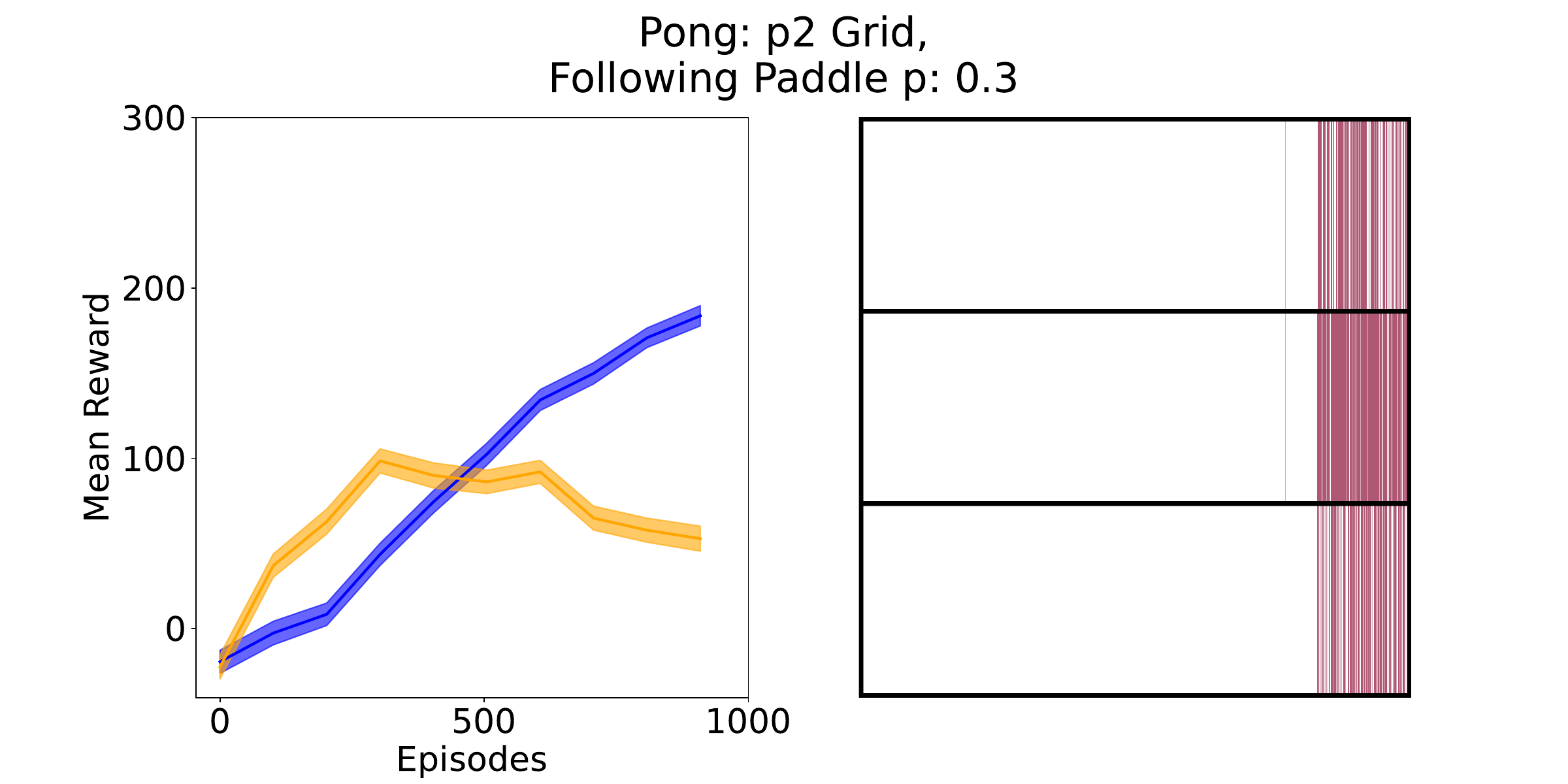}
  \end{subfigure}
  \hfill
  \begin{subfigure}{0.3\textwidth}
    \includegraphics[width=\linewidth]{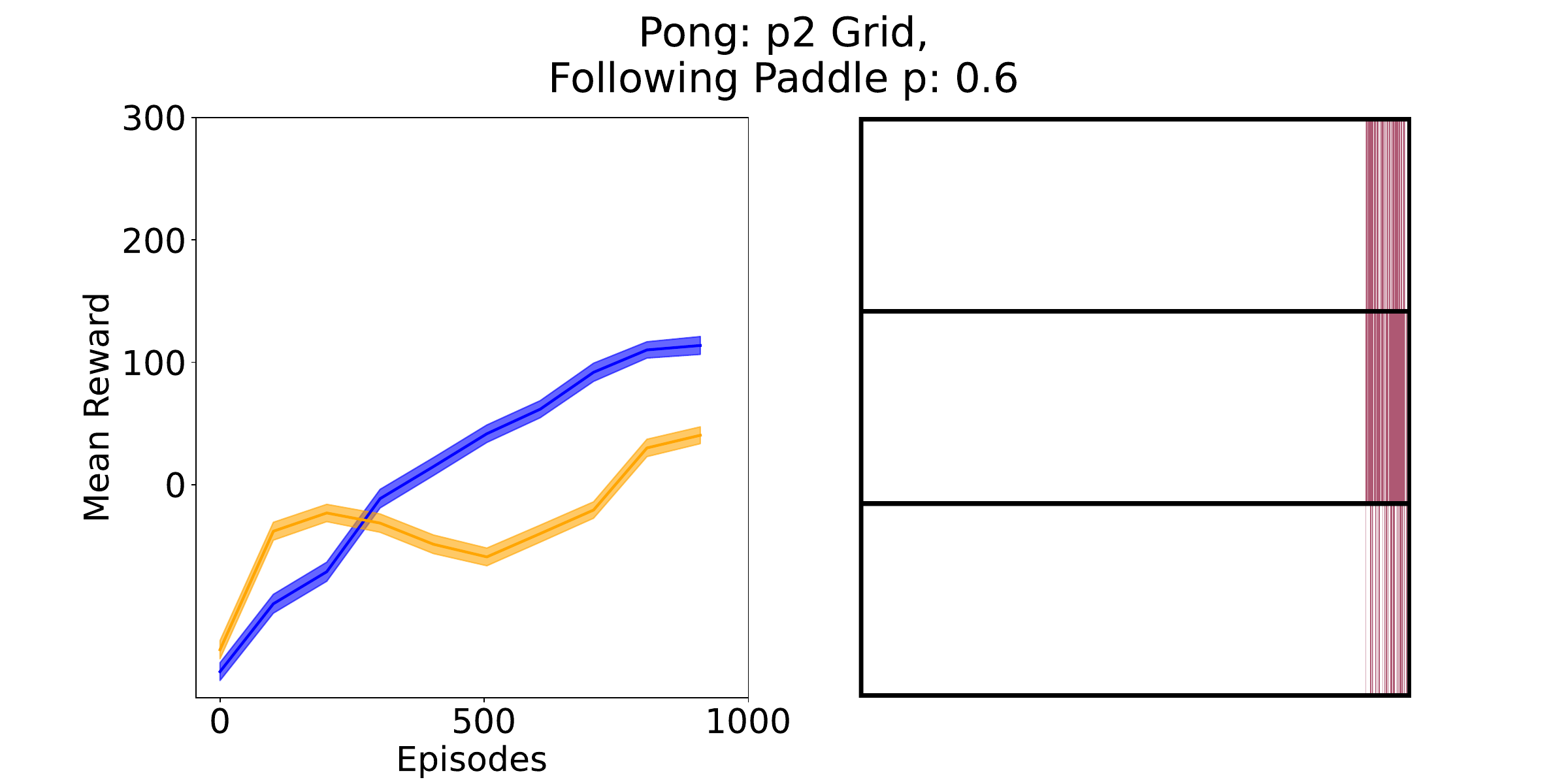}
  \end{subfigure}

  \caption{\emph{Q-learning Agent with Boltzmann exploration strategy}: The \textit{exploration grid} visualizing the difference in State-Action (S-A) pairs explored by these agents ($D_{LG}$). Results for Pong p1, p2 grids, the agent is trained on non-noisy variations of different environments (reported in the headings) and tested in the Low-Noise regime. Rows in the right figure represents agent's actions Left, Right, Stop.}
  \label{fig:atari_variations-exploration-pong-qlearning-boltzmann}
\end{figure*}

\begin{figure*}[t]
  \centering
  \begin{subfigure}{0.3\textwidth}
    \includegraphics[width=\linewidth]{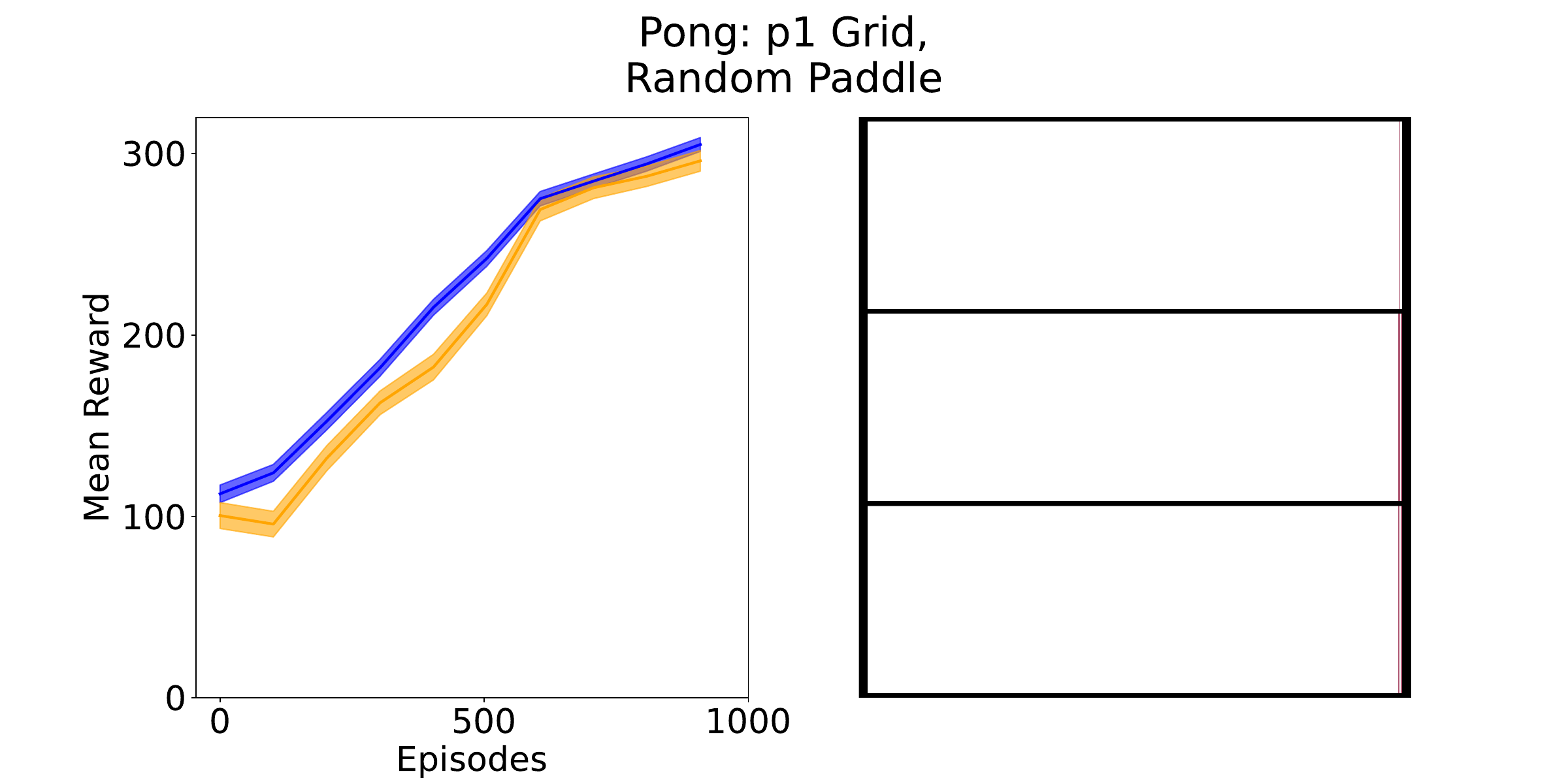}
  \end{subfigure}
  \hfill
  \begin{subfigure}{0.3\textwidth}
    \includegraphics[width=\linewidth]{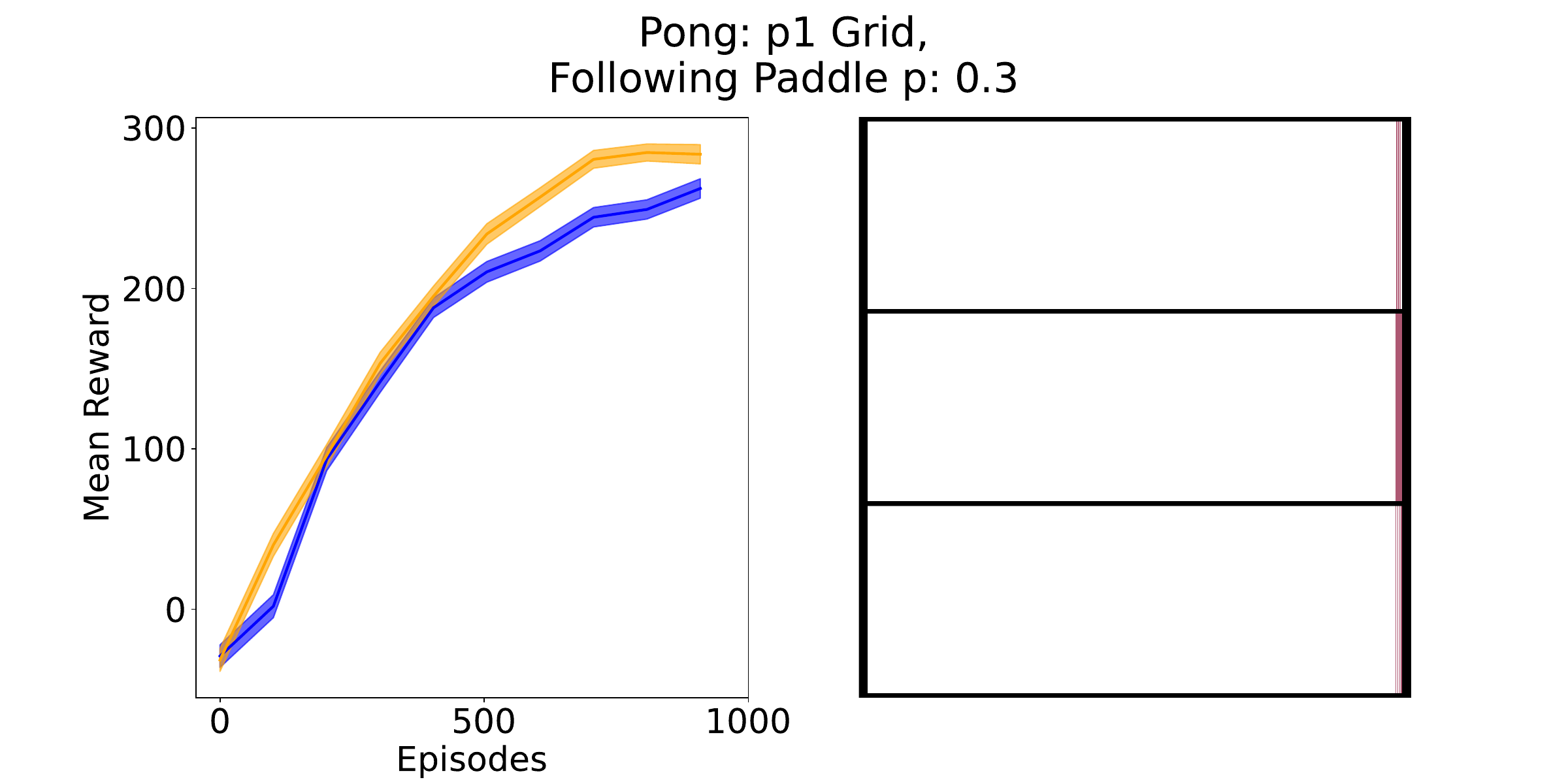}
  \end{subfigure}
  \hfill
  \begin{subfigure}{0.3\textwidth}
    \includegraphics[width=\linewidth]{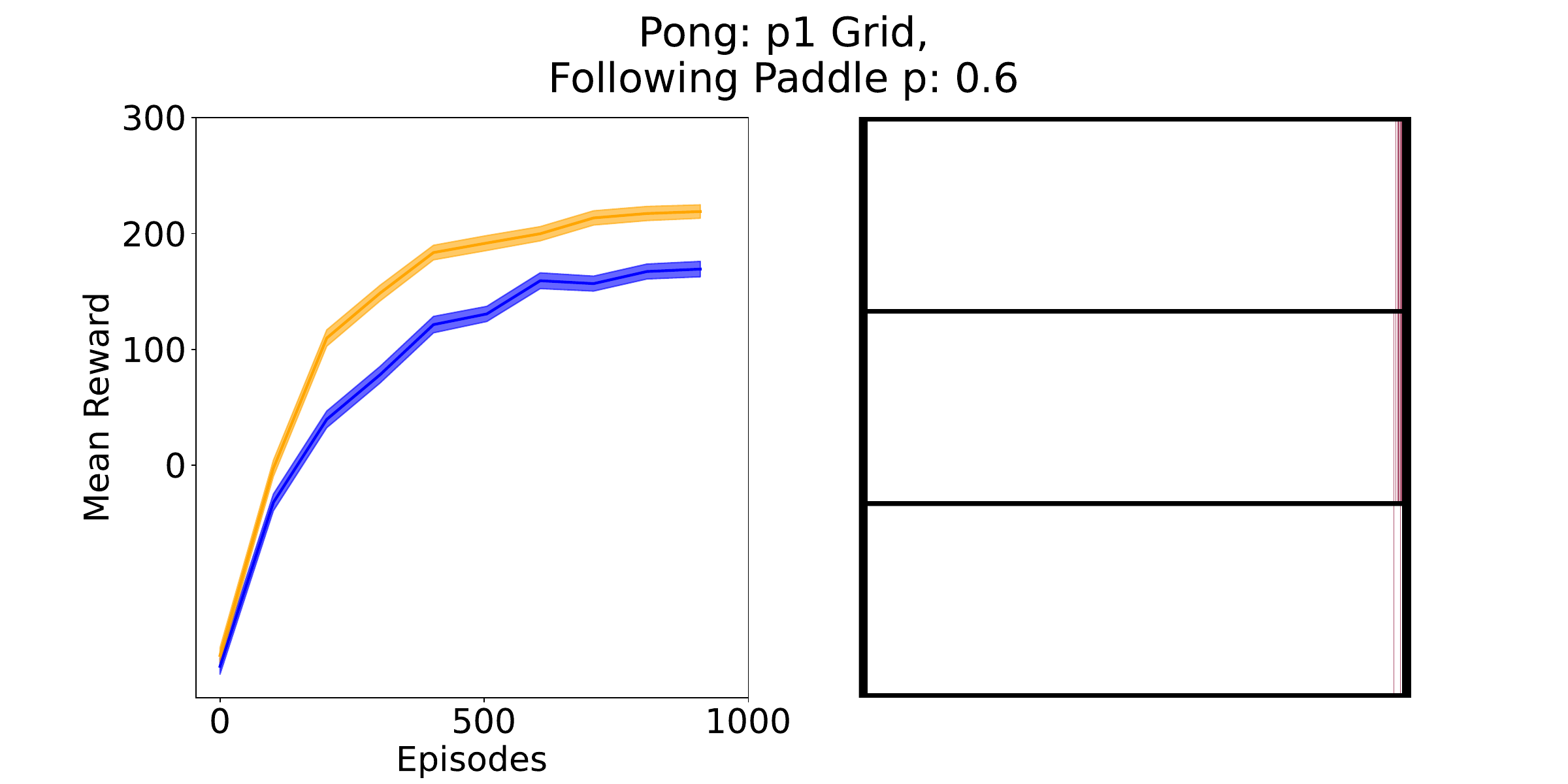}
  \end{subfigure}\\
  
  \begin{subfigure}{0.3\textwidth}
    \includegraphics[width=\linewidth]{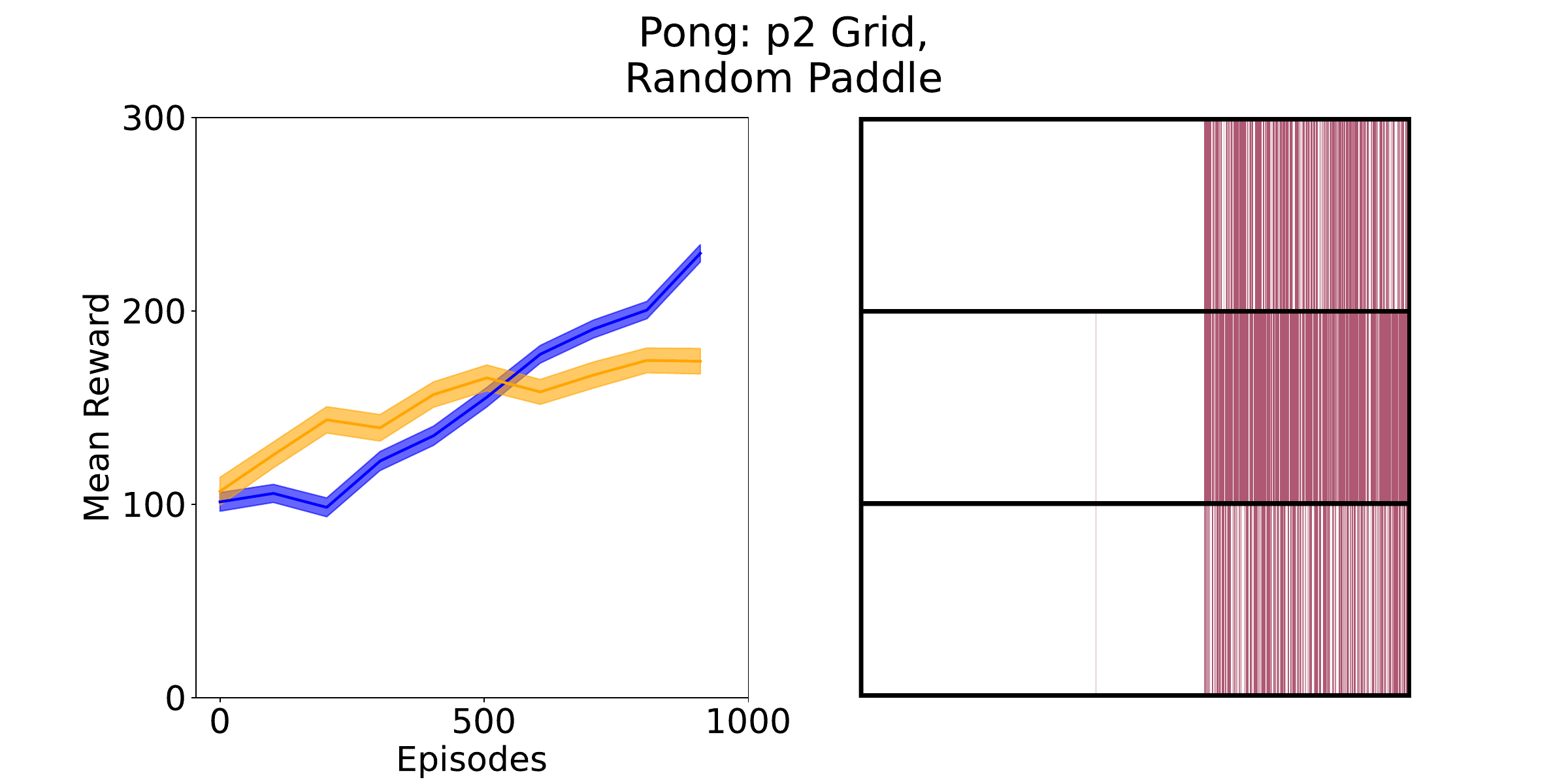}
  \end{subfigure}
  \hfill
  \begin{subfigure}{0.3\textwidth}
    \includegraphics[width=\linewidth]{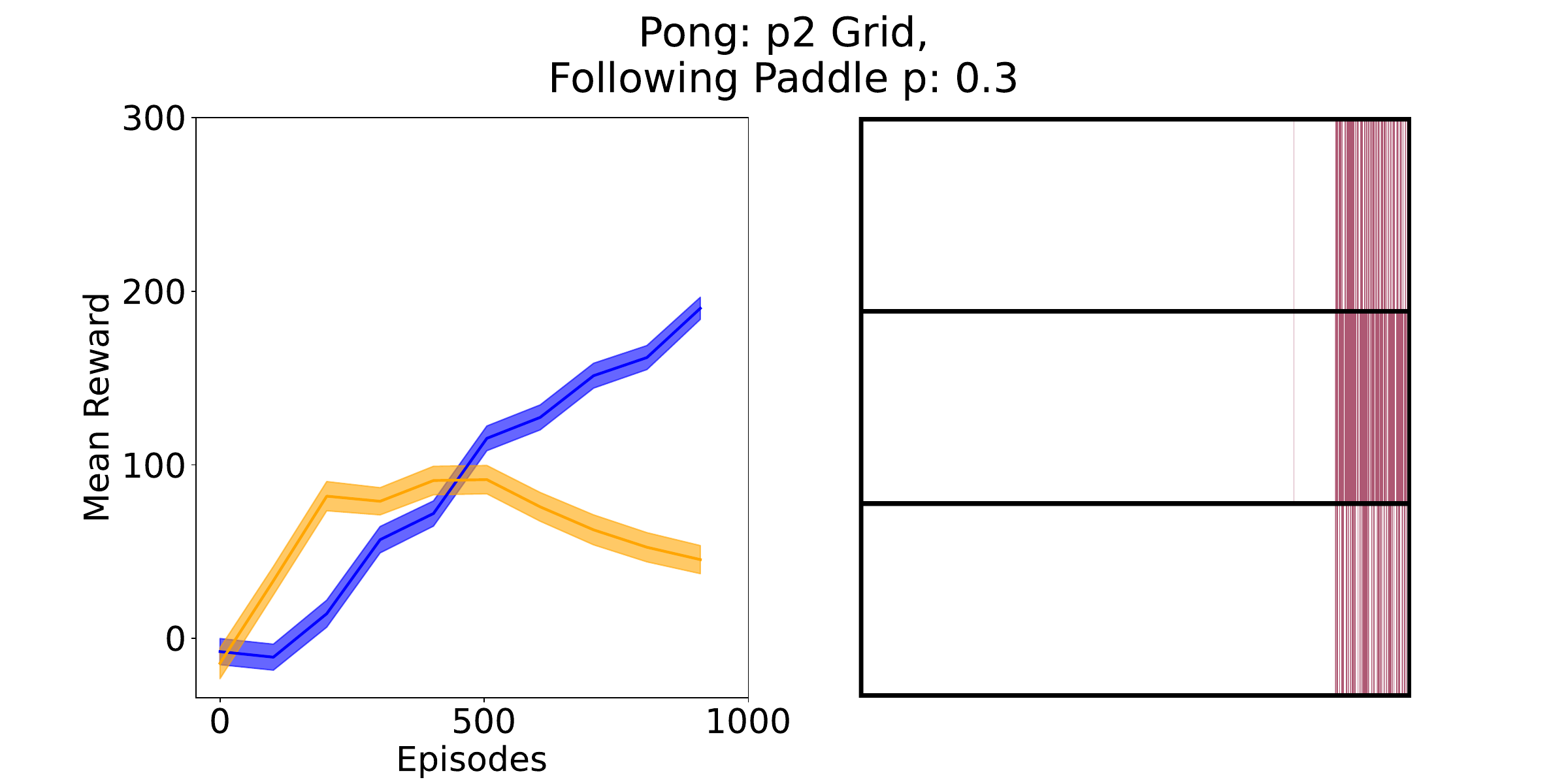}
  \end{subfigure}
  \hfill
  \begin{subfigure}{0.3\textwidth}
    \includegraphics[width=\linewidth]{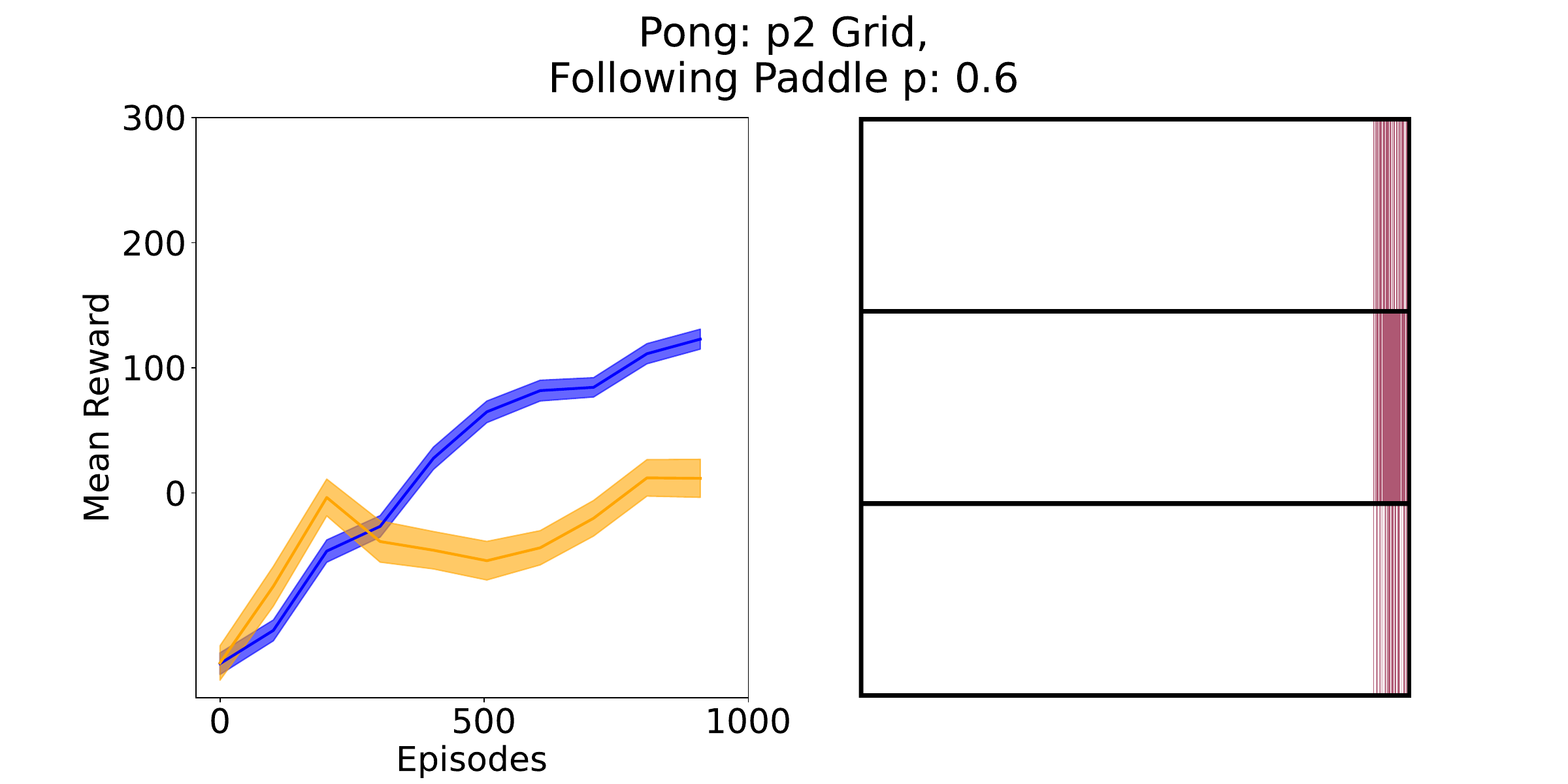}
  \end{subfigure}

  \caption{\emph{Q-learning Agent with $\epsilon\text{-}$greedy exploration strategy}: The \textit{exploration grid} visualizing the difference in State-Action (S-A) pairs explored by these agents ($D_{LG}$). Results for Pong p1, p2 grids, the agent is trained on non-noisy variations of different environments (reported in the headings) and tested in the Low-Noise regime. Rows in the right figure represents agent's actions Left, Right, Stop.}
  \label{fig:atari_variations-exploration-pong-qlearning-egreedy}
\end{figure*}

\begin{figure*}[t]
  \centering
  \begin{subfigure}{0.3\textwidth}
    \includegraphics[width=\linewidth]{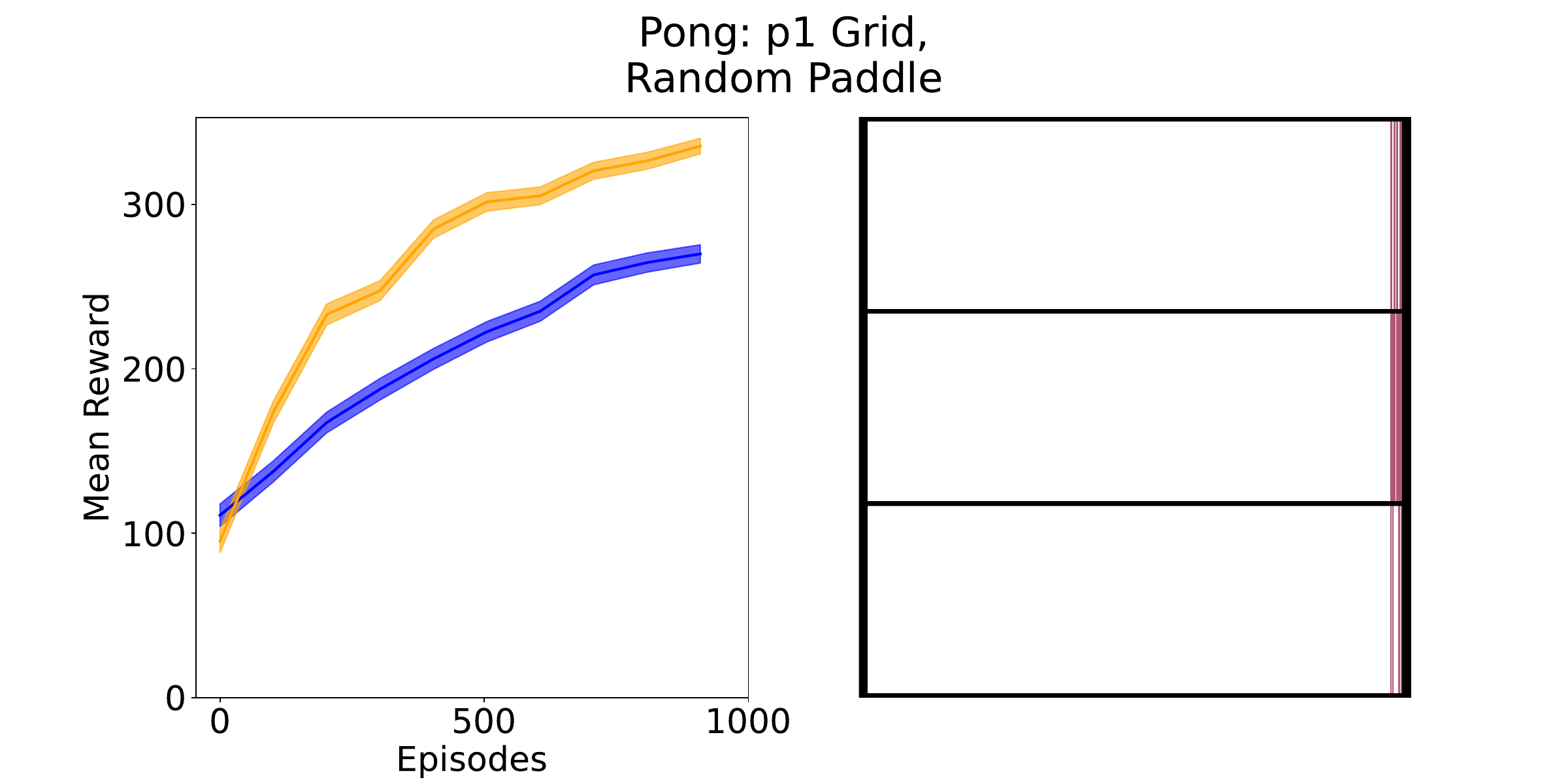}
  \end{subfigure}
  \hfill
  \begin{subfigure}{0.3\textwidth}
    \includegraphics[width=\linewidth]{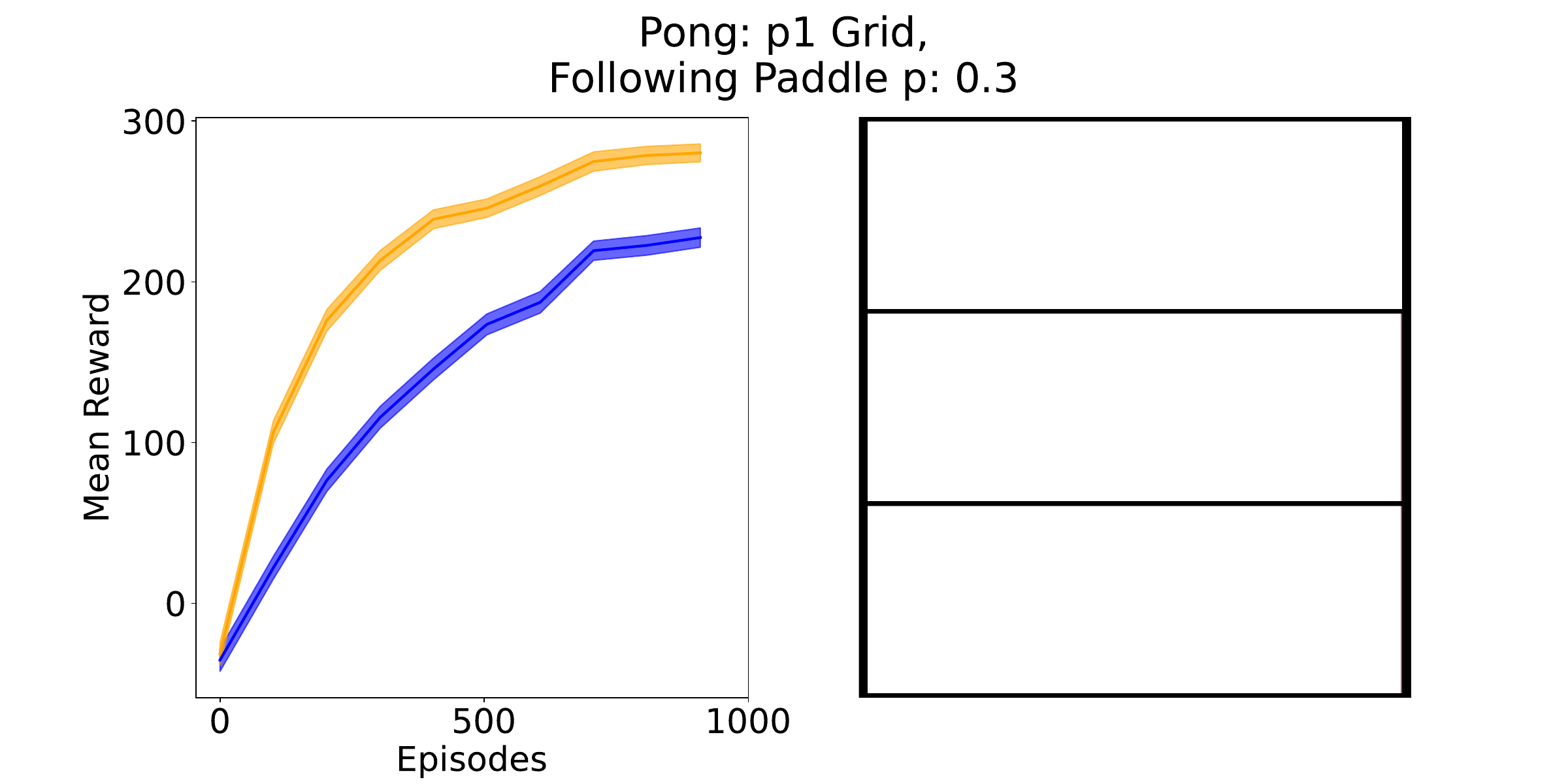}
  \end{subfigure}
  \hfill
  \begin{subfigure}{0.3\textwidth}
    \includegraphics[width=\linewidth]{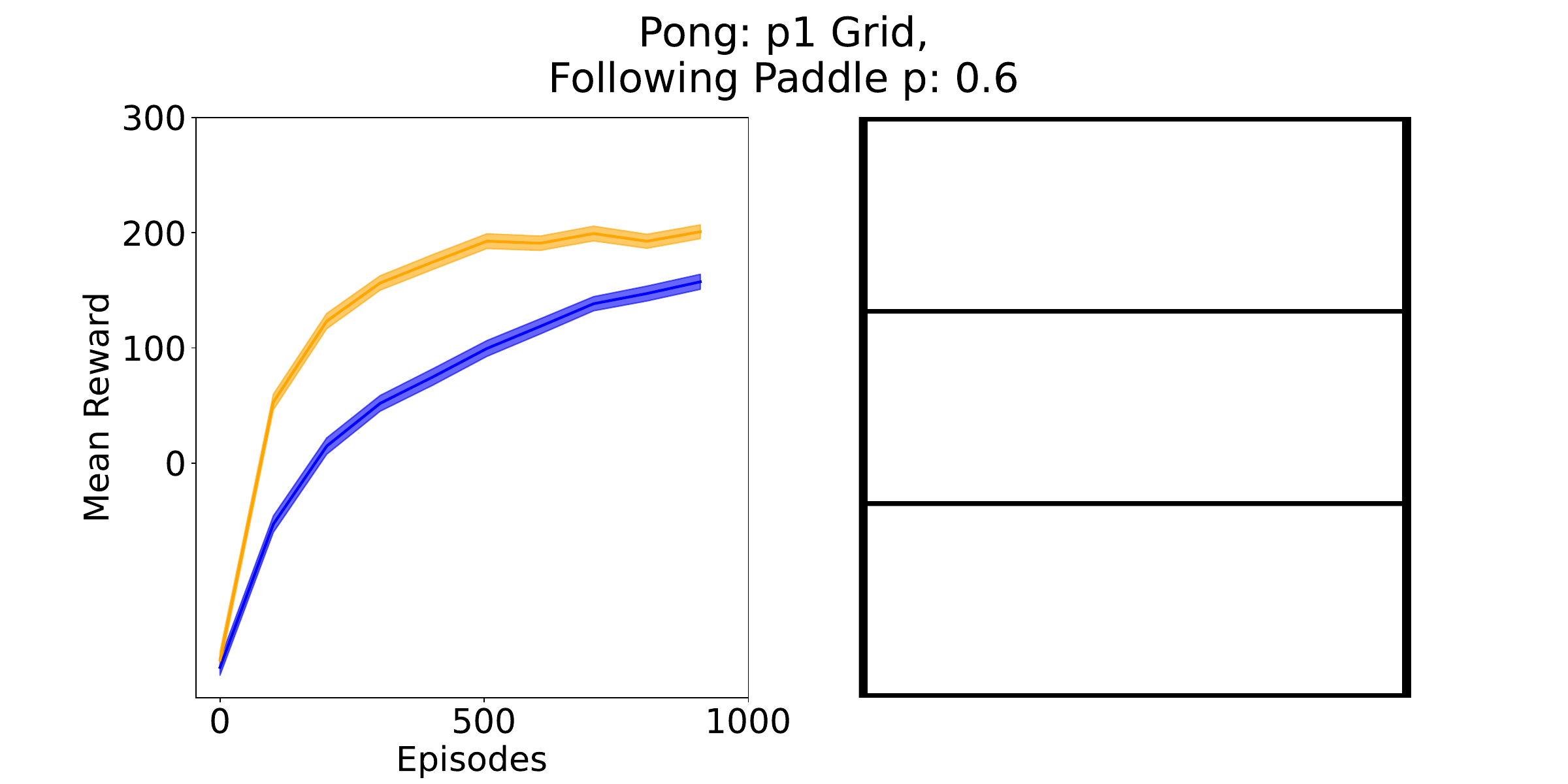}
  \end{subfigure}\\
  
  \begin{subfigure}{0.3\textwidth}
    \includegraphics[width=\linewidth]{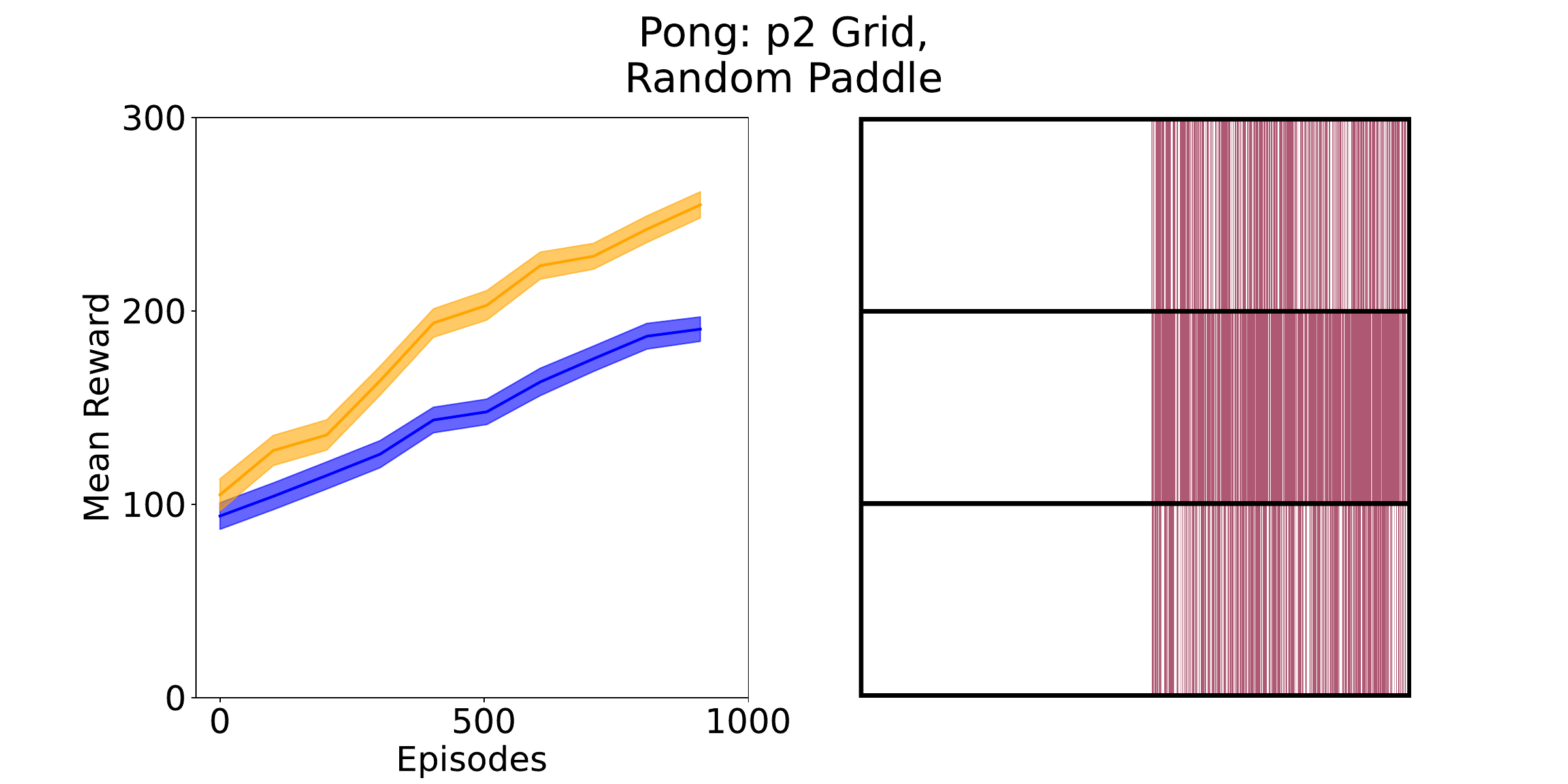}
  \end{subfigure}
  \hfill
  \begin{subfigure}{0.3\textwidth}
    \includegraphics[width=\linewidth]{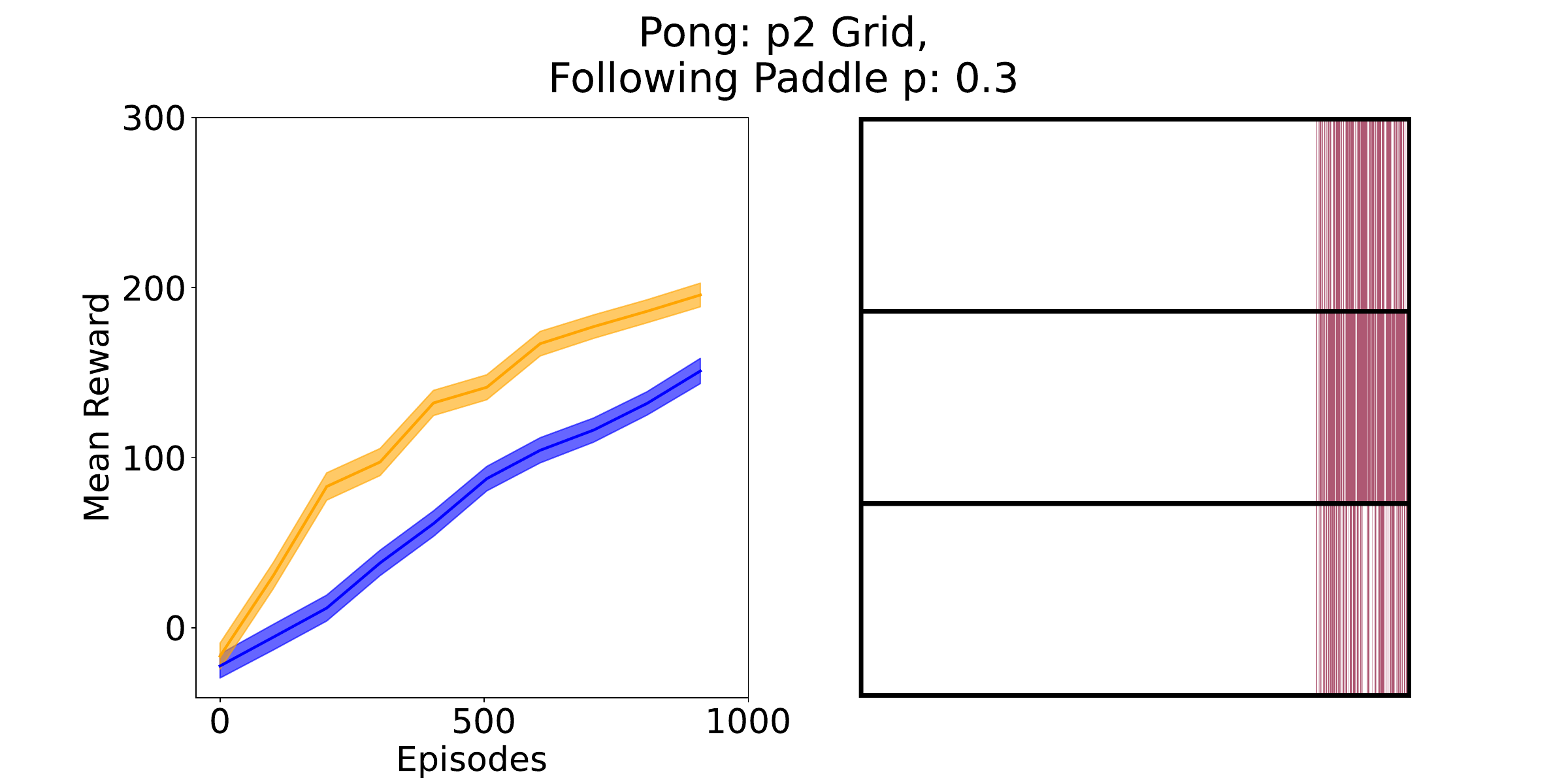}
  \end{subfigure}
  \hfill
  \begin{subfigure}{0.3\textwidth}
    \includegraphics[width=\linewidth]{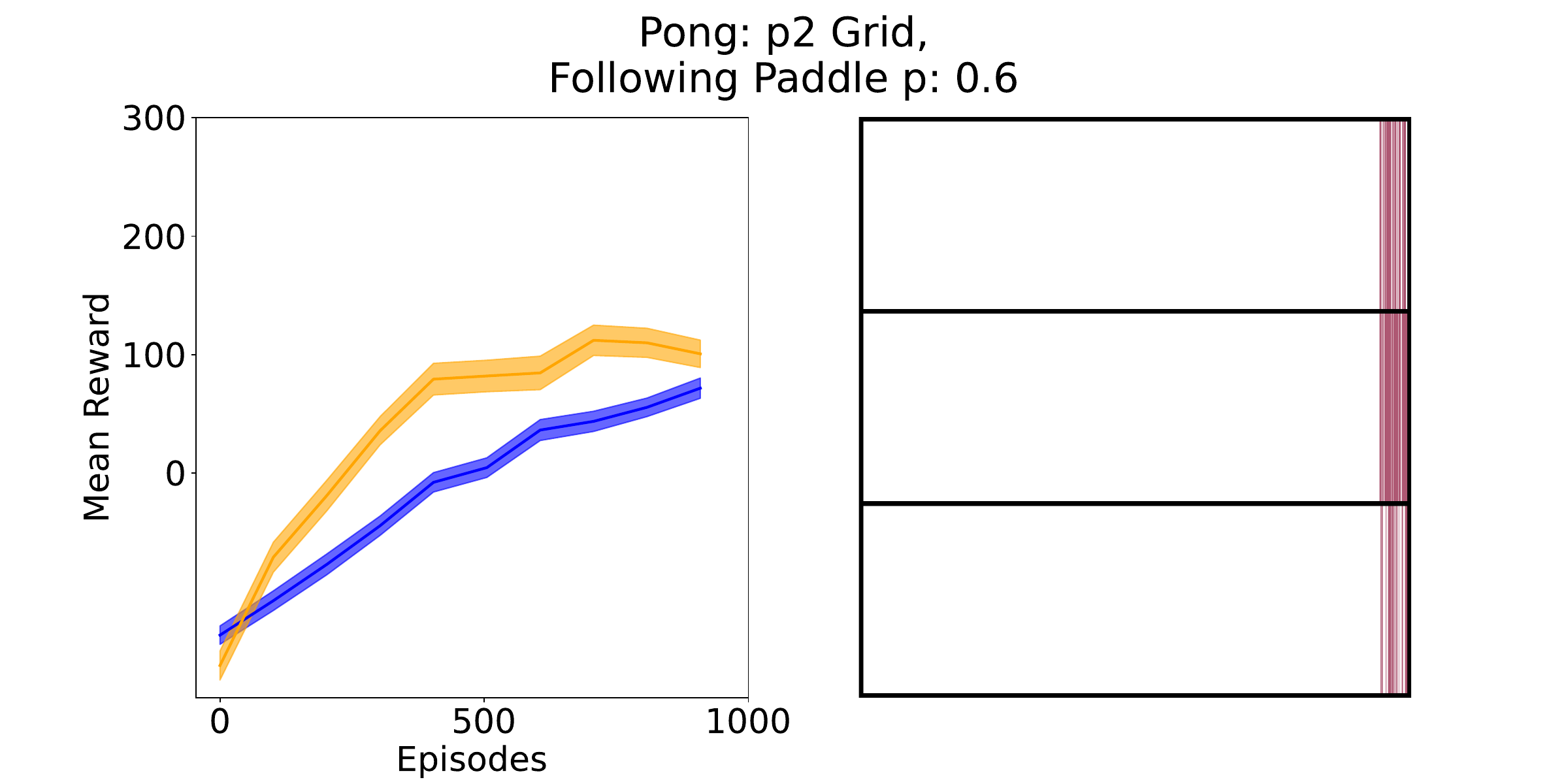}
  \end{subfigure}

  \caption{\emph{SARSA Agent with Boltzmann exploration strategy}: The \textit{exploration grid} visualizing the difference in State-Action (S-A) pairs explored by these agents ($D_{LG}$). Results for Pong p1, p2 grids, the agent is trained on non-noisy variations of different environments (reported in the headings) and tested in the Low-Noise regime. Rows in the right figure represents agent's actions Left, Right, Stop.}
  \label{fig:atari_variations-exploration-pong-sarsa-boltzmann}
\end{figure*}

\begin{figure*}[t]
  \centering
  \begin{subfigure}{0.3\textwidth}
    \includegraphics[width=\linewidth]{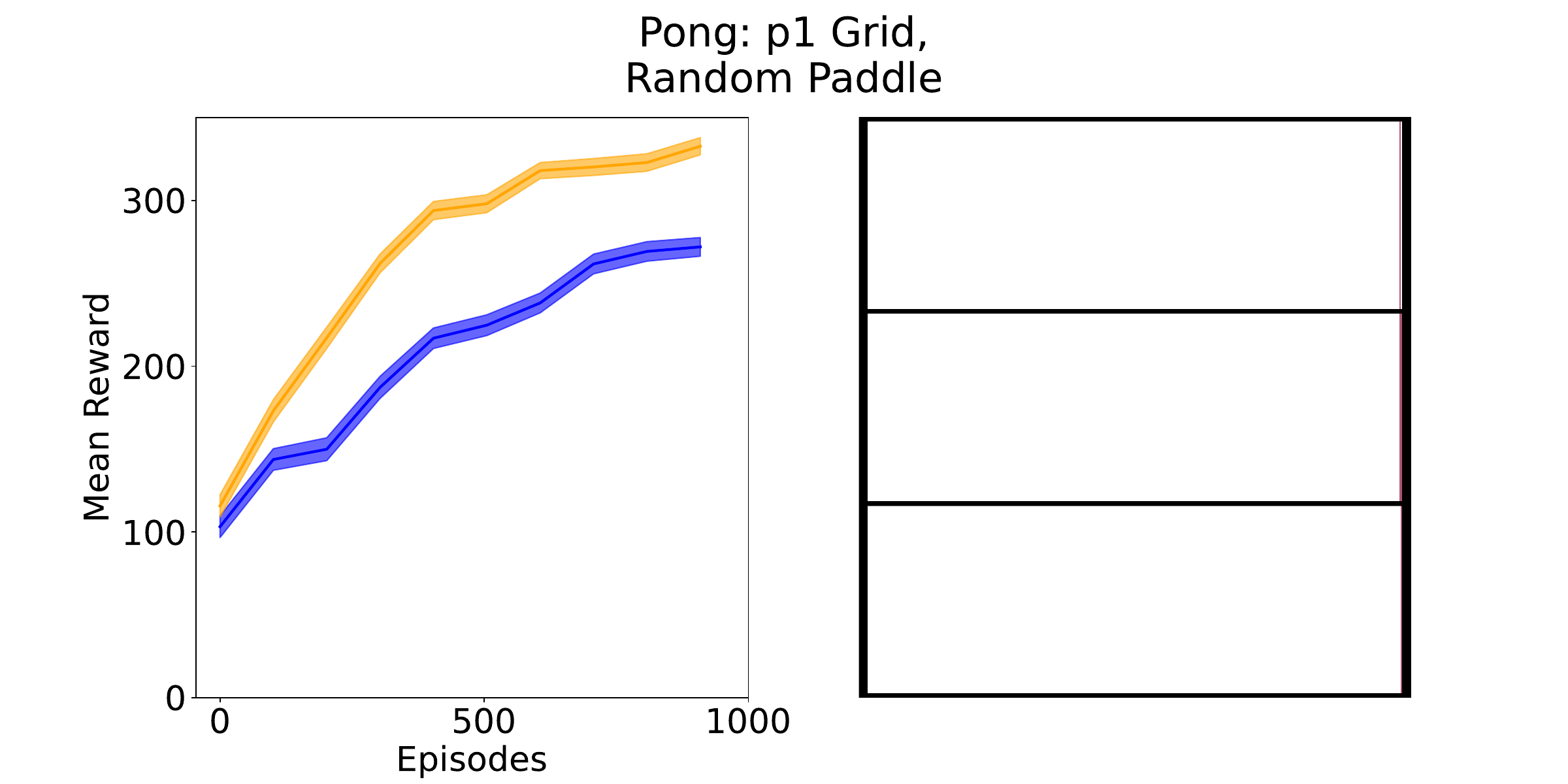}
  \end{subfigure}
  \hfill
  \begin{subfigure}{0.3\textwidth}
    \includegraphics[width=\linewidth]{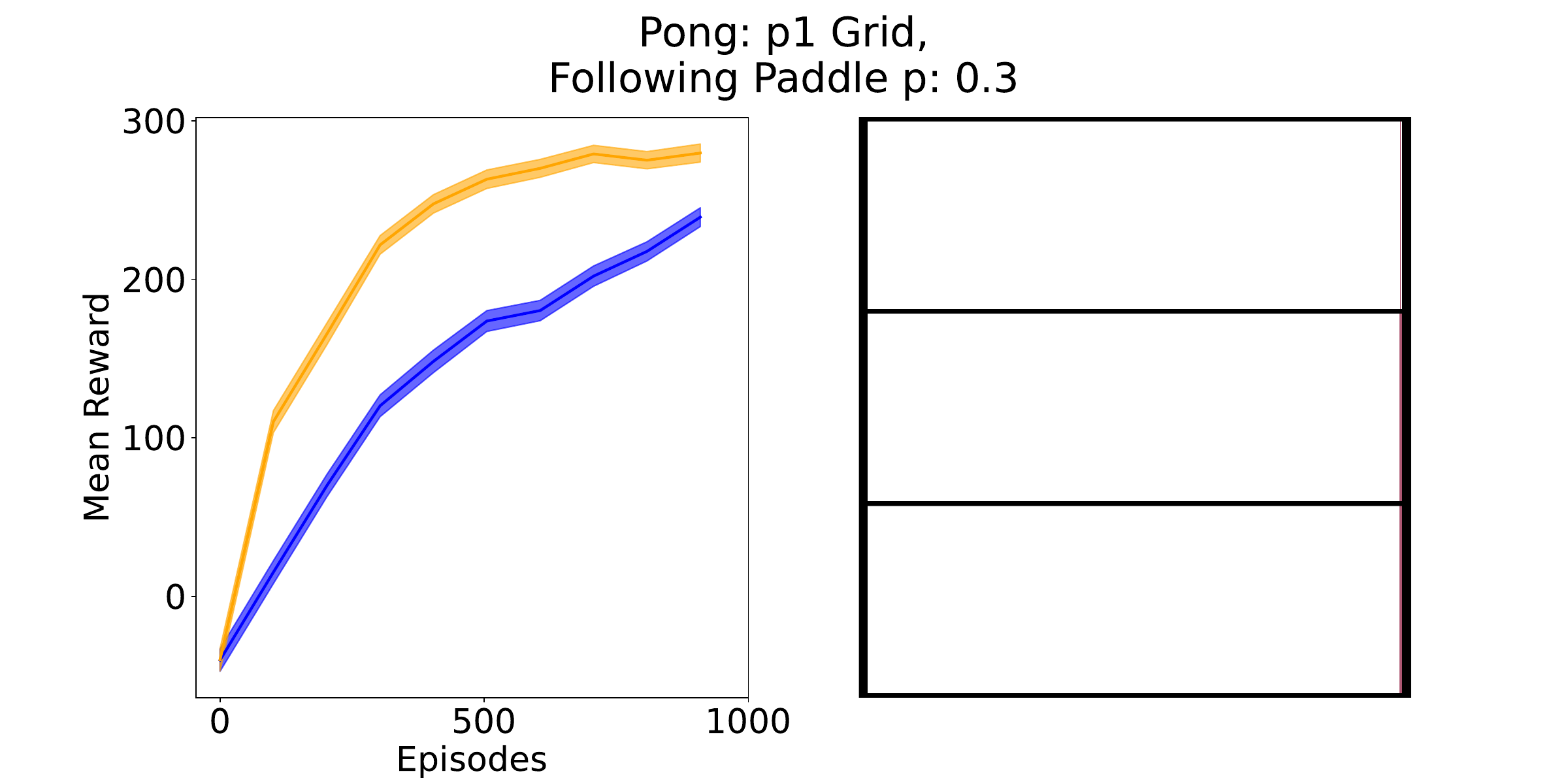}
  \end{subfigure}
  \hfill
  \begin{subfigure}{0.3\textwidth}
    \includegraphics[width=\linewidth]{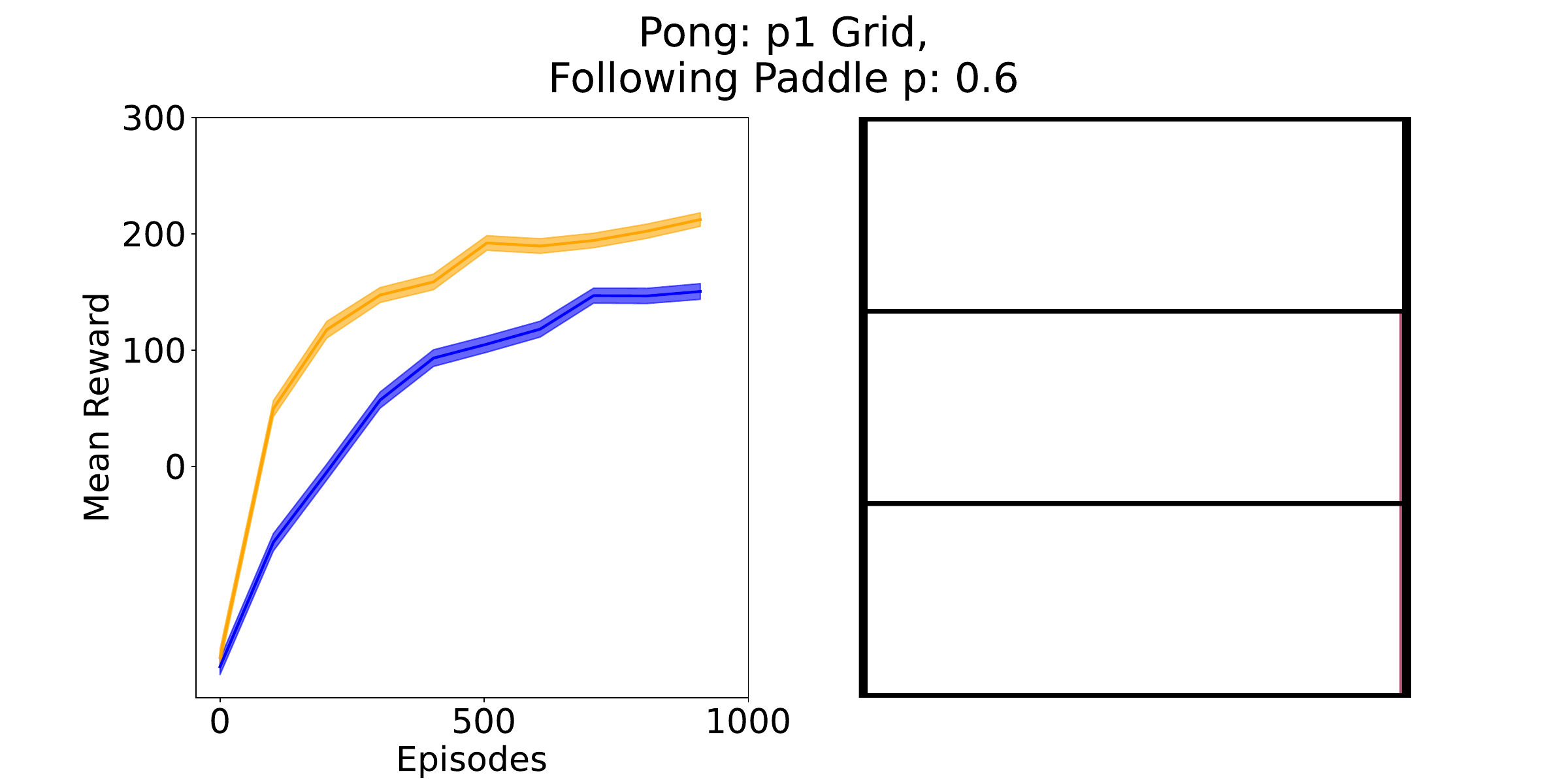}
  \end{subfigure}\\
  
  \begin{subfigure}{0.3\textwidth}
    \includegraphics[width=\linewidth]{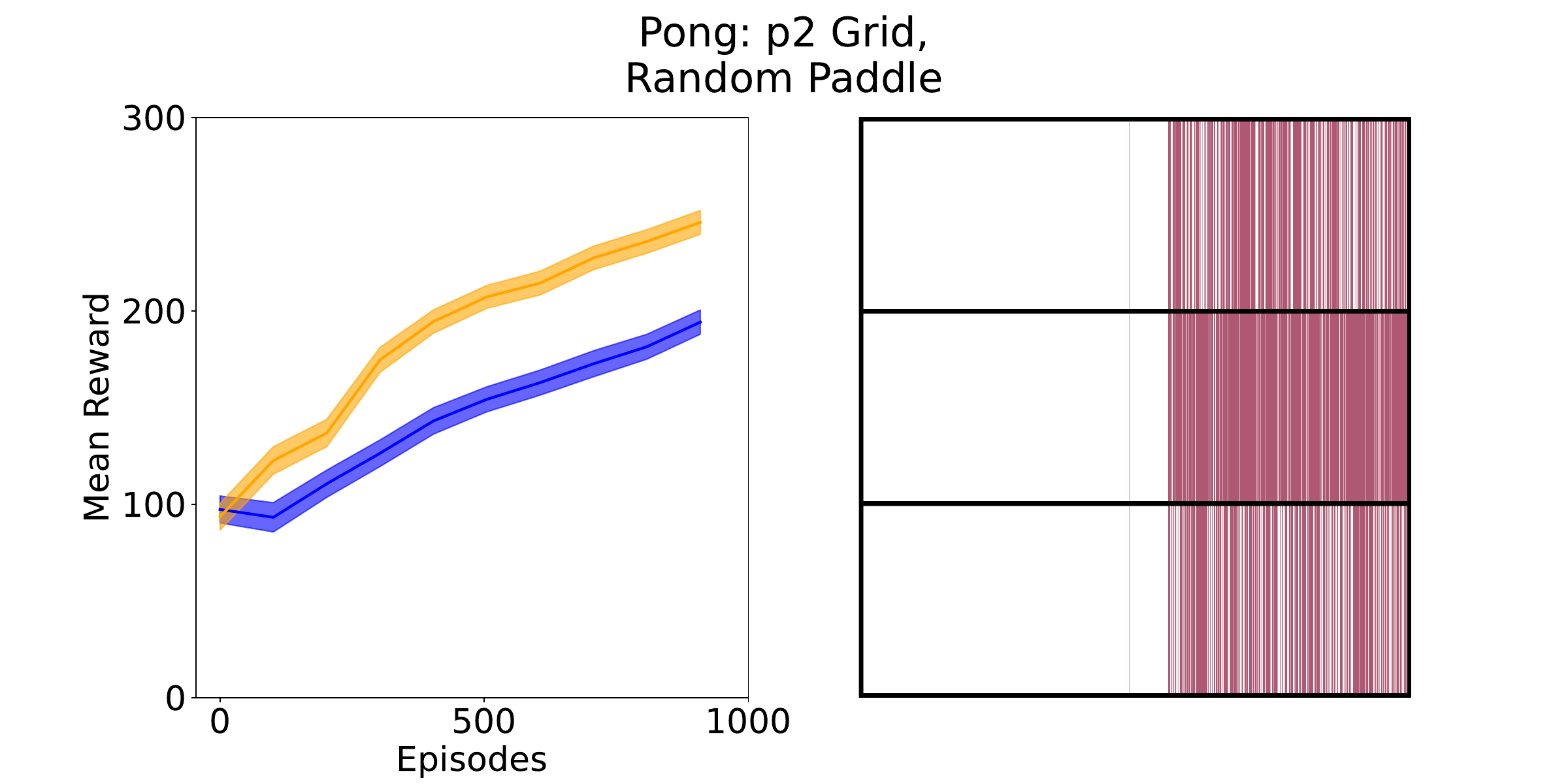}
  \end{subfigure}
  \hfill
  \begin{subfigure}{0.3\textwidth}
    \includegraphics[width=\linewidth]{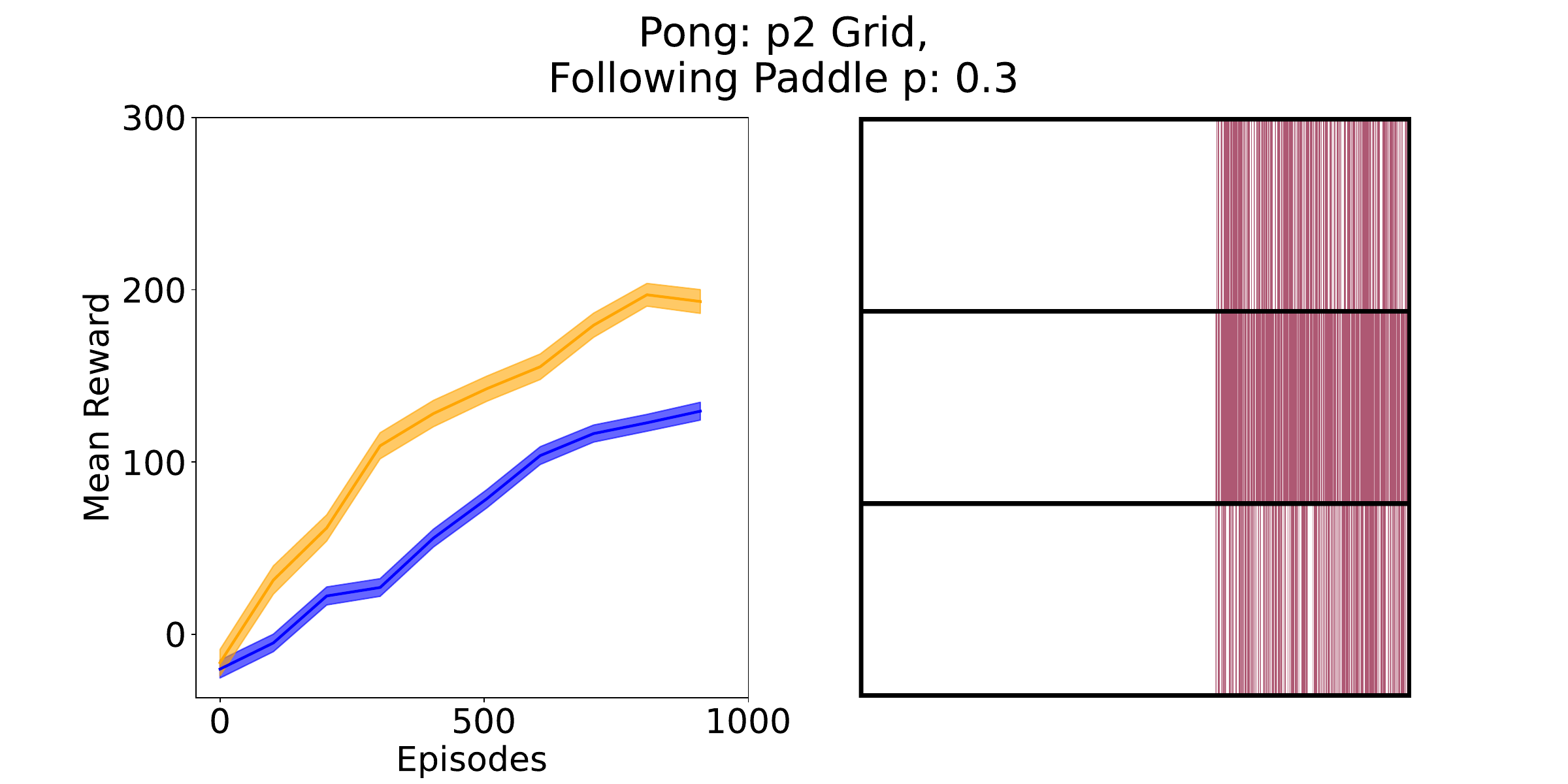}
  \end{subfigure}
  \hfill
  \begin{subfigure}{0.3\textwidth}
    \includegraphics[width=\linewidth]{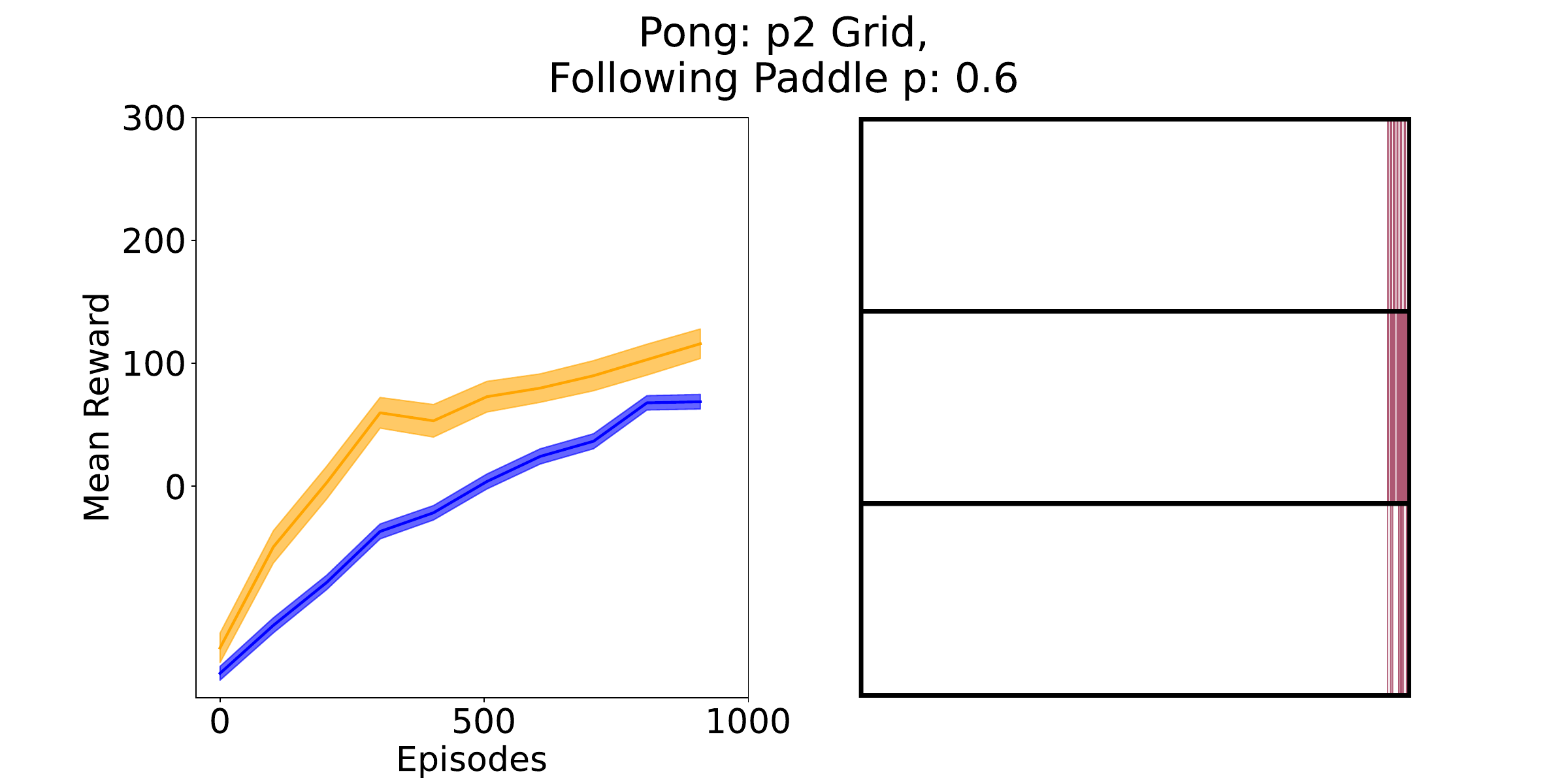}
  \end{subfigure}

  \caption{\emph{SARSA Agent with $\epsilon\text{-}$greedy exploration strategy}: The \textit{exploration grid} visualizing the difference in State-Action (S-A) pairs explored by these agents ($D_{LG}$). Results for Pong p1, p2 grids, the agent is trained on non-noisy variations of different environments (reported in the headings) and tested in the Low-Noise regime. Rows in the right figure represents agent's actions Left, Right, Stop.}
  \label{fig:atari_variations-exploration-pong-sarsa-egreedy}
\end{figure*}

\begin{figure*}[t]
  \centering
  \begin{subfigure}{0.32\textwidth}
    \includegraphics[width=\linewidth]{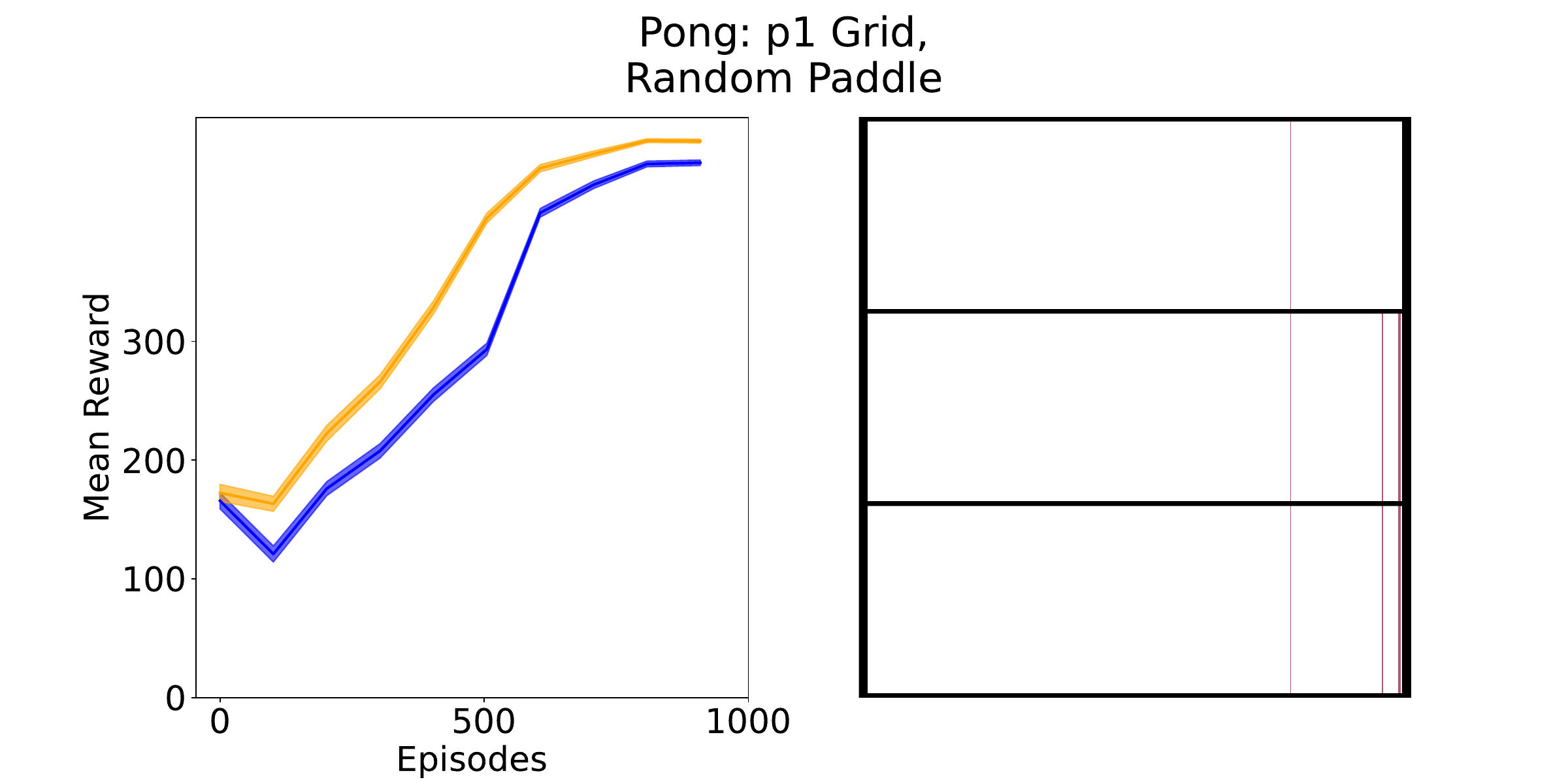}
  \end{subfigure}
  \hspace{0.6mm}
  \begin{subfigure}{0.32\textwidth}
    \includegraphics[width=\linewidth]{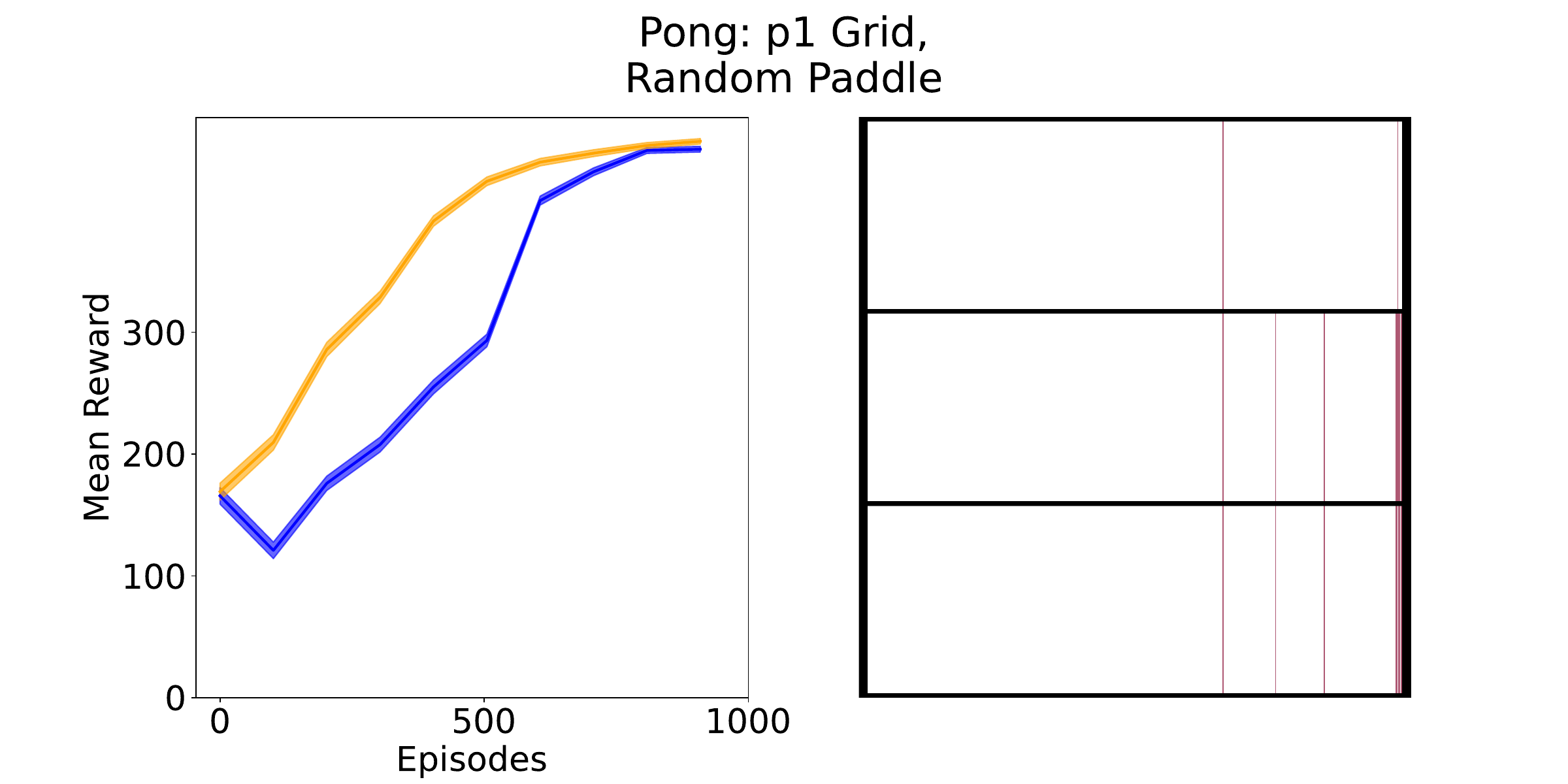}
  \end{subfigure}\\
  
  \begin{subfigure}{0.32\textwidth}
    \includegraphics[width=\linewidth]{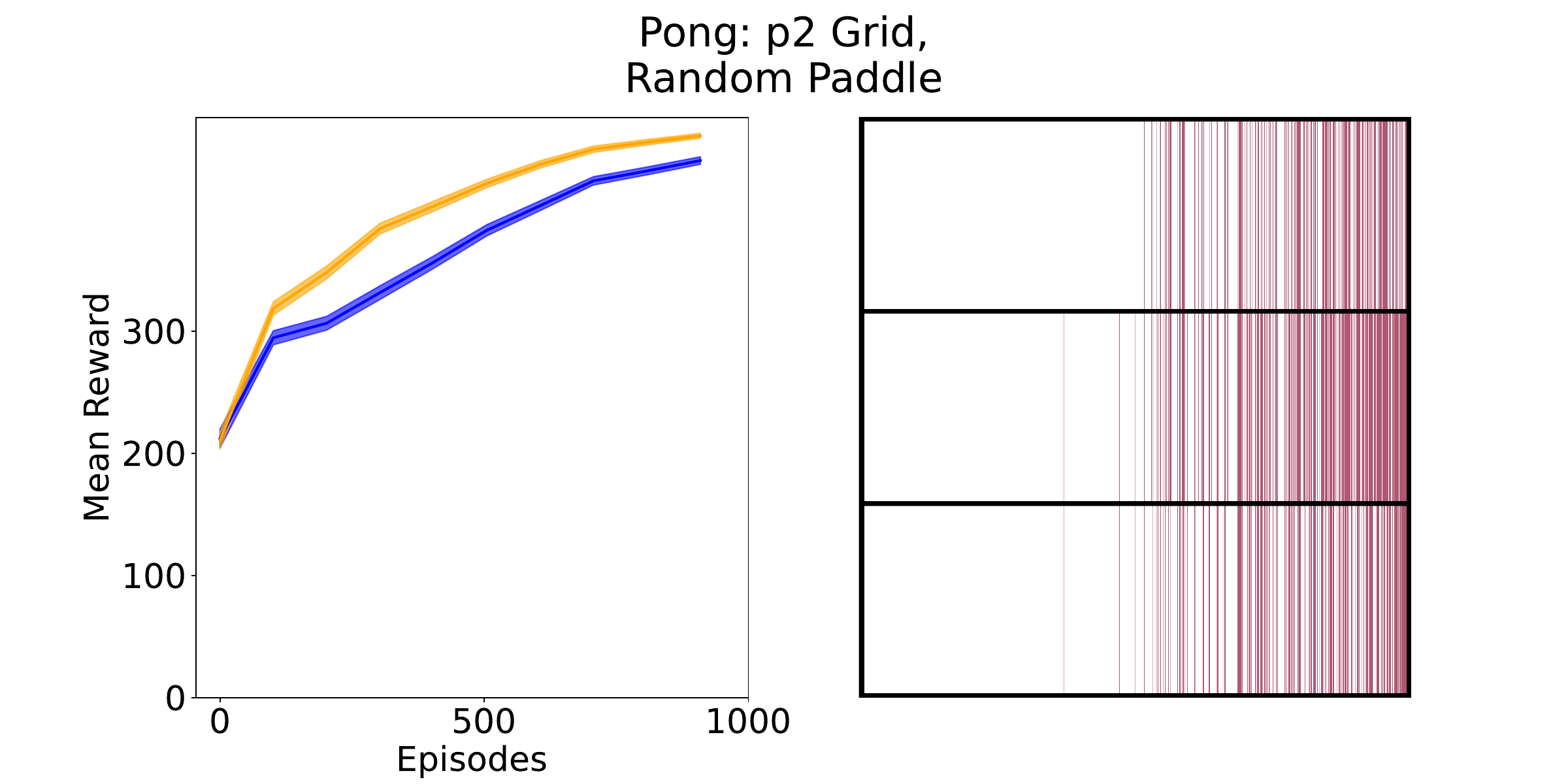}
  \end{subfigure}
  \hspace{0.6mm}
  \begin{subfigure}{0.32\textwidth}
    \includegraphics[width=\linewidth]{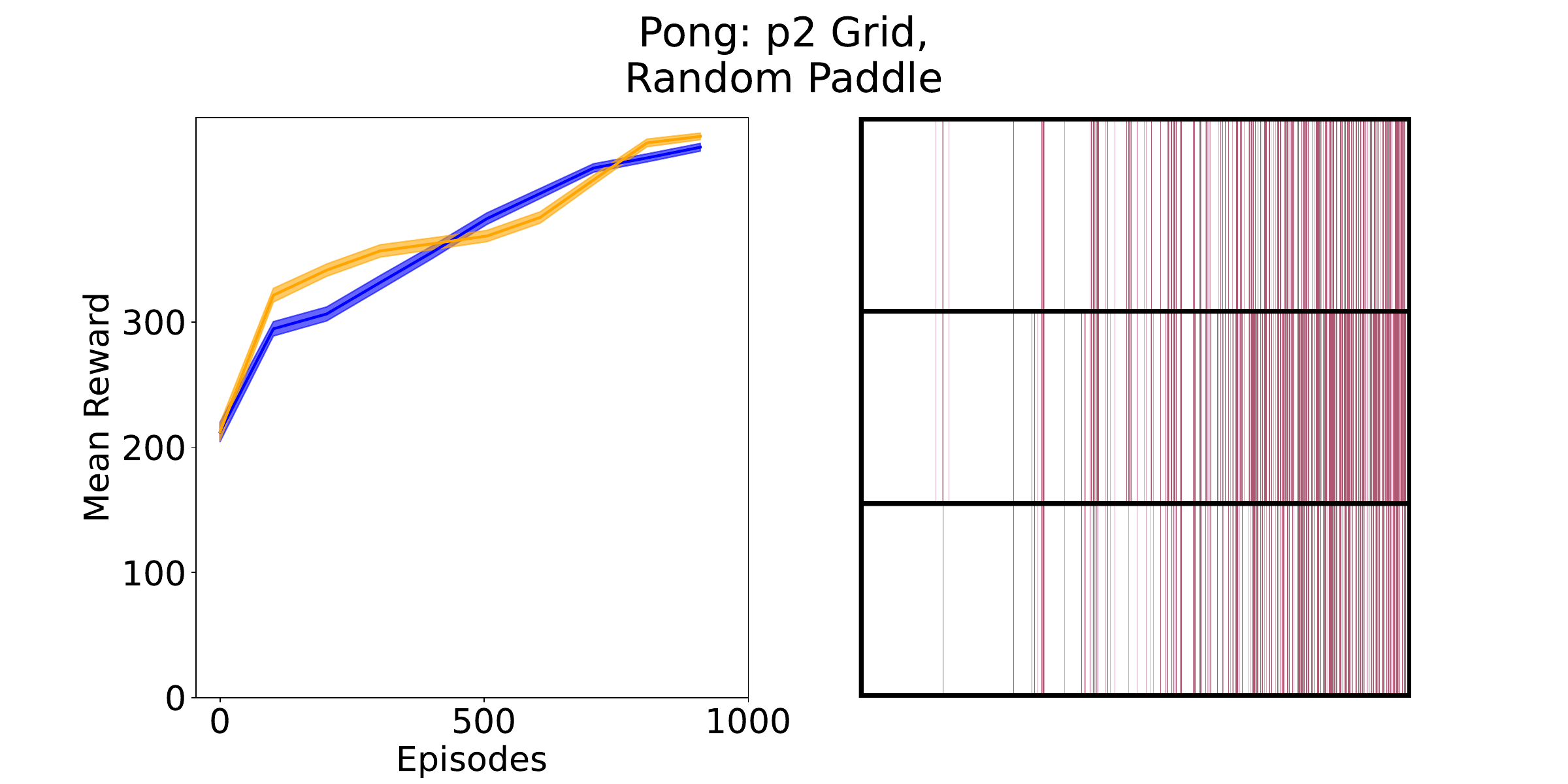}
  \end{subfigure}\\

  \caption{\emph{Q-learning Agent with Boltzmann exploration strategy}: The \textit{exploration grid} visualizing the difference in State-Action (S-A) pairs explored by these agents ($D_{LG}$). Results for Pong p1, p2 grids, the agent is trained on Directional paddle ($p=0.3$ top, $p=0.6$, bottom) variation and tested in the Random paddle environment. Rows in the right figure represents agent's actions Left, Right, Stop.}
  \label{fig:atari_variations-exploration-semantic-pong-qlearning-boltzmann}
\end{figure*}

\begin{figure*}[t]
  \centering
  \begin{subfigure}{0.32\textwidth}
    \includegraphics[width=\linewidth]{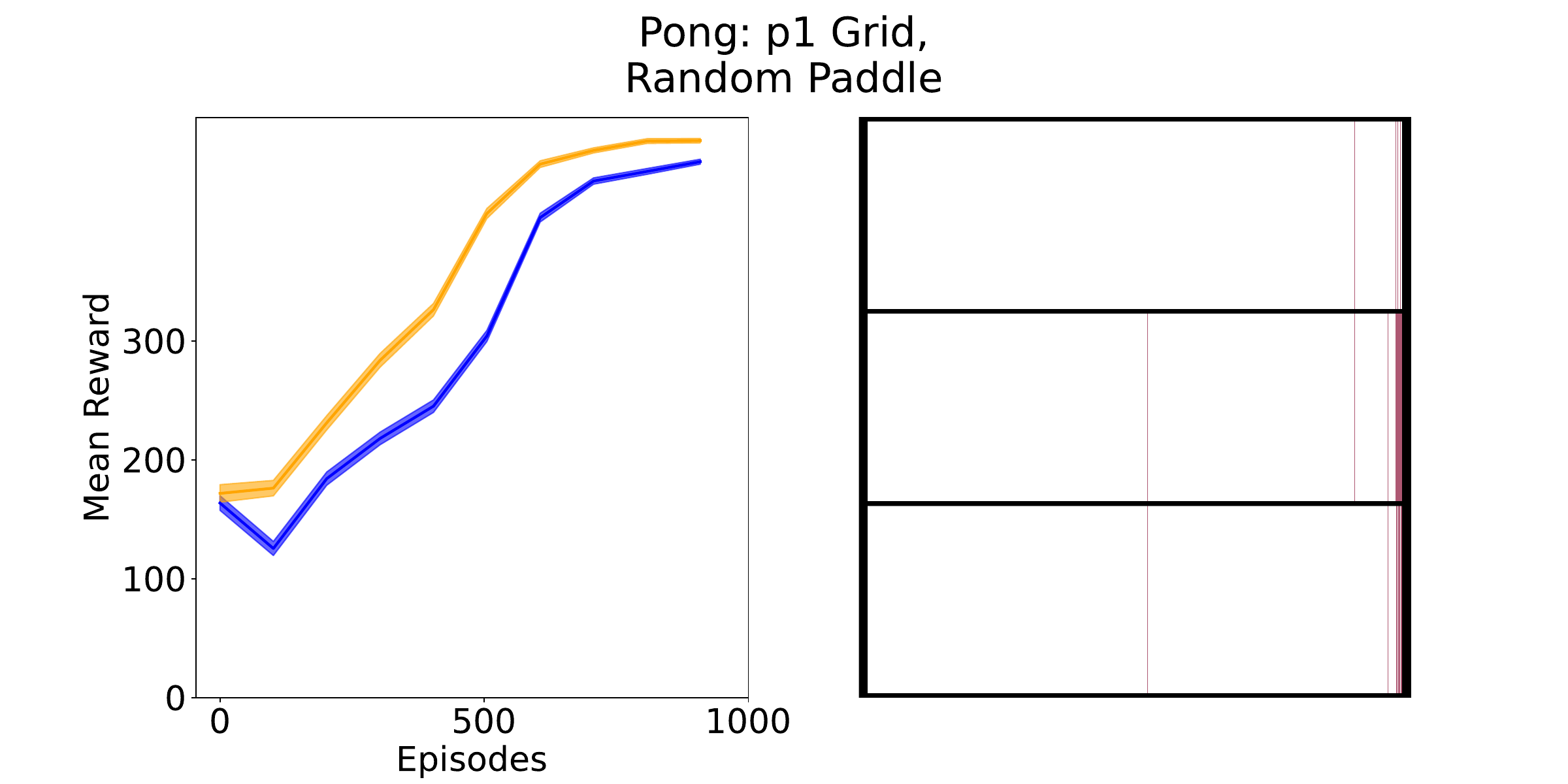}
  \end{subfigure}
  \hspace{0.6mm}
  \begin{subfigure}{0.32\textwidth}
    \includegraphics[width=\linewidth]{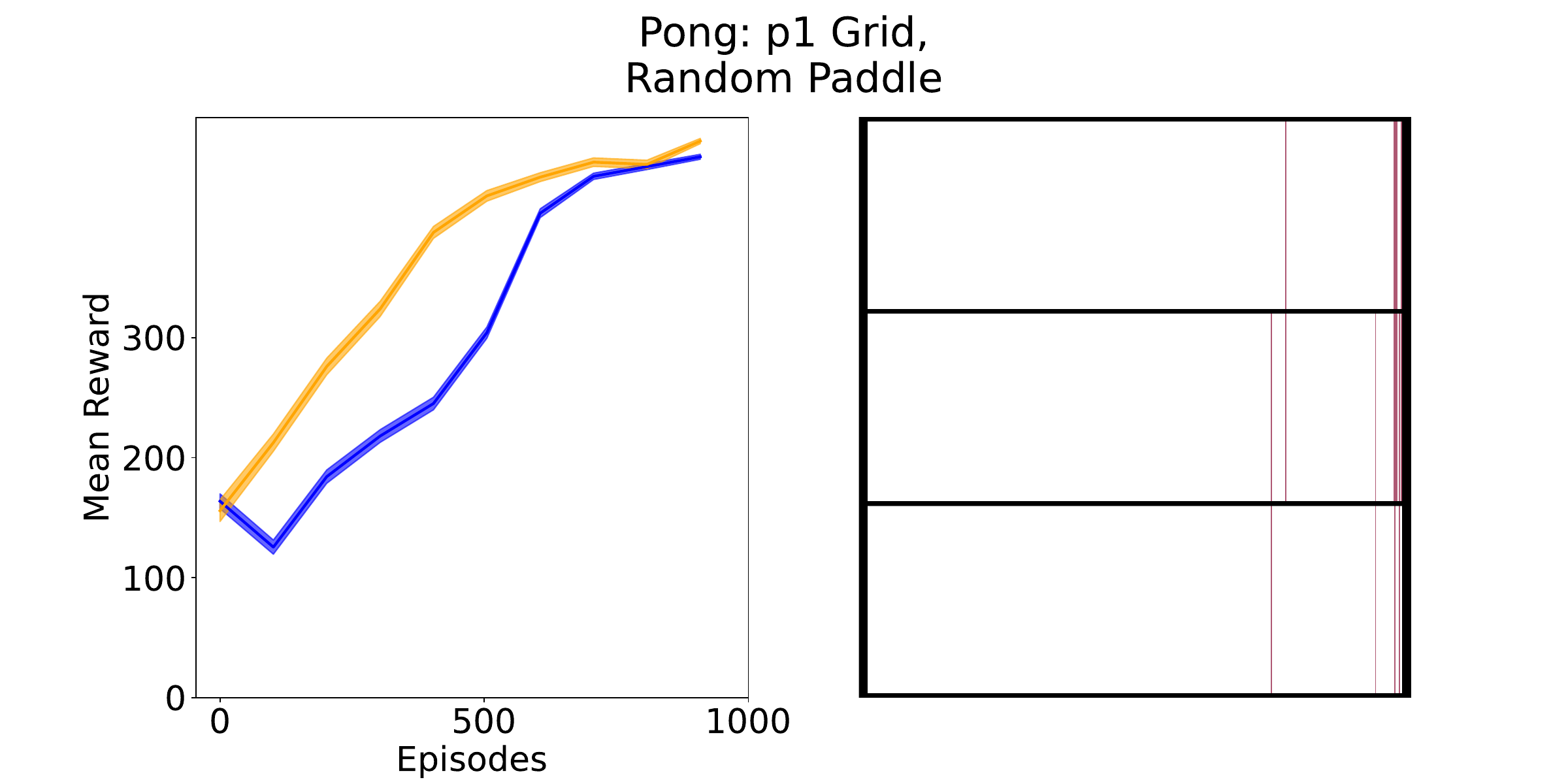}
  \end{subfigure}\\
  
  \begin{subfigure}{0.32\textwidth}
    \includegraphics[width=\linewidth]{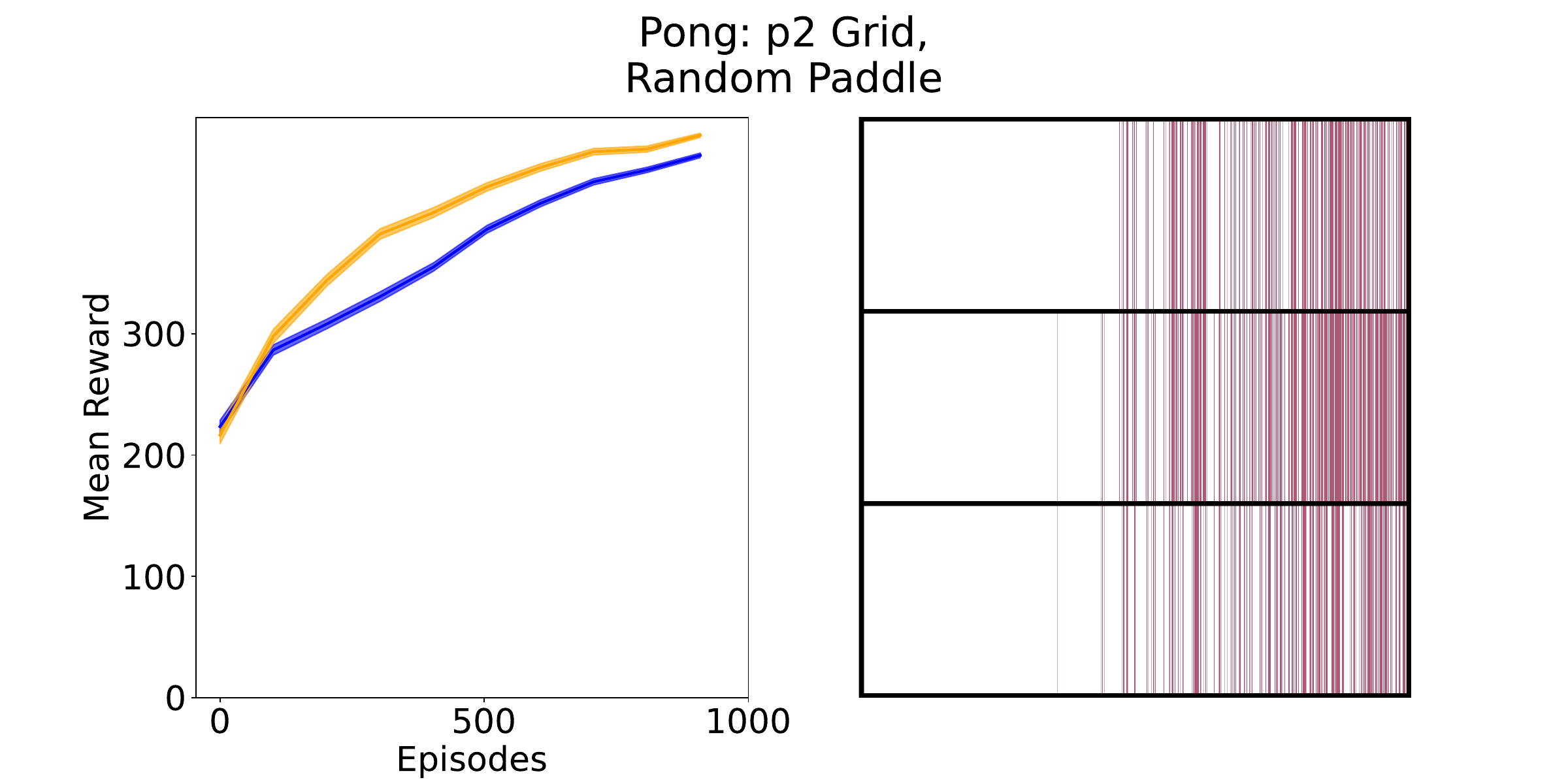}
  \end{subfigure}
  \hspace{0.6mm}
  \begin{subfigure}{0.32\textwidth}
    \includegraphics[width=\linewidth]{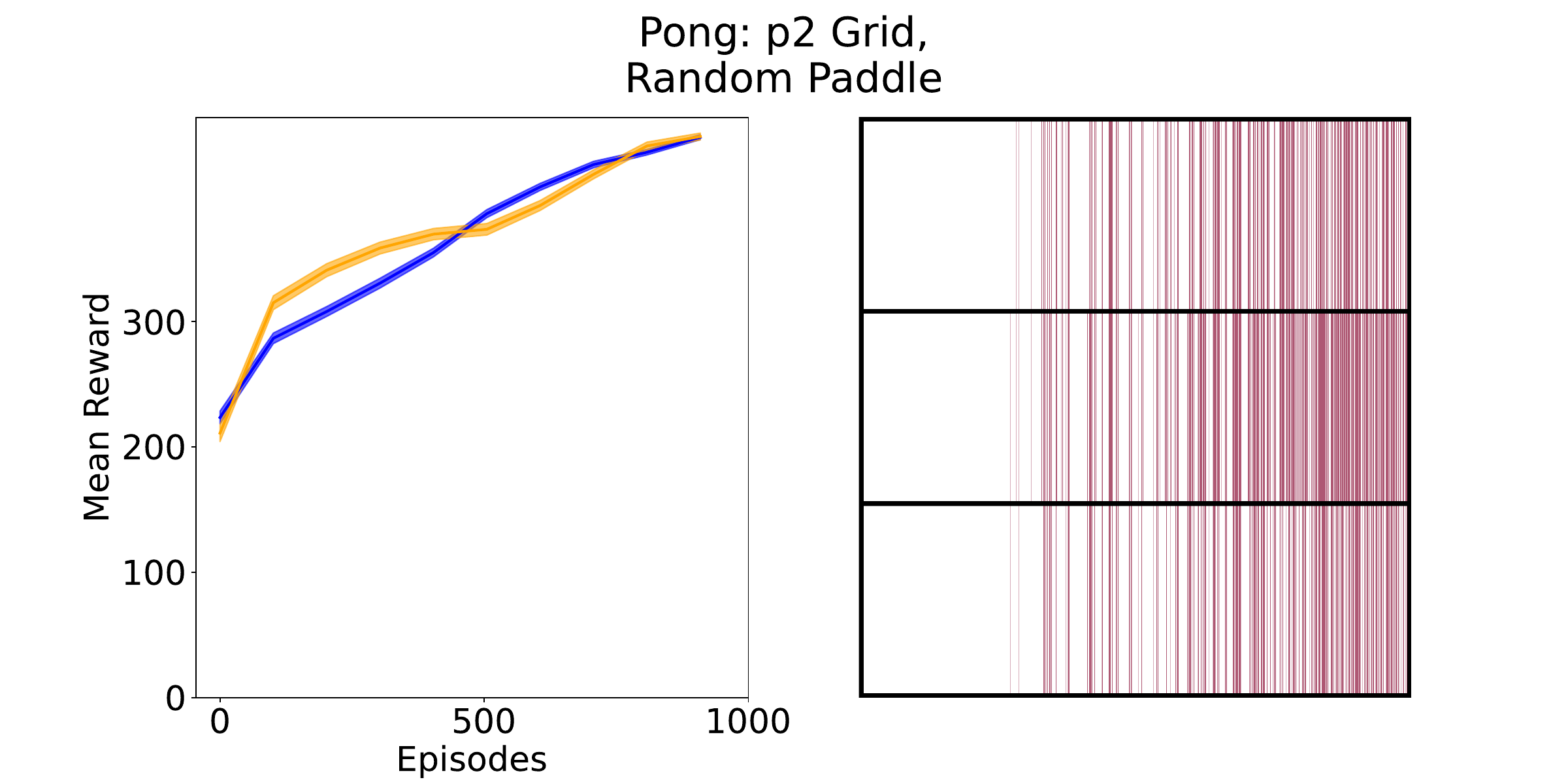}
  \end{subfigure}\\

  \caption{\emph{Q-learning Agent with $\epsilon\text{-}$greedy exploration strategy}: The \textit{exploration grid} visualizing the difference in State-Action (S-A) pairs explored by these agents ($D_{LG}$). Results for Pong p1, p2 grids, the agent is trained on Directional Paddle ($p=0.3$ top, $p=0.6$, bottom) variation and tested in the Random Paddle environment. Rows in the right figure represents agent's actions Left, Right, Stop.}
  \label{fig:atari_variations-exploration-semantic-pong-qlearning-egreedy}
\end{figure*}

\begin{figure*}[t]
  \centering
  \begin{subfigure}{0.32\textwidth}
    \includegraphics[width=\linewidth]{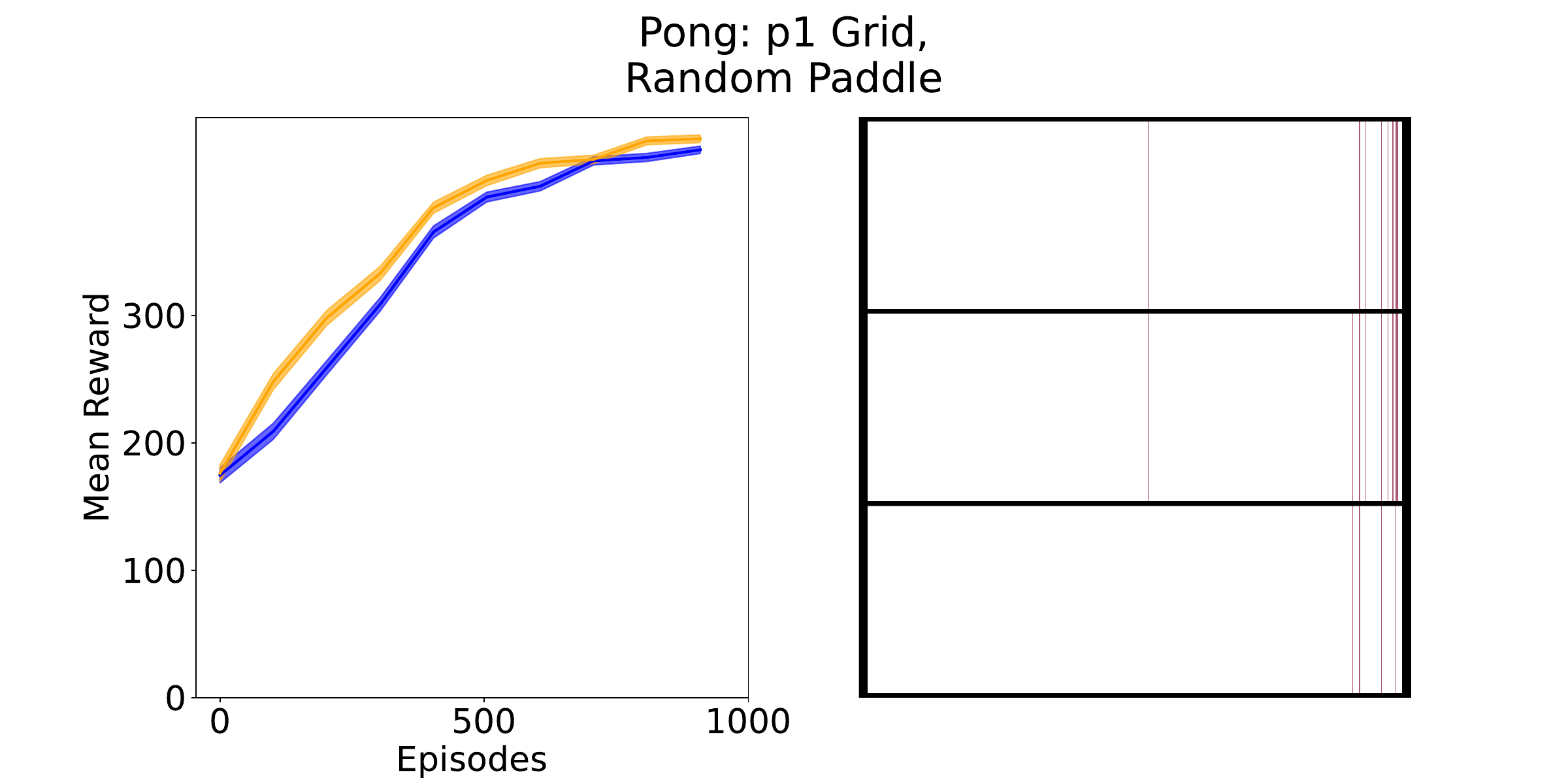}
  \end{subfigure}
  \hspace{0.6mm}
  \begin{subfigure}{0.32\textwidth}
    \includegraphics[width=\linewidth]{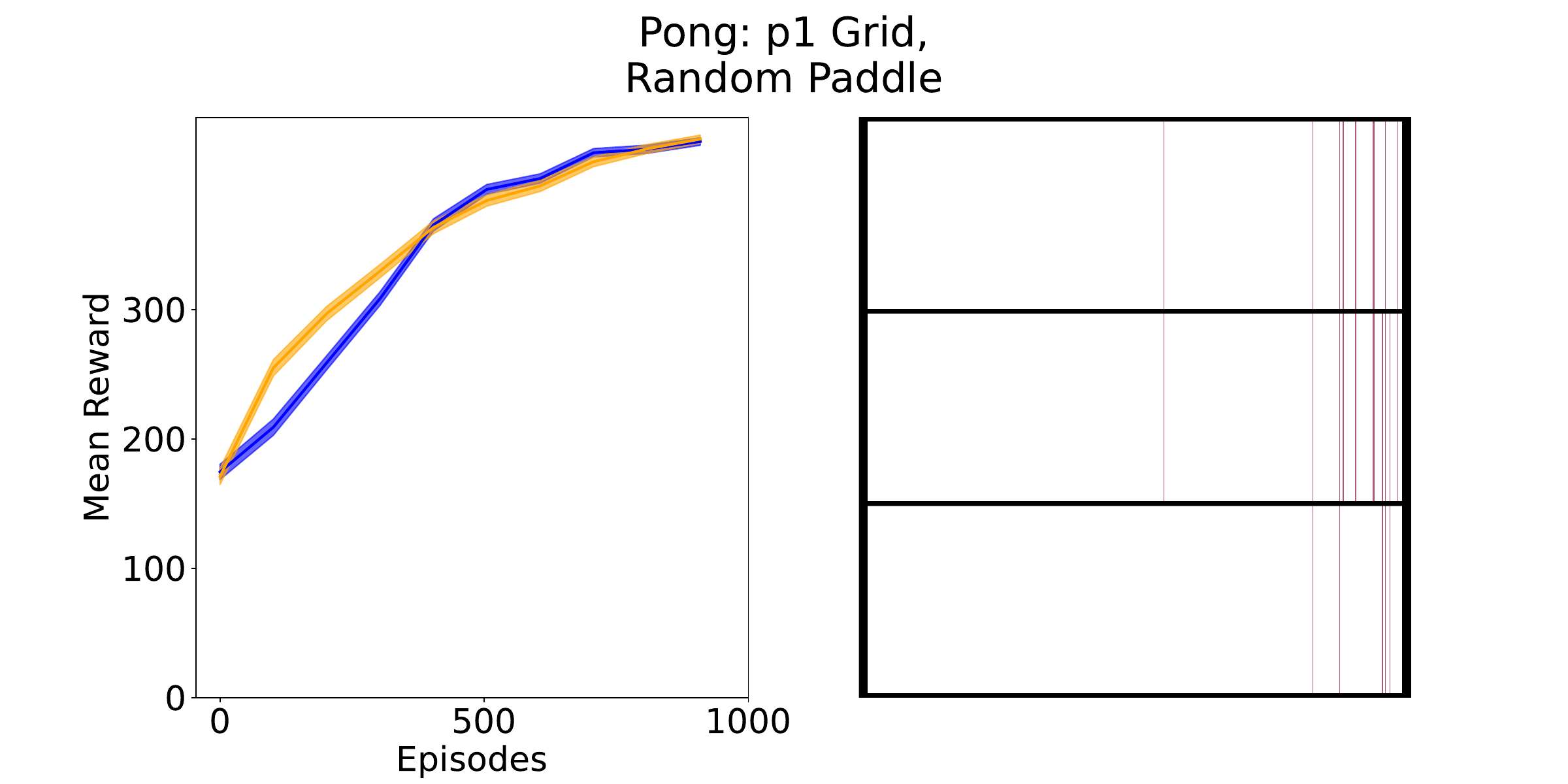}
  \end{subfigure}\\
  
  \begin{subfigure}{0.32\textwidth}
    \includegraphics[width=\linewidth]{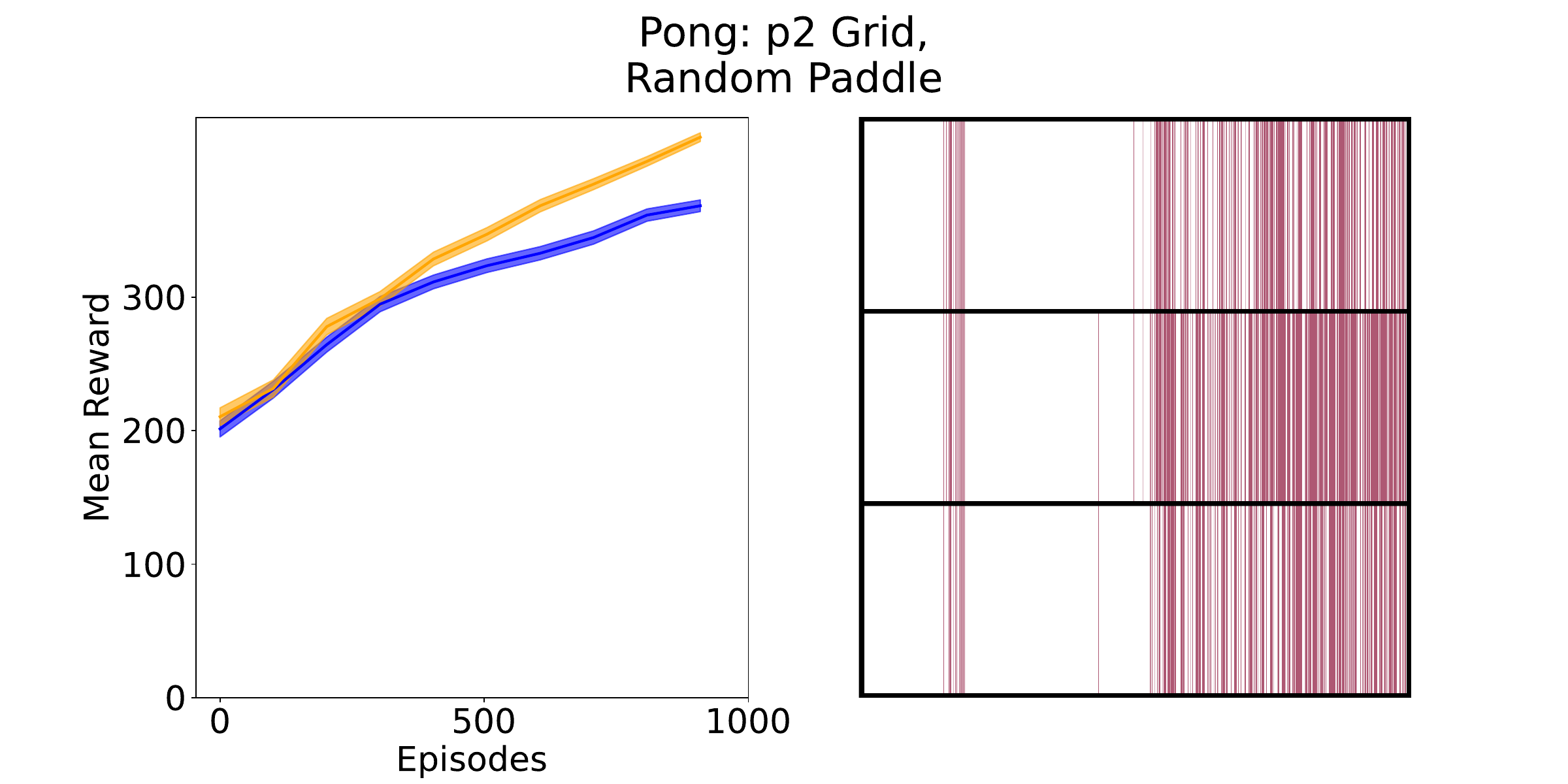}
  \end{subfigure}
  \hspace{0.6mm}
  \begin{subfigure}{0.32\textwidth}
    \includegraphics[width=\linewidth]{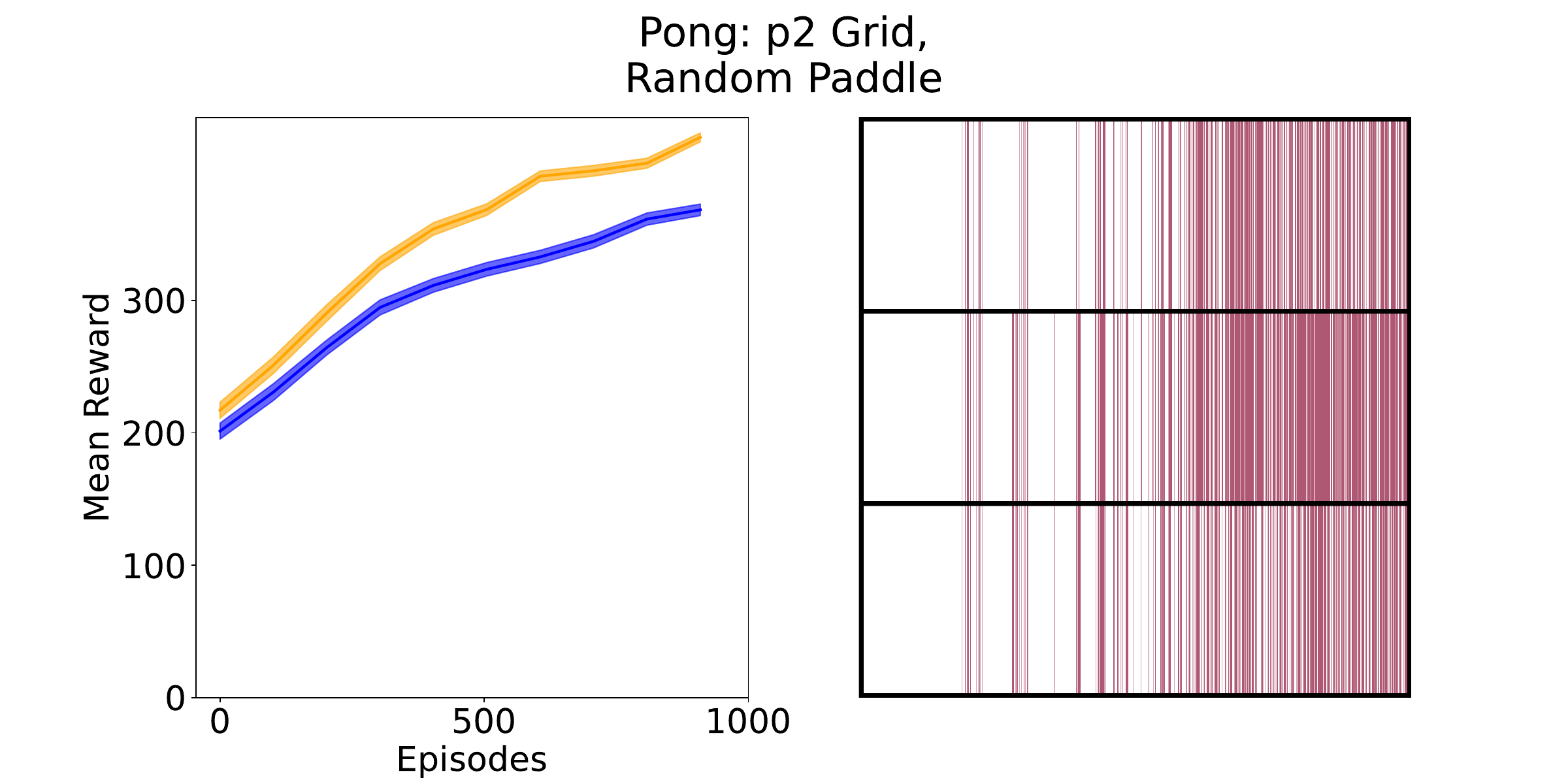}
  \end{subfigure}\\

  \caption{\emph{SARSA Agent with Boltzmann exploration strategy}: The \textit{exploration grid} visualizing the difference in State-Action (S-A) pairs explored by these agents ($D_{LG}$). Results for Pong p1, p2 grids, the agent is trained on Directional Paddle ($p=0.3$ top, $p=0.6$, bottom) variation and tested in the Random Paddle environment. Rows in the right figure represents agent's actions Left, Right, Stop.}
  \label{fig:atari_variations-exploration-semantic-pong-sarsa-boltzmann}
\end{figure*}

\begin{figure*}[t]
  \centering
  \begin{subfigure}{0.32\textwidth}
    \includegraphics[width=\linewidth]{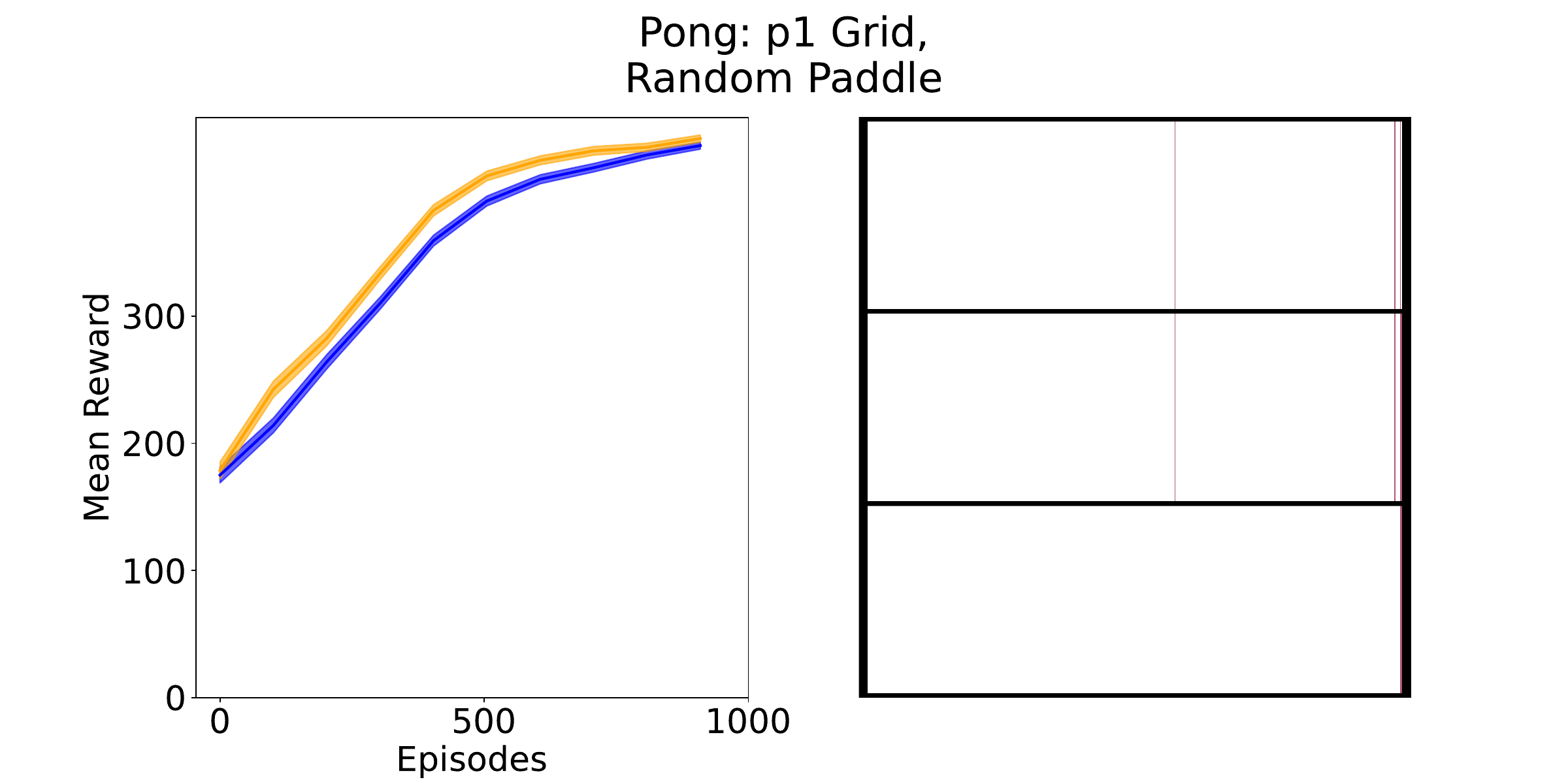}
  \end{subfigure}
  \hspace{0.6mm}
  \begin{subfigure}{0.32\textwidth}
    \includegraphics[width=\linewidth]{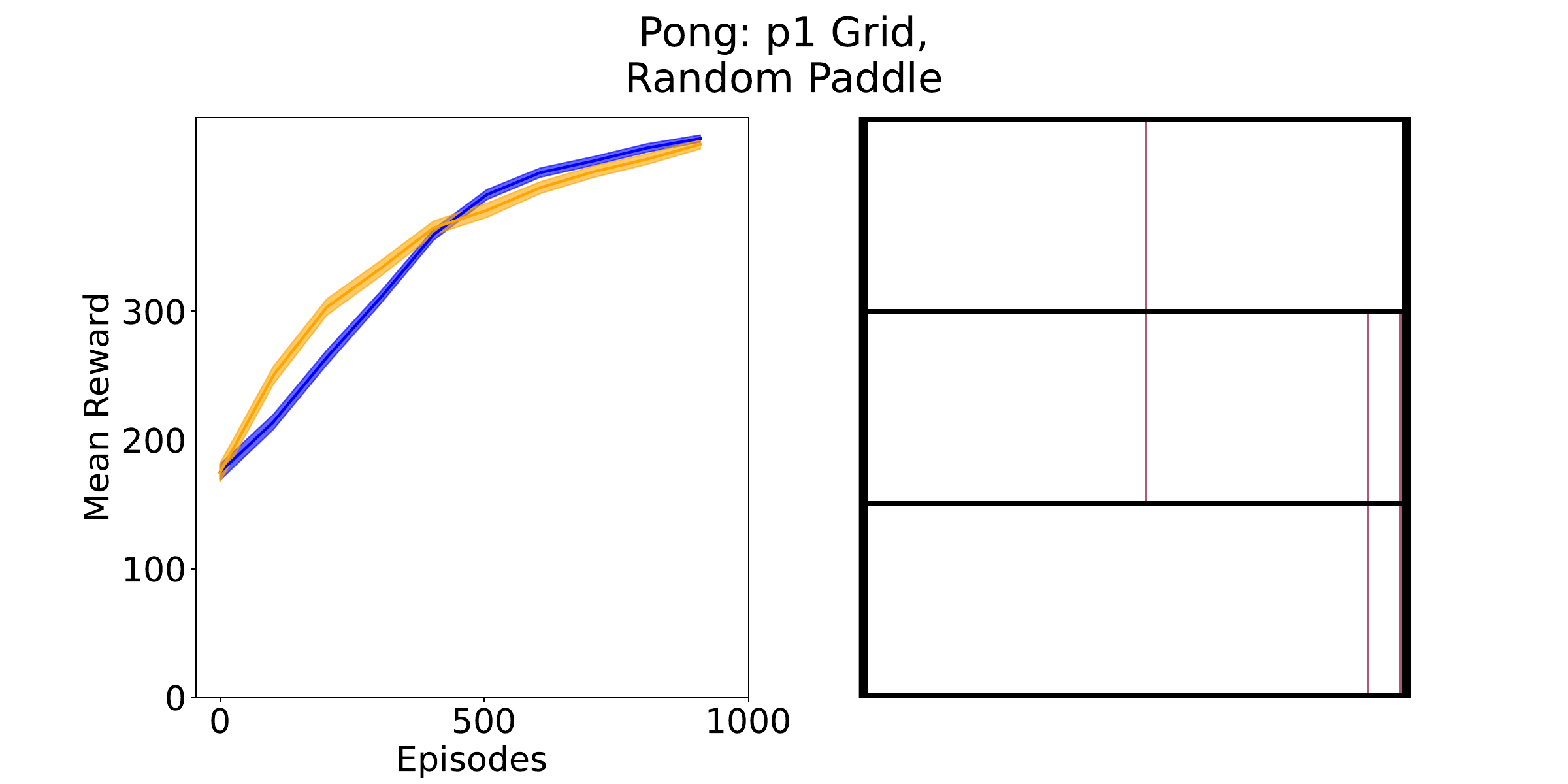}
  \end{subfigure}\\
  
  \begin{subfigure}{0.32\textwidth}
    \includegraphics[width=\linewidth]{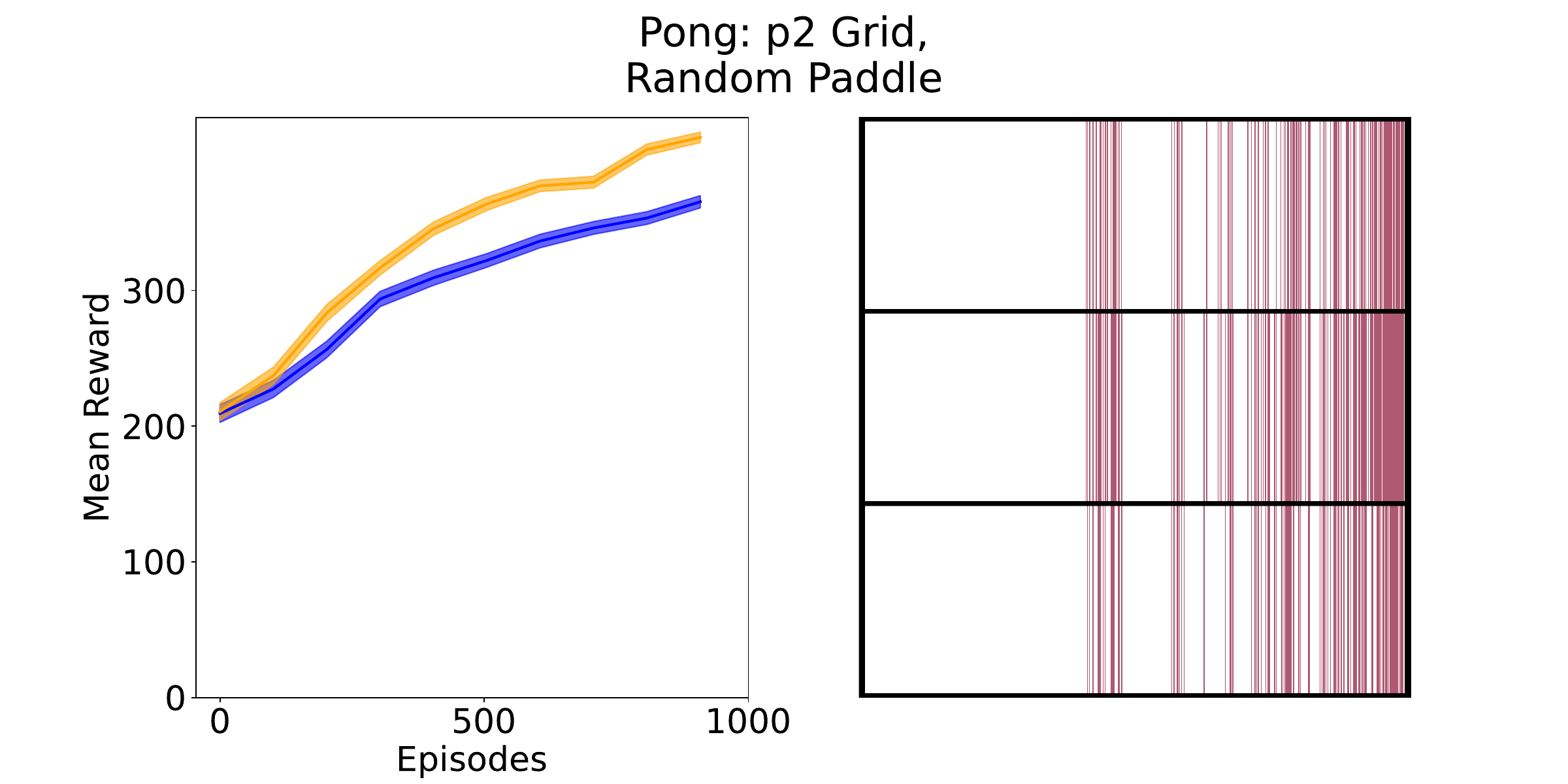}
  \end{subfigure}
  \hspace{0.6mm}
  \begin{subfigure}{0.32\textwidth}
    \includegraphics[width=\linewidth]{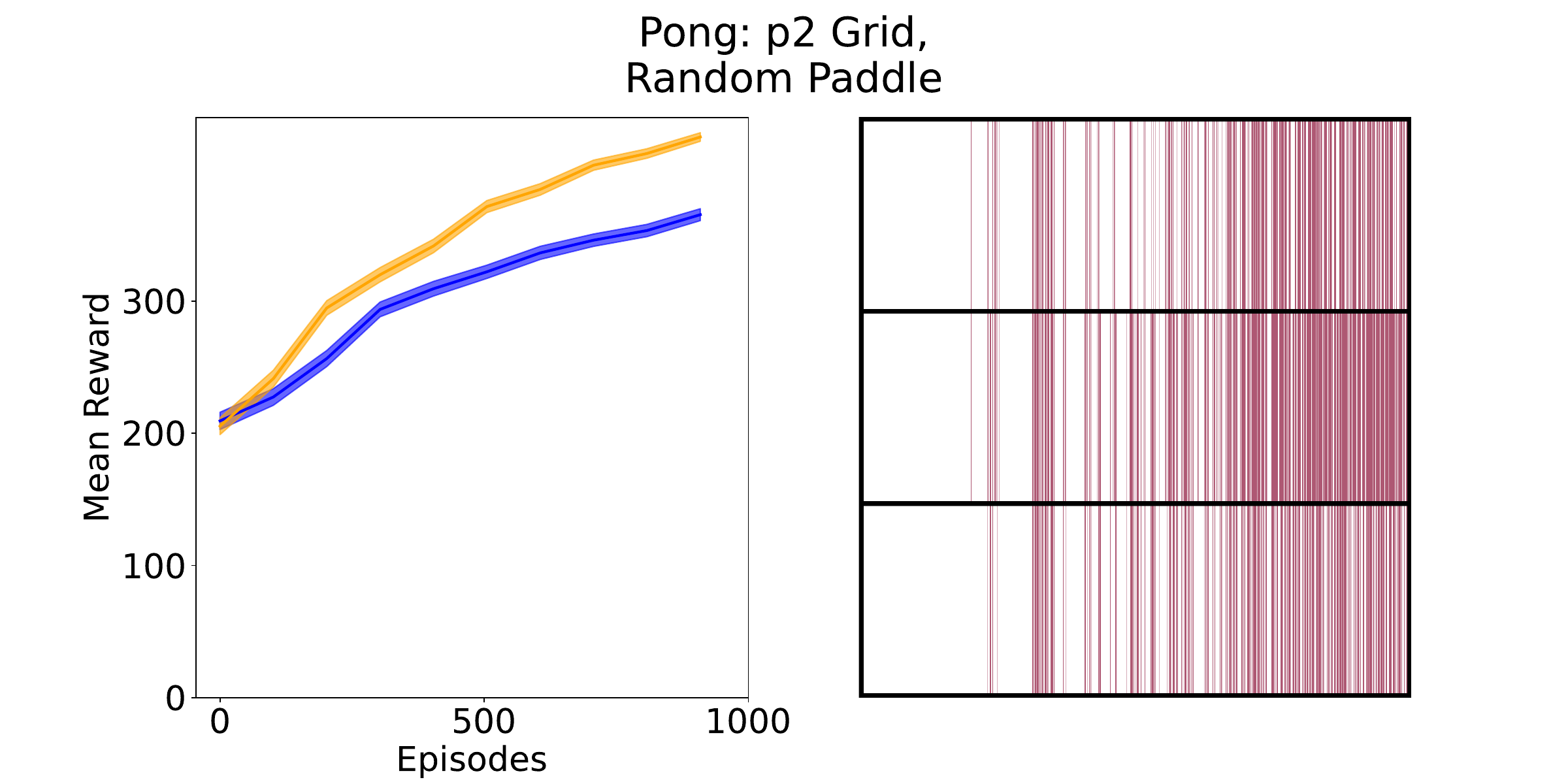}
  \end{subfigure}\\

  \caption{\emph{SARSA Agent with $\epsilon\text{-}$greedy exploration strategy}: The \textit{exploration grid} visualizing the difference in State-Action (S-A) pairs explored by these agents ($D_{LG}$). Results for Pong p1, p2 grids, the agent is trained on Directional Paddle ($p=0.3$ top, $p=0.6$, bottom) variation and tested in the Random Paddle environment. Rows in the right figure represents agent's actions Left, Right, Stop.}
  \label{fig:atari_variations-exploration-semantic-pong-sarsa-egreedy}
\end{figure*}

Similarly, for pong \emph{p1} and \emph{p2}, \ref{fig:atari_variations-exploration-pong-qlearning-boltzmann}, \ref{fig:atari_variations-exploration-pong-qlearning-egreedy}, \ref{fig:atari_variations-exploration-pong-sarsa-boltzmann}, and \ref{fig:atari_variations-exploration-pong-sarsa-egreedy} report grid exploration graphs for non-semantic variations of Q-learning Agent and SARSA Agent respectively, while \ref{fig:atari_variations-exploration-semantic-pong-qlearning-boltzmann}, \ref{fig:atari_variations-exploration-semantic-pong-qlearning-egreedy}, \ref{fig:atari_variations-exploration-semantic-pong-sarsa-boltzmann}, and \ref{fig:atari_variations-exploration-semantic-pong-sarsa-egreedy} for semantic variations.
\subsection{Breakout}
\begin{figure*}[t]
  %\centering
  \begin{subfigure}{0.32\textwidth}
    \includegraphics[width=\linewidth]{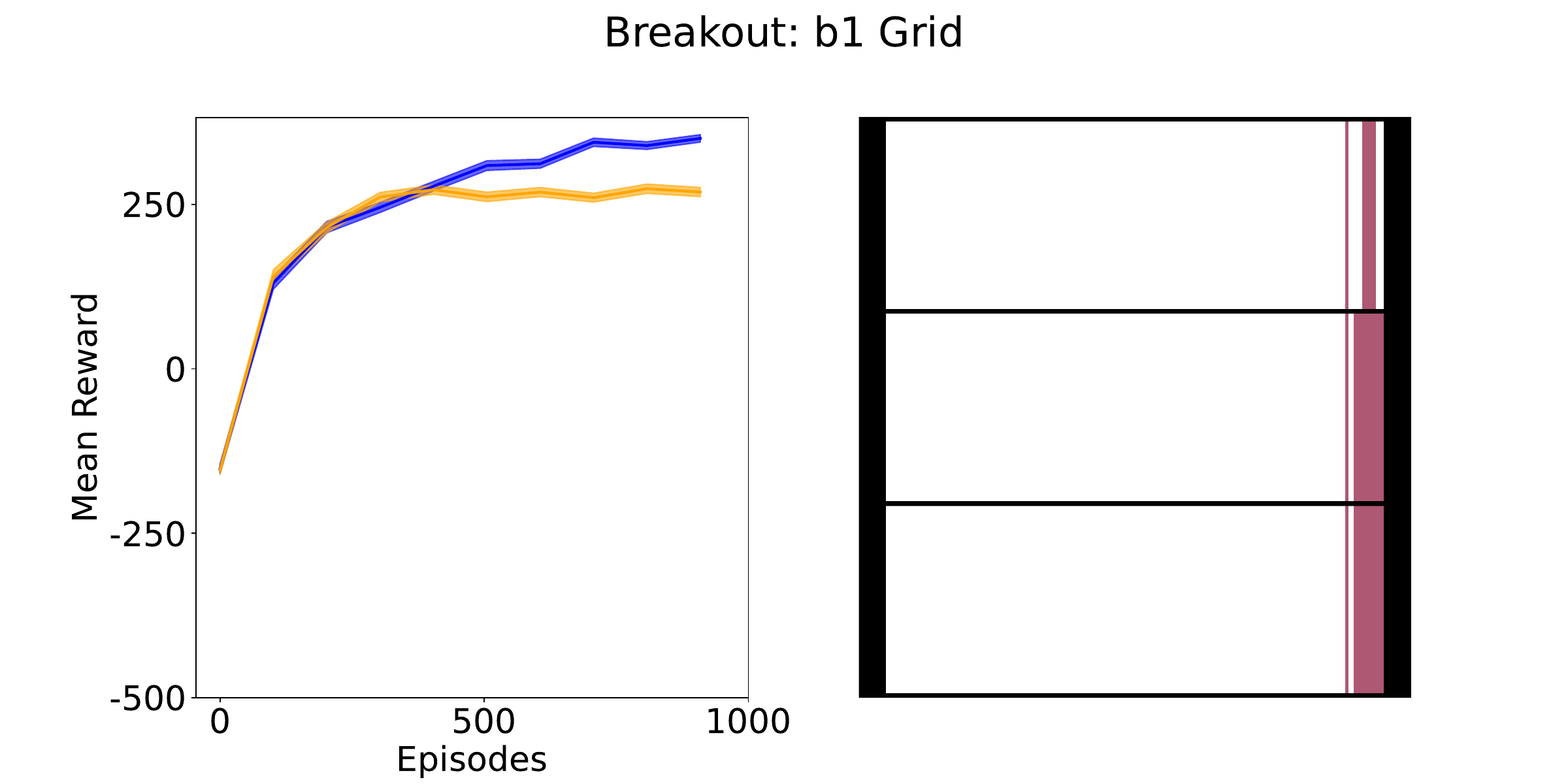}
  \end{subfigure}
  \hfill
  \begin{subfigure}{0.32\textwidth}
    \includegraphics[width=\linewidth]{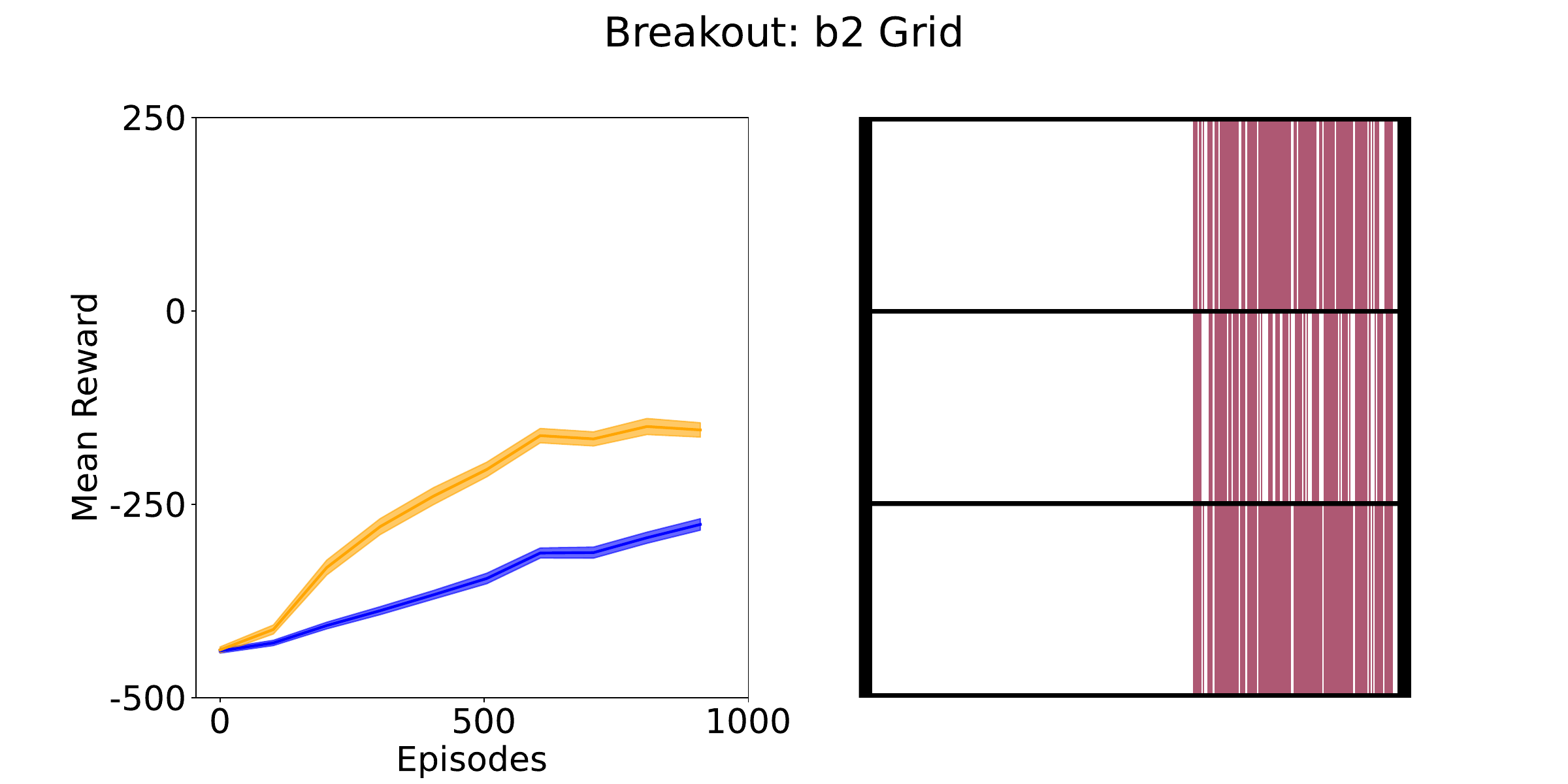}
  \end{subfigure}
  \hfill
  \begin{subfigure}{0.32\textwidth}
    \includegraphics[width=\linewidth]{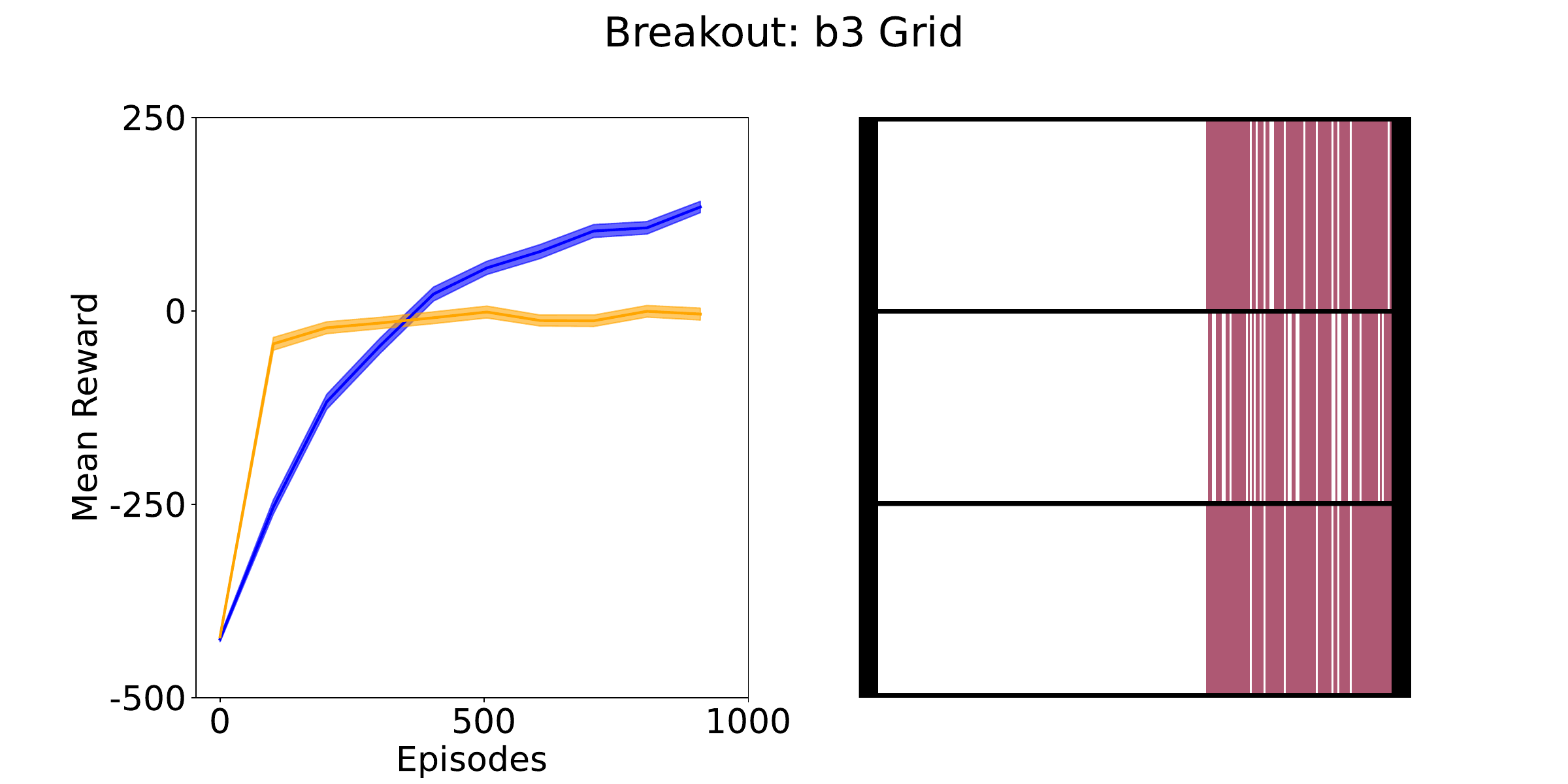}
  \end{subfigure}

  \caption{\emph{Q-learning Agent with $\epsilon\text{-}$greedy exploration strategy}: The \textit{exploration grid} visualizing the difference in State-Action (S-A) pairs explored by these agents ($D_{LG}$). Results for Breakout b1, b2, b3 grids, the agent is trained on non-noisy variations of different environments (reported in the headings) and tested in the Low-Noise regime. Rows in the right figure represents agent's actions Left, Right, Stop.}
  \label{fig:atari_variations-exploration-breakout-qlearning-egreedy}
\end{figure*}

\begin{figure*}[t]
  %\centering
  \begin{subfigure}{0.32\textwidth}
    \includegraphics[width=\linewidth]{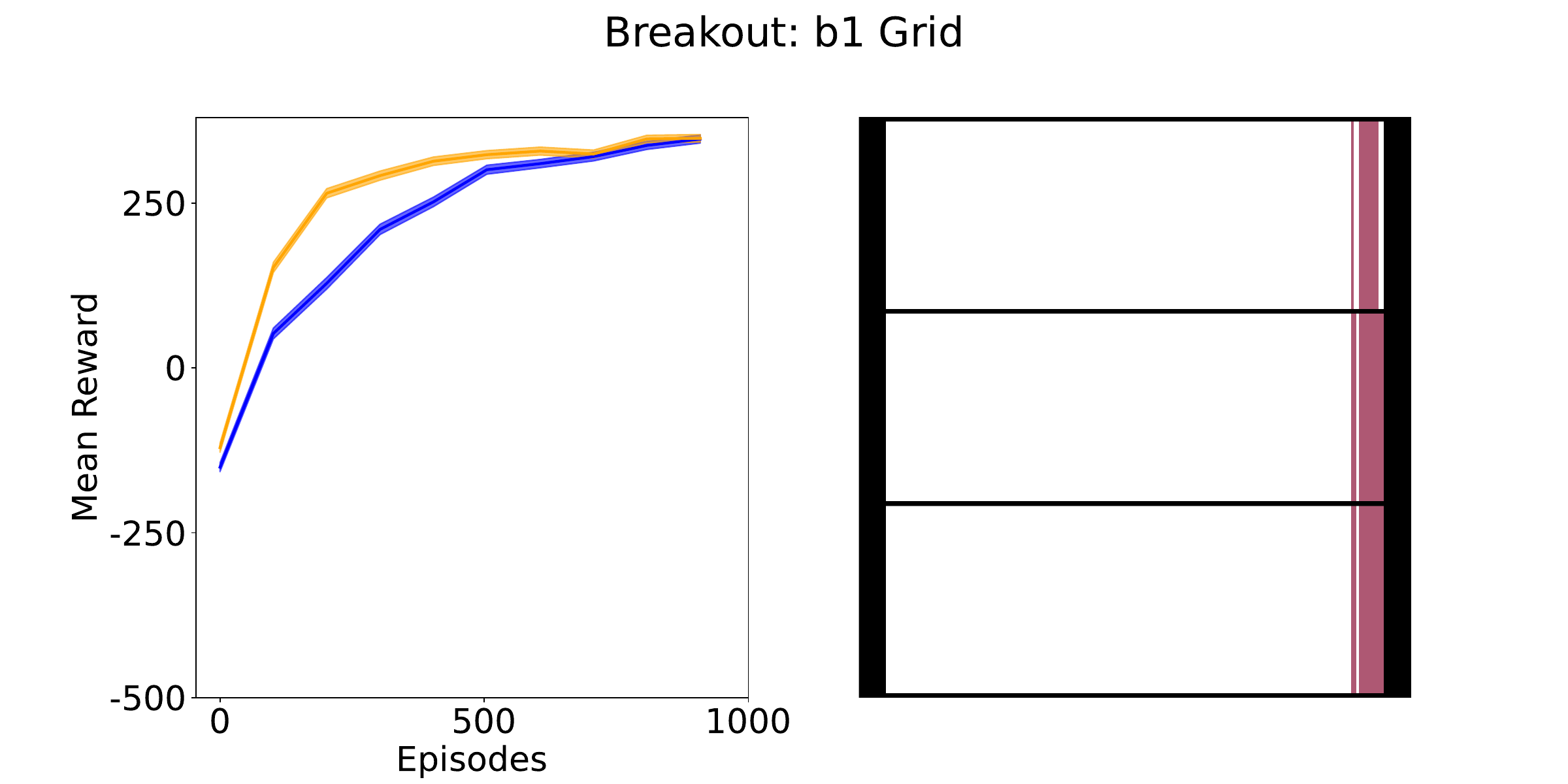}
  \end{subfigure}
  \hfill
  \begin{subfigure}{0.32\textwidth}
    \includegraphics[width=\linewidth]{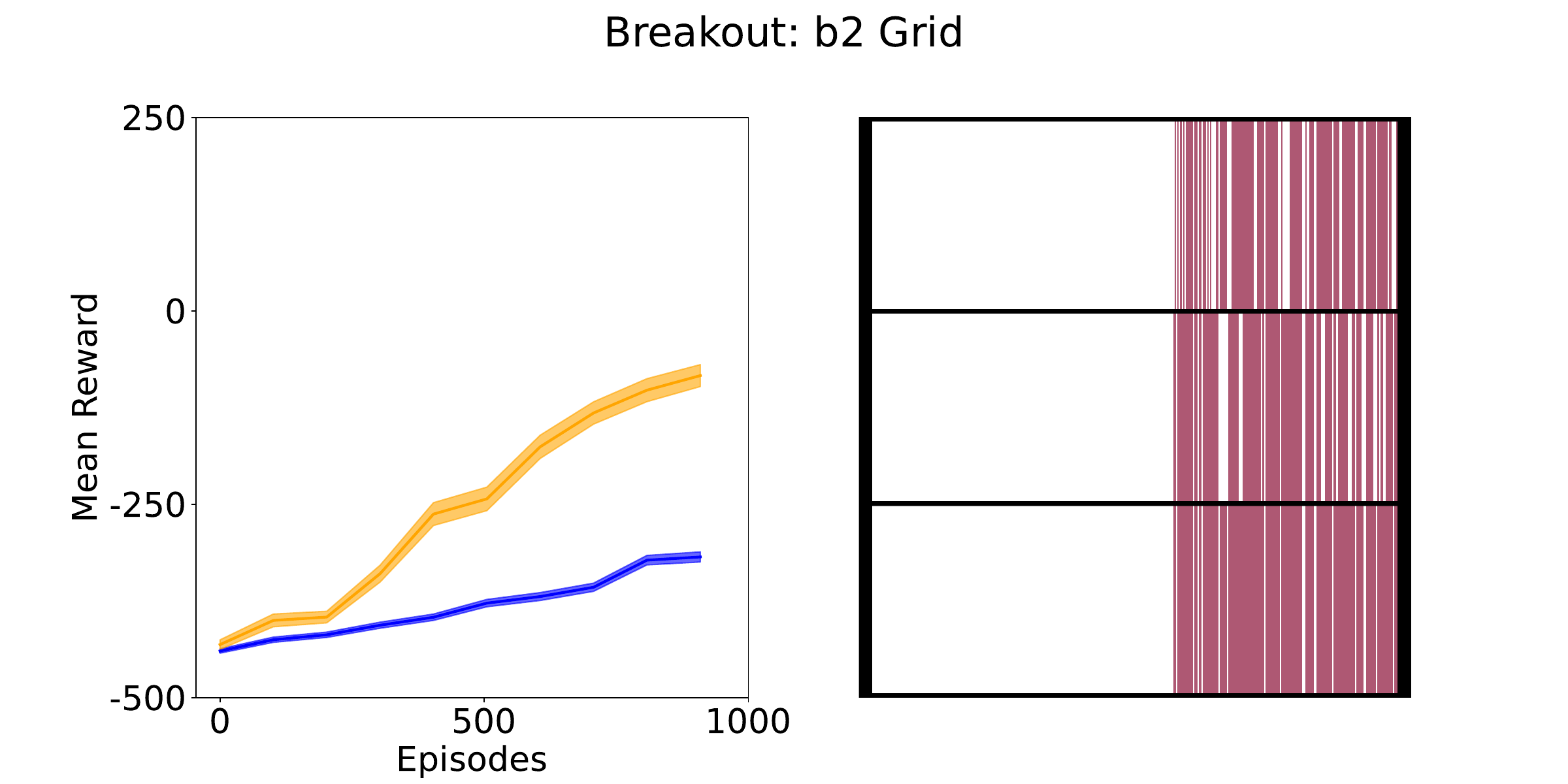}
  \end{subfigure}
  \hfill
  \begin{subfigure}{0.32\textwidth}
    \includegraphics[width=\linewidth]{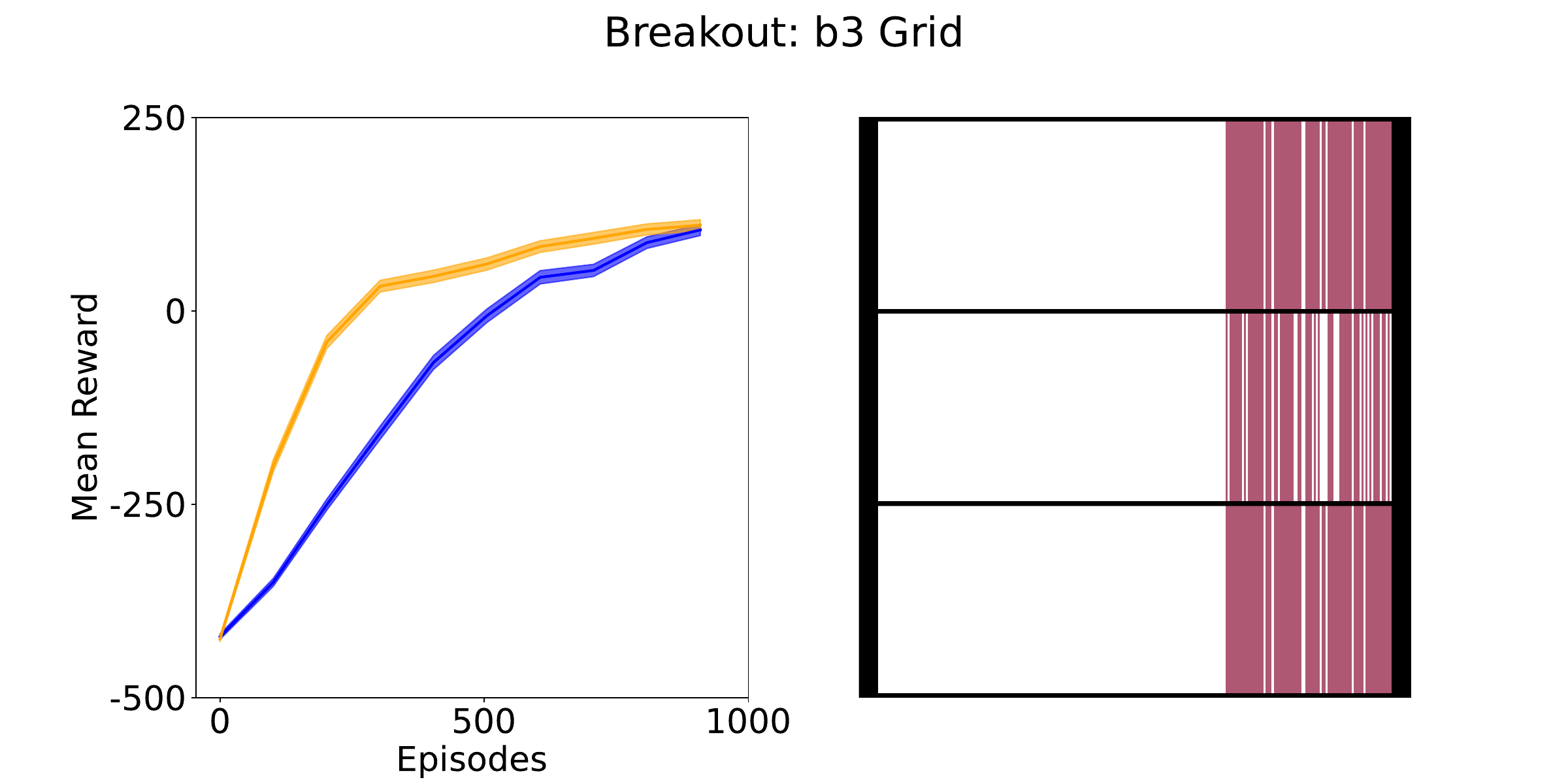}
  \end{subfigure}

  \caption{\emph{SARSA Agent with Boltzmann exploration strategy}:  The \textit{exploration grid} visualizing the difference in State-Action (S-A) pairs explored by these agents ($D_{LG}$). Results for Breakout b1, b2, b3 grids, the agent is trained on non-noisy variations of different environments (reported in the headings) and tested in the Low-Noise regime. Rows in the right figure represents agent's actions Left, Right, Stop.}
  \label{fig:atari_variations-exploration-breakout-sarsa-boltzmann}
\end{figure*}

\begin{figure*}[t]
  %\centering
  \begin{subfigure}{0.32\textwidth}
    \includegraphics[width=\linewidth]{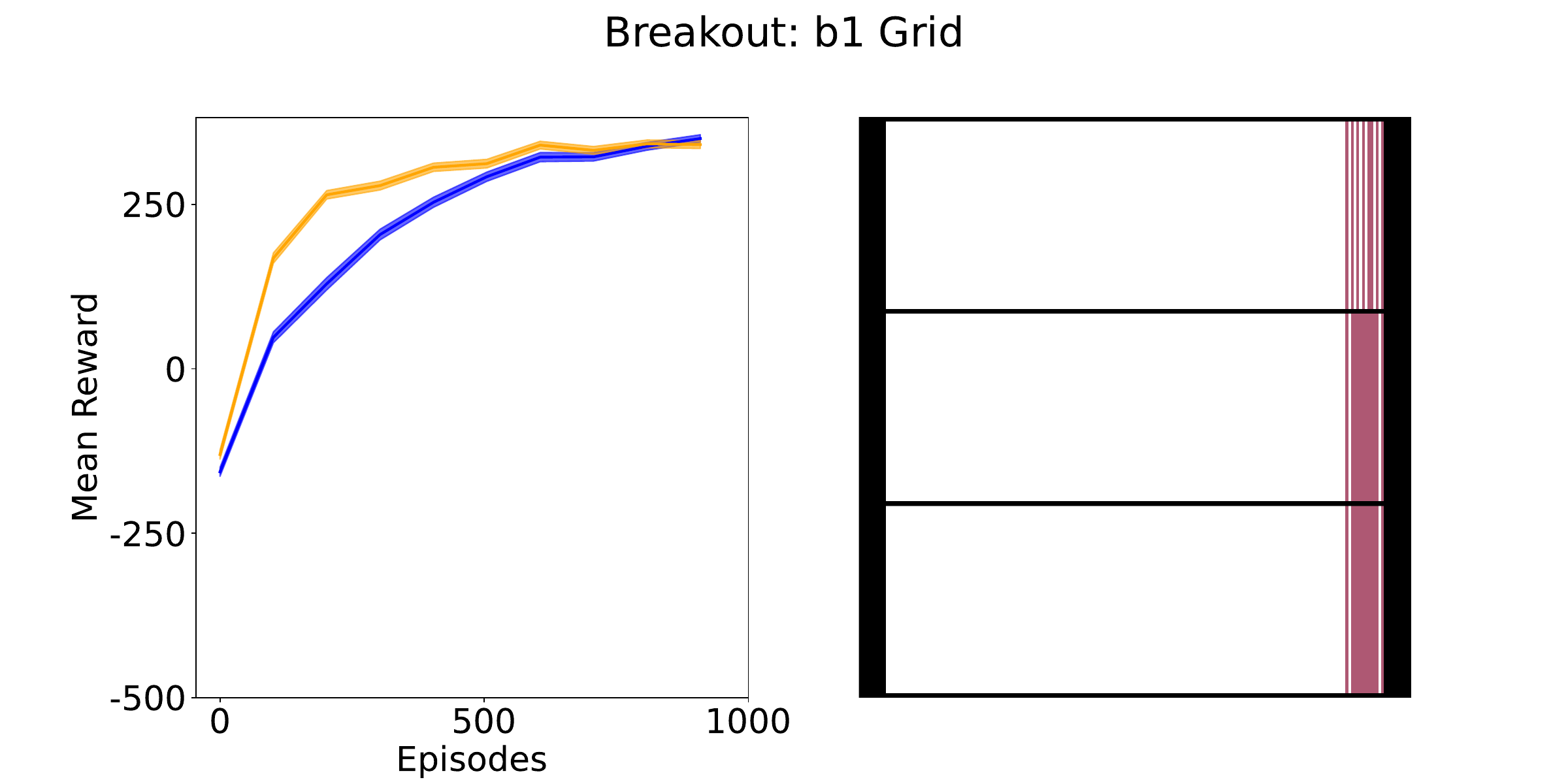}
  \end{subfigure}
  \hfill
  \begin{subfigure}{0.32\textwidth}
    \includegraphics[width=\linewidth]{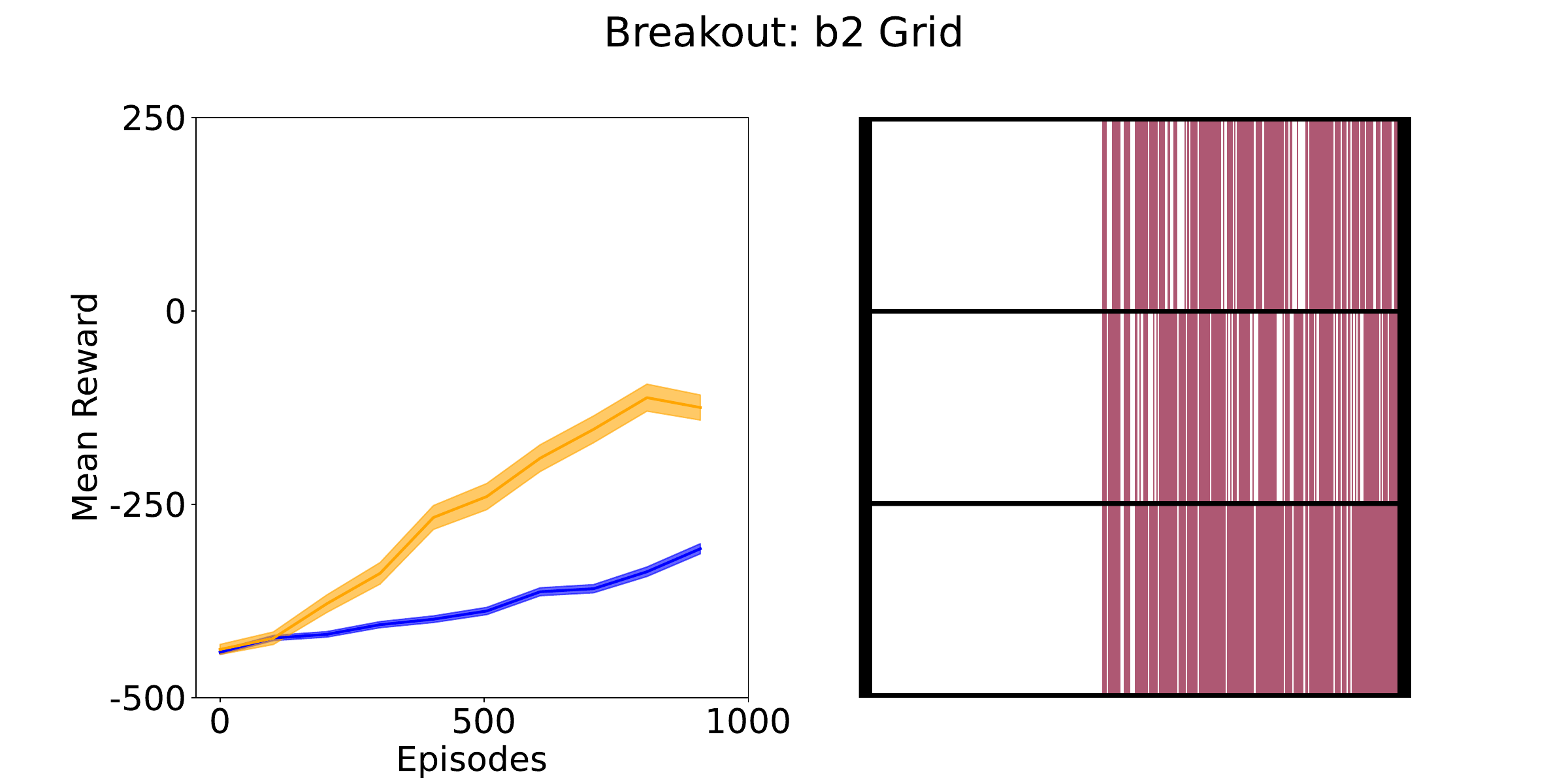}
  \end{subfigure}
  \hfill
  \begin{subfigure}{0.32\textwidth}
    \includegraphics[width=\linewidth]{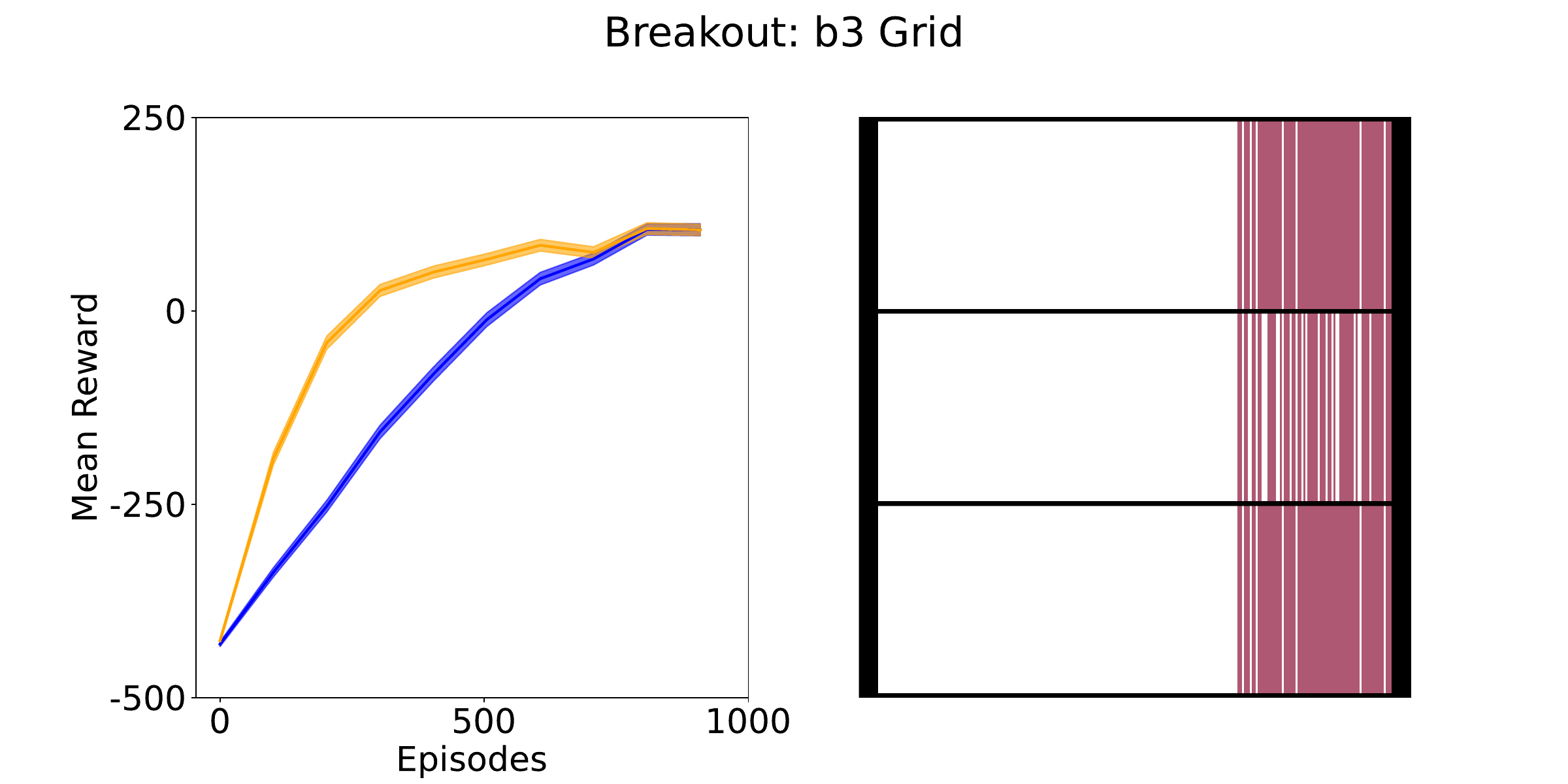}
  \end{subfigure}

  \caption{\emph{SARSA Agent with $\epsilon\text{-}$greedy exploration strategy}: The \textit{exploration grid} visualizing the difference in State-Action (S-A) pairs explored by these agents ($D_{LG}$). Results for Breakout b1, b2, b3 grids, the agent is trained on non-noisy variations of different environments (reported in the headings) and tested in the Low-Noise regime. Rows in the right figure represents agent's actions Left, Right, Stop.}
  \label{fig:atari_variations-exploration-breakout-sarsa-egreedy}
\end{figure*}

For Breakout grids \emph{b1},\emph{b2}, and \emph{b3}, exploration graphs for non-semantic variations of Q-learning Agent and SARSA Agent are reported in Supplementary figures \ref{fig:atari_variations-exploration-breakout-qlearning-boltzmann}, \ref{fig:atari_variations-exploration-breakout-qlearning-egreedy}, \ref{fig:atari_variations-exploration-breakout-sarsa-boltzmann}, and \ref{fig:atari_variations-exploration-breakout-sarsa-egreedy}.

\section{Regret Normalization}\label{supp-sec:normalizations}
\begin{figure*}[t]
\centering
        \includegraphics[width=0.9\textwidth]{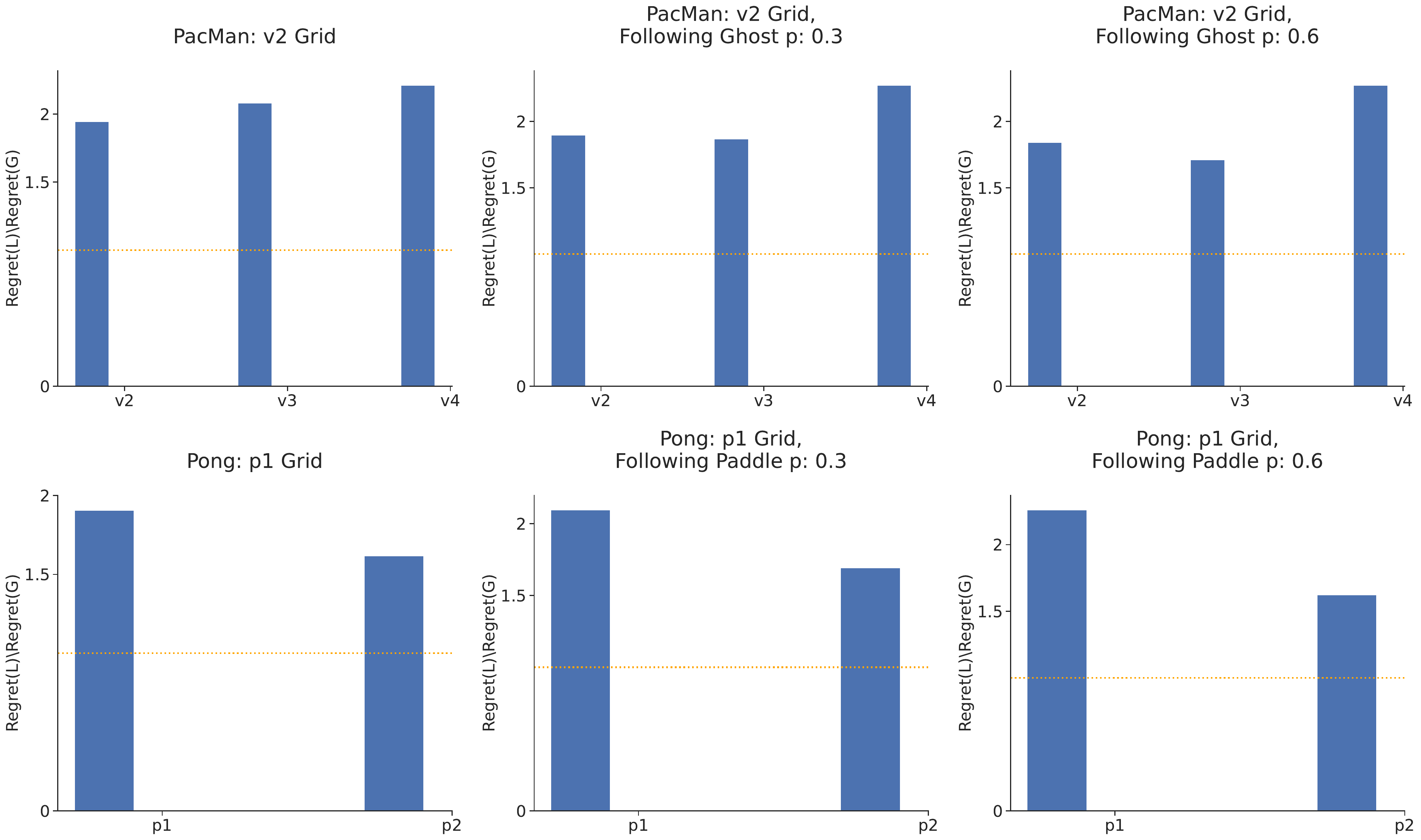} 
        \caption{\emph{Regret for SARSA Agent with $\epsilon\text{-}$greedy exploration strategy across non-semantic game variations}: Results for Pac-Man (v2, v3, v4) and Pong (p1, p2) grids, showing the ratio of regrets between Learnability and Generalization agents (L/G). The agents were trained in the Random Ghost/ Random Bar environment without perturbation and tested in the Low-Noise regime.}
        \label{fig:regrets-pacman-pong-non-semantic-sarsa-egreedy}
\end{figure*}

\begin{figure*}[t]
\centering
        \includegraphics[width=0.9\textwidth]{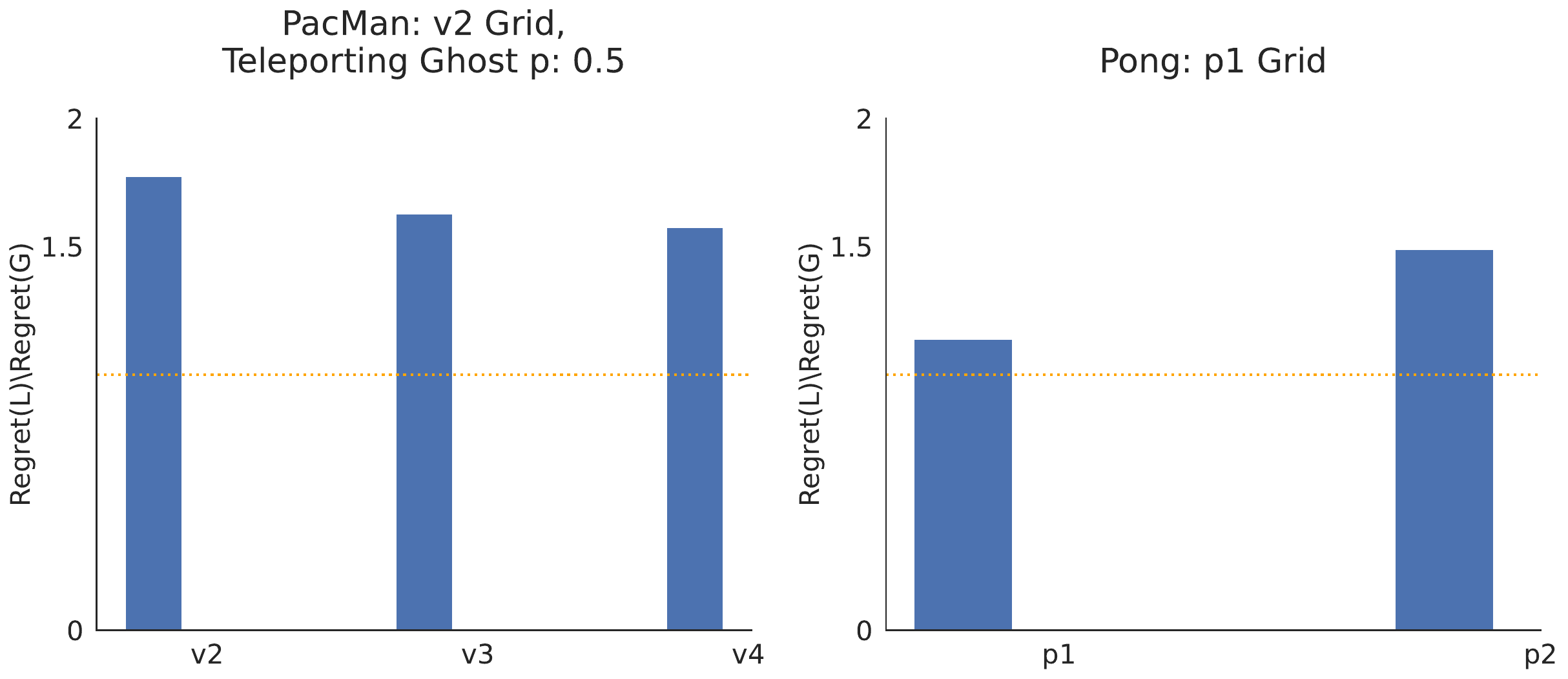} 
        \caption{\emph{Regret for SARSA Agent with $\epsilon\text{-}$greedy exploration strategy across semantic game variations}: Results for Pac-Man (v2, v3, v4) and Pong (p1, p2) grids, showing the ratio of regrets between Learnability and Generalization agents (L/G). The agents were trained in the Random Ghost/ Random Bar environment and tested in the Teleporting Ghost/Directional Paddle ($p=0.3$ top, $p=0.6$, bottom) environments.}
        \label{fig:regrets-pacman-pong-semantic-sarsa-egreedy}
\end{figure*}

\emph{Regret} quantifies the performance gap between an agent's actions and the optimal actions it could have taken. Specifically, it measures the loss incurred due to suboptimal decisions compared to a theoretical best-case scenario. We take the ratio of the learnability and generalization agents' regrets in an effort to provide normalized AUC measures for both non-semantic (\ref{fig:regrets-pacman-pong-non-semantic-sarsa-egreedy}) and semantic (\ref{fig:regrets-pacman-pong-semantic-sarsa-egreedy}) game variations of Pac Man (v1,v2,v3) and Pong (p1,p2).
\section{DQN}\label{supp-sec:DQN}
\begin{figure*}[t]
\centering
        \includegraphics[width=0.9\textwidth]{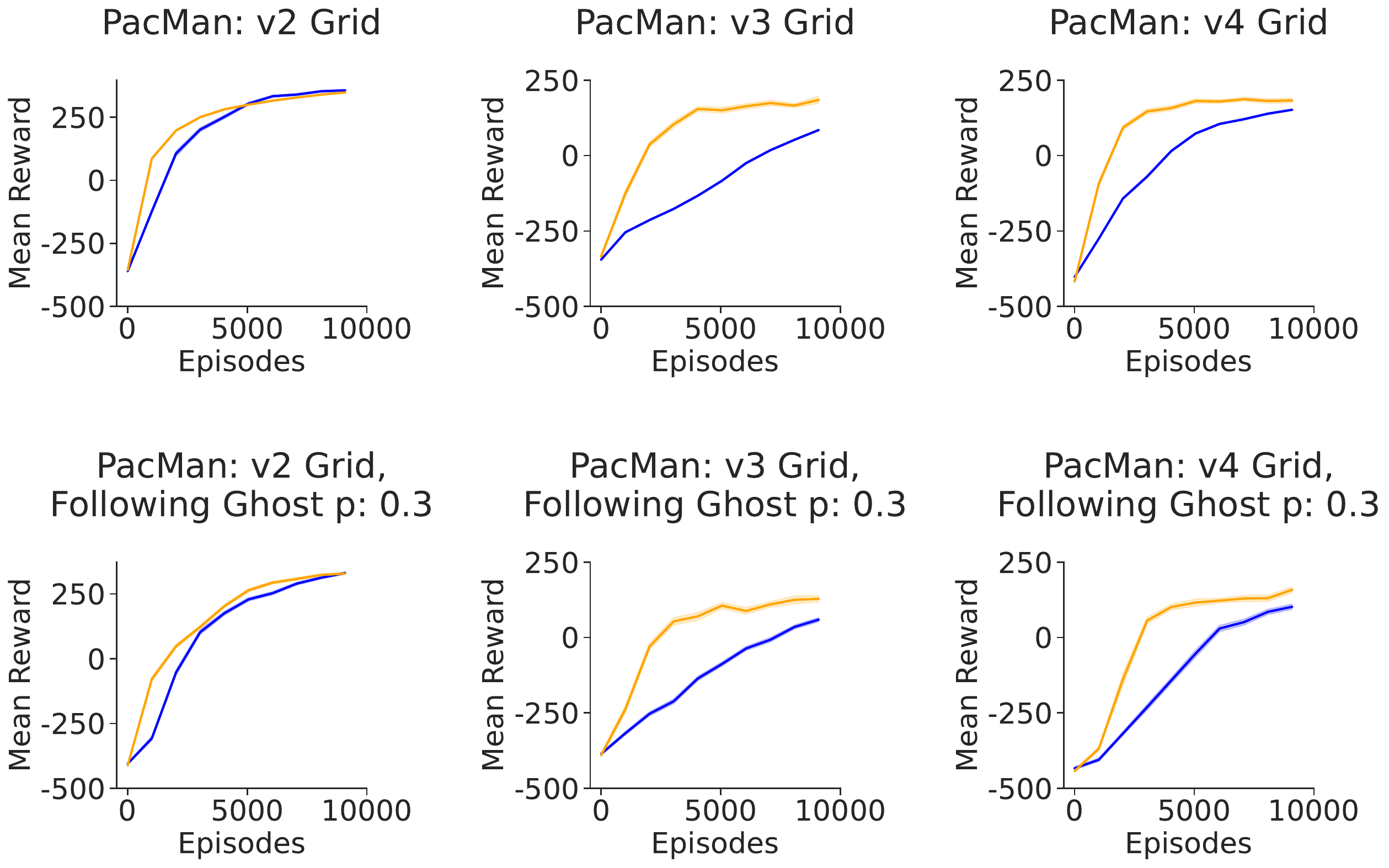} 
        \caption{\emph{DQN Agent with $\epsilon$\text{-}greedy exploration strategy across non-semantic variations}: Results for PacMan v2, v3, v4 grids reporting mean reward as a function of episode number. The agent is trained on the non-noisy version of the environment and tested on different level of noise ($\delta \sim \mathcal{N}(0,0.1)$ in Low-Noise and $\delta \sim \mathcal{N}(0,0.5)$ in High-Noise settings}
        \label{fig:pacman-nonsemantic-PacmanDQN-Egreedy}
\end{figure*}

\begin{figure*}[t]
\centering
        \includegraphics[width=0.9\textwidth]{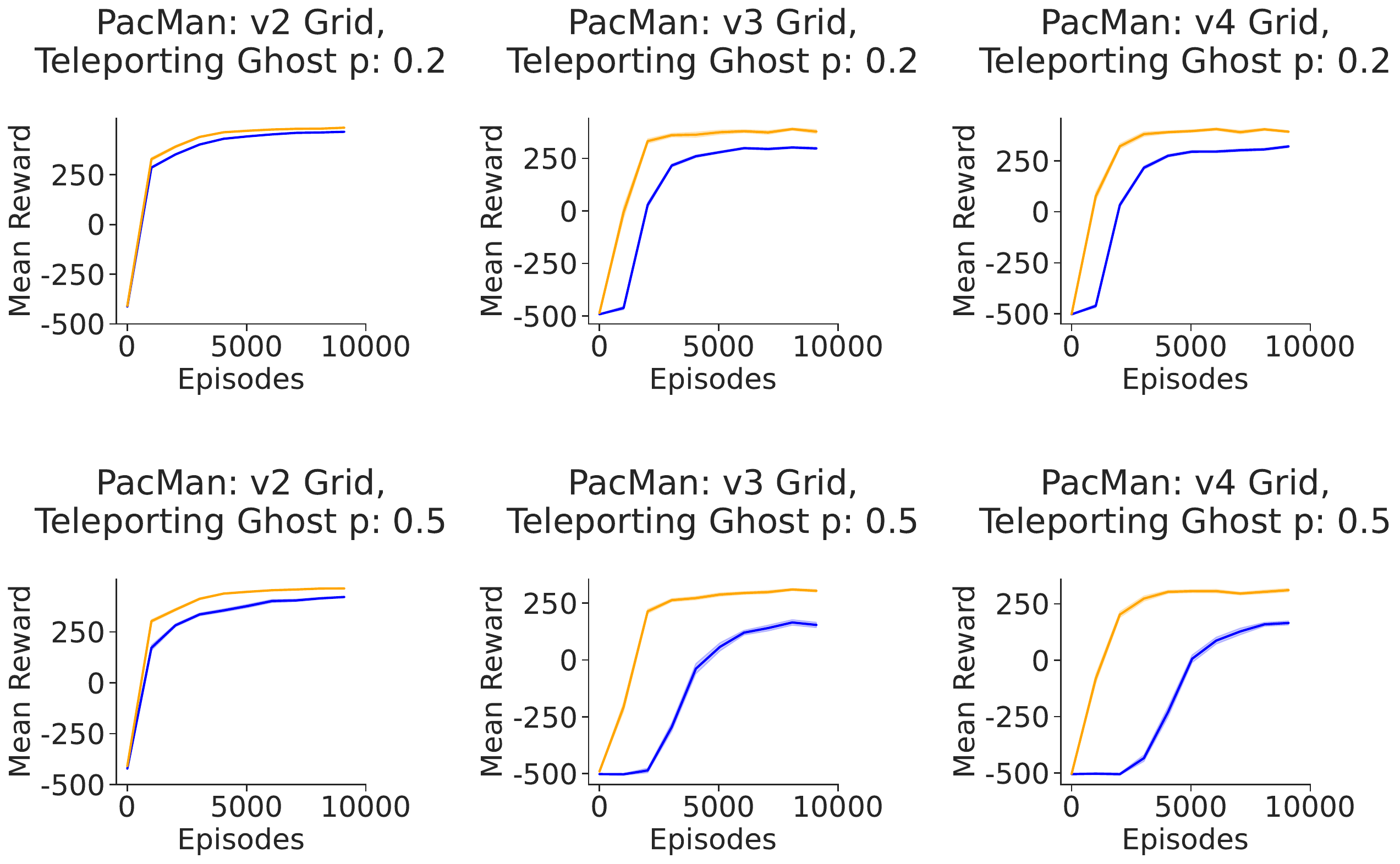} 
        \caption{\emph{DQN Agent with $\epsilon$\text{-}greedy exploration strategy across semantic variations}: Results for PacMan v2, v3, v4 grids reporting mean reward as a function of episode number. The agent is trained on the Random Ghost environment and tested on the Teleporting Ghost variation ($p=0.2$, $p=0.5$)}
        \label{fig:pacman-semantic-PacmanDQN-Egreedy}
\end{figure*}

To address the generalizability of the Indoor-Training Effect to deep learning-based methods, we extended our experiments to include DQN, a widely used deep reinforcement learning algorithm. The results, reported in the appendix, demonstrate the consistency of our findings with DQN as shown in \ref{fig:pacman-nonsemantic-PacmanDQN-Egreedy} for non-semantic variations of Pac Man's grids v1,v2,v3 and \ref{fig:pacman-semantic-PacmanDQN-Egreedy} for semantic variations of the same grids. In particular, we trained agents for $10,000$ episodes and averaged results over $50$ trained agents. After every $100$ training episodes, agents were evaluated using $10$ testing episodes. We report the mean reward curves at convergence.

\section{Perturbation Bounds}\label{supp-sec:perturbations}
\begin{figure*}[t]
\centering
        \includegraphics[width=0.9\textwidth]{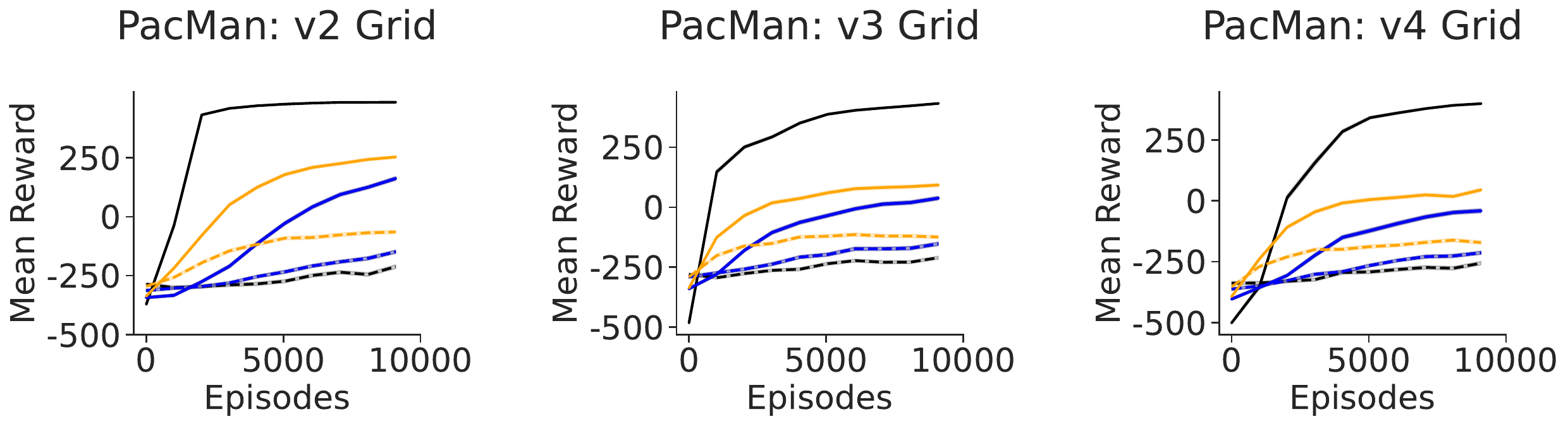} 
        \caption{\emph{SARSA Agent with $\epsilon$\text{-}greedy exploration strategy}: Results for Pac-Man (v2, v3, v4) grids showing the mean reward as a function of episode number. The agent is trained in the non-noisy environment and tested under varying noise levels: Low-Noise ($\delta \sim \mathcal{N}(0, 0.1)$), High-Noise ($\delta \sim \mathcal{N}(0, 0.5)$), and the extremes of maximum noise ($\delta \sim \mathcal{N}(0, 1)$) and no noise ($\delta \sim \mathcal{N}(0, 0)$), represented in black.}
        \label{fig:bounds-pacman-SarsaAgent-Egreed}
\end{figure*}

\begin{figure*}[t]
\centering
        \includegraphics[width=0.9\textwidth]{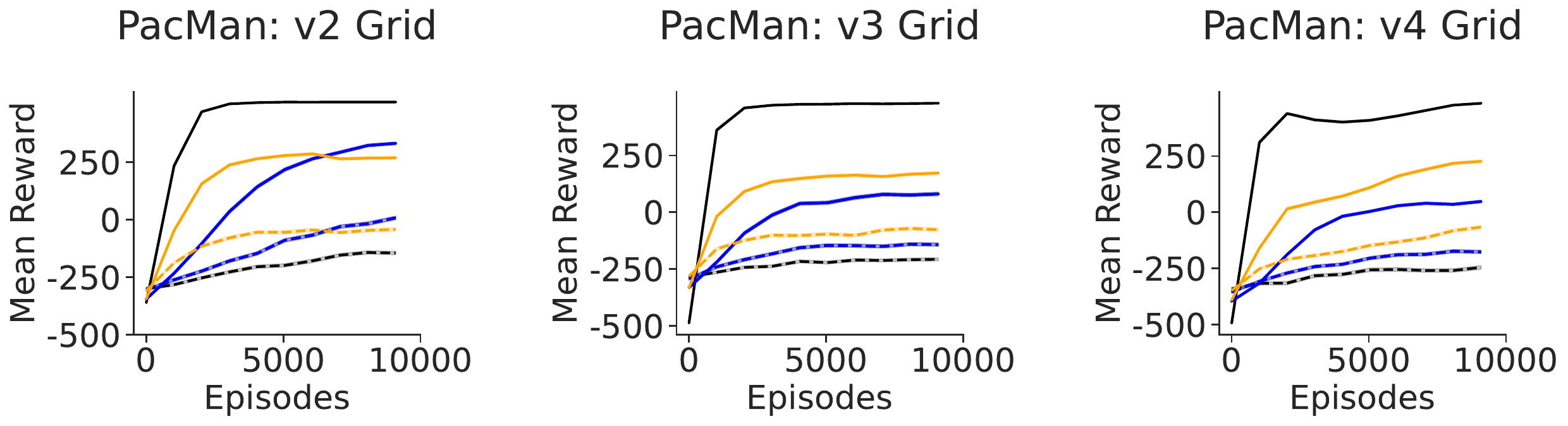} 
        \caption{\emph{Boltzmann Agent with $\epsilon$\text{-}greedy exploration strategy}: Results for Pac-Man (v2, v3, v4) grids showing the mean reward as a function of episode number. The agent is trained in the non-noisy environment and tested under varying noise levels: Low-Noise ($\delta \sim \mathcal{N}(0, 0.1)$), High-Noise ($\delta \sim \mathcal{N}(0, 0.5)$), and the extremes of maximum noise ($\delta \sim \mathcal{N}(0, 1)$) and no noise ($\delta \sim \mathcal{N}(0, 0)$), represented in black.}
        \label{fig:bounds-pacman-BoltzmannAgent-Egreedy}
\end{figure*}

\begin{figure*}[t]
\centering
        \includegraphics[width=0.9\textwidth]{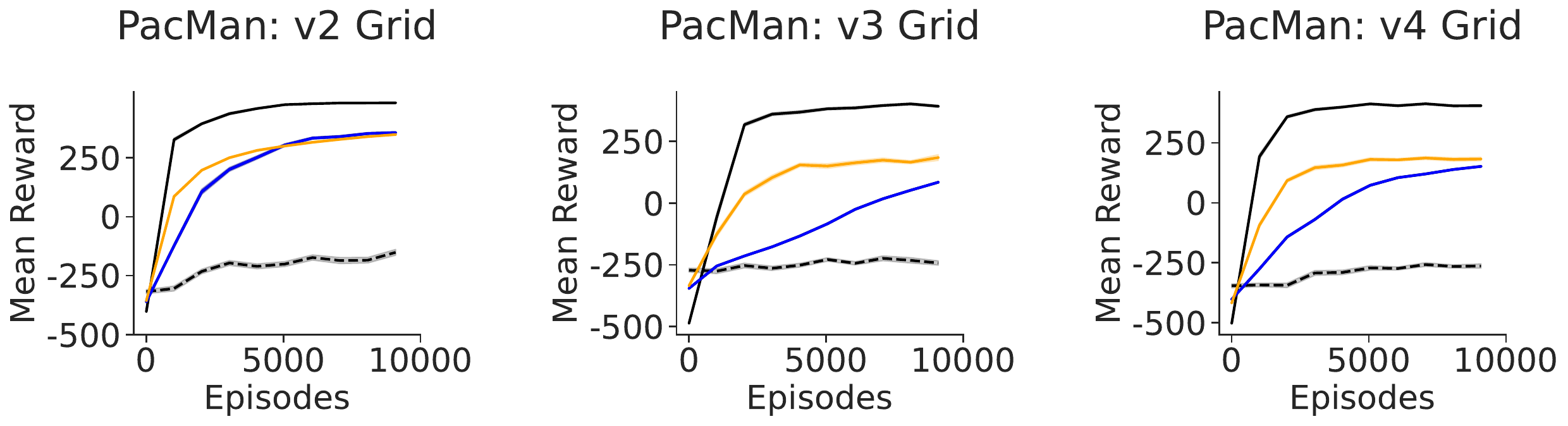} 
        \caption{\emph{DQN Agent with $\epsilon$\text{-}greedy exploration strategy}: Results for Pac-Man (v2, v3, v4) grids showing the mean reward as a function of episode number. The agent is trained in the non-noisy environment and tested under varying noise levels: Low-Noise ($\delta \sim \mathcal{N}(0, 0.1)$), High-Noise ($\delta \sim \mathcal{N}(0, 0.5)$), and the extremes of maximum noise ($\delta \sim \mathcal{N}(0, 1)$) and no noise ($\delta \sim \mathcal{N}(0, 0)$), represented in black.}
        \label{fig:bounds-pacman-PacmanDQN-Egreedy}
\end{figure*}

We include additional graphs illustrating the performance bounds under maximum and minimum perturbation settings. Specifically, we tested agents in the extreme scenarios of no noise ($\delta \sim \mathcal{N}(0, 0)$) and maximum noise ($\delta \sim \mathcal{N}(0, 1)$) and visualize it with the learnability and generalization agents under the Low-Noise regime. Results across the different agents are reported in \ref{fig:bounds-pacman-SarsaAgent-Egreed}, \ref{fig:bounds-pacman-BoltzmannAgent-Egreedy}, \ref{fig:bounds-pacman-PacmanDQN-Egreedy}


\begin{figure*}[t]
    \centering
        \includegraphics[width=0.9\textwidth]{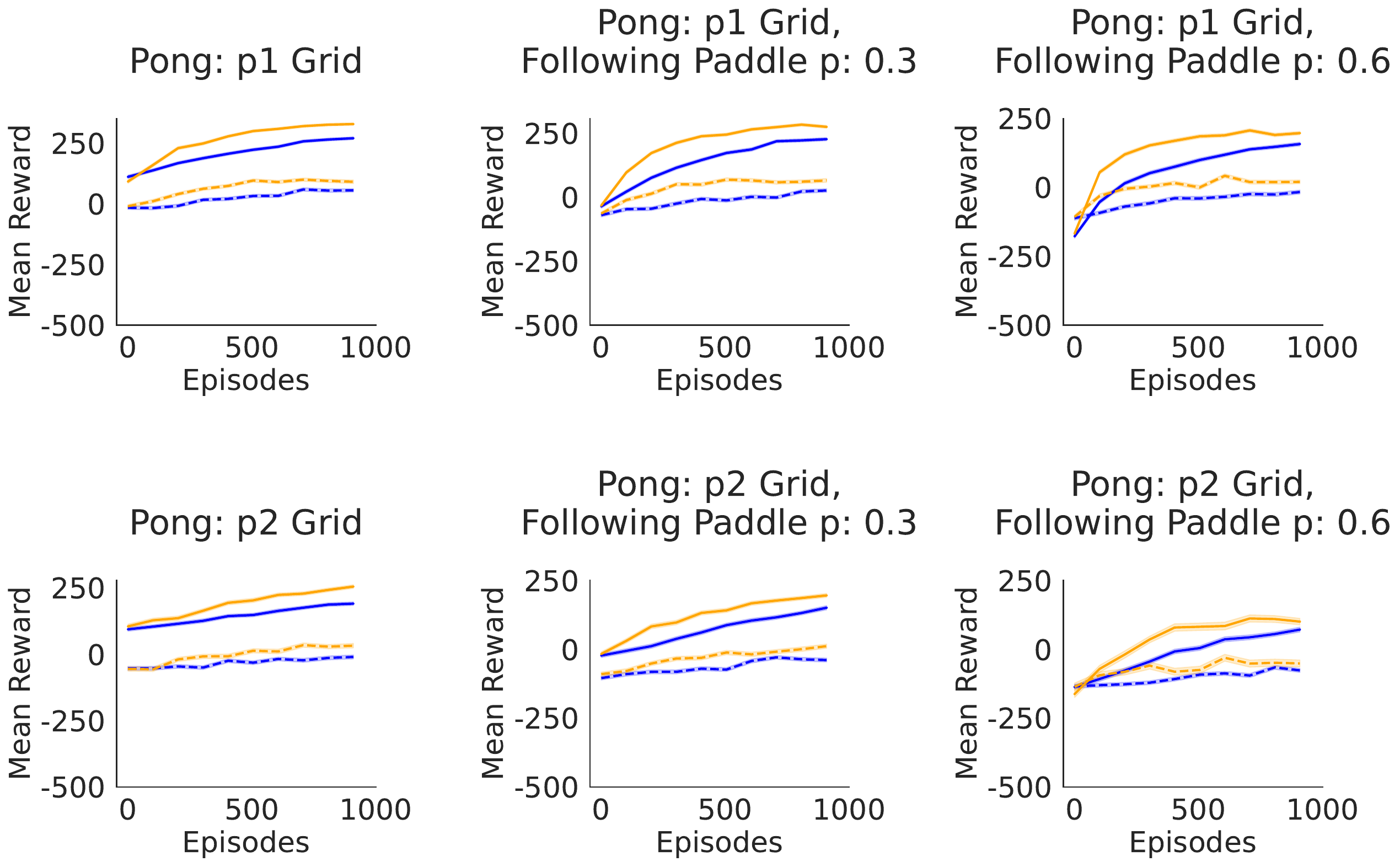} 
        \caption{\emph{SARSA Agent with Boltzmann exploration strategy}: Results for Pong p1, p2 grids reporting mean reward as a function of episode number. The agent is trained on the non-noisy version of the environment and tested on different level of noise ($\delta \sim \mathcal{N}(0,0.1)$ in Low-Noise and $\delta \sim \mathcal{N}(0,0.5)$ in High-Noise settings).
        }
        \label{fig:atari_variations-pong-sarsa-boltzmann}
\end{figure*}

\begin{figure*}[t]
\centering
        \includegraphics[width=0.9\textwidth]{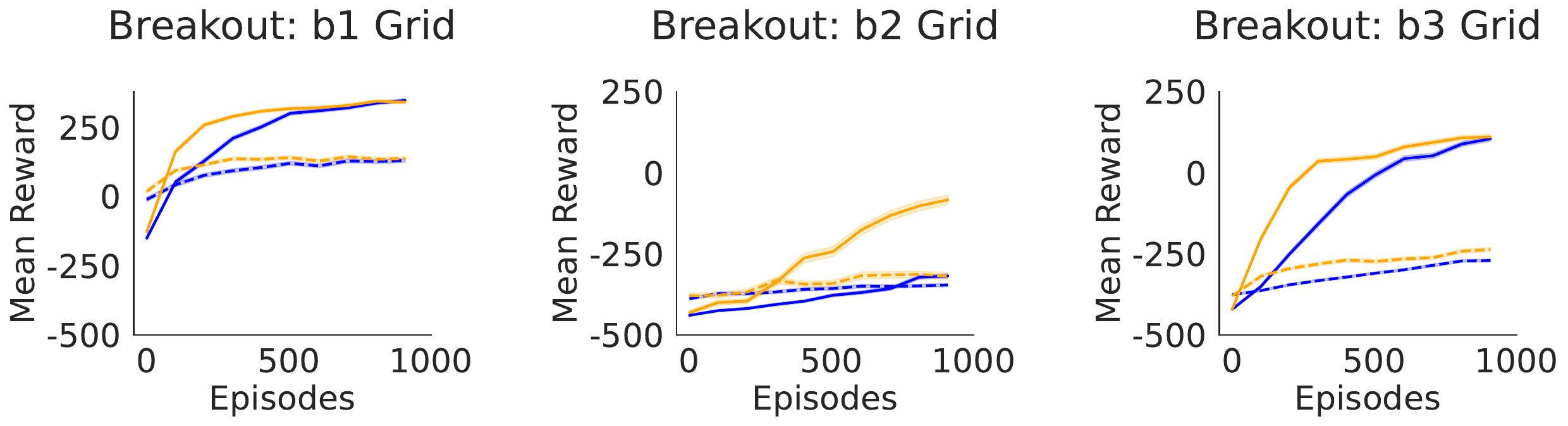} 
        \caption{\emph{SARSA Agent with Boltzmann exploration strategy}: Results for Breakout b1, b2, b3 grids reporting mean reward as a function of episode number. The agent is trained on the non-noisy version of the environment and tested on different level of noise ($\delta \sim \mathcal{N}(0,0.1)$ in Low-Noise and $\delta \sim \mathcal{N}(0,0.5)$ in High-Noise settings).
        }
        \label{fig:atari_variations-breakout-sarsa-boltzmann}
\end{figure*}

\begin{figure*}[t]
\centering
        \includegraphics[width=0.9\textwidth]{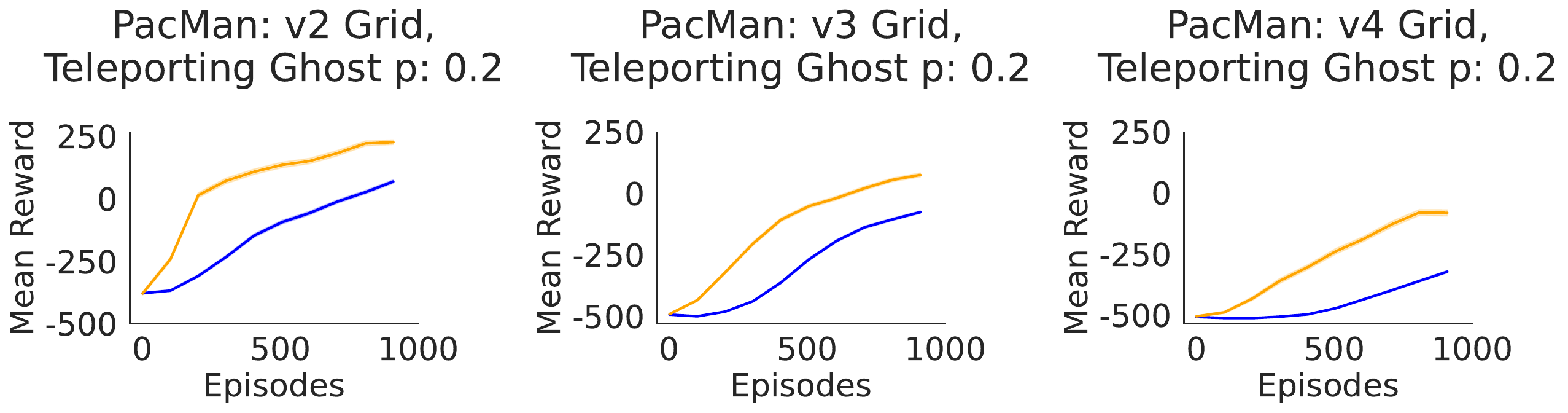} 
        \caption{\emph{SARSA Agent with Boltzmann exploration strategy}: Results for PacMan v2, v3, v4 grids reporting mean reward as a function of episode number. The agent is trained on the Random Ghost environment and tested on the Teleporting Ghost variation ($p=0.2$, $p=0.5$)}
        \label{fig:atari_variations-semantic-pacman-sarsa-boltzmann}
\end{figure*}


\begin{figure*}[t]
\centering
        \includegraphics[width=0.65\textwidth]{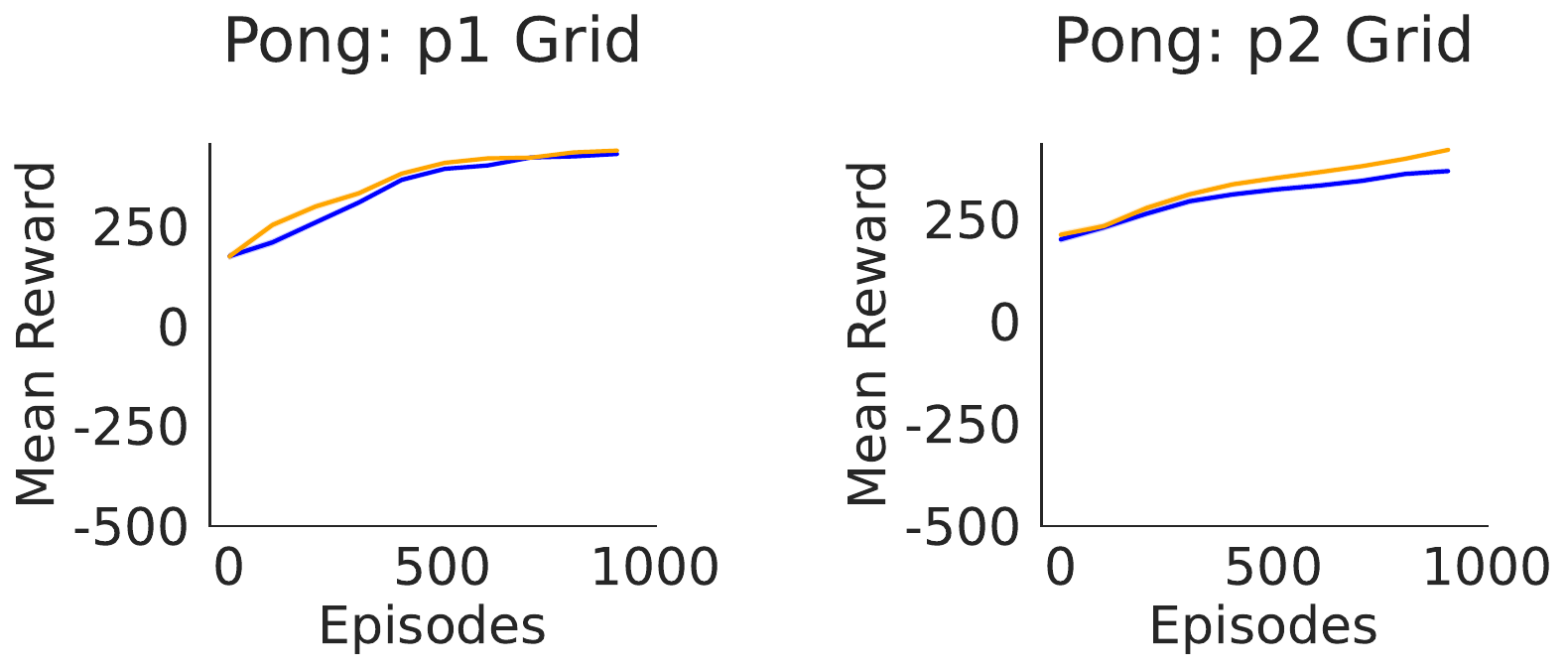} 
        \caption{\emph{SARSA Agent with $\epsilon\text{-}$greedy exploration strategy}: Results for Pong p1, p2 grids reporting mean reward as a function of episode number. The agent is trained on the Directional Ghost ($p=0.3$) environment and tested on the Random Ghost variation.}
        \label{fig:atari_variations-semantic-pong-sarsa-boltzmann}
\end{figure*}

\begin{figure*}[t]
\centering
        \includegraphics[width=0.65\textwidth]{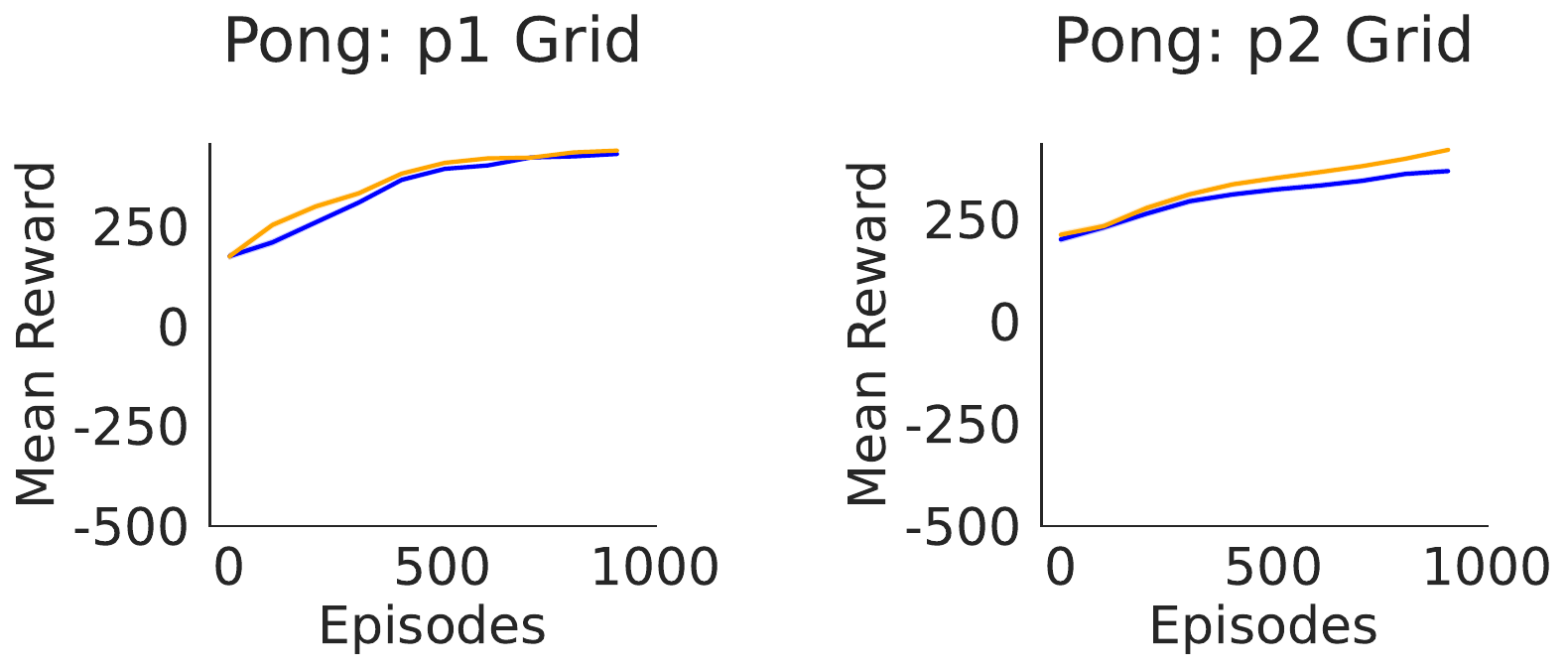} 
        \caption{\emph{SARSA Agent with $\epsilon\text{-}$greedy exploration strategy}: Results for Pong p1, p2 grids reporting mean reward as a function of episode number. The agent is trained on the Directional Ghost ($p=0.3$) environment and tested on the Random Ghost variation.}
        \label{fig:atari_variations-semantic-pong-sarsa-boltzmann}
\end{figure*}

%%%%%

%%%%%%

\begin{figure*}[t]
  %\centering
  \begin{subfigure}{0.32\textwidth}
    \includegraphics[width=\linewidth]{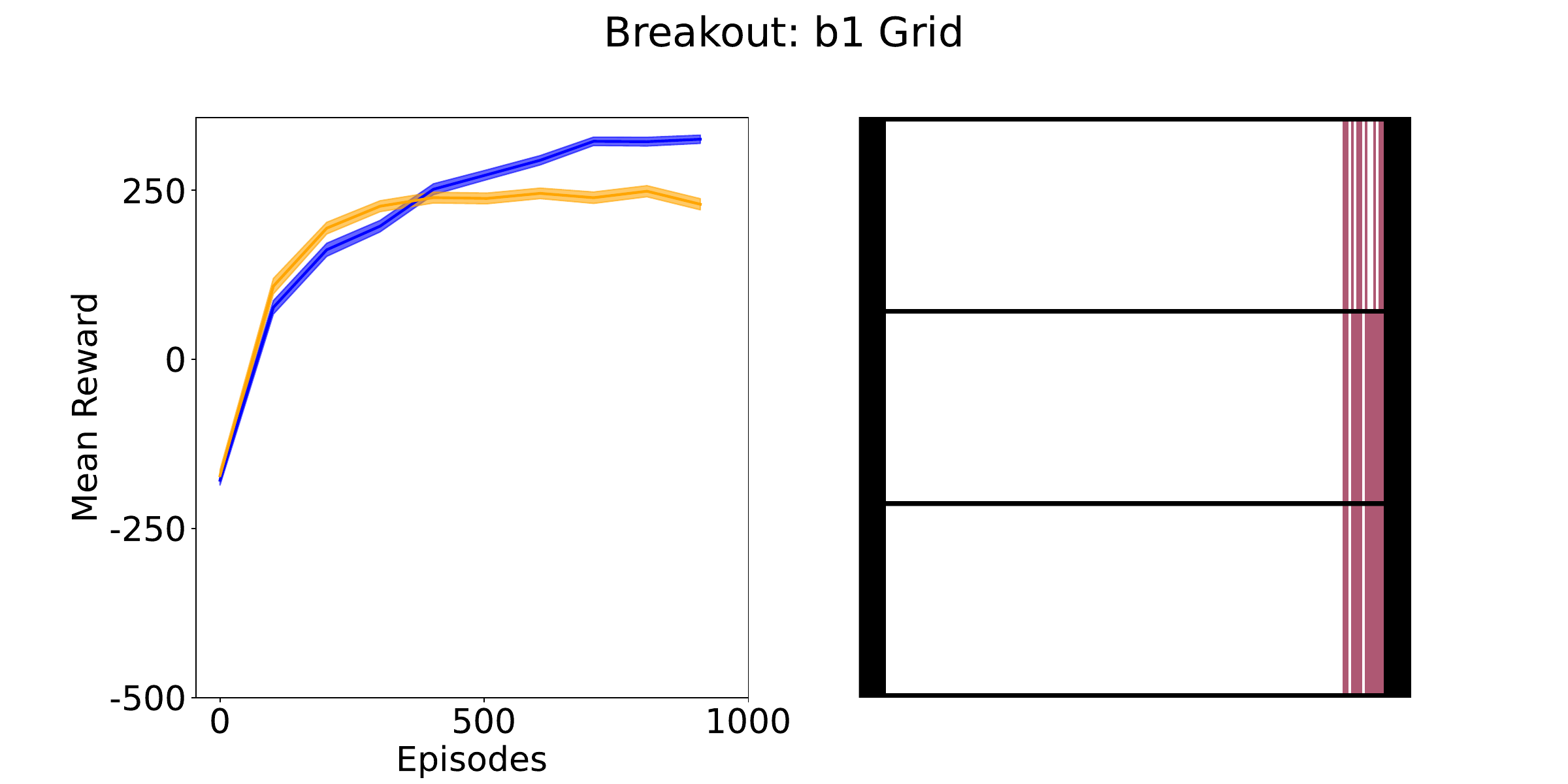}
  \end{subfigure}
  \hfill
  \begin{subfigure}{0.32\textwidth}
    \includegraphics[width=\linewidth]{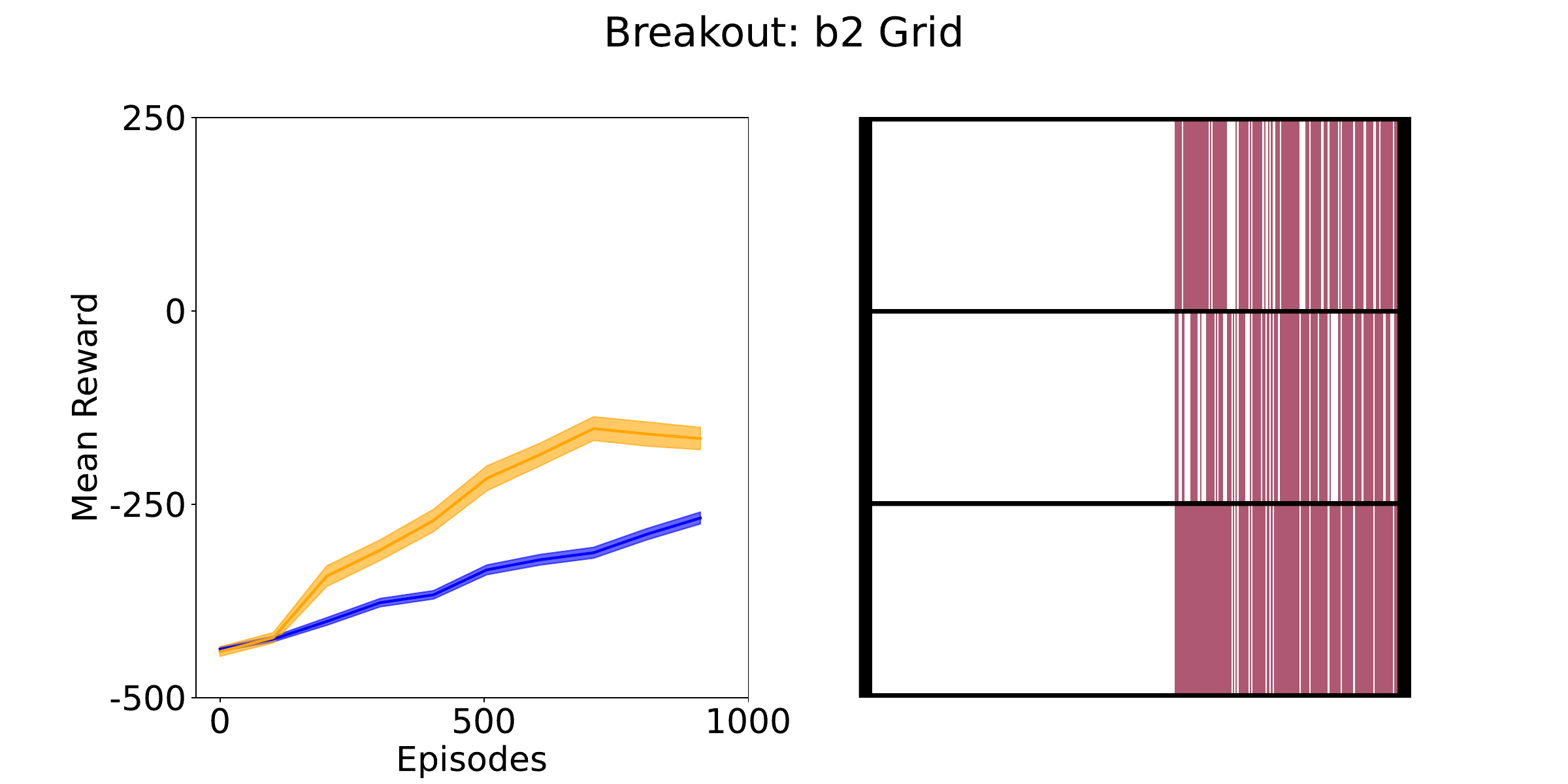}
  \end{subfigure}
  \hfill
  \begin{subfigure}{0.32\textwidth}
    \includegraphics[width=\linewidth]{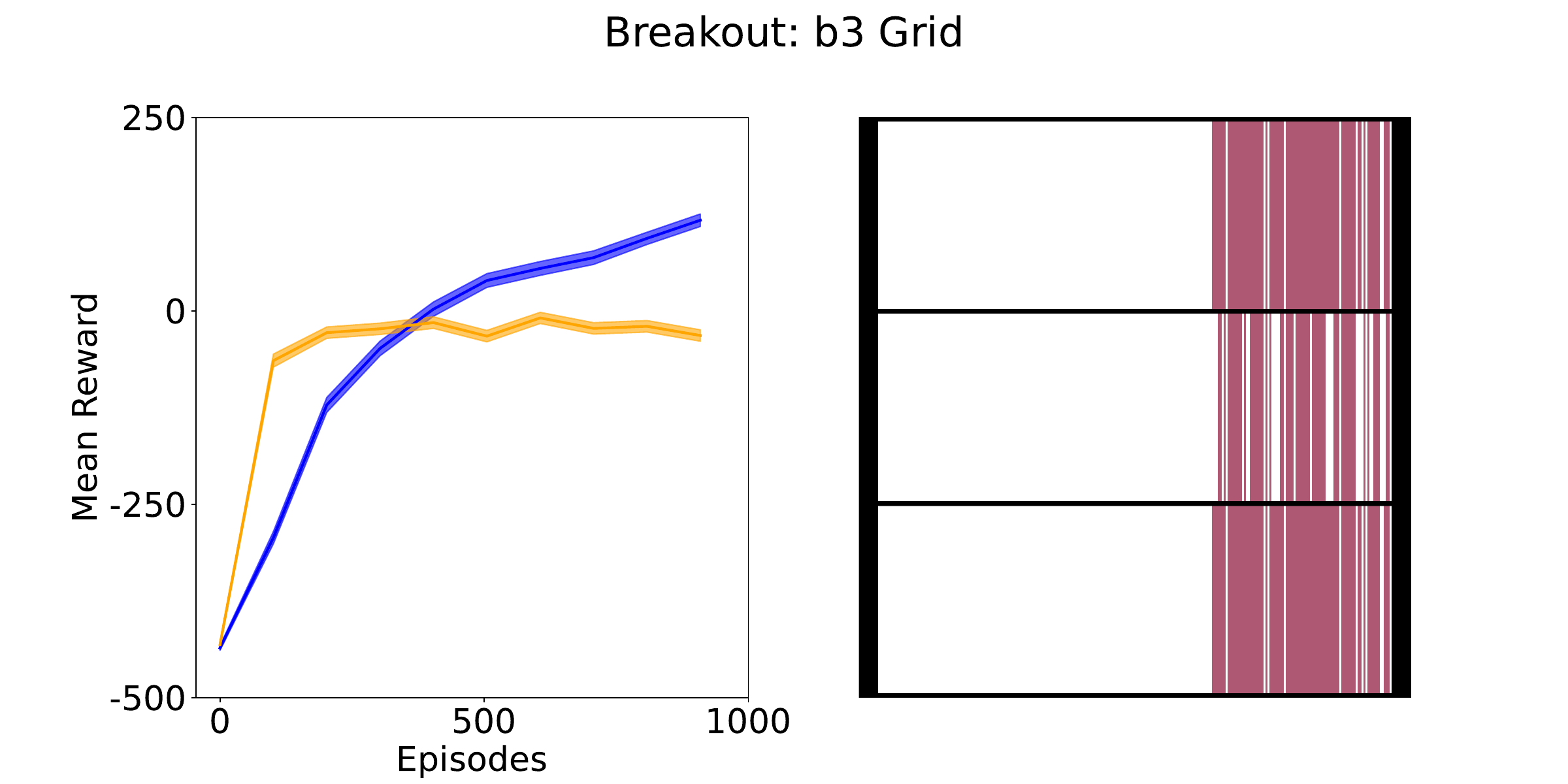}
  \end{subfigure}

  \caption{\emph{Q-learning Agent with Boltzmann exploration strategy}: The \textit{exploration grid} visualizing the difference in State-Action (S-A) pairs explored by these agents ($D_{LG}$). Results for Breakout b1, b2, b3 grids, the agent is trained on non-noisy variations of different environments (reported in the headings) and tested in the Low-Noise regime. Rows in the right figure represents agent's actions Left, Right, Stop.}
  \label{fig:atari_variations-exploration-breakout-qlearning-boltzmann}
\end{figure*}

% \end{document}

% \end{document}
\end{document}